 \documentclass[iicol,sn-basic]{sn-jnl}
\usepackage{graphicx}%
\usepackage{multirow}%
\usepackage{amsmath,amssymb,amsfonts}%
\usepackage{amsthm}%
\usepackage{mathrsfs}%
\usepackage[title]{appendix}%
\usepackage{xcolor}%
\usepackage{textcomp}%
\usepackage{manyfoot}%
\usepackage{booktabs}%
\usepackage{algorithm}
\usepackage{algorithmic}
\usepackage{cite}
\usepackage{listings}%
\usepackage{relsize}
\usepackage{threeparttable}
\usepackage{subfig} 
\definecolor{revision}{RGB}{31,33,175}
\usepackage{hyperref}
\usepackage{soul}
\usepackage{caption}

\theoremstyle{thmstyleone}%
\newtheorem{theorem}{Theorem}%  meant for continuous numbers
\newtheorem{proposition}{Proposition}% to get separate numbers for theorem and proposition etc.
\newtheorem{corollary}{Corollary}
\theoremstyle{thmstyletwo}%

\theoremstyle{thmstylethree}%

\newtheorem{lemma}{Lemma}
\newtheorem{assumption}{Assumption}
\usepackage{arydshln}

\usepackage{hyperref}
\definecolor{mydarkblue}{rgb}{0,0.08,0.45}
\hypersetup{ %
    pdfkeywords={},
    pdfborder=0 0 0,
    pdfpagemode=UseNone,
    colorlinks=true,
    linkcolor=mydarkblue,
    citecolor=mydarkblue,
    filecolor=mydarkblue,
    urlcolor=mydarkblue,
}
\usepackage{bbding}
\usepackage{xspace}
\makeatletter
\DeclareRobustCommand\bmvaOneDot{\futurelet\@let@token\bmv@onedotaux}
\def\bmv@onedotaux{\ifx\@let@token.\else.\null\fi\xspace}
\makeatother

\usepackage{mathtools}

\begin{document}

\title[Article Title]{Curvature Learning for Generalization of Hyperbolic Neural Networks}

%%=============================================================%%
%% GivenName	-> \fnm{Joergen W.}
%% Particle	-> \spfx{van der} -> surname prefix
%% FamilyName	-> \sur{Ploeg}
%% Suffix	-> \sfx{IV}
%% \author*[1,2]{\fnm{Joergen W.} \spfx{van der} \sur{Ploeg} 
%%  \sfx{IV}}\email{iauthor@gmail.com}
%%=============================================================%%

\author[1]{Xiaomeng Fan}\email{fanxiaomeng@bit.edu.cn}

\author[1]{Yuwei Wu}\email{wuyuwei@bit.edu.cn}
% \equalcont{These authors contributed equally to this work.}

\author*[1]{Zhi Gao}\email{zhi.gao@bit.edu.cn}
% \equalcont{These authors contributed equally to this work.}

\author[3]{Mehrtash Harandi}\email{mehrtash.harandi@monash.edu}
% \equalcont{These authors contributed equally to this work.}

\author*[2]{and Yunde Jia}\email{jiayunde@bit.edu.cn}
% \equalcont{These authors contributed equally to this work.}

\affil[1]{\orgdiv{Beijing Laboratory of Intelligent
Information Technology}, \orgname{School of Computer Science, Beijing
Institute of Technology (BIT)}, \orgaddress{ \city{Beijing}, \postcode{100081}, \country{P.R. China}}}

\affil[2]{\orgdiv{Guangdong Laboratory of Machine
Perception and Intelligent Computing}, 
\\
\orgname{Shenzhen MSU-BIT University}, \orgaddress{ \city{Shenzhen}, \postcode{518172}, \country{P.R. China}}}

\affil[3]{\orgdiv{Department of Electrical and Computer
Systems Eng.}, \orgname{Monash University, and Data61}, \orgaddress{ \country{Australia}}}

%%==================================%%
%% Sample for unstructured abstract %%
%%==================================%%

\abstract{Hyperbolic neural networks (HNNs) have demonstrated notable efficacy in representing real-world data with hierarchical structures via exploiting the geometric properties of hyperbolic spaces characterized by negative curvatures.
Curvature plays a crucial role in optimizing HNNs. Inappropriate curvatures may cause HNNs to converge to suboptimal parameters, degrading overall performance. 
So far, the theoretical foundation of the effect of curvatures on HNNs has not been developed.
In this paper, we derive a PAC-Bayesian generalization bound of HNNs, highlighting the role of curvatures in the generalization of HNNs via their effect on the smoothness of the loss landscape.
Driven by the derived bound, we propose a sharpness-aware curvature learning method to smooth the loss landscape, thereby improving the generalization of HNNs. 
In our method, 
  we design a scope sharpness measure for curvatures, which is minimized through a bi-level optimization process. Then, we introduce an implicit differentiation algorithm that efficiently solves the bi-level optimization by approximating gradients of curvatures. We present the approximation error and convergence analyses of the proposed method, showing that the approximation error is upper-bounded, and the proposed method can converge by bounding gradients of HNNs. Experiments on four settings:  classification, learning from long-tailed data, learning from noisy data, and few-shot learning show that our method can improve the performance of HNNs.}

\keywords{Hyperbolic neural networks,  Curvature learning,  PAC-Bayesian generalization bound, Implicit differentiation.}

%%\pacs[JEL Classification]{D8, H51}

%%\pacs[MSC Classification]{35A01, 65L10, 65L12, 65L20, 65L70}

\maketitle

\section{Introduction}
	Hyperbolic neural networks (HNNs) work well for modeling real-world data with inherent hierarchical  structures~\citep{khrulkov2020hyperbolic, guo2022clipped}. The efficacy of HNNs stems from their ability to capitalize on the distinct representational capabilities of hyperbolic spaces, characterized by negative curvatures~\citep{peng2021hyperbolic, mettes2024hyperbolic}.
Hyperbolic spaces offer an effective way to capture hierarchical structures because their volume grows exponentially with the radius, 
consistent with the exponential increase of data numbers in the hierarchical structures with depth~\citep{fang2021kernel}.
The curvature is significant for optimizing HNNs, and many works have empirically shown that curvature affects the performance of HNNs across various tasks, such as image classification~\citep{gao2023curvature} and node classification~\citep{fu2023hyperbolic}.
So far, the theoretical foundation of the effect of curvatures on HNNs has not been developed.

In this paper, we present a theoretical framework  for exploring the effect of curvatures in hyperbolic neural networks (HNNs). 
We analyze the Lipschitz continuity of hyperbolic operations and derive the corresponding Lipschitz constants.
Then, we establish a Lipschitz bound analysis that quantifies the approximation error between HNNs under different tangent points. 
We further derive a PAC-Bayesian generalization bound for HNNs, which indicates that the minima of HNNs with a smoother loss landscape generalize better than those with a sharper loss landscape.

We observe that Euclidean studies~\citep{foret2020sharpness,dinh2017sharp} have reached similar conclusions, and our derivation augments them.
Along the way, we further find that curvatures of hyperbolic spaces affect the smoothness of the loss landscape in the derived theorem, highlighting the significance of curvatures for generalization in HNNs.
% This finding drives us to learn optimal curvatures to smooth the loss landscape, thereby improving the generalization of HNNs.
This finding drives us to investigate a curvature learning method to smooth the loss landscape, thereby improving the generalization of HNNs.

To this end, we propose a sharpness-aware curvature learning method. 
Specifically, we design a scope sharpness measure to capture the sharpness of a local minimum in HNNs within a given scope around the minimum.
Then, the scope sharpness measure is minimized to learn optimal curvatures, providing a feasible solution to smooth the loss landscape and enhance the generalization of HNNs.
Minimizing the measure is formulated as a bi-level optimization problem (\textit{i.e.}, a nested optimization problem with inner and outer levels), where HNNs are trained in the inner level, and curvatures are optimized in the outer level. 
In the bi-level optimization, computing gradients of curvatures poses computational challenges due to the need to calculate multiple Hessian matrices and unroll the inner-level optimization. We introduce an implicit differentiation algorithm that presents an approximated solution for gradients of curvatures, reducing the computational load.

% Specifically, we design a scope sharpness measure to capture the sharpness across a given scope of a local minima in HNNs.
% of minima in HNNs within a given scope.

% Solving the bi-level optimization problem poses computational challenges  due to the need to calculate multiple Hessian matrices and unroll the inner-level optimization. We introduce an implicit differentiation algorithm that derives an approximated solution for gradients of curvatures, significantly reducing the computational load.

% 
% derive an approximated solution for gradients of curvatures by making use of the implicit differentiation theory~\citep{lorraine2020optimizing}, significantly reducing the computational load.

The effectiveness of our method is demonstrated both theoretically and empirically. 
Theoretically, we prove that the approximation error in the proposed method has an upper bound, showing the rationality of approximation ways in the proposed method. We also present convergence guarantees by bounding gradients of parameters and curvatures in HNNs.
Empirically, we conduct a comprehensive set of experiments to evaluate the proposed method on four tasks: classification, learning from long-tailed data, learning from noisy data, and few-shot learning. Experimental results show that the proposed method can efficiently smooth the loss landscape and thus improve the generalization of HNNs.
% achieving superior performance in comparison to existing methods.
Our contributions are summarized as follows:

\begin{itemize}
    \item 
        We derive the PAC-Bayesian generalization bound of HNNs, 
        highlighting the significance of curvatures for the generalization in HNNs.
       
    \item  We present foundational theoretical analyses of HNNs by characterizing the Lipschitz continuity of hyperbolic operations and quantifying the Lipschitz bound of HNNs across different tangent points.

    \item  We propose a sharpness-aware curvature learning method that optimizes curvatures of HNNs to smooth the loss landscape, enhancing the generalization of HNNs.

     \item  We introduce an implicit differentiation that brings an efficient bi-level optimization process by approximating gradients of curvatures. 
     
    \item  We present the upper bound and convergence guarantees of our method.

 % connecting the generalization ability of HNNs with their curvatures.
    % capability
    % which reduces the time and memory consumption by avoiding the unrolling procedure and simplifying the complex computation
    %
      % We develop the theoretical analysis of the effect of curvatures on HNNs.
    % demonstrate that approximating gradients of curvatures are 
   
    % \MH{I think the implicit simplification is also a significant contribution}
\end{itemize}

 \section{Related Work}
\subsection{Hyperbolic Neural Networks}
% hyperbolic spaces have shown superior performance in many tasks~\citep{hong2023hyperbolic, li2023euclidean, long2023cross}. Hyperbolic neural networks~\citep{ganea2018hyperbolic} incorporate several hyperbolic operations to obtain hyperbolic embeddings and employ the hyperbolic multinomial logistic regression (MLR) if classification is deemed. Subsequent efforts to improve HNNs include Clipped HNNs (C-HNNs)~\citep{guo2022clipped} to address the vanishing gradient in HNNs by clipping features, studies such as~\citep{khrulkov2020hyperbolic, yan2021unsupervised, atigh2022hyperbolic} to learn better embeddings and extension of various neural architectures to hyperbolic spaces~\citep{shimizu2020hyperbolic, dai2021hyperbolic, gulcehre2018hyperbolic, skopek2019mixed, van2023poincar}.
% concentrate on extending famous Euclidean neural networks to hyperbolic spaces, such as hyperbolic convolutional network and hyperbolic graph network.
% , such as hyperbolic convolutional network~\citep{shimizu2020hyperbolic}, hyperbolic graph network~\citep{dai2021hyperbolic}, hyperbolic attention network~\citep{gulcehre2018hyperbolic}, and hyperbolic variational autoencoder~\citep{skopek2019mixed}. 
% Different from them, we develop the generalization bound for HNNs and focus on improving the generalization ability of HNNs.

Modeling via hyperbolic spaces has shown superior performances in many tasks due to their capabilities in encoding data with hierarchical structures. Hyperbolic neural networks~\citep{ganea2018hyperbolic} incorporate several hyperbolic operations on the top of a neural network to obtain hyperbolic embeddings. 
% and design the hyperbolic MLR as a classifier. 
~\citet{guo2022clipped} further improve the training procedure of HNNs by introducing feature clipping to prevent vanishing gradients.
% After this work, Guo~\textit{et al.}~\citep{guo2022clipped} design a more powerful hyperbolic classifier via clipping the features to avoid vanishing gradients.
Efforts to learn powerful embeddings in the hyperbolic spaces have been explored for various applications, including but not limited to image retrieval~\citep{khrulkov2020hyperbolic}, medical image recognition~\citep{yu2022skin}, action recognition~\citep{long2020searching}, anomaly recognition~\citep{hong2023curved}, audio-visual learning~\citep{hong2023hyperbolic}, image segmentation~\citep{atigh2022hyperbolic}, anomaly detection~\citep{Li_2024_CVPR}, and 3D visual grounding~\citep{Wang_2024_CVPR}.
Moreover, extensions of various neural architectures from Euclidean to hyperbolic spaces have been actively studied, such as hyperbolic convolutional~\citep{shimizu2020hyperbolic}, graph network~\citep{dai2021hyperbolic}, attention mechanism~\citep{gulcehre2018hyperbolic}, and variational autoencoder~\citep{skopek2019mixed}. 
In addition, several works develop the well-known learning paradigms from Euclidean spaces to hyperbolic spaces, such as metric learning~\citep{yan2021unsupervised}, contrastive learning~\citep{ge2023hyperbolic},  self-supervised learning~\citep{franco2023hyperbolic_ss}, active learning~\citep{franco2023hyperbolic_al}, clustering learning~\citep{lin2023mhcn}, and continual learning~\citep{gao2023exploring}. Different from them, we are the first to develop a generalization error bound of HNNs and further improve the generalization of HNNs by learning curvatures.

% In contrast, we develop the PAC Bayesian generalization error bound and improve the generalization ability for HNNs.

% Existing researches of learning on the hyperbolic spaces can be divided into two categories: learning hyperbolic embeddings and designing hyperbolic neural networks structure. The first category method incorporate several hyperbolic operations on the top of conventional neural network to obtain hyperbolic embeddings and utilize hyperbolic MLR as classifier

\subsection{Curvature Learning Methods}

Existing curvature learning methods aim to capture geometrical structures of data. In computer vision, 
curvature learning methods aim to generate curvatures to match the characteristics of image data. 
The methods proposed by~\citet{gao2021curvature,gao2023curvature} automatically generate task-specific curvatures to adapt models to diverse geometric structures. A Riemannian AdamW optimizer is introduced to learn curvatures for improving training stability and computational efficiency~\citep{bdeir2024optimizing}.
In data mining, curvature learning methods design graph curvatures to capture the geometric embeddings of graph data. 
Hyperbolic graph convolutional networks~\citep{fu2021ace} treat curvatures as trainable parameters that are optimized jointly with model weights. A reinforcement learning-based strategy proposed by~\citet{chami2019hyperbolic} dynamically learns task-specific curvatures to better represent complex graph structures. ~\citet{fu2023hyperbolic} design class-aware graph curvatures that  measure the effect of label information and neighborhood connectivity to address the hierarchy imbalance issue in node classification. 
$\kappa$hgcn method~\citep{yang2023kappahgcn} integrates learnable graph curvatures and curvature-based filtering mechanisms to enhance performance.
Instead of focusing on the geometrical structures of data, we theoretically explore the effect of curvatures on HNNs, demonstrating that the generalization of HNNs can be improved by curvature learning.

\subsection{Flat Minima and Optimization Methods}
\citet{hochreiter1994simplifying} are the first to demonstrate that the smoothness of the loss landscape affects the generalization of neural networks.
Subsequent efforts to improve generalization by smoothing the loss landscape have largely fallen into two categories. 
Methods of the first category opt for designing sharpness measures and minimizing them to improve the generalization. ~\citet{dinh2017sharp} propose the sharpness measure by utilizing the spectrum of the Hessian. Other works~\citep{liang2019fisher, petzka2021relative} present the scale-invariance sharpness measure, while the method proposed by~\citet{jang2022reparametrization} introduces a reparametrization-invariant sharpness measure. 

Methods of the second category focused on finding flat minima through sharpness-aware minimization. Sharpness-aware minimization (SAM) proposed by~\citet{foret2020sharpness} is the pioneering 
work, which seeks parameters that lie in neighborhoods having uniformly low loss value.
Subsequently,  some works focus on reducing the complexity of SAM by adaptively applying SAM (instead of using it at every iteration), thereby speeding up SAM~\citep{jiang2023adaptive}.
Some works focus on designing improved variants for SAM. For example, the adaptive sharpness-aware minimization method~\citep{kwon2021asam} adaptively adjusts maximization regions to realize scale-invariant parameters, and the surrogate gap minimization method~\citep{zhuang2022surrogate}  
models a surrogate gap as an equivalent measure of sharpness to improve the generalization.
Some work aim to refine the SAM method for specific tasks, such as the sharpness-aware MAML method for few-shot learning~\citep{abbas2022sharp}, the data augmented flatness-aware gradient projection for continual learning~\citep{yang2023data}, 
the SAM method for federated learning~\citep{qu2022generalized}, and the class-conditional SAM method for long-tailed recognition~\citep{zhou2023class}. 
In contrast to prior studies predominantly centered on Euclidean spaces, we focus on the effect of sharpness on generalization in hyperbolic spaces and derive the generalization bound of HNNs.

\section{Lipschitz Continuity and Lipschitz Bound in HNNs}
\label{section:lip_hyper_opera}
% To explore the effect the 
To construct the theoretical framework for exploring the effect of curvatures in HNNs, we analyze the Lipschitz continuity and Lipschitz bound in HNNs.
Specifically, we present the details of HNNs (shown in Section~\ref{section:hnns}), and then we analyze the Lipschitz continuity of several hyperbolic operations (shown in Section~\ref{section:lip_continuous}).
Based on the Lipschitz continuity, we develop the Lipschitz bound for changes in tangent point (shown in Section~\ref{section:approximation_error_bound}). 

\subsection{Hyperbolic Neural Networks}
\label{section:hnns}
In this subsection, we first present denotations of hyperbolic spaces and hyperbolic operations, and then introduce details of HNNs. The utilized mathematical notations are summarized in Table~\ref{tabel:notations}.
Unlike Euclidean spaces with the fixed zero curvature,
a hyperbolic spaces $\mathcal{H}^{d,c}$ is a smooth Riemannian manifold with a constant negative curvature $-c$ ~\citep{lee2006riemannian}. 
We choose the $d-$ dimensional Poincar{\'e} ball model~\citep{cannon1997hyperbolic} of constant negative curvature $-c$ to work with. It is denoted as $\mathcal{H}^{d,c} = \{ \boldsymbol{x} \in \mathbb{R}^{d},  c\| \boldsymbol{x} \|^{2} <1  \} $, where $\Vert \cdot \Vert$ is the Euclidean norm. 
The tangent space to $\mathcal{H}$ at a tangent point $\boldsymbol{x}$, denoted as $T_{\boldsymbol{x}}\mathcal{H}$, consists of all tangent vectors at that tangent point. 
% The tangent space to $\mathcal{H}$ at $\boldsymbol{X}$ denoted as $T_{\boldsymbol{X}}\mathcal{H}$ is the set of all tangent vectors. 
The following hyperbolic operations will be used in our work.

\noindent \textbf{Möbius Addition.} For $\boldsymbol{x}, \boldsymbol{y} \in \mathcal{H}^{d,c}$, the Möbius addition of $\boldsymbol{x} $ and $\mathcal{H}^{d,c}$ is
\begin{equation}
\small
	\begin{aligned}
		\boldsymbol{x} \oplus_{c} \boldsymbol{y} = \frac{(1+2c\langle \boldsymbol{x},\boldsymbol{y} \rangle + c\Vert \boldsymbol{y} \Vert^{2})\boldsymbol{x} + (1-c\Vert \boldsymbol{x} \Vert^{2})\boldsymbol{y}}{1 + 2c \langle \boldsymbol{x},\boldsymbol{y}  \rangle +c^{2}\Vert\boldsymbol{x} \Vert^{2}\Vert\boldsymbol{y} \Vert^{2}  }.
	\end{aligned}
\end{equation}

\noindent \textbf{Distance measure.} For $\boldsymbol{x}, \boldsymbol{y} \in \mathcal{H}^{d,c}$, the measure of $\boldsymbol{x} $ and $\mathcal{H}^{d,c}$ is
\begin{equation}
	\begin{aligned}
		&d^{c}(\boldsymbol{x}, \boldsymbol{y}) = \\
            &\frac{1}{\sqrt{|c|}} \text{cosh}^{-1}\left(1-2c\frac{\Vert  \boldsymbol{x}- \boldsymbol{y}\Vert^{2}}{(1+c\Vert  \boldsymbol{x}\Vert^{2})(1+c\Vert  \boldsymbol{y}\Vert^{2})}\right).
	\end{aligned}
\end{equation}
\noindent \textbf{Exponential map.}  The exponential map $\text{expm}_{\boldsymbol{y}}^{c}(\boldsymbol{x})$ projects a vector $\boldsymbol{x}$ from the tangent space at the tangent point $\boldsymbol{y}$ to the manifold $\mathcal{H}^{d,c}$
% $T_{\boldsymbol{y}}\mathcal{H}^{d,c}$ 
\begin{equation}
\small
	\text{expm}_{\boldsymbol{y}}^{c}(\boldsymbol{x}) = \boldsymbol{y} \oplus_{c}\left(\text{tanh}(\sqrt{|c|}\frac{\lambda_{\boldsymbol{y}}^{c}\Vert \boldsymbol{x} \Vert}{2})\frac{\boldsymbol{x}}{\sqrt{|c|}\Vert \boldsymbol{x} \Vert}\right),
\end{equation}
where $\lambda_{\boldsymbol{y}}^{c} = 2/(1-c\Vert\boldsymbol{y}\Vert^{2})$ is the conformal factor.
Setting $\boldsymbol{y} = 0$, the exponential map is computed as 
\begin{equation}
\small
    \begin{aligned}
        \text{expm}_{\boldsymbol{0}}^{c}(\boldsymbol{s}) = \text{tanh}(\sqrt{|c|} \Vert \boldsymbol{s} \Vert )\frac{\boldsymbol{s}}{\sqrt{|c|}\Vert \boldsymbol{s} \Vert}.
    \end{aligned}
\end{equation}

\noindent \textbf{Logarithmic map.}  The Logarithmic map $\text{logm}_{\boldsymbol{y}}^{c}(\boldsymbol{x})$ projects a vector $\boldsymbol{x}$ from the manifold $\mathcal{H}^{d,c}$ to the tangent space $T_{\boldsymbol{y}}\mathcal{H}$,

\begin{equation}
\small
	\begin{aligned}
	   & \text{logm}_{\boldsymbol{y}}^{c}(\boldsymbol{x}) = \\
         & \frac{2}{\sqrt{|c|}\lambda_{\boldsymbol{y}}^{c}} \text{tanh}^{-1}(\sqrt{|c|} \Vert -\boldsymbol{y} \oplus_{c} \boldsymbol{x}  \Vert) \frac{-\boldsymbol{y} \oplus_{c} \boldsymbol{x} }{\Vert -\boldsymbol{y} \oplus_{c} \boldsymbol{x}  \Vert}.
	\end{aligned}
\end{equation}

Hyperbolic neural networks (HNNs)~\citep{ganea2018hyperbolic} combine the formalism of Möbius gyrovector spaces~\citep{ungar2008analytic, ungar2001hyperbolic} with the Riemannian geometry of hyperbolic spaces, which derive hyperbolic versions of deep learning tools: feed-forward networks and multinomial logistic regression.
The  feed-forward technology in HNNs aims to obtain the hyperbolic embeddings.
Given the hyperbolic inputs $\boldsymbol{x} \in \mathcal{H}^{d,c}$, the feed-forward technology in HNNs is modeled as
\begin{equation}
\small
\label{equation:hyperbolic_forward}
    \begin{aligned}
        \boldsymbol{x} \leftarrow {\rm expm}_{\boldsymbol{y}}^{c}(f_{\boldsymbol{a}}({\rm logm}^{c}_{\boldsymbol{y}}(\boldsymbol{x}))) \oplus_{c} \boldsymbol{b},
    \end{aligned}
\end{equation}
where $f_{a}(\cdot)$ is the feed-forward function parameterized as $\boldsymbol{a} \in T_{\boldsymbol{y}}\mathcal{H}$ being the linear mapping weight, and $\boldsymbol{b} \in \mathcal{H}^{d,c}$ is the bias.
By the feed-forward technology, one can obtain the hyperbolic embeddings $\boldsymbol{x}$.
In practice, the tangent point $\boldsymbol{y}$ is set to $\boldsymbol{0}$ to simplify the computations.

\begin{table*}[]
    \centering
    \small
    \caption{Table of notations. }
\label{tabel:notations}
\setlength{\tabcolsep}{0.1mm}{
    \begin{tabular}{ccc}
    \midrule
          \bfseries Notation & \bfseries Section & \bfseries  Definition \\
          \toprule
          $\mathcal{H}^{d,c}$  & 3.1 & d-dimensional hyperbolic spaces \\
          $\mathbb{R}^{d}$  & 3.1 & d-dimensional Euclidean spaces \\
          $T_{\boldsymbol{x}}\mathcal{H} $  & 3.1 & tangent space of hyperbolic spaces \\
          $c$ & 3.1 & curvature of hyperbolic spaces \\
          $\oplus_{c}$ & 3.1 & Möbius addition in hyperbolic spaces \\
          $d^{c}(\cdot)$ & 3.1 & distance measure in hyperbolic spaces \\
          $\text{expm}_{\boldsymbol{y}}^{c}(\boldsymbol{x})$ & 3.1 & exponential map in hyperbolic spaces \\
          $\text{logm}_{\boldsymbol{y}}^{c}(\boldsymbol{x})$ & 3.1 & Logarithmic map in hyperbolic spaces \\
          % $\boldsymbol{w}^{\mathcal{H}}$ & 3.1 & parameters of HNNs in hyperbolic spaces \\
          % $\boldsymbol{w}$ & 3.1 &  parameters in Euclidean spaces \\
          % $\mathcal{L}$ & 3.1 & loss function of HNNs \\
           % $\mathcal{L}(\boldsymbol{w})$ & 3.1 & loss on Euclidean spaces \\
          % $\mathcal{L}(\text{expm}_{\boldsymbol{0}}^{c}(\boldsymbol{w}))$ & 3.1 & loss on hyperbolic spaces \\
          $\mathcal{L}(\boldsymbol{w}, c, \boldsymbol{y})$ & 3.1 & loss on hyperbolic neural networks \\
          % $G_{0}$ & 3.1 & Lipschitz constant of loss function \\
           % $\mathcal{C}_{8}, \mathcal{C}_{9}$ & 3.1 & constants in Lemma 1 \\
           $L_{\oplus_{c}}, L_{\oplus_{x}}, L_{\oplus_{y}}, L_{{\rm expm}_{x}}, L_{{\rm expm}_{y}}, L_{{\rm logm}_{y}}, L_{p}$ & 3.2 & Lipschitz constants of operations in hyperbolic spaces\\
           $L_{\mathcal{L}}^{'}, L_{\mathcal{L}}, L_{f}$ & 3.3 & Lipschitz constants of loss function and feed-forward function\\
           $L_{\rm tangent}$ & 3.3 & Lipschitz constants of tangent point in HNNs\\
           $\mathcal{E}_{l_{y}}$ & 3.3 & Lipschitz bound of changing tangent points in HNNs\\
           %  & 3.3 & Lipschitz constants of exponential map\\
           % $L_{{\rm logm}_{y}}$ & 3.4 & Lipschitz constants of logarithmic map\\
            \midrule
          $\mathcal{D}, \mathcal{S}$ & 4 & real distribution, training set \\
          % $\mathcal{S}$ & 3.2 & training set \\
          $\boldsymbol{\epsilon}, \rho$ & 4 & perturbation of weights, perturbation radius \\
          $\mathcal{L}_{\mathcal{S}}^{sharp}, \mathcal{E}_{l_{gen}}$ & 4 & generalization error bound of HNNs \\
          % $\rho$ & 3.2 & perturbation radius \\
          % $\mathcal{C}_{1}, \mathcal{C}_{10}$ & 3.2 & constants in generalization bound of HNNs \\
          \midrule
          $\text{SN}(\cdot)$ & 5 & scope sharpness measure \\
          $\mathcal{F}(\cdot)$ & 5 & outer-level objective of bi-level optimization \\
          $\mathcal{L}(\boldsymbol{w}, c)$ & 5 & loss on hyperbolic spaces after simplified notation \\
          $\text{sn}(\cdot)$ & 5.1 & reparametrization-invariant sharpness measure \\
          $\nabla$ & 5.1 & gradients \\
          $\hat{\text{sn}}(\cdot)$ & 5.1 & approximated reparametrization-invariant sharpness measure \\
          $\hat{\text{sn}}(\hat{\boldsymbol{w}}(c))$ & 5.1 & approximated scope sharpness measure \\
          $\boldsymbol{U}_{1}, \boldsymbol{U}_{2}$ & 5.2 & part of gradient with respect to curvatures $c$  \\
          $\boldsymbol{U}_{1}^{'}, \boldsymbol{U}_{2}^{'}$ & 5.2 & approximation of $\boldsymbol{U}_{1}, \boldsymbol{U}_{2}$ \\
          % $\boldsymbol{U}_{1}^{'}, \boldsymbol{U}_{2}^{'}$ & 4.2 & approximation of $\boldsymbol{U}_{1}, \boldsymbol{U}_{2}$ \\
          \midrule
          $L_{G}, L_{\bar{G}}, L_{H}, L_{H_{2}}, L_{G^{'}}, L_{G_{2}}, L_{G_{3}}, L_{G_{4}}, L_{H_{c}}$ & 6.1 & Lipschitz constants for theoretical analysis\\
           $\mathcal{E}_{mea}, \mathcal{E}_{cur_{1}}, \mathcal{E}_{cur_{2}}$ & 6.2 & constants in approximation error analyses\\
           $\mathcal{C}_{w_{1}}, \mathcal{C}_{w_{2}}, \mathcal{C}_{c}$ & 6.3 & constants in convergence analyses\\
           \midrule
    \end{tabular}}
\end{table*}
To classify the hyperbolic embeddings $\boldsymbol{x}$, HNNs propose the hyperbolic Multinomial Logistic Regression (MLR) as the classifier.  In HNNs, the probability that a hyperbolic embedding $\boldsymbol{x} \in \mathcal{H}^{d,c}$ classified to the $k$-th class is defined as
\begin{equation}
\label{equation:hyperbolic_mlr}
\scriptsize
    \begin{aligned}
        & p(z = k|\boldsymbol{x}) \propto \text{exp} \\
        & \left(\frac{\lambda^{c}_{\boldsymbol{b}_{k}^{'}} \Vert \boldsymbol{a}_{k}^{'}\Vert}{\sqrt{c}} \text{sinh}^{-1}  
        \left( \frac{2\sqrt{c}\langle -\boldsymbol{b}_{k}^{'} \oplus_{c}\boldsymbol{x}, \boldsymbol{a}_{k}^{'} \rangle}{(1-c\Vert-\boldsymbol{b}_{k}^{'} \oplus_{c} \boldsymbol{x}\Vert^{2})\Vert \boldsymbol{a}_{k}^{'} \Vert}\right)\right),
    \end{aligned}
\end{equation}
where $\boldsymbol{b}_{k}^{'} \in \mathcal{H}^{d,c}, \boldsymbol{a}_{k}^{'} \in T_{\boldsymbol{b}_{k}^{'}} \mathcal{H}$ are parameters of the classifier in HNNs, $\boldsymbol{a}_{k}^{'}$ is the normal vector, and $\boldsymbol{b}_{k}^{'}$ is the shift.
Denote the parameters of HNNs as $\boldsymbol{w}$ that includes 
$\boldsymbol{a}_{k}^{'}, \boldsymbol{b}_{k}^{'}$, and parameters for extracting hyperbolic features, \emph{i.e.}, $\boldsymbol{a}, \boldsymbol{b}$.
% {\color{green}
% % Due to some parameters of HNNs are located in Hyperbolic spaces, such as $\boldsymbol{b}_{k}$, 
% }
% where parameters located on either Euclidean or Hyperbolic spaces.
Note that some parameters of HNNs are located in hyperbolic spaces, such as $\boldsymbol{b}$ and $\boldsymbol{b}^{'}$.
Generally, to obtain parameters on the hyperbolic spaces,  one utilizes $\rm{expm}_{\boldsymbol{y}}^{c}(\cdot)$ to project parameters from the Euclidean spaces to hyperbolic spaces.   
In practice, $\boldsymbol{y}$ is also set to $\boldsymbol{0}$ to simplify the computations, which has been empirically demonstrated to have little impact on the obtained results~\citep{khrulkov2020hyperbolic,hu2024rethinking}.
% We denote all parameters of HNNs as $\boldsymbol{w}$ that includes $\boldsymbol{w}\triangleq \{\boldsymbol{a}, \boldsymbol{b}, \boldsymbol{a}^{'}, \boldsymbol{b}^{'}\}$, and other parameters.

The loss function is defined as $\mathcal{L}(\boldsymbol{w}, c, \boldsymbol{y})  \triangleq \frac{1}{n} \sum_{i=1}^{n} [l(\boldsymbol{w}, c, \boldsymbol{y}, \boldsymbol{x}_{i}, \iota_{i})]$.
The loss function takes as input the parameters of hyperbolic neural networks $\boldsymbol{w}$, curvature $c$, tangent point $\boldsymbol{y}$, data point $\boldsymbol{x}$, and label $\iota$, and outputs the loss value for optimizing the parameters $\boldsymbol{w}$. The process can be summarized in three stages: (1) Utilize the hyperbolic neural network to extract the representation in hyperbolic spaces via Eq.~\eqref{equation:hyperbolic_forward}; (2) Apply hyperbolic MLR in Eq.~\eqref{equation:hyperbolic_mlr} to compute the probability; (3) Compute the final loss value by comparing the predicted probability with the ground truth label (\textit{e.g.}, using the cross-entropy loss).

By setting $c=0$ and $\boldsymbol{y} = \boldsymbol{0}$, the loss function $\mathcal{L}(\boldsymbol{w},0,\boldsymbol{0})$ is degenerated to loss functions in Euclidean spaces~\citep{ganea2018hyperbolic}.

% In this way, the loss function of optimizing HNNs can be denoted as $\mathcal{L}(\boldsymbol{w}, c, \boldsymbol{y})$, which represents the embeddings and parameters are mapped to hyperbolic spaces via $\rm{expm}_{\boldsymbol{y}}^{c}(\cdot)$  and are operated on $\mathcal{H}^{d,c}$.

\subsection{Lipschitz Continuity of Hyperbolic Operations}
\label{section:lip_continuous}
% For the further analysis of generalization error bound in HNNs, 
To theoretically analyze the effect of curvatures in HNNs,
we first analyze the Lipschitz continuity of the utilized hyperbolic operations, \textit{i.e.}, Möbius addition $\boldsymbol{x} \oplus_{c} \boldsymbol{y}$, 
exponential map $\text{expm}_{\boldsymbol{y}}^{c}(\boldsymbol{x})$, and logarithmic map $\text{logm}_{\boldsymbol{y}}^{c}(\boldsymbol{x})$.
Note that for brevity and ease of reading, we present the proofs of this subsection in GitHub \footnote{\href{https://github.com/XiaomengFanmcislab/ijcv\_curvature\_learning\_for\_generalization/}{https://github.com/XiaomengFanmcislab/ijcv\_curvature\_\newline learning\_for\_generalization/}}.

\subsubsection{Lipschitz Continuity of Möbius Addition}
We analyze the Lipschitz continuity of the Möbius addition $\boldsymbol{x}\oplus_{c}\boldsymbol{y}$ with respect to curvatures $c$, right input $\boldsymbol{y}$,  and left input $\boldsymbol{x}$ in Theorems~\ref{theorem:lip_continous_oplus_c}, ~\ref{theorem:lip_continous_oplus_y}, and~\ref{theorem:lip_continous_oplus_x}, respectively.

% In theorem~\ref{theorem:lip_continous_oplus_c}, we present 
% In theorem~\ref{theorem:lip_continous_oplus_y}, we present the lipschitz property analysis of the addition operations $\oplus_{c}$ with respect to the right term $\boldsymbol{y}$.

\begin{theorem}
\label{theorem:lip_continous_oplus_c}
    Given two points on hyperbolic spaces $\boldsymbol{x}$ and $\boldsymbol{y}$,
    the Möbius addition $\boldsymbol{x} \oplus_{c} \boldsymbol{y}$ is Lipschitz continuous with respect to curvatures $c$, \textit{i.e.}, 
    \begin{equation}
    \small
        \begin{aligned}
            \Vert \boldsymbol{x} \oplus_{c_{1}} \boldsymbol{y}  -   \boldsymbol{x} \oplus_{c_{2}} \boldsymbol{y}  \Vert \le 
            L_{\oplus_{c}} |c_{1} - c_{2}|.
        \end{aligned}
    \end{equation}
    % Denote $\mathcal{N}(c, \boldsymbol{x}, \boldsymbol{y}) \triangleq |(1 + 2c \langle \boldsymbol{x}, \boldsymbol{y} \rangle + c^2 \|\boldsymbol{x}\|^2 \|\boldsymbol{y}\|^2)$,  
    $L_{\oplus_{c}}$ is computed as
    \begin{equation}
   \small
       \begin{aligned}
            &L_{\oplus_{c}} \triangleq
           \frac{
           2  \| \boldsymbol{x} \| \| \boldsymbol{y} \|  +
             3 \| \boldsymbol{x} \|^{2} \| \boldsymbol{y} \| +  
             2 \| \boldsymbol{x} \| \| \boldsymbol{y} \|^{2}  
             }{\mathcal{N}(c_{1}, \boldsymbol{x}, \boldsymbol{y}) \mathcal{N}(c_{2}, \boldsymbol{x}, \boldsymbol{y}) } \\\
             &+
              \frac{ 
              (c_{1}+c_{2})  \| \boldsymbol{x} \|^2 \| \boldsymbol{y} \|^3 +  (c_{1}+c_{2})  \| \boldsymbol{x} \|^3 \| \boldsymbol{y} \|^2
             }{\mathcal{N}(c_{1}, \boldsymbol{x}, \boldsymbol{y}) \mathcal{N}(c_{2}, \boldsymbol{x}, \boldsymbol{y}) }
             \\
             &+
             \frac{ c_1 c_2  \| \boldsymbol{x} \|^3 \| \boldsymbol{y} \|^4  + 3c_1 c_2  \| \boldsymbol{x} \|^4 \| \boldsymbol{y} \|^3}
             {\mathcal{N}(c_{1}, \boldsymbol{x}, \boldsymbol{y}) \mathcal{N}(c_{2}, \boldsymbol{x}, \boldsymbol{y})}.
       \end{aligned}
   \end{equation}
    $\mathcal{N}(c_{1}, \boldsymbol{x}, \boldsymbol{y})$ and $\mathcal{N}(c_{2}, \boldsymbol{x}, \boldsymbol{y})$ are given by
   \begin{equation}
   \small
       \begin{aligned}
           &\mathcal{N}(c_{1}) = |(1 + 2c_1 \langle \boldsymbol{x}, \boldsymbol{y} \rangle + c_1^2 \|\boldsymbol{x}\|^2 \|\boldsymbol{y}\|^2) |, \\
           & \mathcal{N}(c_{2}) = |(1 + 2c_2 \langle \boldsymbol{x}, \boldsymbol{y} \rangle + c_2^2 \|\boldsymbol{x}\|^2 \|\boldsymbol{y}\|^2)|.
       \end{aligned}
   \end{equation}
   We also derive that
    \begin{equation}
    \small
     \begin{aligned}
         \Vert  \boldsymbol{x} \oplus_{c} \boldsymbol{y} - (\boldsymbol{x} + \boldsymbol{y})  \Vert \le  L_{\oplus_{c0}},
     \end{aligned}
 \end{equation}
    where
     \begin{equation}
     \small
       \begin{aligned}
           L_{\oplus_{c0}} &\triangleq
            \frac{2 c \| \boldsymbol{x} \| \| \boldsymbol{y} \|  +
             3 c\| \boldsymbol{x} \|^{2} \| \boldsymbol{y} \| +  
             2 c\| \boldsymbol{x} \| \| \boldsymbol{y} \|^{2} 
             }{ |(1 + 2c \langle \boldsymbol{x}, \boldsymbol{y} \rangle + c^2 \|\boldsymbol{x}\|^2 \|\boldsymbol{y}\|^2) | }
             \\
             &
             +\frac{ c^{2} \| \boldsymbol{x} \|^2 \| \boldsymbol{y} \|^3
             +
             c^{2}  \| \boldsymbol{x} \|^3 \| \boldsymbol{y} \|^2}{|(1 + 2c \langle \boldsymbol{x}, \boldsymbol{y} \rangle + c^2 \|\boldsymbol{x}\|^2 \|\boldsymbol{y}\|^2) |}.
       \end{aligned}
   \end{equation}
    Moreover, $L_{\oplus_{c0}}$ satisfies that
    \begin{equation}
    \small
        \begin{aligned}
          \underset{c \rightarrow 0}{\lim}  L_{\oplus_{c0}} = 0.
        \end{aligned}
    \end{equation}

    Denote the angle between $\boldsymbol{x}$ and $\boldsymbol{y}$ as $\theta$.
    Suppose that  $\theta$  satisfy $\cos(\theta) \ge \cos{\Tilde{\theta}}$. By utilizing the hyperbolic constraint of $\boldsymbol{x}$ and $\boldsymbol{y}$,  $L_{\oplus_{x}}$ can be further modeled as
    \begin{equation}
    \small
        \begin{aligned}
            L_{\oplus_{c}} =  |c_{2} - c_{1}| 
           \frac{
            \left(
            \frac{6}{c^{3/2}} +  \frac{2}{c} +    \frac{ 4c_{1}c_{2} }{c^{7/2}} + \frac{2(c_{1} + c_{2})}{c^{5/2}}
            \right)
           }{(1- \cos(\Tilde{\theta})^{2})^{2}},
        \end{aligned}
    \end{equation}
    and  $L_{\oplus_{c0}}$ can be modeled as
    \begin{equation}
    \small
     \begin{aligned}
       L_{\oplus_{c0}} \triangleq  \frac{
           \frac{8}{c^{1/2}} + 2
           }{(1- \cos(\Tilde{\theta})^{2})^{2}}.
     \end{aligned}
 \end{equation}

\end{theorem}

\begin{theorem}
\label{theorem:lip_continous_oplus_y}
    Given points on hyperbolic spaces $\boldsymbol{x}$ and $\boldsymbol{y}$,
    the Möbius addition $\boldsymbol{x} \oplus_{c} \boldsymbol{y}$ is Lipschitz continuous with respect to the right input $\boldsymbol{y}$, \textit{i.e.}, 
    \begin{equation}
    \small
        \begin{aligned}
            \Vert \boldsymbol{x} \oplus_{c} \boldsymbol{y}_{1}  -   \boldsymbol{x} \oplus_{c} \boldsymbol{y}_{2}  \Vert \le 
            L_{\oplus_{y}} \Vert \boldsymbol{y}_{1} - \boldsymbol{y}_{2} \Vert,
        \end{aligned}
    \end{equation}
      $L_{\oplus_{y}}$ is computed as
      \begin{equation}
      \footnotesize
           \begin{aligned}
                &L_{\oplus_{y}}\triangleq \frac{
                1 + c\left( 5 \Vert \boldsymbol{x} \Vert^{2} 
               +5 \Vert \boldsymbol{x} \Vert\| \boldsymbol{y}_1 \|
               +   \Vert \boldsymbol{x} \Vert\| \boldsymbol{y}_2 \|
               \right)
               }{\mathcal{N}(c, \boldsymbol{x}, \boldsymbol{y}_{1}) \mathcal{N}(c, \boldsymbol{x}, \boldsymbol{y}_{2})} +\\
               &
                \frac{
               c^{2}\Vert \boldsymbol{x} \Vert^{2} \left(13  \Vert \boldsymbol{x} \Vert  \| \boldsymbol{y}_1 \|
               +   \Vert \boldsymbol{x} \Vert  \| \boldsymbol{y}_2 \| 
               +  6   \| \boldsymbol{y}_1 \|^{2} 
               +  3  \| \boldsymbol{y}_1 \| \| \boldsymbol{y}_2 \|
               \right)
               }{\mathcal{N}(c, \boldsymbol{x}, \boldsymbol{y}_{1}) \mathcal{N}(c, \boldsymbol{x}, \boldsymbol{y}_{2})} + 
               \\
               &\frac{ c^{3} \Vert \boldsymbol{x} \Vert^{3} \left( 6 \Vert \boldsymbol{x} \Vert  \| \boldsymbol{y}_1 \|^{2} 
             +  3 \Vert \boldsymbol{x} \Vert \| \boldsymbol{y}_1 \| \| \boldsymbol{y}_2 \| + 2  \| \boldsymbol{y}_1 \|^{3} + 2  \| \boldsymbol{y}_1 \|^{2}  \| \boldsymbol{y}_2 \| \right) }{\mathcal{N}(c, \boldsymbol{x}, \boldsymbol{y}_{1}) \mathcal{N}(c, \boldsymbol{x}, \boldsymbol{y}_{2})}.
           \end{aligned}
       \end{equation}
        $\mathcal{N}(c, \boldsymbol{x}, \boldsymbol{y}_{1})$ and $\mathcal{N}(c, \boldsymbol{x}, \boldsymbol{y}_{2})$ are given by
   \begin{equation}
   \small
       \begin{aligned}
           &\mathcal{N}(c, \boldsymbol{x}, \boldsymbol{y}_{1}) = |(1 + 2c \langle \boldsymbol{x}, \boldsymbol{y}_{1} \rangle + c^2 \|\boldsymbol{x}\|^2 \|\boldsymbol{y}_{1}\|^2) | \\
           & \mathcal{N}(c, \boldsymbol{x}, \boldsymbol{y}_{2}) = |(1 + 2c \langle \boldsymbol{x}, \boldsymbol{y}_{2} \rangle + c^2 \|\boldsymbol{x}\|^2 \|\boldsymbol{y}_{2}\|^2)|.
       \end{aligned}
   \end{equation}
       Moreover, $L_{\oplus_{y}}$ satisfies that
       \begin{equation}
       \small
        \begin{aligned}
            \underset{c \rightarrow 0}{\lim}  L_{\oplus_{y}}  = 1.
        \end{aligned}
    \end{equation}

        Denote the angle between $\boldsymbol{x}$ and $\boldsymbol{y}_{1}$, and $\boldsymbol{x}$ and $\boldsymbol{y}_{2}$ as $\theta_{1}$ and $\theta_{2}$, respectively. Suppose that  $\theta_{1}$ and $\theta_{2}$  satisfy $\cos(\theta_{1}), \cos(\theta_{2}) \ge \cos{\Tilde{\theta}}$.
    By utilizing the hyperbolic constraint of $\boldsymbol{x}$ and $\boldsymbol{y}$,  $L_{\oplus_{y}}$ can be further modeled as
    \begin{equation}
    \small
        \begin{aligned}
           L_{\oplus_{y}} =   \frac{
         48
          }{
          (1- \cos(\Tilde{\theta})^{2})^{2}
          }.
        \end{aligned}
    \end{equation}
\end{theorem}

% For the further analysis, we present a  corollary  from Theorem~\ref{theorem:lip_continous_oplus_y}.
% \begin{corollary}
% \label{corollary:addition_lip_y}
%   Supposing that $\Vert \boldsymbol{x} \Vert \le \frac{1}{c}, \Vert \boldsymbol{y} \Vert \le \frac{1}{\sqrt{c}}$, 
%     the Möbius addition $\boldsymbol{x} \oplus_{c} \boldsymbol{y}$ is $L_{\oplus_{y}}-$Lipschitz continuous with respect to $\boldsymbol{y}$, \textit{i.e.},
%      \begin{equation}
%      \small
%         \begin{aligned}
%             \Vert \boldsymbol{x} \oplus_{c} \boldsymbol{y}_{1}  -   \boldsymbol{x} \oplus_{c} \boldsymbol{y}_{2}  \Vert \le 
%             L_{\oplus_{y}} \Vert \boldsymbol{y}_{1} - \boldsymbol{y}_{2} \Vert,
%         \end{aligned}
%     \end{equation}
%     and  $L_{\oplus_{y}}$ can be re-formulated as 
%     \begin{equation}
%     \small
%         \begin{aligned}
%            L_{\oplus_{y}} \triangleq \frac{
%         \left(
%            \frac{9}{c^{2}} + \frac{18}{c^{\frac{3}{2}}}+ \frac{14}{c}+ \frac{6}{\sqrt{c}} + 1
%             \right)
%           }{
%          (1- \cos(\Tilde{\theta})^{2})^{2}
%           }.
%         \end{aligned}
%     \end{equation}
    
% \end{corollary}

\begin{theorem}
\label{theorem:lip_continous_oplus_x}
    Given points on hyperbolic spaces $\boldsymbol{x}$ and $\boldsymbol{y}$,
    the Möbius addition $\boldsymbol{x} \oplus_{c} \boldsymbol{y}$ is Lipschitz continuous with respect to $\boldsymbol{x}$, \textit{i.e.}, 
    \begin{equation}
    \small
        \begin{aligned}
            \Vert \boldsymbol{x}_{1} \oplus_{c} \boldsymbol{y}  -   \boldsymbol{x}_{2} \oplus_{c} \boldsymbol{y}  \Vert \le 
            L_{\oplus_{x}} \Vert \boldsymbol{x}_{1} - \boldsymbol{x}_{2} \Vert,
        \end{aligned}
    \end{equation}
        where  $L_{\oplus_{x}}$ is computed as
     \begin{equation}
        \footnotesize
        \begin{aligned}
           & L_{\oplus_{x}} \triangleq 
             \frac{1+ 3c\Vert  \boldsymbol{y} \Vert^{2} +  3c \| \boldsymbol{x}_{1} \| \Vert  \boldsymbol{y}  \Vert   + 7c  \| \boldsymbol{x}_{2} \| \Vert  \boldsymbol{y}  \Vert }{\mathcal{N}(c, \boldsymbol{x}_{1}, \boldsymbol{y})\mathcal{N}(c, \boldsymbol{x}_{2}, \boldsymbol{y})} + \\
        &   \frac{ c^{2}\Vert \boldsymbol{y} \Vert^{2} 
        \left(7 \Vert \boldsymbol{x}_{1} \Vert  \Vert \boldsymbol{x}_{2} \Vert  
        + \| \boldsymbol{x}_{1} \|\Vert \boldsymbol{y} \Vert
        +5 \|\boldsymbol{x}_{2} \|\Vert \boldsymbol{y}\Vert
        +6\|\boldsymbol{x}_{2}\|^{2}
        \right)}{\mathcal{N}(c, \boldsymbol{x}_{1}, \boldsymbol{y})\mathcal{N}(c, \boldsymbol{x}_{2}, \boldsymbol{y})} + \\
       &   \frac{ c^{3} \| \boldsymbol{x}_{2} \| \| \boldsymbol{y} \|^3  \left(
        4\| \boldsymbol{x}_{1} \|^2  
         +4\| \boldsymbol{x}_{1} \|\| \boldsymbol{x}_{2} \|
         +2\| \boldsymbol{x}_{2} \| \| \boldsymbol{y} \|
         +\| \boldsymbol{x}_{1} \|  \| \boldsymbol{y} \|
       \right) }{\mathcal{N}(c, \boldsymbol{x}_{1}, \boldsymbol{y})\mathcal{N}(c, \boldsymbol{x}_{2}, \boldsymbol{y})}.
        \end{aligned}
    \end{equation}
     $\mathcal{N}(c, \boldsymbol{x}_{1}, \boldsymbol{y})$ and $\mathcal{N}(c, \boldsymbol{x}_{2}, \boldsymbol{y})$ are given by
   \begin{equation}
   \small
       \begin{aligned}
           &\mathcal{N}(c, \boldsymbol{x}_{1}, \boldsymbol{y}) = |(1 + 2c \langle \boldsymbol{x}_{1}, \boldsymbol{y} \rangle + c^2 \|\boldsymbol{x}_{1}\|^2 \|\boldsymbol{y}\|^2) |, \\
           & \mathcal{N}(c, \boldsymbol{x}_{2}, \boldsymbol{y}) = |(1 + 2c \langle \boldsymbol{x}_{2}, \boldsymbol{y} \rangle + c^2 \|\boldsymbol{x}_{2}\|^2 \|\boldsymbol{y}\|^2)|.
       \end{aligned}
   \end{equation}
       Moreover, $L_{\oplus_{x}}$ satisfies that
       \begin{equation}
       \small
        \begin{aligned}
            \underset{c \rightarrow 0}{\lim}  L_{\oplus_{x}}  = 1.
        \end{aligned}
    \end{equation}

    Denote the angle between $\boldsymbol{x}_{1}$ and $\boldsymbol{y}$, and $\boldsymbol{x}_{2}$ and $\boldsymbol{y}$ as $\theta_{1}$ and $\theta_{2}$, respectively.
    Suppose that  $\theta_{1}$ and $\theta_{2}$ satisfy $\cos(\theta_{1}), \cos(\theta_{2}) \ge \cos{\Tilde{\theta}}$. By utilizing the hyperbolic constraint of $\boldsymbol{x}$ and $\boldsymbol{y}$,  $L_{\oplus_{x}}$ can be further modeled as
    \begin{equation}
    \small
        \begin{aligned}
           L_{\oplus_{x}} \triangleq  \frac{44
          }{
          (1- \cos(\Tilde{\theta})^{2})^{2}
          }.
        \end{aligned}
    \end{equation}

\end{theorem}

% For further analysis, we present a  corollary  from Theorem~\ref{theorem:lip_continous_oplus_x}.
% \begin{corollary}
% \label{corollary:addition_lip_x}
%     Supposing that $\Vert \boldsymbol{x} \Vert  \le \frac{1}{c}, \Vert \boldsymbol{y} \Vert \le \frac{1}{\sqrt{c}}$, 
%     the Möbius addition $\boldsymbol{x} \oplus_{c} \boldsymbol{y}$ is $L_{\oplus_{x}}-$Lipschitz continuous with respect to the left input $\boldsymbol{x}$, \textit{i.e.}, 
%     \begin{equation}
%     \small
%         \begin{aligned}
%             \Vert \boldsymbol{x}_{1} \oplus_{c} \boldsymbol{y}  -   \boldsymbol{x}_{2} \oplus_{c} \boldsymbol{y}  \Vert \le 
%             L_{\oplus_{x}} \Vert \boldsymbol{x}_{1} - \boldsymbol{x}_{2} \Vert,
%         \end{aligned}
%     \end{equation}
%     and  $L_{\oplus_{x}}$ is re-formulated as
%     \begin{equation}
%     \small
%     \begin{aligned}
%          L_{\oplus_{x}} = \frac{(4 + \frac{16}{c} + \frac{16}{\sqrt{c}}  +   \frac{8}{c^{\frac{3}{2}}})}{(1- \cos(\Tilde{\theta})^{2})^{2}},
%     \end{aligned}
% \end{equation}
% \end{corollary}

\subsubsection{Lipschitz Continuity of Exponential Map}

In this section, we analyze the Lipschitz continuity of exponential map $\text{expm}_{\boldsymbol{y}}^{c}(\boldsymbol{x}) $ with respect to $\boldsymbol{x}$ and $\boldsymbol{y}$ in Theorems~\ref{theorem:lip_continous_expm_y} and~\ref{theorem:lip_continous_expm_x}.
\begin{theorem}
\label{theorem:lip_continous_expm_y}
    The exponential map $ \text{expm}_{\boldsymbol{y}}^{c}(\boldsymbol{x}) $ is $L_{{\rm expm}_{y}}-$Lipschitz continuous  with respect to $\boldsymbol{y}$.
    Mathematically,
    for any $\boldsymbol{y}_{1}$ and $\boldsymbol{y}_{2}$, we have that
      \begin{equation}
      \small
      \label{equation:expm_lip_0}
        \begin{aligned}
            \Vert {\rm expm}_{\boldsymbol{y}_{1}}^{c}(\boldsymbol{x}) - {\rm expm}_{\boldsymbol{y}_{2}}^{c}(\boldsymbol{x})  \Vert \le L_{{\rm expm}_{y}} \Vert \boldsymbol{y}_{1} - \boldsymbol{y}_{2} \Vert,
        \end{aligned}
    \end{equation}
        where $L_{{\rm expm}_{y}}$ is computed as 
        \begin{equation}
        \small
            \begin{aligned}
               L_{{\rm expm}_{y}} \triangleq  \frac{L_{\oplus_{y}} c\Vert \boldsymbol{x}  \Vert (\Vert \boldsymbol{y}_{2} \Vert + \Vert \boldsymbol{y}_{1} \Vert)}{(1 - \sqrt{c})^{2}} + L_{\oplus_{x}}.
            \end{aligned}
        \end{equation}
        Moreover, $ L_{{\rm expm}_{y}}$ satisfies that
        \begin{equation}
        \small
            \begin{aligned}
                \underset{c \rightarrow 0}{\lim}  L_{{\rm expm}_{y}} = 1.
            \end{aligned}
        \end{equation}
        By utilizing the hyperbolic constraint of $\boldsymbol{y}$, $ L_{{\rm expm}_{y}}$ can be further modeled as
        \begin{equation}
        \small
            \begin{aligned}
                L_{{\rm expm}_{y}} \triangleq \frac{2 \sqrt{c}L_{\oplus_{y}} \Vert \boldsymbol{x}  \Vert}{(1 - \sqrt{c})^{2}} + L_{\oplus_{x}}.
            \end{aligned}
        \end{equation}
\end{theorem}

\begin{theorem}
\label{theorem:lip_continous_expm_x}
    The exponential map $ {\rm expm}_{\boldsymbol{y}}^{c}(\boldsymbol{x}) $ is $L_{{\rm expm}_{x}}-$Lipschitz continuous with respect to $\boldsymbol{x}$.
    Mathematically,
    for any $\boldsymbol{x}_{1}$ and $\boldsymbol{x}_{2}$, we have that
      \begin{equation}
      \small
      \label{equation:expm_lip_0}
        \begin{aligned}
            \Vert {\rm expm}_{\boldsymbol{y}}^{c}(\boldsymbol{x}_{1}) - {\rm expm}_{\boldsymbol{y}}^{c}(\boldsymbol{x}_{2})  \Vert \le L_{{\rm expm}_{x}} \Vert \boldsymbol{x}_{1} - \boldsymbol{x}_{2} \Vert,
        \end{aligned}
    \end{equation}
        where $L_{{\rm expm}_{x}}$ is computed as 
        \begin{equation}
        \small
        \begin{aligned}
            L_{{\rm expm}_{x}} \triangleq L_{\oplus_{y}} \left( 
       \left|  \frac{1}{1-c\Vert \boldsymbol{y} \Vert^{2}} \right|
        +
        \frac{2\text{tanh}\left( \frac{\sqrt{c} \Vert \boldsymbol{x}_{2} \Vert}{1-c\Vert \boldsymbol{y} \Vert^{2}} \right)}{\sqrt{c}\Vert \boldsymbol{x}_{2} \Vert} 
        \right),
        \end{aligned}
        \end{equation}
    and $ L_{{\rm expm}_{x}}$ can satisfy that
    \begin{equation}
    \small
    \begin{aligned}
       \underset{ c \rightarrow 0 }{ \lim } L_{{\rm expm}_{x}} = 1.
    \end{aligned}
\end{equation}
        
        From the hyperbolic constraint of $\boldsymbol{y}$, 
we can further model $L_{{\rm expm}_{x}}$ as
\begin{equation}
\small
    \begin{aligned}
        L_{{\rm expm}_{x}} = 
     L_{\oplus_{y}} \left|  \frac{1}{1-\sqrt{c}} \right| + 
        \frac{2L_{\oplus_{y}}}{\sqrt{c} \Vert \boldsymbol{x}_{2} \Vert }.
    \end{aligned}
    \end{equation}
\end{theorem}

\subsubsection{Lipschitz Continuity of Logarithmic Map}
In Theorem~\ref{theorem:lip_continous_logm_y}, we analyze the Lipschitz continuity of logarithmic map $ {\rm logm}_{\boldsymbol{y}}^{c}(\boldsymbol{x}) $ with respect to $\boldsymbol{y}$.
\begin{theorem}
\label{theorem:lip_continous_logm_y}
      The logarithmic map $ {\rm logm}_{\boldsymbol{y}}^{c}(\boldsymbol{x}) $ is Lipschitz continuous with
    the constant $L_{{\rm logm}_{y}}$ with respect to $\boldsymbol{y}$.
    Mathematically,
    for any $\boldsymbol{y}_{1}$ and $\boldsymbol{y}_{2}$, we have that
      \begin{equation}
      \small
    \label{equation:expm_lip_0}
        \begin{aligned}
            \Vert {\rm logm}_{\boldsymbol{y}_{1}}^{c}(\boldsymbol{x}) - {\rm logm}_{\boldsymbol{y}_{2}}^{c}(\boldsymbol{x})  \Vert \le L_{{\rm logm}_{y}} \Vert \boldsymbol{y}_{1} - \boldsymbol{y}_{2} \Vert,
        \end{aligned}
    \end{equation}
        where $L_{{\rm logm}_{y}}$ is computed as 
        \begin{equation}
        \small
            \begin{aligned}
                L_{{\rm logm}_{y}} & \triangleq  L_{\oplus_{x}} (1 + c\Vert\boldsymbol{x} \Vert \Vert\boldsymbol{y}_{1} \Vert)^{2}  \\&+ \frac{\sqrt{c}}{4} \Vert \boldsymbol{y}_{1} + \boldsymbol{y}_{2}\Vert +  \sqrt{c} L_{\oplus_{x}},
            \end{aligned}
        \end{equation}
        and $L_{{\rm logm}_{y}}$ satisfies that
    \begin{equation}
    \small
    \begin{aligned}
        \underset{c \rightarrow 0}{\lim} L_{{\rm logm}_{y}}  = 1.
    \end{aligned}
\end{equation}
From the hyperbolic constraint of $\boldsymbol{x}$ and $ \boldsymbol{y}$,
$ L_{{\rm logm}_{y}}$ can be modeled as
\begin{equation}
\small
    \begin{aligned}
         L_{{\rm logm}_{y}} =  4
         L_{\oplus_{x}}
         +
        \frac{1}{2}
        +
        \sqrt{c} L_{\oplus_{x}}.
    \end{aligned}
\end{equation}
\end{theorem}

\subsection{Lipschitz Bound for Changes in Tangent Points}
\label{section:approximation_error_bound}
In practice, one sets the tangent point $\boldsymbol{y}$ as $\boldsymbol{0}$ for HNNs for the scale of simplicity. 
In this section, we analyze the error bound between $\mathcal{L}(\boldsymbol{w}, c, \boldsymbol{y})$ and $\mathcal{L}(\boldsymbol{w}, c, \boldsymbol{0})$.
To this end, we first analyze the Lipschitz continuity of loss function $\mathcal{L}(\boldsymbol{w}, c, \boldsymbol{y})$ with respect to $\boldsymbol{y}$, shown in Theorem~\ref{theorem:error_bound_geometric}.
By utilizing the developed theorem, we derive the corollary to present the Lipschitz bound for changing tangent point into $\boldsymbol{0}$.

% In this section, we analyze the approximation error bound of loss functions in HNNs.
% Specifically, we first present two assistant lemmas.
% Then,
% we present the approximation error bound between $\mathcal{L}(\boldsymbol{w}, c, \boldsymbol{y})$ and $\mathcal{L}(\boldsymbol{w}, c, \boldsymbol{0})$ in Theorem~\ref{theorem:error_bound_geometric}, which maps parameters and embeddings to hyperbolic spaces at $\boldsymbol{y}$ and the origin $\boldsymbol{0}$ via $ \text{expm}_{\boldsymbol{y}}^{c}(\boldsymbol{w})$ and $ \text{expm}_{\boldsymbol{0}}^{c}(\boldsymbol{w})$, respectively.
% We also present the approximation error bound between $\mathcal{L}(\boldsymbol{w}, c, \boldsymbol{0})$ and $\mathcal{L}(\boldsymbol{w}, 0, \boldsymbol{0})$ in Theorem~\ref{theorem:error_bound_curvature}, which 
% executes operations on hyperbolic spaces $\mathcal{H}^{d,c}$ and Euclidean spaces, respectively.

\begin{theorem}
\label{theorem:error_bound_geometric} 
 Suppose that the loss function $\mathcal{L}(\cdot)$ is $L_{\mathcal{L}}^{'}-$Lipschitz continuous with respect
to the predicted probability, the hyperbolic MLR in Eq.~\eqref{equation:hyperbolic_mlr} is $L_{p}-$Lipschitz continuous with respect to its input $\boldsymbol{x}$, and $f_{\boldsymbol{a}}(\cdot)$ is  $L_{f}-$Lipschitz continuous.
    % Suppose that $\mathcal{L}(\cdot)$ is Lipschitz continuous with constant $L_{\mathcal{L}}$, and $f_{\boldsymbol{a}}(\cdot)$ is  Lipschitz continuous with constant $L_{f}$.
	For any tangent point $\boldsymbol{y}_{1}, \boldsymbol{y}_{2} \in \mathcal{H}^{d,c}$, it holds that
    \begin{equation}
    \small
        \begin{aligned}
            |\mathcal{L}(\boldsymbol{w}, c, \boldsymbol{y}_{1}) - \mathcal{L}(\boldsymbol{w}, c, \boldsymbol{y}_{2}) | \le L_{\rm tangent} \Vert \boldsymbol{y}_{1} - \boldsymbol{y}_{2} \Vert,
        \end{aligned}
    \end{equation}
    where $L_{\rm tangent}$ is a constant and is computed as
    \begin{equation}
   \small
       \begin{aligned}
           L_{\rm tangent}  \triangleq  L_{\mathcal{L}} L_{\oplus_{x}}
           (L_{{\rm expm}_{y}}  +  L_{{\rm expm}_{x}} L_f L_{{\rm logm}_{y}} \Vert \boldsymbol{a} \Vert ).
       \end{aligned}
   \end{equation}
   $\boldsymbol{a}$,  a subset of $\boldsymbol{w}$, is the parameters of the function $f_{\boldsymbol{a}}(\cdot)$.
    When $c \rightarrow 0$, $L_{\rm tangent}$ can be further modeled as 
    \begin{equation}
    \small
        \begin{aligned}
        \underset{c \rightarrow 0}{\lim}    L_{\rm tangent} =   L_{\mathcal{L}}+ 
            L_{\mathcal{L}}  L_f  \Vert \boldsymbol{a} \Vert,
        \end{aligned}
    \end{equation} 
    which is consistent with the property in Euclidean spaces.

   % Moreover, it holds that
   %   \begin{equation}
   %   \small
   %      \begin{aligned}
   %       |\mathcal{L}(\boldsymbol{w}, c, \boldsymbol{y}) - \mathcal{L}(\boldsymbol{w}, c, \boldsymbol{0}) | \le \mathcal{E}_{l_{y}},
   %      \end{aligned}
   %  \end{equation}
   %  where $\mathcal{E}_{l_{y}}$ is given by
   %  \begin{equation}
   %  \small
   %      \begin{aligned}
   %       \mathcal{E}_{l_{y}} \triangleq   L_{\mathcal{L}} L_{\oplus_{x}} (L_{{\rm expm}_{y}}+ L_{{\rm expm}_{x}} L_f L_{{\rm logm}_{y}}  \Vert \boldsymbol{a} \Vert   ) \Vert \boldsymbol{y} \Vert.
   %      \end{aligned}
   %  \end{equation}
   %  When $\boldsymbol{y} \rightarrow \boldsymbol{0}$, we obtain that
   %  \begin{equation}
   %  \small
   %      \begin{aligned}
   %          \underset{\boldsymbol{y} \rightarrow \boldsymbol{0}}{\lim}  \mathcal{E}_{l_{y}}  
   %          = 0,
   %      \end{aligned}
   %  \end{equation} 
   %  which also conforms to common sense. 
   % By utilizing the hyperbolic property of $\boldsymbol{y}$, $\mathcal{E}_{l_{y}}$ can be further modeled as 
   %  \begin{equation}
   %  \small
   %      \begin{aligned}
   %          \mathcal{E}_{l_{y}} \triangleq L_{\mathcal{L}} L_{\oplus_{x}} \frac{1}{c}(L_{{\rm expm}_{y}} + L_{{\rm expm}_{x}} L_f L_{{\rm logm}_{y}}\Vert \boldsymbol{a} \Vert).
   %      \end{aligned}
   %  \end{equation}
\end{theorem}

\begin{proof}
We define $\boldsymbol{x}_{1}$ and $\boldsymbol{x}_{2}$ as the hyperbolic representations computed by Eq.~\eqref{equation:hyperbolic_forward}, using the tangent points $\boldsymbol{y}_{1}$ and $\boldsymbol{y}_{2}$, respectively.
Mathematically, $\boldsymbol{x}_{1}$ and $\boldsymbol{x}_{2}$ are computed as
\begin{equation}
\label{equation:x1_x2_tangent}
    \begin{aligned}
         &\boldsymbol{x}_{1} = {\rm expm}_{\boldsymbol{y}_{1}}^{c}(f_{\boldsymbol{a}}({\rm logm}^{c}_{\boldsymbol{y}_{1}}(\boldsymbol{x}))) \oplus_{c} \boldsymbol{b}, \\
         &\boldsymbol{x}_{2} = {\rm expm}_{\boldsymbol{y}_{2}}^{c}(f_{\boldsymbol{a}}({\rm logm}^{c}_{\boldsymbol{y}_{2}}(\boldsymbol{x}))) \oplus_{c} \boldsymbol{b}.
    \end{aligned}
\end{equation}
Considering that the loss function is $L_{\mathcal{L}}^{'}-$Lipschitz continuous with respect to the predicted probability, and the hyperbolic MLR in Eq.~\eqref{equation:hyperbolic_mlr} is $L_{p}-$Lipschitz continuous with respect to its input $\boldsymbol{x}$,
we can obtain that
\begin{equation}
\small
\label{equation:Lip_lossfunction_specific}
       \begin{aligned}
            &| \mathcal{L}(\boldsymbol{w}, c, \boldsymbol{y}_{1}) - \mathcal{L}(\boldsymbol{w}, c, \boldsymbol{y}_{2}) | \\
            &\le  L_{\mathcal{L}}^{'}  | p(z = k|\boldsymbol{x}_{1}) - p(z = k|\boldsymbol{x}_{2})| \\
            &\le  L_{\mathcal{L}}^{'} L_{p} \Vert \boldsymbol{x}_{1} - \boldsymbol{x}_{2} \Vert.
       \end{aligned}
\end{equation}
By denoting $L_{\mathcal{L}} \triangleq L_{\mathcal{L}}^{'} L_{p}$ and substituting Eq.~\eqref{equation:x1_x2_tangent} into  Eq.~\eqref{equation:Lip_lossfunction_specific}, we can obtain that
   \begin{equation}
   \small
       \begin{aligned}
            &| \mathcal{L}(\boldsymbol{w}, c, \boldsymbol{y}_{1}) - \mathcal{L}(\boldsymbol{w}, c, \boldsymbol{y}_{2}) |  \le L_{\mathcal{L}} \Vert \boldsymbol{x}_{1} - \boldsymbol{x}_{2} \Vert
            \\
            & \le
            L_{\mathcal{L}} \Vert
            {\rm expm}_{\boldsymbol{y}_{1}}^{c} \left(
            f_{\boldsymbol{a}}({\rm logm}_{\boldsymbol{y}_{1}}^{c}(\boldsymbol{x}))  
            \right) \oplus_{c} \boldsymbol{b} \\
            &-
            {\rm expm}_{\boldsymbol{y}_{2}}^{c} \left(
            f_{\boldsymbol{a}}({\rm logm}_{\boldsymbol{y}_{2}}^{c}(\boldsymbol{x})) 
            \right)\oplus_{c} \boldsymbol{b} 
            \Vert.
       \end{aligned}
   \end{equation}

From the  Lipschitz continuity in Theorem~\ref{theorem:lip_continous_oplus_x}, it holds that
   \begin{equation}
   % \footnotesize
   \small
       \begin{aligned}
            &| \mathcal{L}(\boldsymbol{w}, c, \boldsymbol{y}_{1}) - \mathcal{L}(\boldsymbol{w}, c, \boldsymbol{y}_{2}) |
            \\
            & \le
            L_{\mathcal{L}} \Vert
            {\rm expm}_{\boldsymbol{y}_{1}}^{c} \left(
            f_{\boldsymbol{a}}({\rm logm}_{\boldsymbol{y}_{1}}^{c}(\boldsymbol{x}))  
            \right) \oplus_{c} \boldsymbol{b} \\
            &-
            {\rm expm}_{\boldsymbol{y}_{2}}^{c} \left(
            f_{\boldsymbol{a}}({\rm logm}_{\boldsymbol{y}_{2}}^{c}(\boldsymbol{x})) 
            \right)\oplus_{c} \boldsymbol{b} 
            \Vert\\
            & \le
             L_{\mathcal{L}} L_{\oplus_{x}} \Vert
            {\rm expm}_{\boldsymbol{y}_{1}}^{c} \left(
            f_{\boldsymbol{a}}({\rm logm}_{\boldsymbol{y}_{1}}^{c}(\boldsymbol{x}))  
            \right) \\
            &-
            {\rm expm}_{\boldsymbol{y}_{2}}^{c} \left(
            f_{\boldsymbol{a}}({\rm logm}_{\boldsymbol{y}_{2}}^{c}(\boldsymbol{x})) 
            \right)
             \Vert
            \\
            & \le
            L_{\mathcal{L}} L_{\oplus_{x}}
            \Vert
            {\rm expm}_{\boldsymbol{y}_{1}}^{c} \left(
            f_{\boldsymbol{a}}({\rm logm}_{\boldsymbol{y}_{1}}^{c}(\boldsymbol{x}))
            \right) \\
            &-
            {\rm expm}_{\boldsymbol{y}_{2}}^{c} \left(
            f_{\boldsymbol{a}}({\rm logm}_{\boldsymbol{y}_{1}}^{c}(\boldsymbol{x}))
            \right)
            \Vert
            \\
            &+ L_{\mathcal{L}} L_{\oplus_{x}}
             \Vert
            {\rm expm}_{\boldsymbol{y}_{2}}^{c} \left(
            f_{\boldsymbol{a}}({\rm logm}_{\boldsymbol{y}_{1}}^{c}(\boldsymbol{x})) 
            \right) \\
            &-
            {\rm expm}_{\boldsymbol{y}_{2}}^{c} \left(
            f_{\boldsymbol{a}}({\rm logm}_{\boldsymbol{y}_{2}}^{c}(\boldsymbol{x})) 
            \right)
             \Vert.
       \end{aligned}
   \end{equation}
   From the  Lipschitz continuity in Theorems~\ref{theorem:lip_continous_expm_y} and ~\ref{theorem:lip_continous_expm_x}, we can obtain that
   \begin{equation}
   \small
       \begin{aligned}
           &\Vert
            {\rm expm}_{\boldsymbol{y}_{1}}^{c} \left(
            f_{\boldsymbol{a}}({\rm logm}_{\boldsymbol{y}_{1}}^{c}(\boldsymbol{x})) 
            \right)
            -
            {\rm expm}_{\boldsymbol{y}_{2}}^{c} \left(
            f_{\boldsymbol{a}}({\rm logm}_{\boldsymbol{y}_{1}}^{c}(\boldsymbol{x})) 
            \right)
            \Vert\\
            &\le
            L_{{\rm expm}_{y}} \Vert \boldsymbol{y}_{1} - \boldsymbol{y}_{2} \Vert,
       \end{aligned}
   \end{equation}
   and
   \begin{equation}
   \small
       \begin{aligned}
           &\Vert
            {\rm expm}_{\boldsymbol{y}_{2}}^{c} \left(
            f_{\boldsymbol{a}}({\rm logm}_{\boldsymbol{y}_{1}}^{c}(\boldsymbol{x})) 
            \right)
            -
            {\rm expm}_{\boldsymbol{y}_{2}}^{c} \left(
            f_{\boldsymbol{a}}({\rm logm}_{\boldsymbol{y}_{2}}^{c}(\boldsymbol{x})) 
            \right)
            \Vert \\
            &\le L_{{\rm expm}_{x}} \Vert  f_{\boldsymbol{a}}({\rm logm}_{\boldsymbol{y}_{1}}^{c}(\boldsymbol{x})) -  f_{\boldsymbol{a}}({\rm logm}_{\boldsymbol{y}_{2}}^{c}(\boldsymbol{x}))  \Vert.
       \end{aligned}
   \end{equation}
   From the Lipschitz continuity assumption of $f_{\boldsymbol{a}}(\cdot)$ and Lipschitz continuity in Theorem~\ref{theorem:lip_continous_logm_y}, it holds that
   \begin{equation}
   \small
       \begin{aligned}
           &\Vert  f_{\boldsymbol{a}}({\rm logm}_{\boldsymbol{y}_{1}}^{c}(\boldsymbol{x})) -  f_{\boldsymbol{a}}({\rm logm}_{\boldsymbol{y}}^{c}(\boldsymbol{x}))  \Vert \\
           &\le 
           L_{f} \Vert \boldsymbol{a} \Vert \Vert {\rm logm}_{\boldsymbol{y}_{1}}^{c}(\boldsymbol{x}) -  {\rm logm}_{\boldsymbol{y}_{2}}^{c}(\boldsymbol{x})  \Vert \\
           &\le L_{f} L_{{\rm logm}_{y}} \Vert \boldsymbol{a} \Vert \Vert \boldsymbol{y}_{1} - \boldsymbol{y}_{2} \Vert.
       \end{aligned}
   \end{equation}
   Overall, we can derive that
   \begin{equation}
   \label{equation:l_w_c_0_2}
   \small
       \begin{aligned}
            &| \mathcal{L}(\boldsymbol{w}, c, \boldsymbol{y}_{1}) - \mathcal{L}(\boldsymbol{w}, c, \boldsymbol{y}_{2}) | 
            \le
            L_{\mathcal{L}} L_{\oplus_{x}} L_{{\rm expm}_{y}} \Vert \boldsymbol{y}_{1} - \boldsymbol{y}_{2} \Vert 
            \\
            &+ 
            L_{\mathcal{L}}   L_{\oplus_{x}} L_{{\rm expm}_{x}} L_f L_{{\rm logm}_{y}} \Vert \boldsymbol{a} \Vert \Vert \boldsymbol{y}_{1} - \boldsymbol{y}_{2}  \Vert.
       \end{aligned}
   \end{equation}
   By denoting 
   \begin{equation}
   \small
       \begin{aligned}
           L_{\rm tangent} \triangleq  L_{\mathcal{L}} L_{\oplus_{x}}
           (L_{{\rm expm}_{y}}   +  L_{{\rm expm}_{x}} L_f L_{{\rm logm}_{y}} \Vert \boldsymbol{a} \Vert ),
       \end{aligned}
   \end{equation}
   we have proved that
   \begin{equation}
   \small
        \begin{aligned}
            |\mathcal{L}(\boldsymbol{w}, c, \boldsymbol{y}) - \mathcal{L}(\boldsymbol{w}, c, \boldsymbol{0}) | \le L_{\rm tangent}\Vert \boldsymbol{y}_{1} - \boldsymbol{y}_{2}  \Vert.
        \end{aligned}
    \end{equation}

       Because $\underset{c \rightarrow 0}{\lim}L_{\oplus_{x}} = 1$, $\underset{c \rightarrow 0}{\lim} L_{{\rm expm}_{x}} = 1$,  $\underset{c \rightarrow 0}{\lim} L_{{\rm expm}_{y}} = 1$, and $\underset{c \rightarrow 0}{\lim} L_{{\rm logm}_{y}} = 1$,
    when $c \rightarrow 0$, Eq.~\eqref{equation:l_w_c_0_2} can be modeled as 
    \begin{equation}
    \small
    % \label{equation:l_w_c_0_2}
        \begin{aligned}
           &\underset{c \rightarrow 0}{\lim} | \mathcal{L}(\boldsymbol{w}, c, \boldsymbol{y}_{1}) - \mathcal{L}(\boldsymbol{w}, c, \boldsymbol{y}_{2}) | \\
           & \le
            ( L_{\mathcal{L}}+ 
            L_{\mathcal{L}}  L_f  \Vert \boldsymbol{a} \Vert)
            \Vert \boldsymbol{y}_{1} -  \boldsymbol{y}_{2} \Vert.
        \end{aligned}
    \end{equation}
   The exponential map and logarithmic map satisfy that
   \begin{equation}
   \small
       \begin{aligned}
           \underset{c \rightarrow 0}{\lim} {\rm expm}_{\boldsymbol{y}}^{c}(\boldsymbol{x}) =  \boldsymbol{x} + \boldsymbol{y}, 
           \underset{c \rightarrow 0}{\lim} {\rm logm}_{\boldsymbol{y}}^{c}(\boldsymbol{x}) =  \boldsymbol{x} - \boldsymbol{y}. 
       \end{aligned}
   \end{equation}
   By utilizing this property, we can obtain the expression of 
   $| \mathcal{L}(\boldsymbol{w}, c, \boldsymbol{y}_{1}) - \mathcal{L}(\boldsymbol{w}, c, \boldsymbol{y}_{2}) |$ in Euclidean spaces, \textit{i.e.},
   \begin{equation}
   \small
       \begin{aligned}
         & \underset{c \rightarrow 0}{\lim}
          | \mathcal{L}(\boldsymbol{w}, c, \boldsymbol{y}_{1}) - \mathcal{L}(\boldsymbol{w}, c, \boldsymbol{y}_{2}) | \\
         & = | \mathcal{L}(f_{\boldsymbol{a}}(\boldsymbol{x} - \boldsymbol{y}_{1}) + \boldsymbol{y}_{1} + \boldsymbol{b} )
          -
           \mathcal{L}(f_{\boldsymbol{a}}(\boldsymbol{x} - \boldsymbol{y}_{2}) + \boldsymbol{y}_{2} + \boldsymbol{b} )
          |.
       \end{aligned}
   \end{equation}
   Moreover, it holds that
   \begin{equation}
    \small
       \begin{aligned}
           &| \mathcal{L}(f_{\boldsymbol{a}}(\boldsymbol{x} - \boldsymbol{y}_{1}) + \boldsymbol{y}_{1} + \boldsymbol{b} )
          -
           \mathcal{L}(f_{\boldsymbol{a}}(\boldsymbol{x} - \boldsymbol{y}_{2}) + \boldsymbol{y}_{2} + \boldsymbol{b} )
          |
          \\&\le L_{\mathcal{L}} \Vert f_{\boldsymbol{a}}(\boldsymbol{x} - \boldsymbol{y}_{1}) -  f_{\boldsymbol{a}}(\boldsymbol{x}- \boldsymbol{y}_{2}) + \boldsymbol{y}_{1} - \boldsymbol{y}_{2} \Vert \\
          &\le  L_{\mathcal{L}}\left( \Vert f_{\boldsymbol{a}}(\boldsymbol{x} - \boldsymbol{y}_{1})  -  f_{\boldsymbol{a}}(\boldsymbol{x} - \boldsymbol{y}_{2}) \Vert 
          +
          \Vert
          \boldsymbol{y}_{1} - \boldsymbol{y}_{2} 
          \Vert
          \right)\\
          & \le 
           L_{\mathcal{L}}  \Vert
          \boldsymbol{y}_{1} - \boldsymbol{y}_{2} 
          \Vert+ 
            L_{\mathcal{L}} L_f  \Vert \boldsymbol{a} \Vert   \Vert
          \boldsymbol{y}_{1} - \boldsymbol{y}_{2} 
          \Vert.
       \end{aligned}
   \end{equation}
     In this way, we have proved that when $c \rightarrow 0$, our proof is consistent with the expression in Euclidean spaces.

\end{proof}

\begin{corollary}
\label{corollary:tangen_point_y}
    By setting $\boldsymbol{y}_{1} = \boldsymbol{y}$ and $\boldsymbol{y}_{2} = \boldsymbol{0}$, it holds that
    \begin{equation}
    \small
        \begin{aligned}
              |\mathcal{L}(\boldsymbol{w}, c, \boldsymbol{y}) - \mathcal{L}(\boldsymbol{w}, c, \boldsymbol{0}) | \le \mathcal{E}_{l_{y}},
        \end{aligned}
    \end{equation}
    where $\mathcal{E}_{l_{y}}$ is a constant and is computed as
    \begin{equation}
   \small
       \begin{aligned}
           \mathcal{E}_{l_{y}} =  L_{\mathcal{L}} L_{\oplus_{x}}\Vert \boldsymbol{y} \Vert
           (L_{{\rm expm}_{y}}   +  L_{{\rm expm}_{x}} L_f L_{{\rm logm}_{y}} \Vert \boldsymbol{a} \Vert ).
       \end{aligned}
   \end{equation}
   When $\boldsymbol{y} \rightarrow \boldsymbol{0}$, it holds that
    \begin{equation}
    \small
        \begin{aligned}
            \underset{\boldsymbol{y} \rightarrow \boldsymbol{0}}{\lim}  \mathcal{E}_{l_{y}}
            = 0,
        \end{aligned}
    \end{equation} 
    which also conforms to common sense. 
   By utilizing the hyperbolic property of $\boldsymbol{y}$, $\mathcal{E}_{l_{y}}$ can be further modeled as 
    \begin{equation}
    \small
        \begin{aligned}
            \mathcal{E}_{l_{y}} \triangleq L_{\mathcal{L}} L_{\oplus_{x}} \frac{1}{c}(L_{{\rm expm}_{y}} +  L_{{\rm expm}_{x}} L_f L_{{\rm logm}_{y}} \Vert \boldsymbol{a} \Vert).
        \end{aligned}
    \end{equation}
\end{corollary}
From the corollary, we have proved that the approximation error of setting the tangent point $\boldsymbol{y}$ as $\boldsymbol{0}$ is upper-bounded.

\noindent \textbf{Discussion about assumptions.} The assumptions in our theoretical analysis are generally mild. For example, the angular constraint $(\cos \theta \ge \cos{\Tilde{\theta}})$ essentially requires that the angle between two vectors does not approach 90 degrees. In high-dimensional spaces, two randomly sampled vectors are rarely exactly orthogonal, making this condition naturally satisfied in most cases.  As for the assumptions related to the Lipschitz continuity of the network, existing theoretical analysis~\citep{gouk2021regularisation} shows that common neural network modules, including fully connected layers, convolutional layers, ReLU, and BatchNorm, are Lipschitz continuous, with their Lipschitz constants computable or controllable via the spectral norms of the weights.

In fact,  it is common for many well-established methods that rely on certain assumptions for theoretical tractability, while still achieving strong empirical performance in real-world scenarios. 
% Many widely-used machine learning methods rely on assumptions to enable theoretical analysis.
For example, convergence guarantees for optimization algorithms such as Adam~\citep{kingma2014adam} typically require assumptions like Lipschitz continuity and smoothness of the loss function. Similarly, theoretical studies on Low-Rank Adaptation (LoRA)~\citep{zengexpressive, hu2022lora} introduce mild structural assumptions to ensure its adaptation capability.
Similar to other theoretical frameworks, the assumptions in our work do not necessarily undermine the real-world applicability of our method.

In future work, we intend to extend our theoretical analysis by relaxing the current assumptions. To this end, we plan to investigate localized or parameterized Lipschitz bounds, and we will develop new approximation techniques and bounding, thereby broadening the theoretical applicability of our results.

\section{Generalization Analyses in HNNs}

% \subsection{Sharpness and Generalization in HNNs}
\label{section:sharpness_generalization}
% \subsubsection{Theoretical study}
% The HNNs are usually trained by minimizing a loss function $\mathcal{L}(\cdot)$. Optimizing $\mathcal{L}(\cdot)$ is a typically non-convex problem that exhibits multiple minima with various generalization abilities.

% \MH{it would be better if you define the loss, the sample $\mathcal{S}$ first}

In order to develop the theoretical foundation of the effect of curvatures on HNNs, we derive the PAC-Bayesian generalization bound of HNNs in Theorem~\ref{Theorem:generalization_error_bound}.
% , based on generalization bound~\citep{foret2020sharpness}.
Note that the generalization error bound in Euclidean spaces cannot be directly applied to HNNs.   
The reason is that the generalization error bound in Euclidean spaces assumes the parameters $\boldsymbol{w}$ obeying the normal distribution~\citep{foret2020sharpness}, which does not hold for the parameters of HNNs.
Moreover, the generalization error bound analysis in Euclidean spaces does not consider the hyperbolic operations in hyperbolic spaces that are essential for HNNs.
% which parameters of HNNs can not satisfy.
This discrepancy indicates that the existing derivation does not fully account for the geometric properties of hyperbolic spaces, and thus, a tailored derivation to accommodate the characteristics of hyperbolic spaces in HNNs is necessary.
% ~\citep{mcallester1999pac, dziugaite2017computing}. \zhi{remove the two papers}
% Before giving the generalization bound, we first present the approximation error bounds between using the fixed tangent space  and using a dynamic tangent space.

To establish the generalization error bound analysis, we first introduce two auxiliary lemmas.
\begin{lemma}
\label{lemma:exponential_map}
The exponential map operation satisfies that
    $\Vert \boldsymbol{x} - {\rm expm}_{\boldsymbol{0}}^{c}(\boldsymbol{x}) \Vert = |\Vert \boldsymbol{x} \Vert - {\rm tanh}(\sqrt{c} \Vert \boldsymbol{x} \Vert ) \frac{1}{\sqrt{c}}| $,
    and
    $\underset{c \rightarrow 0}{ \lim } \Vert \boldsymbol{x} - {\rm expm}_{\boldsymbol{0}}^{c}(\boldsymbol{x}) \Vert = 0$.
\end{lemma}

\begin{proof}
    Recall that the exponential map is computed as 
    \begin{equation}
    \small
    \begin{aligned}
        \text{expm}_{\boldsymbol{0}}^{c}(\boldsymbol{x}) = \text{tanh}(\sqrt{|c|} \Vert \boldsymbol{x} \Vert )\frac{\boldsymbol{x}}{\sqrt{|c|}\Vert \boldsymbol{x} \Vert}.
    \end{aligned}
\end{equation}
Therefore,
\begin{equation}
\small
    \begin{aligned}
        \Vert \boldsymbol{x} - {\rm expm}_{\boldsymbol{0}}^{c}(\boldsymbol{x}) \Vert 
        &=
         \left\Vert \boldsymbol{x} -\text{tanh}(\sqrt{|c|} \Vert \boldsymbol{x} \Vert )\frac{\boldsymbol{x}}{\sqrt{|c|}\Vert \boldsymbol{x} \Vert}
         \right\Vert
         \\
         &=  \Vert \boldsymbol{x} \Vert \left| 1- \frac{\text{tanh}(\sqrt{|c|} \Vert \boldsymbol{x} \Vert )}{\sqrt{|c|}\Vert \boldsymbol{x} \Vert} \right| \\
         & = \left | \Vert \boldsymbol{x} \Vert - \frac{\text{tanh}(\sqrt{|c|} \Vert \boldsymbol{x} \Vert )}{\sqrt{|c|}} \right|. 
    \end{aligned}
\end{equation}
As to $\frac{\text{tanh}(\sqrt{|c|} \Vert \boldsymbol{x} \Vert )}{\sqrt{|c|}}$, we observe that
\begin{equation}
\small
    \begin{aligned}
        &\underset{c \rightarrow 0}{\lim}
        \frac{\text{tanh}(\sqrt{|c|} \Vert \boldsymbol{x} \Vert )}{\sqrt{|c|}}
        =
        \underset{c \rightarrow 0}{\lim}
        \frac{{\rm sech}^{2}(\sqrt{c}\Vert \boldsymbol{x} \Vert) \frac{\Vert \boldsymbol{x} \Vert}{2 \sqrt{c}} }{\frac{1}{2 \sqrt{c}}} \\
       & =
        \underset{c \rightarrow 0}{\lim} {\rm sech}^{2}(\sqrt{c}\Vert \boldsymbol{x} \Vert) \Vert \boldsymbol{x} \Vert
        =
        \Vert \boldsymbol{x} \Vert.
    \end{aligned}
\end{equation}
Therefore, it holds that
\begin{equation}
\small
    \begin{aligned}
        \underset{c \rightarrow 0}{ \lim } \Vert \boldsymbol{x} - {\rm expm}_{\boldsymbol{0}}^{c}(\boldsymbol{x}) \Vert = 0.
    \end{aligned}
\end{equation}

\end{proof}

\begin{lemma}
\label{lemma:logarithmic_map}
The logarithmic map operation satisfies that
    $\Vert \boldsymbol{x} - {\rm logm}_{\boldsymbol{0}}^{c}(\boldsymbol{x}) \Vert = |\Vert \boldsymbol{x} \Vert - {\rm tanh}^{-1}(\sqrt{c} \Vert \boldsymbol{x} \Vert ) \frac{1}{\sqrt{c}}| $,
    and
    $\underset{c \rightarrow 0}{ \lim } \Vert \boldsymbol{x} - {\rm logm}_{\boldsymbol{0}}^{c}(\boldsymbol{x}) \Vert = 0$.
\end{lemma}

\begin{proof}
    Recall that the exponential map is computed as 
    \begin{equation}
    \small
    \begin{aligned}
        \text{logm}_{\boldsymbol{0}}^{c}(\boldsymbol{x}) = \text{tanh}^{-1}(\sqrt{|c|} \Vert \boldsymbol{x} \Vert )\frac{\boldsymbol{x}}{\sqrt{|c|}\Vert \boldsymbol{x} \Vert}.
    \end{aligned}
\end{equation}
Therefore, we can obtain that
\begin{equation}
\small
    \begin{aligned}
        \Vert \boldsymbol{x} - {\rm logm}_{\boldsymbol{0}}^{c}(\boldsymbol{x}) \Vert 
        &=
         \left\Vert \boldsymbol{x} -\text{tanh}^{-1}(\sqrt{|c|} \Vert \boldsymbol{x} \Vert )\frac{\boldsymbol{x}}{\sqrt{|c|}\Vert \boldsymbol{x} \Vert}
         \right\Vert
         \\
         &=  \Vert \boldsymbol{x} \Vert \left| 1- \frac{\text{tanh}^{-1}(\sqrt{|c|} \Vert \boldsymbol{x} \Vert )}{\sqrt{|c|}\Vert \boldsymbol{x} \Vert} \right| \\
         & = \left | \Vert \boldsymbol{x} \Vert - \frac{\text{tanh}^{-1}(\sqrt{|c|} \Vert \boldsymbol{x} \Vert )}{\sqrt{|c|}} \right|. 
    \end{aligned}
\end{equation}
As to $\frac{\text{tanh}^{-1}(\sqrt{|c|} \Vert \boldsymbol{x} \Vert )}{\sqrt{|c|}}$, we denote $t \triangleq \sqrt{c}$ and  
observe that
\begin{equation}
\small
    \begin{aligned}
        \underset{t \rightarrow 0}{\lim}
        \frac{\text{tanh}^{-1}(t \Vert \boldsymbol{x} \Vert )}{t}
        &=
         \underset{t \rightarrow 0}{\lim}
        \frac{ \frac{d}{dt} \text{tanh}^{-1}(t \Vert \boldsymbol{x} \Vert ) }{\frac{d}{dt} t}\\
        &=
         \underset{t \rightarrow 0}{\lim}
        \frac{ \frac{\Vert x \Vert}{1 - t^{2}\Vert x \Vert^{2}} }{1}
        =
        \Vert \boldsymbol{x} \Vert,
    \end{aligned}
\end{equation}
and thus
\begin{equation}
\small
    \begin{aligned}
       \frac{\text{tanh}^{-1}(\sqrt{|c|} \Vert \boldsymbol{x} \Vert )}{\sqrt{|c|}}
        =
        \Vert \boldsymbol{x} \Vert,
    \end{aligned}
\end{equation}
Therefore, 
\begin{equation}
\small
    \begin{aligned}
        \underset{c \rightarrow 0}{ \lim } \Vert \boldsymbol{x} - {\rm logm}_{\boldsymbol{0}}^{c}(\boldsymbol{x}) \Vert = 0.
    \end{aligned}
\end{equation}

\end{proof}

After the auxiliary lemmas, we establish the generalization analysis in HNNs shown in Theorem~\ref{Theorem:generalization_error_bound}, which quantifies the error bound between loss on any distribution and loss on training set.
\begin{theorem}
	\label{Theorem:generalization_error_bound} 
    % Suppose that $\mathcal{L}(\cdot)$ is Lipschitz continuous with constant $L_{\mathcal{L}}$, and $f_{\boldsymbol{a}}(\cdot)$ is  Lipschitz continuous with constant $L_{f}$.
    % Suppose that the loss function $\mathcal{L}(\cdot)$ is $L_{\mathcal{L}}^{'}-$Lipschitz continuous to the logits, the hyperbolic MLR in Eq.~\eqref{equation:hyperbolic_mlr} is $L_{p}-$Lipschitz continuous to its input $\boldsymbol{x}$, and $f_{\boldsymbol{a}}(\cdot)$ is  $L_{f}-$Lipschitz continuous.
    % Here, we also denote $\mathcal{L}_{\mathcal{D}}$ as the loss function on the real distribution $\mathcal{D}$, and denote $\mathcal{L}_{\mathcal{S}}$ as the empirical loss on the training set $\mathcal{S}$.
    Suppose that the loss function $\mathcal{L}(\cdot)$ is $L_{\mathcal{L}}^{'}-$Lipschitz continuous with respect
to the predicted probability, the hyperbolic MLR in Eq.~\eqref{equation:hyperbolic_mlr} is $L_{p}-$Lipschitz continuous with respect to its input $\boldsymbol{x}$, and $f_{\boldsymbol{a}}(\cdot)$ is  $L_{f}-$Lipschitz continuous.
	For any $\rho>0$ and any real distribution $\mathcal{D}$, with probability $1-\delta$ over the choice of the training set $\mathcal{S} \sim \mathcal{D}$, the generalization  bound on distribution $\mathcal{D}$ is
 \begin{equation}
            \small
            \label{equation:ge_error_bound}
	    \begin{aligned}
	        \mathcal{L}_{\mathcal{D}}(\boldsymbol{w}, c, \boldsymbol{y}) 
         \le 
                \mathcal{L}_{\mathcal{S}}(\boldsymbol{w}, c, \boldsymbol{y}) 
                +
        \mathcal{L}_{\mathcal{S}}^{\text{sharp}}
                 +  \mathcal{E}_{l_{gen}},
	    \end{aligned}
	\end{equation}
where 
\begin{equation}
\small
    \begin{aligned}
        \mathcal{L}_{\mathcal{S}}^{\text{sharp}} = \underset{\Vert \boldsymbol{\epsilon} \Vert \le \rho}{\max} \mathcal{L}_{\mathcal{S}}(\boldsymbol{w}+\boldsymbol{\epsilon}, c, \boldsymbol{y})  - \mathcal{L}_{\mathcal{S}}(\boldsymbol{w}, c, \boldsymbol{y}),
    \end{aligned}
\end{equation}
$\mathcal{L}_{\mathcal{D}}(\cdot)$ is the loss function on $\mathcal{D}$, and $\mathcal{L}_{\mathcal{S}}(\cdot)$ is the loss function on $\mathcal{S}$. $\mathcal{E}_{l_{gen}}^{'}$ is given by
\begin{equation}
\small
    \begin{aligned}
         \mathcal{E}_{l_{gen}} \triangleq \mathcal{E}_{l_{gen}}^{'} +  \mathcal{E}_{l_{y}} +  \mathcal{E}_{l_{c}} + \mathcal{E}_{l_{y}}^{'} + \mathcal{E}_{l_{c}}^{'},
    \end{aligned}
\end{equation}
where $\mathcal{E}_{l_{gen}}^{'}$ is computed as
   \vspace{-1.5em}
\begin{center}
\begin{equation}
\resizebox{0.47\textwidth}{!}{$
\begin{aligned}
      \mathcal{E}_{l_{gen}}^{'} \triangleq \sqrt{\frac{d\log(1+\frac{\Vert w \Vert^{2}}{\rho^{2}}(1+\sqrt{\frac{\log(n)}{d}})^{2})+4\log \frac{n}{\delta}+8\log(6n+3d)}{n-1}},
\end{aligned}
$}
\end{equation}
\end{center}
  \begin{equation}
  \small
        \begin{aligned}
         &\mathcal{E}_{l_{c}} =  L_{\mathcal{L}} L_{\oplus_{c0}}+L_{\mathcal{L}} L_{\oplus_{x}} \Vert \boldsymbol{a} \Vert \frac{{\rm tanh}^{-1}(c^{1/4})}{\sqrt{c}} (L_{f} + 1),\\
            &\mathcal{E}_{l_{c}}^{'} =  L_{\mathcal{L}} L_{\oplus_{c0}}+L_{\mathcal{L}} L_{\oplus_{x}}  \frac{{\rm tanh}^{-1}(c^{1/4})}{\sqrt{c}} (L_{f} + 1)(\Vert \boldsymbol{a} \Vert + \rho), \\
            & \mathcal{E}_{l_{y}}^{'} =  L_{\mathcal{L}} L_{\oplus_{x}} \frac{1}{c}(L_{{\rm expm}_{y}} + L_{{\rm expm}_{x}} L_f L_{{\rm logm}_{y}} (\Vert \boldsymbol{a} \Vert + \rho) ), \\
        \end{aligned}
    \end{equation}
$n = |\mathcal{S}|$, and $d$ denotes numbers of parameters.
$\boldsymbol{a}$,  a subset of $\boldsymbol{w}$, is the parameters of the function $f_{\boldsymbol{a}}(\cdot)$.
% $\boldsymbol{a}$ denotes the parameters of function $f_{\boldsymbol{a}}(\cdot)$, which are parts of $\boldsymbol{w}$.
The generalization error bound of loss function with respect to HNNs satisfies that
    \begin{equation}
    \small
        \begin{aligned}
              \underset{\boldsymbol{y} \rightarrow 0, c \rightarrow 0}{\lim} \mathcal{E}_{l_{gen}} = \mathcal{E}_{l_{gen}}^{'}.
        \end{aligned}
    \end{equation}
\end{theorem}

\begin{proof}
We aim to derive an upper bound for the loss given a distribution $\mathcal{D}$.
\emph{The main challenge lies in the absence of important mathematical theorems, such as the KL divergence on hyperbolic spaces, which does not have an analytic expression~\citep{cho2024hyperbolic}. }
To overcome this challenge, we base our theoretical analyses on the PAC-Bayesian generalization error bound, leveraging the Lipschitz continuity and Lipschitz bound in HNNS in Section~\ref{section:lip_hyper_opera}, as well as Lipschitz continuity of the loss function.
% To overcome this challenge, we base our theoretical analysis on the PAC-Bayesian generalization error bound and leverage the analytical expressions of hyperbolic operations and the Lipschitz continuity of the loss function.

  Follow the PAC-Bayesian generalization error bound~\citep{foret2020sharpness}, with probability $1-\frac{1}{\sqrt{n}}$, the loss function $\mathcal{L}(\boldsymbol{w}, 0, \boldsymbol{0})$ satisfies that
   \vspace{-1.5em}
\begin{center}
\begin{equation}
\resizebox{0.47\textwidth}{!}{$
\begin{aligned}
\label{equation:generalization_bound}
 & \mathcal{L}_{\mathcal{D}}(\boldsymbol{w}, 0, \boldsymbol{0}) \le \underset{\Vert \boldsymbol{\epsilon} \Vert \le \rho}{\max} \mathcal{L}_{\mathcal{S}}(\boldsymbol{w} + \boldsymbol{\epsilon}, 0, \boldsymbol{0}) + \\
            &\sqrt{\frac{d\log(1+\frac{\Vert w \Vert^{2}}{\rho^{2}}(1+\sqrt{\frac{\log(n)}{d}})^{2})+4\log \frac{n}{\delta}+8\log(6n+3d)}{n-1}}.
\end{aligned}
$}
\end{equation}
\end{center}
We denote the constant term in Eq.~\eqref{equation:generalization_bound} as $\mathcal{E}_{l_{gen}}^{'}$, 
   \vspace{-1.5em}
\begin{center}
\begin{equation}
\resizebox{0.47\textwidth}{!}{$
\begin{aligned}
      \mathcal{E}_{l_{gen}}^{'} \triangleq \sqrt{\frac{d\log(1+\frac{\Vert w \Vert^{2}}{\rho^{2}}(1+\sqrt{\frac{\log(n)}{d}})^{2})+4\log \frac{n}{\delta}+8\log(6n+3d)}{n-1}}.
\end{aligned}
$}
\end{equation}
\end{center}

We define $\boldsymbol{x}_{1}$ and $\boldsymbol{x}_{2}$ as the hyperbolic representations computed by Eq.~\eqref{equation:hyperbolic_forward}, setting the curvature as $c$ and $0$, respectively.
Mathematically, $\boldsymbol{x}_{1}$ and $\boldsymbol{x}_{2}$ are computed as
\begin{equation}
\small
    \begin{aligned}
         &\boldsymbol{x}_{1} = {\rm expm}_{\boldsymbol{0}}^{c}(f_{\boldsymbol{a}}({\rm logm}^{c}_{\boldsymbol{0}}(\boldsymbol{x}))) \oplus_{c} \boldsymbol{b} \\
         &\boldsymbol{x}_{2} = {\rm expm}_{\boldsymbol{0}}^{0}(f_{\boldsymbol{a}}({\rm logm}^{0}_{\boldsymbol{0}}(\boldsymbol{x}))) \oplus_{0} \boldsymbol{b} =f_{\boldsymbol{a}}(\boldsymbol{x}) + \boldsymbol{b}.
    \end{aligned}
\end{equation}
From Eq.~\eqref{equation:Lip_lossfunction_specific} and $L_{\mathcal{L}} = L_{\mathcal{L}}^{'} L_{p}$, we can obtain that
    \begin{equation}
	    \begin{aligned}
	       &| \mathcal{L}(\boldsymbol{w}, c, \boldsymbol{0}) -
            \mathcal{L}(\boldsymbol{w}, 0, \boldsymbol{0})  |  \le L_{\mathcal{L}} \Vert \boldsymbol{x}_{1} - \boldsymbol{x}_{2} \Vert\\
            &\le 
          L_{\mathcal{L}} \Vert
            {\rm expm}_{\boldsymbol{0}}^{c} \left(
            f_{\boldsymbol{a}}({\rm logm}_{\boldsymbol{0}}^{c}(\boldsymbol{x}))  
            \right) \oplus_{c} \boldsymbol{b}
            -
            (f_{\boldsymbol{a}} (\boldsymbol{x}) + \boldsymbol{b})
            \Vert.
	    \end{aligned}
	\end{equation}
From Lipschitz continuity in Theorems~\ref{theorem:lip_continous_oplus_c} and~\ref{theorem:lip_continous_oplus_x}, it holds that
    \begin{equation}
    \small
	    \begin{aligned}
	       &| \mathcal{L}(\boldsymbol{w}, c, \boldsymbol{0}) -
            \mathcal{L}(\boldsymbol{w}, 0, \boldsymbol{0})  | \\
         &\le 
          L_{\mathcal{L}} \Vert
            {\rm expm}_{\boldsymbol{0}}^{c} \left(
            f_{\boldsymbol{a}}({\rm logm}_{\boldsymbol{0}}^{c}(\boldsymbol{x}))  
            \right) \oplus_{c} \boldsymbol{b}
            -
            (f_{\boldsymbol{a}} (\boldsymbol{x}) + \boldsymbol{b})
            \Vert\\
          % & \le
          % L_{\mathcal{L}} \Vert
          %   {\rm expm}_{\boldsymbol{0}}^{c} \left(
          %   f_{\boldsymbol{a}}({\rm logm}_{\boldsymbol{0}}^{c}(\boldsymbol{x}))  
          %   \right) \oplus_{c} \boldsymbol{b}
          %   -
          %   (f_{\boldsymbol{a}} (\boldsymbol{x}) \oplus_{c} \boldsymbol{b})
          %   \Vert\\
          %   & 
          %   +L_{\mathcal{L}}
          %   \Vert
          %   (f_{\boldsymbol{a}} (\boldsymbol{x}) \oplus_{c} \boldsymbol{b})
          %   -
          %   (f_{\boldsymbol{a}} (\boldsymbol{x}) + \boldsymbol{b})
          %   \Vert
          %   \\
            & \le L_{\mathcal{L}} L_{\oplus_{x}} \Vert {\rm expm}_{\boldsymbol{0}}^{c} \left(
            f_{\boldsymbol{a}}({\rm logm}_{\boldsymbol{0}}^{c}(\boldsymbol{x}))  
            \right) - f_{\boldsymbol{a}}(\boldsymbol{x})) \Vert + L_{\mathcal{L}} L_{\oplus_{c0}}.
	    \end{aligned}
	\end{equation}
            We can further derive that
         \begin{equation}
         \small
        \label{equation:expm_0_f_a}
            \begin{aligned}
                &L_{\mathcal{L}} L_{\oplus_{x}} \Vert {\rm expm}_{\boldsymbol{0}}^{c} \left(
        f_{\boldsymbol{a}}({\rm logm}_{\boldsymbol{0}}^{c}(\boldsymbol{x}))  
        \right) - f_{\boldsymbol{a}}(\boldsymbol{x})) \Vert  \\
        &\le
        L_{\mathcal{L}} L_{\oplus_{x}} 
        \left(
        \Vert
         {\rm expm}_{\boldsymbol{0}}^{c} \left(
        f_{\boldsymbol{a}}({\rm logm}_{\boldsymbol{0}}^{c}(\boldsymbol{x}))  
        \right) 
        -
         f_{\boldsymbol{a}}({\rm logm}_{\boldsymbol{0}}^{c}(\boldsymbol{x}))  
         \Vert\right. \\
         &\left.+ 
         \Vert
         f_{\boldsymbol{a}}({\rm logm}_{\boldsymbol{0}}^{c}(\boldsymbol{x}))   - f_{\boldsymbol{a}}(\boldsymbol{x})
         \Vert
        \right).
            \end{aligned}
        \end{equation}
        From Lemma~\ref{lemma:exponential_map}, left terms in Eq.~\eqref{equation:expm_0_f_a} can be further modeled as 
        \begin{equation}
        \small
            \begin{aligned}
            &    \Vert
         {\rm expm}_{\boldsymbol{0}}^{c} \left(
        f_{\boldsymbol{a}}({\rm logm}_{\boldsymbol{0}}^{c}(\boldsymbol{x}))  
        \right) 
        -
         f_{\boldsymbol{a}}({\rm logm}_{\boldsymbol{0}}^{c}(\boldsymbol{x}))  
         \Vert 
         \\
         &\le  \bigg|\Vert f_{\boldsymbol{a}}({\rm logm}_{\boldsymbol{0}}^{c}(\boldsymbol{x})) \Vert
        - {\rm tanh}(\sqrt{c} \Vert f_{\boldsymbol{a}}({\rm logm}_{\boldsymbol{0}}^{c}(\boldsymbol{x})) \Vert) \frac{1}{\sqrt{c}}
        \bigg|.
            \end{aligned}
        \end{equation}
        Because 
        \begin{equation}
        \small
            \begin{aligned}
                \Vert f_{\boldsymbol{a}}({\rm logm}_{\boldsymbol{0}}^{c}(\boldsymbol{x})) \Vert > {\rm tanh}(\sqrt{c} \Vert f_{\boldsymbol{a}}({\rm logm}_{\boldsymbol{0}}^{c}(\boldsymbol{x})) \Vert) \frac{1}{\sqrt{c}},
            \end{aligned}
        \end{equation}
        and $f(x)-{\rm tanh}^{-1}(\sqrt{c}x)/\sqrt{c}$ is an increasing function,
        it holds that
        \begin{equation}
        \small
            \begin{aligned}
                &\bigg| \Vert f_{\boldsymbol{a}}({\rm logm}_{\boldsymbol{0}}^{c}(\boldsymbol{x})) \Vert - {\rm tanh}(\sqrt{c} \Vert f_{\boldsymbol{a}}({\rm logm}_{\boldsymbol{0}}^{c}(\boldsymbol{x})) \Vert) \frac{1}{\sqrt{c}} \bigg| \\
                &\le \Tilde{N} -  \frac{{\rm tanh}(\sqrt{c} \Tilde{N})}{\sqrt{c}},
            \end{aligned}
        \end{equation}
        where $\Tilde{N}$ is denoteds as the upper bound of $f_{\boldsymbol{a}}({\rm logm}_{\boldsymbol{0}}^{c}(\boldsymbol{x}))$, \textit{i.e.}, $\Vert f_{\boldsymbol{a}}({\rm logm}_{\boldsymbol{0}}^{c}(\boldsymbol{x})) \Vert \le \Tilde{N}$.
        
        From the Lipschitz continuity assumption of $f_{\boldsymbol{a}}(\cdot)$ and Lemma~\ref{lemma:logarithmic_map}, the right term in Eq.~\eqref{equation:expm_0_f_a} can be further modeled as
         \begin{equation}
         \small
         \label{equation:f_a_logm_f_a}
            \begin{aligned}
                &\Vert
         f_{\boldsymbol{a}}({\rm logm}_{\boldsymbol{0}}^{c}(\boldsymbol{x}))   - f_{\boldsymbol{a}}(\boldsymbol{x})
         \Vert \le
           L_f \Vert \boldsymbol{a} \Vert \Vert {\rm logm}_{\boldsymbol{0}}^{c}(\boldsymbol{x}) - \boldsymbol{x} \Vert\\
           & \le L_f \Vert \boldsymbol{a} \Vert \bigg| \Vert \boldsymbol{x} \Vert - {\rm tanh}^{-1}(\sqrt{c} \Vert \boldsymbol{x} \Vert ) \frac{1}{\sqrt{c}}  \bigg|
           .
            \end{aligned}
        \end{equation}
        Besides, because 
        \begin{equation}
        \small
            \begin{aligned}
                {\rm tanh}^{-1}(\sqrt{c} \Vert \boldsymbol{x} \Vert ) \frac{1}{\sqrt{c}}  > \Vert \boldsymbol{x} \Vert,
            \end{aligned}
        \end{equation}
        Eq.~\eqref{equation:f_a_logm_f_a} is converted into 
        \begin{equation}
        \small
            \begin{aligned}
              &  \Vert  f_{\boldsymbol{a}}({\rm logm}_{\boldsymbol{0}}^{c}(\boldsymbol{x}))   - f_{\boldsymbol{a}}(\boldsymbol{x})
         \Vert \\
         &\le L_f \Vert \boldsymbol{a} \Vert  \left(   {\rm tanh}^{-1}(\sqrt{c} \Vert \boldsymbol{x} \Vert ) \frac{1}{\sqrt{c}} - \Vert \boldsymbol{x} \Vert \right).
            \end{aligned}
        \end{equation}
        In this way, Eq.~\eqref{equation:expm_0_f_a} can be transformed as 
        \begin{equation}
        \small
        \label{equation:expm_0_f_a_2}
            \begin{aligned}
                &L_{\mathcal{L}} L_{\oplus_{x}} \Vert {\rm expm}_{\boldsymbol{0}}^{c} \left(
        f_{\boldsymbol{a}}({\rm logm}_{\boldsymbol{0}}^{c}(\boldsymbol{x}))  
        \right) - f_{\boldsymbol{a}}(\boldsymbol{x})) \Vert  \\
        &\le
         L_{\mathcal{L}} L_{\oplus_{x}} 
        \left(
        \Tilde{N} -  \frac{{\rm tanh}(\sqrt{c} \Tilde{N})}{\sqrt{c}}
        \right.
        \\ &\left.+
        L_f \Vert \boldsymbol{a} \Vert \big( {\rm tanh}^{-1}(\sqrt{c} \Vert \boldsymbol{x} \Vert ) \frac{1}{\sqrt{c}} -  \Vert \boldsymbol{x} \Vert \big)
        \right).
            \end{aligned}
        \end{equation}
        In conclusion,
        the loss function $\mathcal{L}(\cdot)$ satisfies that
        \begin{equation}
        \small
            \begin{aligned}
                 &| \mathcal{L}(\boldsymbol{w}, c, \boldsymbol{0}) -
            \mathcal{L}(\boldsymbol{w}, 0, \boldsymbol{0})  |  \le L_{\mathcal{L}} L_{\oplus_{c0}}
            \\ & + L_{\mathcal{L}} L_{\oplus_{x}} 
            \left(\Tilde{N} -  \frac{{\rm tanh}(\sqrt{c} \Tilde{N})}{\sqrt{c}} \right.
            \\
         & +
        \left.
       L_f \Vert \boldsymbol{a} \Vert \big( {\rm tanh}^{-1}(\sqrt{c} \Vert \boldsymbol{x} \Vert ) \frac{1}{\sqrt{c}} -  \Vert \boldsymbol{x} \Vert \big)
        \right).
            \end{aligned}
        \end{equation}
        By denoting 
        \begin{equation}
        \small
            \begin{aligned}
               & \mathcal{E}_{l_{c}} \triangleq L_{\mathcal{L}} L_{\oplus_{c0}} + L_{\mathcal{L}} L_{\oplus_{x}} 
            \left(\Tilde{N} -  \frac{{\rm tanh}(\sqrt{c} \Tilde{N})}{\sqrt{c}} \right.
            \\
         & +
        \left.
       L_f \Vert \boldsymbol{a} \Vert \big( {\rm tanh}^{-1}(\sqrt{c} \Vert \boldsymbol{x} \Vert ) \frac{1}{\sqrt{c}} -  \Vert \boldsymbol{x} \Vert \big)
        \right),
            \end{aligned}
        \end{equation}
        it can be proved that
        \begin{equation}
        \small
        \label{equation:l_w_c_l_w_0}
	    \begin{aligned}
	       | \mathcal{L}(\boldsymbol{w}, c, \boldsymbol{0}) -
            \mathcal{L}(\boldsymbol{w}, 0, \boldsymbol{0})  |
         \le 
         \mathcal{E}_{l_{c}}.
	    \end{aligned}
	\end{equation}
    
        From the established theorems, we have that
        \begin{equation}
        \small
        \left\{
            \begin{aligned}
               & \underset{c \rightarrow 0}{\lim}
                  \Tilde{N} -  \frac{{\rm tanh}(\sqrt{c} \Tilde{N})}{\sqrt{c}}
         = 0 \\
         & \underset{c \rightarrow 0}{\lim}  {\rm tanh}^{-1}(\sqrt{c} \Vert \boldsymbol{x} \Vert ) \frac{1}{\sqrt{c}} -  \Vert \boldsymbol{x} \Vert  = 0\\
         & \underset{c \rightarrow 0}{\lim} L_{\oplus_{c0}}^{'} = 0
            \end{aligned}
            \right..
        \end{equation}
    Therefore, $\mathcal{E}_{l_c}$ satisfies that
    \begin{equation}
    \small
        \begin{aligned}
            \underset{c \rightarrow 0}{\lim}  \mathcal{E}_{l_c}  = 0.
        \end{aligned}
    \end{equation}
    By constraining the upper bound of $\Vert \boldsymbol{x} \Vert$, \textit{i.e.}, $\Vert \boldsymbol{x} \Vert \le \frac{1}{c^{1/4}}$, $ \mathcal{E}_{l_{c}}$ can be modeled as 
    \begin{equation}
    \small
        \begin{aligned}
           & \mathcal{E}_{l_{c}} \le  L_{\mathcal{L}} L_{\oplus_{x}} 
        \left( 
        \Vert f_{\boldsymbol{a}}({\rm logm}_{\boldsymbol{0}}^{c}(\boldsymbol{x})) \Vert\right.
        \\&\left.+
         L_f \Vert \boldsymbol{a} \Vert {\rm tanh}^{-1}(\sqrt{c} \Vert \boldsymbol{x} \Vert ) \frac{1}{\sqrt{c}}
        \right)
        +
        L_{\mathcal{L}} L_{\oplus_{c0}},
        \end{aligned}
    \end{equation}
    where
    \begin{equation}
    \small
        \begin{aligned}
            \Vert f_{\boldsymbol{a}}({\rm logm}_{\boldsymbol{0}}^{c}(\boldsymbol{x})) \Vert 
            &\le \Vert \boldsymbol{a} \Vert \Vert  {\rm logm}_{\boldsymbol{0}}^{c}(\boldsymbol{x}) \Vert
            \\
            &\le \frac{\Vert \boldsymbol{a} \Vert{\rm tanh}^{-1}(\sqrt{c}\Vert \boldsymbol{x} \Vert)}{\sqrt{c}}.
        \end{aligned}
    \end{equation}
    Therefore, the error bound can be further modeled as 
    \begin{equation}
    \small
        \begin{aligned}
            \mathcal{E}_{l_{c}} =  L_{\mathcal{L}} L_{\oplus_{c0}}+L_{\mathcal{L}} L_{\oplus_{x}} \Vert \boldsymbol{a} \Vert \frac{{\rm tanh}^{-1}(c^{1/4})}{\sqrt{c}} (L_{f} + 1).
        \end{aligned}
    \end{equation}

From the above analysis, we have
 \begin{equation}
 \small
	    \begin{aligned}
	       | \mathcal{L}(\boldsymbol{w}, c, \boldsymbol{0}) -
            \mathcal{L}(\boldsymbol{w}, c, \boldsymbol{0})  |
         \le 
         \mathcal{E}_{l_{c}}.
	    \end{aligned}
	\end{equation}
     By substituting $\mathcal{L}_{\mathcal{D}}$ and $\mathcal{L}_{\mathcal{S}}$ into Eq.~\eqref{equation:l_w_c_l_w_0}, we have that
     \begin{equation}
     \small
	    \begin{aligned}
	        &\mathcal{L}_{\mathcal{D}}(\boldsymbol{w}, c, \boldsymbol{0}) -
            \mathcal{L}_{\mathcal{D}}(\boldsymbol{w}, 0, \boldsymbol{0})  
         \le 
         \mathcal{E}_{l_{c}},
         \\
         & \mathcal{L}_{\mathcal{S}}(\boldsymbol{w}, c, \boldsymbol{0}) -
            \mathcal{L}_{\mathcal{S}}(\boldsymbol{w}, 0, \boldsymbol{0})  
         \le 
         \mathcal{E}_{l_{c}}.
	    \end{aligned}
	\end{equation}
    Moreover, the loss function $\mathcal{L}_{\mathcal{S}}(\cdot)$ satisfies that
    \begin{equation}
    \small
    \label{equation:l_s_0_0_l_s_c_0}
        \begin{aligned}
            \mathcal{L}_{\mathcal{S}}(\boldsymbol{w}+ \boldsymbol{\epsilon}, 0, \boldsymbol{0})  
            \le 
             \mathcal{L}_{\mathcal{S}}(\boldsymbol{w} + \boldsymbol{\epsilon}, c, \boldsymbol{0}) 
             +
         \mathcal{E}_{l_{c}}^{'}.
        \end{aligned}
    \end{equation}
    Here, $\mathcal{E}_{l_{c}}^{'}$ is computed as
    \begin{equation}
    \small
        \begin{aligned}
            &\mathcal{E}_{l_{c}}^{'} \triangleq 
        L_{\mathcal{L}} L_{\oplus_{c0}}+
        L_{\mathcal{L}} L_{\oplus_{x}}\left(
        \Tilde{N} - \frac{{\rm tanh}(\sqrt{c} \Tilde{N})}{\sqrt{c}} \right.\\
        &\left.+
        L_f (\Vert \boldsymbol{a} \Vert + \Vert \boldsymbol{\epsilon} \Vert) ( \frac{{\rm tanh}^{-1}(\sqrt{c} \Vert \boldsymbol{x} \Vert )}{\sqrt{c}} - \Vert \boldsymbol{x} \Vert )
        \right)\\
            &=  L_{\mathcal{L}} L_{\oplus_{c0}}+L_{\mathcal{L}} L_{\oplus_{x}}  \frac{{\rm tanh}^{-1}(c^{1/4})}{\sqrt{c}} (L_{f} + 1)(\Vert \boldsymbol{a} \Vert + \Vert \boldsymbol{\epsilon} \Vert),
        \end{aligned}
    \end{equation}
    where $\mathcal{E}_{l_{c}}^{'} $ also satisfies that
    \begin{equation}
    \small
        \begin{aligned}
            \underset{c \rightarrow 0}{\lim}  \mathcal{E}_{l_{c}}^{'} = 0.
        \end{aligned}
    \end{equation}

    From Corollary~\ref{corollary:tangen_point_y}, we have proved that
    \begin{equation}
    \small
        \begin{aligned}
            \mathcal{L}_{\mathcal{D}}(\boldsymbol{w}, c, \boldsymbol{y}) - \mathcal{L}_{\mathcal{D}}(\boldsymbol{w}, c, \boldsymbol{0})  \le \mathcal{E}_{l_{y}},
        \end{aligned}
    \end{equation}
    and thus we can prove that
    \begin{equation}
    \small
        \begin{aligned}
            \mathcal{L}_{\mathcal{D}}(\boldsymbol{w}, c, \boldsymbol{y}) &\le
            \mathcal{L}_{\mathcal{D}}(\boldsymbol{w}, c, \boldsymbol{0}) + \mathcal{E}_{l_{y}} \\
            &\le
             \mathcal{L}_{\mathcal{D}}(\boldsymbol{w}, 0, \boldsymbol{0}) + \mathcal{E}_{l_{c}} + \mathcal{E}_{l_{y}}.
        \end{aligned}
    \end{equation}
    As to $\mathcal{L}_{\mathcal{S}}(\cdot)$, we also have that
    \begin{equation}
    \small
        \begin{aligned}
        \mathcal{L}_{\mathcal{S}}(\boldsymbol{w}, c, \boldsymbol{0}) 
            \le  \mathcal{L}_{\mathcal{S}}(\boldsymbol{w}, c, \boldsymbol{y})  + \mathcal{E}_{l_{y}},
        \end{aligned}
    \end{equation}
    and thus
     \begin{equation}
     \small
     \label{equation:l_s_c_0_l_s_c_y}
        \begin{aligned}
        \mathcal{L}_{\mathcal{S}}(\boldsymbol{w} + \boldsymbol{\epsilon}, c, \boldsymbol{0}) 
            \le  \mathcal{L}_{\mathcal{S}}(\boldsymbol{w} + \boldsymbol{\epsilon}, c, \boldsymbol{y})  + \mathcal{E}_{l_{y}}^{'}.
        \end{aligned}
    \end{equation}
    Here, $\mathcal{E}_{l_{y}}^{'}$ is given by 
    \begin{equation}
    \small
        \begin{aligned}
            & \mathcal{E}_{l_{y}}^{'} \triangleq  L_{\mathcal{L}} L_{\oplus_{x}}
           (L_{{\rm expm}_{y}} \Vert \boldsymbol{y} \Vert  
           \\
           &+  L_{{\rm expm}_{x}} L_f L_{{\rm logm}_{y}} \Vert \boldsymbol{y} \Vert( \Vert \boldsymbol{a} \Vert + \Vert \boldsymbol{\epsilon} \Vert))\\
             &=  L_{\mathcal{L}} L_{\oplus_{x}} \frac{1}{c}(L_{{\rm expm}_{y}} + L_{{\rm expm}_{x}} L_f L_{{\rm logm}_{y}} (\Vert \boldsymbol{a} \Vert + \Vert \boldsymbol{\epsilon} \Vert) ),
        \end{aligned}
    \end{equation}
    where $ \mathcal{E}_{l_{y}}^{'}$ also satisfies that
    \begin{equation}
     \small
        \begin{aligned}
            \underset{\boldsymbol{y} \rightarrow \boldsymbol{0}}{\lim}  \mathcal{E}_{l_{y}}^{'} = 0.
        \end{aligned}
    \end{equation}
    By combining Eq.~\eqref{equation:l_s_0_0_l_s_c_0} and Eq.~\eqref{equation:l_s_c_0_l_s_c_y}, we derive that
    \begin{equation}
     \small
        \begin{aligned}
            \mathcal{L}_{\mathcal{S}}(\boldsymbol{w}+ \boldsymbol{\epsilon}, 0, \boldsymbol{0})  
            &\le 
             \mathcal{L}_{\mathcal{S}}(\boldsymbol{w} + \boldsymbol{\epsilon}, c, \boldsymbol{0}) 
             +
         \mathcal{E}_{l_{c}}^{'} \\
         &\le  \mathcal{L}_{\mathcal{S}}(\boldsymbol{w} + \boldsymbol{\epsilon}, c, \boldsymbol{y})  + \mathcal{E}_{l_{y}}^{'}  +  \mathcal{E}_{l_{c}}^{'}, 
        \end{aligned}
    \end{equation}
    and thus
    \begin{equation}
    \small
        \begin{aligned}
            \underset{\Vert \epsilon \Vert \le \rho}{\max}  \mathcal{L}_{\mathcal{S}}(\boldsymbol{w}+ \boldsymbol{\epsilon}, 0, \boldsymbol{0})  
         \le \underset{\Vert \epsilon \Vert \le \rho}{\max}   \mathcal{L}_{\mathcal{S}}(\boldsymbol{w} + \boldsymbol{\epsilon}, c, \boldsymbol{y})  + \mathcal{E}_{l_{y}}^{'}  +  \mathcal{E}_{l_{c}}^{'}.
        \end{aligned}
    \end{equation}
    From $\Vert \epsilon \Vert \le \rho$, $\mathcal{E}_{l_{y}}^{'},  \mathcal{E}_{l_{c}}^{'}$ can be further modeled as 
    \begin{equation}
    \small
        \begin{aligned}
            &\mathcal{E}_{l_{c}}^{'} =  L_{\mathcal{L}} L_{\oplus_{c0}}+L_{\mathcal{L}} L_{\oplus_{x}}  \frac{{\rm tanh}^{-1}(c^{1/4})}{\sqrt{c}} (L_{f} + 1)(\Vert \boldsymbol{a} \Vert + \rho), \\
            & \mathcal{E}_{l_{y}}^{'} =  L_{\mathcal{L}} L_{\oplus_{x}} \frac{1}{c}(L_{{\rm expm}_{y}} + L_{{\rm expm}_{x}} L_f L_{{\rm logm}_{y}} (\Vert \boldsymbol{a} \Vert + \rho) ).
        \end{aligned}
    \end{equation}

    In this way, by utilizing Eq.~\eqref{equation:generalization_bound}, we establish the connection between $ \mathcal{L}_{\mathcal{D}}(\boldsymbol{w}, c, \boldsymbol{y})$  and $\underset{\Vert \epsilon \Vert \le \rho}{\max}  \mathcal{L}_{\mathcal{S}}(\boldsymbol{w}+\boldsymbol{\epsilon}, c, \boldsymbol{y})$, \textit{i.e.},
    \begin{equation}
\small
    \begin{aligned}
        & \mathcal{L}_{\mathcal{D}}(\boldsymbol{w}, c, \boldsymbol{y}) \le 
               \mathcal{L}_{\mathcal{D}}(\boldsymbol{w}, 0, \boldsymbol{0}) + \mathcal{E}_{l_{y}} +  \mathcal{E}_{l_{c}} \\
        & \le \underset{\Vert \boldsymbol{\epsilon} \Vert \le \rho}{\max} \mathcal{L}_{\mathcal{S}}(\boldsymbol{w} + \boldsymbol{\epsilon}, 0, \boldsymbol{0}) + \mathcal{E}_{l_{gen}}^{'} +  \mathcal{E}_{l_{y}} +  \mathcal{E}_{l_{c}} \\
            &\le \underset{\Vert \epsilon \Vert \le \rho}{\max}  \mathcal{L}_{\mathcal{S}}(\boldsymbol{w}+\boldsymbol{\epsilon}, c, \boldsymbol{y}) + \mathcal{E}_{l_{gen}}^{'} +  \mathcal{E}_{l_{y}} +  \mathcal{E}_{l_{c}} +  \mathcal{E}_{l_{y}}^{'} +  \mathcal{E}_{l_{c}}^{'}.
              \\
    \end{aligned}.
\end{equation}
Besides, by denoting 
\begin{equation}
\small
    \begin{aligned}
        \mathcal{L}_{\mathcal{S}}^{\text{sharp}} \triangleq \underset{\Vert \boldsymbol{\epsilon} \Vert \le \rho}{\max} \mathcal{L}_{\mathcal{S}}(\boldsymbol{w}+\boldsymbol{\epsilon}, c, \boldsymbol{y}) - \mathcal{L}_{\mathcal{S}}(\boldsymbol{w}, c, \boldsymbol{y}),
    \end{aligned}
\end{equation}
and 
\begin{equation}
\small
    \begin{aligned}
        \mathcal{E}_{l_{gen}} \triangleq  \mathcal{E}_{l_{gen}}^{'} +  \mathcal{E}_{l_{y}} +  \mathcal{E}_{l_{c}} +  \mathcal{E}_{l_{y}}^{'} +  \mathcal{E}_{l_{c}}^{'},
    \end{aligned}
\end{equation}
we have proved the theorem.

 From Corollary~\ref{corollary:tangen_point_y} and the above analysis, we have obtained that
    \begin{equation}
    \small
        \begin{aligned}
            \underset{\boldsymbol{y}  \rightarrow \boldsymbol{0}}{\lim} \mathcal{E}_{l_{y}} = 0,  \underset{c \rightarrow 0}{\lim} \mathcal{E}_{l_{c}} = 0,   \underset{\boldsymbol{y}  \rightarrow \boldsymbol{0}}{\lim} \mathcal{E}_{l_{y}}^{'} = 0,  \underset{c \rightarrow 0}{\lim} \mathcal{E}_{l_{c}}^{'} = 0,
        \end{aligned}
    \end{equation}
    and thus  
    \begin{equation}
    \small
        \begin{aligned}
              \underset{\boldsymbol{y} \rightarrow 0, c \rightarrow 0}{\lim} \mathcal{E}_{l_{gen}} = \mathcal{E}_{l_{gen}}^{'}.
        \end{aligned}
    \end{equation}
    
\end{proof}

In the implementation, we set the tangent point $\boldsymbol{y}$ as $\boldsymbol{0}$ that is a special case of $\boldsymbol{y}$, and thus Theorem~\ref{Theorem:generalization_error_bound} also satisfies when $\boldsymbol{y} = \boldsymbol{0}$.
To minimize $\mathcal{L}_{\mathcal{D}}$, we only need to minimize the first two terms in Eq.~\eqref{equation:ge_error_bound}, since $\mathcal{E}_{l_{gen}}^{'}$ is a constant. The first term $\mathcal{L}_{\mathcal{S}}(\boldsymbol{w}, c, \boldsymbol{y})$ is the loss value of HNNs on the training set.
The second term $\mathcal{L}_{\mathcal{S}}^{\text{sharp}}$  captures the sharpness of $\mathcal{L}_{\mathcal{S}}$ at $\boldsymbol{w}$ by measuring how quickly the training
loss can be increased by moving from  $\boldsymbol{w}$ to a nearby parameter value on hyperbolic spaces, 
where $\{\boldsymbol{w} + \boldsymbol{\epsilon}: \Vert \boldsymbol{\epsilon} \Vert\le \rho\}$ is the neighborhood of $\boldsymbol{w}$ on hyperbolic spaces. 
% In the second term $\mathcal{L}_{\mathcal{S}}^{\text{sharp}}$ , $\{\rm{expm}_{\boldsymbol{0}}^{c}(\boldsymbol{w} + \boldsymbol{\epsilon}): \Vert \boldsymbol{\epsilon} \Vert\le \rho\}$ denotes the neighborhood of $\boldsymbol{w}$ on hyperbolic spaces. $\mathcal{L}_{\mathcal{S}}^{\text{sharp}}$ captures the sharpness of $\mathcal{L}_{\mathcal{S}}$ at $\boldsymbol{w}$ by measuring how quickly the training
% loss can be increased by moving from  $\boldsymbol{w}$ to a nearby parameter value on hyperbolic spaces.  
% In this case, we connect the smoothness (\emph{i.e.}, $\mathcal{L}_{\mathcal{S}}^{\text{sharp}}$) of the loss landscape and the generalization ability of HNNs.
In this case, we demonstrate that the smoothness (\emph{i.e.}, $\mathcal{L}_{\mathcal{S}}^{\text{sharp}}$) of the loss landscape affects the generalization of HNNs.
Specifically, models converging to smoother regions (\textit{i.e.}, with smaller values of $\mathcal{L}_{\mathcal{S}}^{\text{sharp}}$) exhibit better generalization, while those converging to sharper regions (i.e., with larger values of $\mathcal{L}_{\mathcal{S}}^{\text{sharp}}$) generalize worse.

Notably, the sharpness term $\mathcal{L}_{\mathcal{S}}^{\text{sharp}}$ is strongly related to the curvatures $c$. Setting inappropriate curvatures to train HNNs may cause the loss landscape around $\boldsymbol{w}$ to become sharp, limiting the generalization of HNNs. In contrast, appropriate curvatures help smooth the loss landscape around $\boldsymbol{w}$, and improve the generalization of HNNs. 
% Therefore, curvatures are important for improving  generalization  of HNNs from the theoretical perspective.

\begin{figure}[htbp]
	\centering
 % \vspace{-0.35cm}
	% \vspace{-0.5em}
	\begin{center}
		\subfloat[Analysis on the task of classification ]{\label{fig:standard_classification} \includegraphics[width=0.47\textwidth]{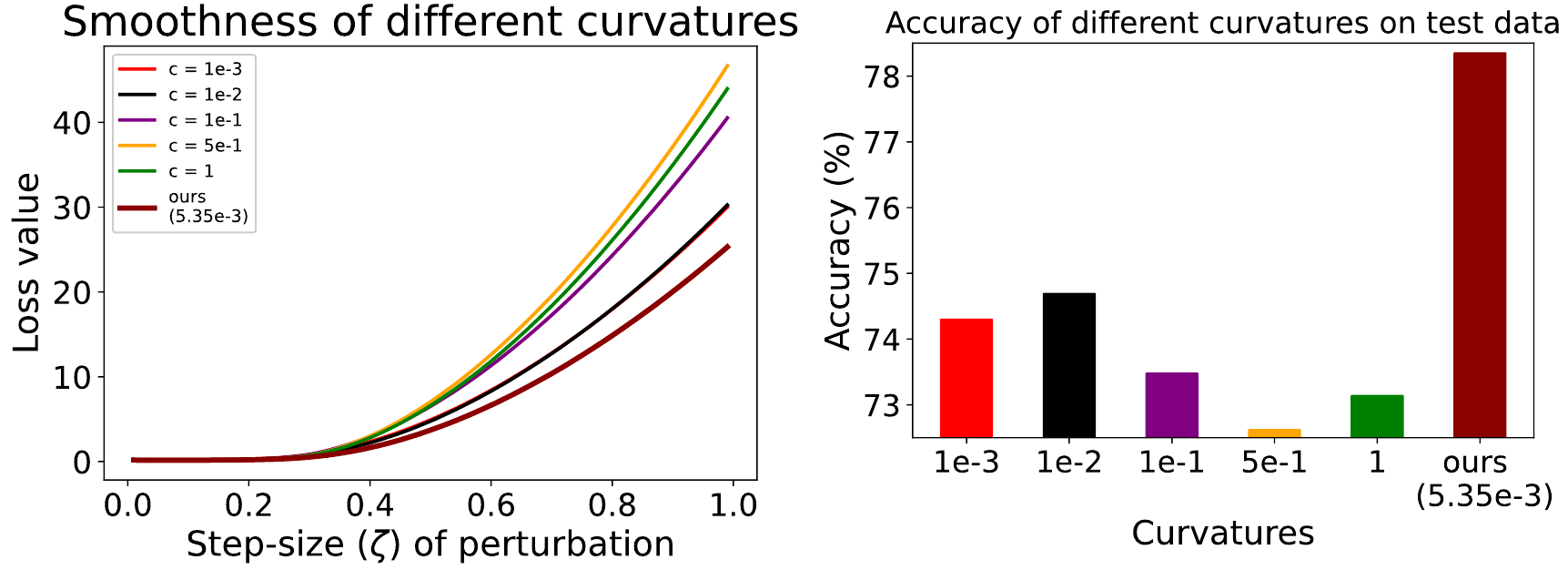}}
  \\
		\subfloat[Analysis on the task of long-tailed classification]{\label{fig:long_tailed_classification} \includegraphics[width=0.47\textwidth]{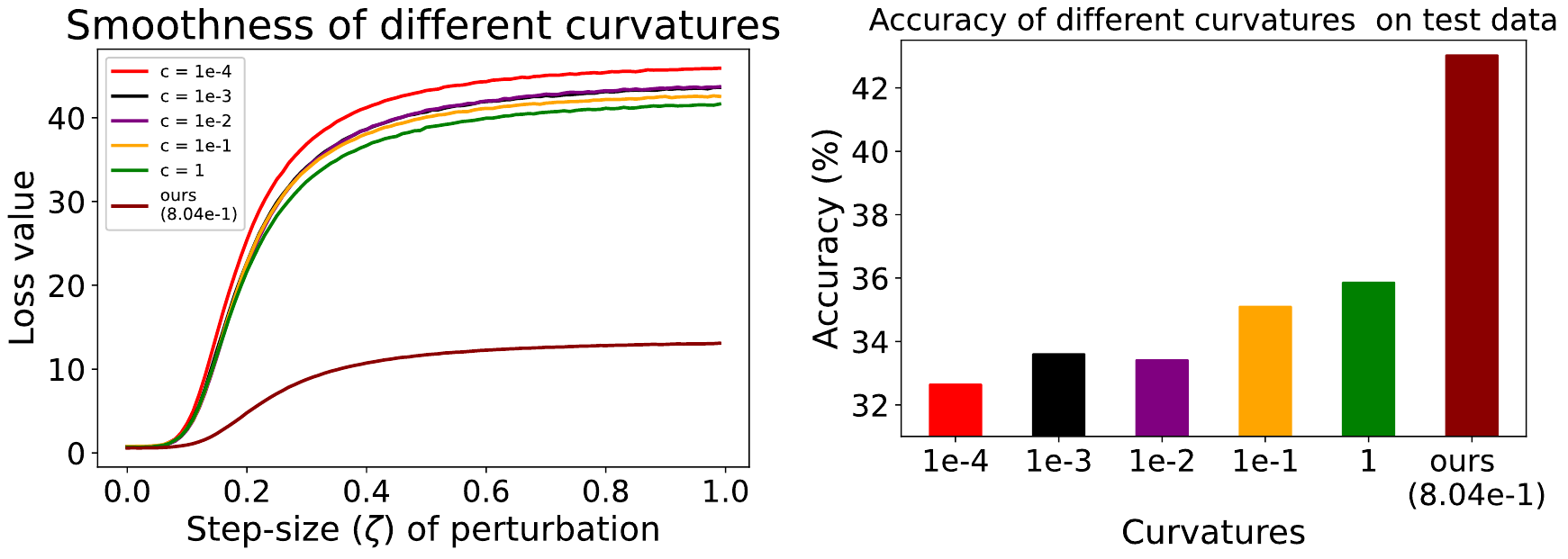}}
		\quad
		% \subfloat{\includegraphics[width=0.44\textwidth]{pdf/legend_unified_end.pdf}}
	\end{center}
 % \vspace{-1.5em}
	\caption{Generalization analysis}
	\label{fig:fig1}
\end{figure}

% \subsubsection{Empirical study}
We further conduct experiments to confirm this point. Specifically, we set multiple different curvatures to train HNNs. After obtaining the trained HNNs, we perturbed parameters of the model as $\Bar{\boldsymbol{w}} = \boldsymbol{w}+ \frac{\zeta \Vert 
\boldsymbol{w} \Vert\boldsymbol{o}}{\Vert \boldsymbol{o} \Vert}$.
% \begin{equation}
% 	\label{equation:perturb_parameter}
% 	\begin{aligned}
% 		\Bar{\boldsymbol{w}} = \boldsymbol{w}+ \frac{\zeta \boldsymbol{Z}}{\Vert \boldsymbol{w} \Vert} ,
% 	\end{aligned}
% \end{equation}
Here $\boldsymbol{o}$ is the direction of perturbation that is the same for all models, and  $\zeta \in (0,1)$ is the step-size of perturbation. The perturbed HNNs are utilized to compute the loss on the training set to analyze the smoothness of HNNs and compute the accuracy on the test set.
As shown in Fig.~\ref{fig:fig1}, 
 HNNs trained with different curvatures exhibit different sharpness and generalization performances.  HNNs trained with $c = 5e-1$ and $c = 1e-4$ converge to the sharpest minima and achieve the worst performance on the test set. On the contrary, HNNs with $c = 1e-2$ and $c = 1$ converge to the smoothest minima and achieve the best performance on the test set among HNNs trained with fixed curvatures.
% As to HNNs trained with the SAM method, models with various curvatures also exhibit various degrees of smoothness.
% These phenomena confirm the significance of curvatures for the generalization in HNNs by affecting the smoothness of the loss landscape.
These phenomena confirm that curvatures are significant for the generalization of HNNs by affecting the smoothness of the loss landscape. 
% These phenomena confirm that curvatures are crucial for the generalization ability of HNNs by affecting the smoothness of the loss landscape.
% the smoothness of the loss landscape. The smooth minima show better generalization ability, compared with the sharp minima.

\section{Method}
The proposed sharpness-aware curvature learning method trains curvatures to improve the generalization of HNNs by smoothing the loss landscape. Concretely, we first design a scope sharpness measure ${\rm SN}(\cdot)$ to measure the sharpness of the local minimum trained with learnable curvatures within a given scope.
Then, we minimize ${\rm SN}(\cdot)$ to train curvatures for smoothing the loss landscape.

The proposed ${\rm SN}(\cdot)$ is not directly affected by the curvature $c$; rather, ${\rm SN}(\cdot)$ is indirectly affected by the curvature $c$ by influencing $\boldsymbol{w}$. Details are in Section 4.1.
As a result, the gradient of ${\rm SN}(\cdot)$ with respect to $c$ is $ \frac{\partial {\rm SN}}{\partial c} = \boldsymbol{0}$.
% \begin{equation}
%     \begin{aligned}
%         \frac{\partial {\rm SN}}{\partial c} = \boldsymbol{0}.
%     \end{aligned}
% \end{equation}
% This implies that the sharpness information cannot propagate to curvature $c$, meaning that the optimized curvature cannot smooth the loss landscape or improve the generalization of HNNs.

To propagate the sharpness information from ${\rm SN}(\cdot)$ to $c$, we view the curvatures as hyper-parameters. Given that optimizing hyper-parameters typically resorts to bi-level optimization~\citep{franceschi2018bilevel}, the minimization process is formulated as a bi-level optimization process,
\begin{equation}
\small
	% \label{equation:bi_level_objective}
	\begin{aligned}
		\min_{c}\mathcal{F}(\boldsymbol{w}^{*}(c),c,\boldsymbol{y}) &\coloneqq \mathcal{L}_{\mathcal{V}}(\boldsymbol{w}^{*}(c),c,\boldsymbol{y})+{\rm SN}(\boldsymbol{w}^{*}(c))\\
		\text{s.t.} \hspace{2mm} \boldsymbol{w}^{*}(c) &= \underset{\boldsymbol{w}}{\arg \min} \mathcal{L}_{\mathcal{S}}(\boldsymbol{w},c,\boldsymbol{y}).
	\end{aligned}
\end{equation}
In implementation, we set $\boldsymbol{y} = 0$ consistently.
\ul{For simplicity and readability, we omit $\boldsymbol{y}$ in $\mathcal{L}(\boldsymbol{w},\boldsymbol{y},c)$ and simplify the denotation of loss function as $\mathcal{L}(\boldsymbol{w},c)$ here and in the following part. } 
Then, the bi-level optimization process is re-denoted as 
\begin{equation}
\small
	\label{equation:bi_level_objective}
	\begin{aligned}
		\min_{c}\mathcal{F}(\boldsymbol{w}^{*}(c),c) &\coloneqq \mathcal{L}_{\mathcal{V}}(\boldsymbol{w}^{*}(c),c)+{\rm SN}(\boldsymbol{w}^{*}(c))\\
		\text{s.t.} \hspace{2mm} \boldsymbol{w}^{*}(c) &= \underset{\boldsymbol{w}}{\arg \min} \mathcal{L}_{\mathcal{S}}(\boldsymbol{w},c).
	\end{aligned}
\end{equation}

% \textbf{Note.} 
% \ul{For simplicity, readability, and distinction, we use $\mathcal{L}(\boldsymbol{w},c)$ to represent  
% $\mathcal{L}(\rm{expm}_{\boldsymbol{0}}^{c}(\boldsymbol{w}))$ here and in the following part.}
%  $\mathcal{L}_{\mathcal{S}}(\cdot)$ and $\mathcal{L}_{\mathcal{V}}(\cdot)$ represent the loss functions on the training set and validation set, respectively, such as the cross-entropy loss and mean square error in the classification task.
This optimization problem can be divided into two sub-problems, \emph{i.e.}, the inner-level and outer-level problems. In the inner-level optimization, HNNs are trained with fixed curvatures, while curvatures are learned in the outer-level optimization.
Then, the gradients of ${\rm SN}(\cdot)$ with respect to $c$ using the chain rule,
\begin{equation}
    \begin{aligned}
        \frac{\partial {\rm SN}(\boldsymbol{w}^{*}(c))}{\partial c} = \frac{\partial {\rm SN}(\boldsymbol{w}^{*}(c))}{\partial \boldsymbol{w}^{*}(c)} \frac{\partial \boldsymbol{w}^{*}(c)}{\partial c}.
    \end{aligned}
\end{equation}
To efficiently compute gradients of curvature in the bi-level optimization, we introduce an implicit differentiation algorithm. 

% Optimizing curvatures incurs the heavy computational load, since computing gradients of curvatures requires unrolling the optimization process of the inner-level process and many complex calculations. Thus we introduce the implicit differentiation algorithm for curvature optimization. 

\subsection{Scope Sharpness Measure}

In this subsection, we first define the proposed scope sharpness measure $\text{SN}(\boldsymbol{w}^{*}(c))$, which measures the sharpness across the neighborhood of the local minimum $\boldsymbol{w}^{\ast}$ trained with the curvature $c$.
Then, we approximate $\text{SN}(\boldsymbol{w}^{*}(c))$  to reduce the time and space complexities by
utilizing the Neumann series, polynomial expansion, and Taylor expansion.

% We propose a scope sharpness measure $\text{SN}(\boldsymbol{w}^{*}(c))$ to measure the sharpness across the neighborhood of the local minimum $\boldsymbol{w}^{\ast}$ trained with the curvature $c$. 
Mathematically, the proposed scope sharpness measure $\text{SN}(\boldsymbol{w}^{\ast})$ is defined as 
\begin{equation}
	\label{equation:our_sharpness_measure}
	\begin{aligned}
		\text{SN}(\boldsymbol{w}^{\ast}(c)) = \underset{\Vert \boldsymbol{\epsilon} \Vert \le \rho}{\max} \text{sn}(\boldsymbol{w}^{\ast}(c)+\boldsymbol{\epsilon}),
	\end{aligned}
\end{equation}
where $\text{sn}(\boldsymbol{w}^{\ast}(c))$ is the reparametrization-invariant sharpness measure~\citep{jang2022reparametrization} to capture the sharpness of the local minima. $\text{sn}(\boldsymbol{w}^{\ast}(c))$ is computed as 
\begin{equation}
	\label{equation:sharpness_measure}
 \small
	\begin{aligned}
		&\text{sn}(\boldsymbol{w}^{\ast}(c)) =  \nabla_{\boldsymbol{w}} \mathcal{L}_{\mathcal{S}}(\boldsymbol{w}^{\ast}(c), c)^{\top}
  \Big(\nabla_{\boldsymbol{w}} \mathcal{L}_{\mathcal{S}}(\boldsymbol{w}^{\ast}(c), c)
  \\
  &\times \nabla_{\boldsymbol{w}} \mathcal{L}_{\mathcal{S}}(\boldsymbol{w}^{\ast}(c), c)^{\top}\Big)^{-1} \nabla_{\boldsymbol{w}} \mathcal{L}_{\mathcal{S}}(\boldsymbol{w}^{\ast}(c), c), \footnotemark
	\end{aligned}
\end{equation}
 \footnotetext{$\times$ denotes standard multiplication of matrices, vectors or scalars, which is added at the line break for clarity and is omitted elsewhere.} where $\nabla$ represents gradients, and $\nabla_{\boldsymbol{w}} \mathcal{L}_{\mathcal{S}}(\boldsymbol{w}, c)  = \frac{\partial \mathcal{L}_{\mathcal{S}}(\boldsymbol{w}, c)}{\partial \boldsymbol{w}}$.
We note that $\rm{sn}(\cdot)$ in Eq.~\eqref{equation:sharpness_measure} contains the inverse of the Fisher matrix $\Big(\nabla_{\boldsymbol{w}} \mathcal{L}_{\mathcal{S}}(\boldsymbol{w}^{\ast}(c), c)\nabla_{\boldsymbol{w}} \mathcal{L}_{\mathcal{S}}(\boldsymbol{w}^{\ast}(c), c)^{\top}\Big)$.
Assuming that the number of parameters $\boldsymbol{w}$ as $d$, the time complexity of the Fisher matrix is $\mathcal{O}(d^{3})$, and the space complexity is $\mathcal{O}(d^{2})$. It is intractable to handle such a notoriously time and memory-consuming operation for HNNs.
In order to address this issue, following existing work~\citep{lorraine2020optimizing}, we utilize the first $K$ terms of Neumann series to approximate this term,
\begin{equation}
	\label{equation:Neumann_series}
 \small
	\begin{aligned}
		&\Big(\nabla_{\boldsymbol{w}} \mathcal{L}_{\mathcal{S}}(\boldsymbol{w}^{\ast}(c), c)\nabla_{\boldsymbol{w}} \mathcal{L}_{\mathcal{S}}(\boldsymbol{w}^{\ast}(c), c)^{\top} \Big)^{-1} \approx \\
  &\sum_{i=0}^{K}\Big(\boldsymbol{I} - \nabla_{\boldsymbol{w}} \mathcal{L}_{\mathcal{S}}(\boldsymbol{w}^{\ast}(c), c)\nabla_{\boldsymbol{w}} \mathcal{L}_{\mathcal{S}}(\boldsymbol{w}^{\ast}(c), c)^{\top}\Big)^{i}.
	\end{aligned}
\end{equation}
By substituting Eq.~\eqref{equation:Neumann_series} into Eq.~\eqref{equation:sharpness_measure}, $\text{sn}(\boldsymbol{w}^{\ast}(c))$ can be approximated as 
\begin{equation}
% \scriptsize
\small
\label{equation:sn_original}
	\begin{aligned}
		&\text{sn}(\boldsymbol{w}^{\ast}(c)) \approx  \hat{\text{sn}}(\boldsymbol{w}^{\ast}(c)) \\ &\triangleq
		\sum_{i=0}^{K}\nabla_{\boldsymbol{w}} \mathcal{L}_{\mathcal{S}}(\boldsymbol{w}^{\ast}(c), c)^{\top}\Big(\boldsymbol{I} 
   -\nabla_{\boldsymbol{w}} \mathcal{L}_{\mathcal{S}}(\boldsymbol{w}^{\ast}(c), c) \\
   & \times
    \nabla_{\boldsymbol{w}} \mathcal{L}_{\mathcal{S}}(\boldsymbol{w}^{\ast}(c), c)^{\top}\Big)^{i}
		\nabla_{\boldsymbol{w}} \mathcal{L}_{\mathcal{S}}(\boldsymbol{w}^{\ast}(c), c).
	\end{aligned}
\end{equation}
Eq.~\eqref{equation:sn_original} can be further simplified, as shown in Proposition~\ref{proposition:sn_derivation}.
\begin{proposition}
    \label{proposition:sn_derivation}
   $\hat{\rm{sn}}(\boldsymbol{w}^{\ast}(c))$ in Eq.~\eqref{equation:sn_original} has an equivalent expression, which is computed as 
\begin{equation}
	\label{equation:sharpness_measure_approx}
	\begin{aligned}
		 \hat{\rm{sn}}(\boldsymbol{w}^{\ast}(c)) = 1-\Big(1-\Big\|
   \nabla_{\boldsymbol{w}} \mathcal{L}_{\mathcal{S}}(\boldsymbol{w}^{\ast}(c), c)
   \Big\|^{2}\Big)^{K}.
	\end{aligned}
\end{equation}
\end{proposition}

\begin{proof}

In this proof, we expand the polynomial in Eq.~\eqref{equation:sn_original} to avoid the computation of $ \nabla_{\boldsymbol{w}} \mathcal{L}_{\mathcal{S}}(\boldsymbol{w}^{\ast}(c), c)
 \nabla_{\boldsymbol{w}} \mathcal{L}_{\mathcal{S}}(\boldsymbol{w}^{\ast}(c), c)^{\top}$ whose time and space complexities are non-trivial.

For simplicity, we omit $\ast$ and $(c)$ from $\boldsymbol{w}^{\ast}(c)$ in the proof.
Intuitively, we substitute $i=0,1,2,...$ into the original  $\hat{\rm{sn}}(\boldsymbol{w})$, and observe that
   \begin{equation}
   \small
    \begin{aligned}
        &i=0 :\\ 
        &( \nabla_{\boldsymbol{w}} \mathcal{L}_{\mathcal{S}}(\boldsymbol{w}, c))^{\top} \left[ \boldsymbol{I} - ( \nabla_{\boldsymbol{w}} \mathcal{L}_{\mathcal{S}}(\boldsymbol{w}, c))( \nabla_{\boldsymbol{w}} \mathcal{L}_{\mathcal{S}}(\boldsymbol{w}, c))^{\top} \right]^{0} \\
        & \times
        ( \nabla_{\boldsymbol{w}} \mathcal{L}_{\mathcal{S}}(\boldsymbol{w}, c)) = ( \nabla_{\boldsymbol{w}} \mathcal{L}_{\mathcal{S}}(\boldsymbol{w}, c))^{\top}  \nabla_{\boldsymbol{w}} \mathcal{L}_{\mathcal{S}}(\boldsymbol{w}, c).
    \end{aligned}
\end{equation}
Beside, when $i = k$, we have that
\begin{equation}
\footnotesize
\label{equation:n_expand}
    \begin{aligned}
        &\left[ \boldsymbol{I} - ( \nabla_{\boldsymbol{w}} \mathcal{L}_{\mathcal{S}}(\boldsymbol{w}, c))( \nabla_{\boldsymbol{w}} \mathcal{L}_{\mathcal{S}}(\boldsymbol{w}, c))^{\top} \right]^{K} = \left(
        \begin{array}{c}
             K  \\
             0 
        \end{array}
        \right) + ... +
        \\
       & \left(
        \begin{array}{c}
             K  \\
             K-2 
        \end{array}
        \right)(-1)^{K-2}\left[ ( \nabla_{\boldsymbol{w}} \mathcal{L}_{\mathcal{S}}(\boldsymbol{w}, c))( \nabla_{\boldsymbol{w}} \mathcal{L}_{\mathcal{S}}(\boldsymbol{w}, c))^{\top}\right]^{K-2} +\\
        &\left(
        \begin{array}{c}
             K  \\
             K-1 
        \end{array}
        \right)(-1)^{K-1}\left[ ( \nabla_{\boldsymbol{w}} \mathcal{L}_{\mathcal{S}}(\boldsymbol{w}, c))( \nabla_{\boldsymbol{w}} \mathcal{L}_{\mathcal{S}}(\boldsymbol{w}, c))^{\top}\right]^{K-1} +\\
       & \left(
        \begin{array}{c}
             K  \\
             K 
        \end{array}
        \right)(-1)^{K}\left[ ( \nabla_{\boldsymbol{w}} \mathcal{L}_{\mathcal{S}}(\boldsymbol{w}, c))( \nabla_{\boldsymbol{w}} \mathcal{L}_{\mathcal{S}}(\boldsymbol{w}, c))^{\top}\right]^{K}.
    \end{aligned}
\end{equation}
Let Eq.~\eqref{equation:n_expand} multiple the vector $\nabla_{\boldsymbol{w}} \mathcal{L}_{\mathcal{S}}(\boldsymbol{w}, c)$ on the left and multiple the vector $( \nabla_{\boldsymbol{w}} \mathcal{L}_{\mathcal{S}}(\boldsymbol{w}, c))^{\top}$. By denoting $( \nabla_{\boldsymbol{w}} \mathcal{L}_{\mathcal{S}}(\boldsymbol{w}, c) )^{\top} ( \nabla_{\boldsymbol{w}} \mathcal{L}_{\mathcal{S}}(\boldsymbol{w}, c)) $ as $\mathcal{A}$,  we have
\begin{equation}
\small
\label{equation:measure_derive_final}
    \begin{aligned}
        &( \nabla_{\boldsymbol{w}} \mathcal{L}_{\mathcal{S}}(\boldsymbol{w}, c) )^{\top} 
        \left[ \boldsymbol{I} - ( \nabla_{\boldsymbol{w}} \mathcal{L}_{\mathcal{S}}(\boldsymbol{w}, c))\right.\\
        &\left.\times
        ( \nabla_{\boldsymbol{w}} \mathcal{L}_{\mathcal{S}}(\boldsymbol{w}, c))^{\top} \right]^{K} ( \nabla_{\boldsymbol{w}} \mathcal{L}_{\mathcal{S}}(\boldsymbol{w}, c)) \\
        % &= \left(
        % \begin{array}{c}
        %      K  \\
        %      0 
        % \end{array}
        % \right)\mathcal{A}  + ... +
        % \left(
        % \begin{array}{c}
        %      K  \\
        %      K-2 
        % \end{array}
        % \right)(-1)^{K-2} \mathcal{A}^{K-1} \\
        % &+  \left(
        % \begin{array}{c}
        %      K  \\
        %      K-1 
        % \end{array}
        % \right)(-1)^{K-1} \mathcal{A}^{K}
        % +  \left(
        % \begin{array}{c}
        %      K  \\
        %      K 
        % \end{array}
        % \right)(-1)^{K} \mathcal{A}^{K+1} \\
        &= \mathcal{A}\left[
       \left(
        \begin{array}{c}
             K  \\
             0 
        \end{array}
       \right) +...+
        \left(
        \begin{array}{c}
             K  \\
             K-2 
        \end{array}
        \right)(-1)^{K-2}\mathcal{A}^{K-2}\right. \\ 
       &\left.+ 
        \left(
        \begin{array}{c}
             K  \\
             K-1 
        \end{array}
        \right)(-1)^{K-1}\mathcal{A}^{K-1}
        +
        \left(
        \begin{array}{c}
             K  \\
             K
        \end{array}
        \right)(-1)^{K}\mathcal{A}^{K}
        \right]\\
        &= \mathcal{A}(1-\mathcal{A})^{K}
    \end{aligned}.
\end{equation}
Then $\text{sn}(\boldsymbol{w}^{\ast}(c))$ is approximated as
\begin{equation}
    \small
	\begin{aligned}
		\text{sn}(\boldsymbol{w}^{\ast}(c)) &\approx \hat{\text{sn}}(\boldsymbol{w}^{\ast}(c))
        \\
        &\triangleq
	 1-\left(1- \Vert  \nabla_{\boldsymbol{w}} \mathcal{L}_{\mathcal{S}}(\boldsymbol{w}^{\ast}(c), c) \Vert^{2}  \right)^{K+1}.
	\end{aligned}
\end{equation}

%     \begin{equation}
%     \small
% 	\begin{aligned}
% 		&\text{sn}(\boldsymbol{w}) \approx \hat{\text{sn}}(\boldsymbol{w}) \\
%     &\triangleq
% 		\sum_{i=0}^{K}( \nabla_{\boldsymbol{w}} \mathcal{L}_{\mathcal{S}}(\boldsymbol{w}, c))^{\top}\Big(\boldsymbol{I} - 
%    \nabla_{\boldsymbol{w}} \mathcal{L}_{\mathcal{S}}(\boldsymbol{w}, c)
%          \nabla_{\boldsymbol{w}} \mathcal{L}_{\mathcal{S}}(\boldsymbol{w}, c)^{\top}\Big)^{i}
% 		( \nabla_{\boldsymbol{w}} \mathcal{L}_{\mathcal{S}}(\boldsymbol{w}, c)) \\
%   % &=\sum_{i=0}^{K}
%   % \mathcal{A}(1-\mathcal{A})^{i} = 1-(1-\mathcal{A})^{K}\\
%   % &=1-\left(1- ( \nabla_{\boldsymbol{w}} \mathcal{L}_{\mathcal{S}}(\boldsymbol{w}, c) )^{\top} ( \nabla_{\boldsymbol{w}} \mathcal{L}_{\mathcal{S}}(\boldsymbol{w}, c)) \right)^{K}
%   % \\
%   &= 1-\left(1- \Vert  \nabla_{\boldsymbol{w}} \mathcal{L}_{\mathcal{S}}(\boldsymbol{w}, c) \Vert^{2}  \right)^{K}.
% 	\end{aligned}
% \end{equation}

\end{proof}

Notably, the time and space complexities of $\hat{\text{sn}}(\boldsymbol{w}^{\ast}(c))$ are only $\mathcal{O}(d)$, which significantly reduce the required complexity and make it feasible for large-scale HNNs.
We substitute Eq.~\eqref{equation:sharpness_measure_approx} into Eq.~\eqref{equation:our_sharpness_measure}, and compute the scope sharpness measure as 
\begin{equation}
	\label{equation:our_sharpness_measure_approx}
	\small
	\begin{aligned}
		&\text{SN}(\boldsymbol{w}^{\ast}(c)) \approx   \underset{\Vert \boldsymbol{\epsilon} \Vert \le \rho}{\max} \hat{\text{sn}}(\boldsymbol{w}^{\ast}(c) + \boldsymbol{\epsilon}) \\
            &=
		\underset{\Vert \boldsymbol{\epsilon} \Vert \le \rho}{\max} 1-\Big(1-\Big\|
        \nabla_{\boldsymbol{w}} \mathcal{L}_{\mathcal{S}}( \boldsymbol{w}^{\ast}(c)
        +\boldsymbol{\epsilon}, c)\Big\|^{2}\Big)^{K+1}.
	\end{aligned}
\end{equation}
% \begin{equation}
% 	\label{equation:our_sharpness_measure_approx}
% 	% \small
% 	\begin{aligned}
% 		\text{SN}(\boldsymbol{w}^{\ast}(c)) &\approx   \underset{\Vert \boldsymbol{\epsilon} \Vert \le \rho}{\max} \hat{\text{sn}}(\boldsymbol{w}^{\ast}(c) + \boldsymbol{\epsilon}) \\
%             &=
% 		\underset{\Vert \boldsymbol{\epsilon} \Vert \le \rho}{\max} 1-\Big(1-\Big\|\frac{\partial \mathcal{L}_{\mathcal{S}}}{\partial \boldsymbol{w}}\Big|_{\boldsymbol{w}^{\ast}(c)+\boldsymbol{\epsilon}}\Big\|^{2}\Big)^{K}.
% 	\end{aligned}
% \end{equation}
The maximization problem in Eq.~\eqref{equation:our_sharpness_measure_approx} is still tedious; hence, we propose an approximation for Eq.~\eqref{equation:our_sharpness_measure_approx}. By the first-order Taylor expansion, Eq.~\eqref{equation:our_sharpness_measure_approx} can be approximated as
\begin{equation}
	\small
 \label{equation:our_sharpness_measure_approx_approx}
	\begin{aligned}
		\text{SN}(\boldsymbol{w}^{\ast}(c))
		&\approx
		\underset{\Vert \boldsymbol{\epsilon} \Vert \le \rho}{ \max} \quad \hat{\text{sn}}(\boldsymbol{w}^{*}({c} )) +  \boldsymbol{\epsilon}^{\top}  \frac{\partial \hat{\text{sn}}(\boldsymbol{w}^{*}({c} ))}{\partial \boldsymbol{w}^{*}}\\
		&= \underset{\Vert \boldsymbol{\epsilon} \Vert \le \rho}{ \max} \quad \boldsymbol{\epsilon}^{\top}   \frac{\partial \hat{\text{sn}}(\boldsymbol{w}^{*}({c} ))}{\partial \boldsymbol{w}^{*}}.
	\end{aligned}
\end{equation}
Eq.~\eqref{equation:our_sharpness_measure_approx_approx} is a classical dual norm problem, which can be solved as
\begin{equation}
\small
	\label{equation:epsilon_approx}
	\begin{aligned}
		\hat{\boldsymbol{\epsilon}} = \rho  \frac{\partial \hat{\text{sn}}(\boldsymbol{w}^{\ast}(c))}{\partial \boldsymbol{w}^{\ast}(c)} / \Big\|  \frac{\partial \hat{\text{sn}}(\boldsymbol{w}^{\ast}(c))}{\partial \boldsymbol{w}^{\ast}(c)} \Big\|.
	\end{aligned}
\end{equation}
By substituting Eq.~\eqref{equation:epsilon_approx} into  Eq.~\eqref{equation:our_sharpness_measure_approx}, the scope sharpness measure  is finally approximated as
% \begin{equation}
% \small
% \label{equation:final_approx_sn}
% 	\begin{aligned}
% 		\text{SN}(\boldsymbol{w}^{\ast}(c)) \approx  \hat{\text{sn}}(\hat{\boldsymbol{w}}(c)) =
% 		1-\Big(1-\Big\|\mathcal{L}_{\mathcal{S}}(\hat{\boldsymbol{w}}(c)) \Big\|^{2}\Big)^{K},
% 	\end{aligned}
% \end{equation}
\begin{equation}
\small
\label{equation:final_approx_sn}
	\begin{aligned}
		\text{SN}(\boldsymbol{w}^{\ast}(c)) &\approx  \hat{\text{sn}}(\hat{\boldsymbol{w}}(c)) 
        \\
        &=
		1-\Big(1-\Big\| 
        \nabla_{\boldsymbol{w}} \mathcal{L}_{\mathcal{S}}(\hat{\boldsymbol{w}}(c), c) 
        \Big\|^{2}\Big)^{K+1},
	\end{aligned}
\end{equation}
where $\hat{\boldsymbol{w}}(c) = \boldsymbol{w}^{\ast}(c)+\hat{\boldsymbol{\epsilon}}$.
By minimizing the sharpness measure to train curvatures, the obtained curvatures can smooth the loss landscape of the neighborhood of local minima and further improve the generalization.

% We substitute Eq.~\eqref{equation:epsilon_approx} into Eq.~\eqref{equation:our_sharpness_measure_approx}, and have

\subsection{Implicit Differentiation for Curvature Optimization }
Based on the approximated scope sharpness measure in Eq.~\eqref{equation:final_approx_sn}, the bi-level optimization objective in Eq.~\eqref{equation:bi_level_objective} can be rewritten as
\begin{equation}
\small
	\label{equation:objective_approx}
	\begin{aligned}
		&\min_{c}\mathcal{F}(\boldsymbol{w}^{*}(c),c) := \mathcal{L}_{\mathcal{V}}(\boldsymbol{w}^{*}(c),c)+\hat{\text{sn}}(\hat{\boldsymbol{w}}(c))\\
		&\text{s.t.} \boldsymbol{w}^{*}(c) = \underset{\boldsymbol{w}}{\arg \min} \mathcal{L}_{\mathcal{S}}(\boldsymbol{w},c).
	\end{aligned}
\end{equation}

In this subsection, we analyze the challenges of solving the bi-level optimization problem in Eq.~\eqref{equation:objective_approx}. Then, we propose an implicit differentiation method to solve the bi-level optimization problem efficiently.

In the inner-level optimization of Eq.~\eqref{equation:objective_approx}, we aim to obtain the optimal parameters $\boldsymbol{w}^{\ast}$ by minimizing the objective $\mathcal{L}_{\mathcal{S}}(\cdot)$ with fixed curvatures. Following SAM~\citep{foret2020sharpness},  $\boldsymbol{w}$ is updated as
\begin{equation}
	% \footnotesize
    \small
	\label{equation:inner_level _update}
	\begin{aligned}
		\boldsymbol{w} &\leftarrow \boldsymbol{w} - \eta  \nabla_{\boldsymbol{w}} \mathcal{L}_{\mathcal{S}}(\boldsymbol{w}^{'}, c),\\
  \text{where } \boldsymbol{w}^{'} = \boldsymbol{w} &+ \hat{\rho}   \nabla_{\boldsymbol{w}} \mathcal{L}_{\mathcal{S}}(\boldsymbol{w}, c) / \Big \|
 \nabla_{\boldsymbol{w}} \mathcal{L}_{\mathcal{S}}(\boldsymbol{w}, c) \Big \|\;.
	\end{aligned}
\end{equation}
Here, $\eta$ is the step-size, and $\hat{\rho}$ is the perturbation radius. 

 We update curvatures by minimizing the objective $\mathcal{F}(\cdot)$ of  Eq.~\eqref{equation:objective_approx}, in the outer-level optimization problem. By using the gradient descent method, $c$ is updated as
\begin{equation}
\small
	\label{equation:curvature_update}
	\begin{aligned}
		c \leftarrow c - \eta_{c} \frac{d \mathcal{F}(\boldsymbol{w}^{*}(c),c)}{d c},
	\end{aligned}
\end{equation}
where $\eta_{c}$ is the step-size for optimizing curvatures.
By the chain rule, the gradient of $c$ is calculated as 
% \begin{equation}
% 	\small
% 	\label{equation:gradient_c_first}
% 	\begin{aligned}
% 		\frac{d \mathcal{F}(\boldsymbol{w}^{\ast}(c), c)}{d c}  = \frac{\partial \mathcal{F}(\boldsymbol{w}^{\ast}(c), c)}{\partial c} + \frac{\partial \mathcal{F}(\boldsymbol{w}^{\ast}(c), c)}{\partial \boldsymbol{w}^{\ast}} \frac{\partial \boldsymbol{w}^{\ast}(c)}{\partial c},
% 	\end{aligned}
% \end{equation}
\begin{equation}
	\small
	\label{equation:gradient_c_first}
	\begin{aligned}
		\frac{d \mathcal{F}(\boldsymbol{w}^{\ast}(c), c)}{d c}  &= \nabla_{c}  \mathcal{F}(\boldsymbol{w}^{\ast}(c), c)
        \\ &+ \nabla_{\boldsymbol{w}} \mathcal{F}(\boldsymbol{w}^{\ast}(c), c) \nabla_{c} \boldsymbol{w}^{\ast}(c),
	\end{aligned}
\end{equation}
% where $\nabla_{c}  \mathcal{F}(\boldsymbol{w}, c) = \frac{\partial  \mathcal{F}(\boldsymbol{w}, c)}{\partial c}$ and  $\nabla_{\boldsymbol{w}}  \mathcal{F}(\boldsymbol{w}, c) = \frac{\partial  \mathcal{F}(\boldsymbol{w}, c)}{\partial \boldsymbol{w}}$\
where $\nabla_{c} \boldsymbol{w}^{\ast}(c) = \frac{\partial \boldsymbol{w}^{\ast}(c)}{\partial c}$.  
$ \nabla_{c}  \mathcal{F}(\boldsymbol{w}^{\ast}(c), c) $ can be computed easily via automatic differentiation technique. Then we focus on $ \nabla_{\boldsymbol{w}} \mathcal{F}(\boldsymbol{w}^{\ast}(c), c) $ and $\nabla_{c} \boldsymbol{w}^{\ast}(c)$ in Eq.~\eqref{equation:gradient_c_first}.
By the chain rule, $ \nabla_{\boldsymbol{w}} \mathcal{F}(\boldsymbol{w}^{\ast}(c), c)$ is expanded as
\begin{equation}
	\small
	\label{equation:gradient_F_W}
	\begin{aligned}
	& 	\nabla_{\boldsymbol{w}} \mathcal{F}(\boldsymbol{w}^{\ast}(c), c)= 
            \nabla_{\boldsymbol{w}} \mathcal{L}_{\mathcal{V}}(\boldsymbol{w}^{*}, c) \\
            &+
        2K\Big(1-\Vert \nabla_{\boldsymbol{w}} \mathcal{L}_{\mathcal{S}}(\hat{\boldsymbol{w}}, c) \Vert\Big)^{K-1} 
      \nabla_{\boldsymbol{w}} \mathcal{L}_{\mathcal{S}}(\hat{\boldsymbol{w}}, c) 
         \nabla^{2}_{\boldsymbol{w}} \mathcal{L}_{\mathcal{S}}(\hat{\boldsymbol{w}}, c),
	\end{aligned}
\end{equation}
where $ \nabla^{2}_{\boldsymbol{w}} \mathcal{L}_{\mathcal{S}}(\hat{\boldsymbol{w}}, c) = \frac{\partial^{2} \mathcal{L}_{\mathcal{S}}(\hat{\boldsymbol{w}}, c) }{ \partial \hat{\boldsymbol{w}}^{2} }$ denotes the Hessian matrices.

 Suppose that $\boldsymbol{w}^{\ast}(c)$ is obtained by inner-level optimization for $T$ steps, 
 $\nabla_{c} \boldsymbol{w}^{\ast}(c)$ can be expanded as 
\begin{equation}
\small
\label{equation:w_c}
\begin{aligned}
      \nabla_{c} \boldsymbol{w}^{\ast}(c) &= -\sum_{j\le T}\Big(\prod_{k<j}\boldsymbol{I}
      - \nabla^{2}_{\boldsymbol{w}} \mathcal{L}_{\mathcal{S}}(\boldsymbol{w}^{(T-k)}, c)\Big)
      \\
      & \times \nabla^{2}_{\boldsymbol{w}c}\mathcal{L}_{\mathcal{S}}(\boldsymbol{w}^{(T-j)}, c),
\end{aligned}
\end{equation}
% \begin{equation}
% \resizebox{0.47\textwidth}{!}{$
% \begin{aligned}
%  \label{equation:w_c}
%   \nabla_{c} \boldsymbol{w}^{\ast}(c) = -\sum_{j\le T}\Big(\prod_{k<j}\boldsymbol{I} - \nabla^{2} \mathcal{L}_{\mathcal{S}}(\boldsymbol{w}^{(T-k)})\Big)\nabla^{2}_{\boldsymbol{w}c}\mathcal{L}_{\mathcal{S}}(\boldsymbol{w}^{(T-j)}),
% \end{aligned}
% $}
% \end{equation}
 where $\nabla^{2}_{\boldsymbol{w}c}\mathcal{L}_{\mathcal{S}}(\boldsymbol{w}, c) = \frac{\partial^{2}\mathcal{L}_{\mathcal{S}}(\boldsymbol{w},c)}{\partial \boldsymbol{w}\partial c}$.
Computing Eq.~\eqref{equation:w_c} needs to unroll the inner-level optimization and compute the product of complex Hessian matrices $\nabla^{2}_{\boldsymbol{w}} \mathcal{L}_{\mathcal{S}}(\boldsymbol{w}^{(T-k)}, c)$, which requires massive time and memory consumption.

Motivated by the work of~\citep{lorraine2020optimizing}, we introduce an implicit differentiation method~\citep{lorraine2020optimizing} to compute  $\nabla_{c} \boldsymbol{w}^{\ast}(c)$  efficiently,
\begin{equation}
\small
	\label{equation:implicit_differentiation_part}
	\begin{aligned}
		\nabla_{c} \boldsymbol{w}^{\ast}(c) = -\Big(\nabla^{2}_{\boldsymbol{w}} \mathcal{L}_{\mathcal{S}}(\boldsymbol{w}^{\ast}, c)\Big)^{-1}  
       \nabla_{\boldsymbol{w}c}^{2} \mathcal{L}_{\mathcal{S}}(\boldsymbol{w}^{\ast}, c).
	\end{aligned}
\end{equation}
% By substitute Eq.~\eqref{equation:implicit_differentiation_part} into Eq.~\eqref{equation:gradient_c_first}, the gradient of $c$ is computed as 
%  \begin{equation}
% \small
% \label{equation:gradient_c_second}
%     \begin{aligned}
%          \frac{d \mathcal{F}(\boldsymbol{w}^{\ast}(c), c)}{d c}  &= \frac{\partial \mathcal{F}(\boldsymbol{w}^{\ast}(c), c)}{\partial c}  \\
%          &+\frac{\partial \mathcal{F}(\boldsymbol{w}^{\ast}(c), c)}{\partial \boldsymbol{w}^{\ast}} (\frac{\partial^{2}\mathcal{L}_{\mathcal{S}}(\boldsymbol{w}^{\ast})}{\partial \boldsymbol{w}^{*} \partial {\boldsymbol{w}^{*}}^{\top}})^{-1} \nabla^{2}_{\boldsymbol{w}c} \mathcal{L}_{\mathcal{S}}(\boldsymbol{w}^{*}, c).
%     \end{aligned}
% \end{equation}
By substituting Eq.~\eqref{equation:implicit_differentiation_part} and Eq.~\eqref{equation:gradient_F_W} into Eq.~\eqref{equation:gradient_c_first}, the second term in Eq.~\eqref{equation:gradient_c_first} is computed as the sum of two terms, \textit{i.e.},
\begin{equation}
	\label{equation:gradient_F_W_1_2}
	\begin{aligned}
		\nabla_{\boldsymbol{w}} \mathcal{F}(\boldsymbol{w}^{\ast}(c), c) \nabla_{c} \boldsymbol{w}^{\ast}(c) = \boldsymbol{U}_{1} + \boldsymbol{U}_{2},
	\end{aligned}
\end{equation}
% \begin{equation}
% 	\label{equation:gradient_F_W_1_2}
% 	\begin{aligned}
% 		\frac{\partial \mathcal{F}(\boldsymbol{w}^{*}(c),c)}{\partial \boldsymbol{w}^{*}} \frac{\partial \boldsymbol{w}^{\ast}(c)}{\partial c} = \boldsymbol{U}_{1} + \boldsymbol{U}_{2},
% 	\end{aligned}
% \end{equation}
where  
\begin{equation}
	\small
	\label{equation:gradient_G_1}
	\begin{aligned}
		\boldsymbol{U}_{1} \triangleq -
  \nabla_{\boldsymbol{w}} \mathcal{L}_{\mathcal{V}}( \boldsymbol{w}^{\ast}, c)
 \Big(\nabla^{2}_{\boldsymbol{w}} \mathcal{L}_{\mathcal{S}}(\boldsymbol{w}^{\ast}, c)\Big)^{-1}
  \nabla^{2}_{\boldsymbol{w}c} \mathcal{L}_{\mathcal{S}}(\boldsymbol{w}^{\ast}, c),
	\end{aligned}
\end{equation}
% \begin{equation}
% 	% \small
% 	\label{equation:gradient_G_1}
% 	\begin{aligned}
% 		\boldsymbol{U}_{1} \triangleq -\frac{\partial \mathcal{L}_{\mathcal{V}}}{\partial \boldsymbol{w}^{*}} \Big(\frac{\partial^{2}\mathcal{L}_{\mathcal{S}}}{\partial \boldsymbol{w}^{*} \partial {\boldsymbol{w}^{*}}^{\top}}\Big)^{-1}\frac{\partial^{2}\mathcal{L}_{\mathcal{S}}}{\partial \boldsymbol{w}^{*} \partial c^{\top}},
% 	\end{aligned}
% \end{equation}
and 
\begin{equation}
\small
	\label{equation:gradient_G_2}
	\begin{aligned}
		&\boldsymbol{U}_{2} \triangleq  -2K\Big(1-\Vert  \nabla_{\boldsymbol{w}} \mathcal{L}_{\mathcal{S}}(\hat{\boldsymbol{w}}, c) \Vert\Big)^{K-1} 
         \nabla_{\boldsymbol{w}} \mathcal{L}_{\mathcal{S}}(\hat{\boldsymbol{w}}, c)
             \\
		&\times
        \nabla^{2}_{\boldsymbol{w}} \mathcal{L}_{\mathcal{S}}(\hat{\boldsymbol{w}}, c)
        \Big(
                 \nabla^{2}_{\boldsymbol{w}} \mathcal{L}_{\mathcal{S}}(\boldsymbol{w}^{\ast}, c)\Big)^{-1} 
		     \nabla^{2}_{\boldsymbol{w}c} \mathcal{L}_{\mathcal{S}}(\boldsymbol{w}^{\ast}, c).
	\end{aligned}
\end{equation}
% \begin{equation}
% 	\label{equation:gradient_G_2}
% 	\begin{aligned}
% 		\boldsymbol{U}_{2} \triangleq & 2K\Big(1-\Vert \frac{\partial \mathcal{L}_{\mathcal{S}}}{\partial \hat{\boldsymbol{w}}} \Vert\Big)^{K-1} \frac{\partial \mathcal{L}_{\mathcal{S}}}{\partial \hat{\boldsymbol{w}}}  \frac{\partial^{2} \mathcal{L}_{\mathcal{S}}}{\partial \hat{\boldsymbol{w}}\partial \hat{\boldsymbol{w}}^{\top}} \\
% 		&\times \Big(\frac{\partial^{2} \mathcal{L}_{\mathcal{S}}}{\partial \boldsymbol{w}^{*}\partial {\boldsymbol{w}^{*}}^{\top}}\Big)^{-1} 
% 		\frac{\partial^{2}\mathcal{L}_{\mathcal{S}}}{\partial \boldsymbol{w}^{*} \partial c^{\top}}.
% 	\end{aligned}
% \end{equation}

Computing $\boldsymbol{U}_{1}$ and $\boldsymbol{U}_{2}$ remains challenging, since they are required to compute the intractable inverse of Hessian matrices in HNNs. Here we present approximation for $\boldsymbol{U}_{1}$ and  $\boldsymbol{U}_{2}$.

\noindent \textit{Approximation of $\boldsymbol{U}_{1}$.} Computing  $\boldsymbol{U}_{1}$ in Eq.~\eqref{equation:gradient_G_1} contains the inverse of the Hessian matrix, which is non-trivial to compute for deep neural networks. Here, we utilize the first $J$ terms of the Neumann series to approximate the inverse of the Hessian matrix.
% , and denote the approximated $\boldsymbol{U}_{1}$ as $\boldsymbol{U}_{1}^{'}$. 
Then $\boldsymbol{U}_{1}$ can be approximated as 
\begin{equation}
	% \footnotesize
    \small
	\label{equation:G_1:approx}
	\begin{aligned}
		\boldsymbol{U}_{1} &\approx \boldsymbol{U}_{1}^{'} \triangleq - \nabla_{\boldsymbol{w}} \mathcal{L}_{\mathcal{V}}
  (\boldsymbol{w}^{\ast}, c)\\
    &\times \sum_{j=0}^{J}\Big(\boldsymbol{I} - 
    \nabla^{2}_{\boldsymbol{w}} \mathcal{L}_{\mathcal{S}}(\boldsymbol{w}^{\ast}, c)
    \Big)^{j}	\nabla^{2}_{\boldsymbol{w}c} \mathcal{L}_{\mathcal{S}}(\boldsymbol{w}^{\ast}, c).
	\end{aligned}
\end{equation}
To compute Eq.~\eqref{equation:G_1:approx}, we first initialize two intermediate variables $\boldsymbol{v}^{0}$ and $\boldsymbol{p}^{0}$ both  as $\nabla_{\boldsymbol{w}} \mathcal{L}_{\mathcal{V}}
  (\boldsymbol{w}^{\ast})$. For every iteration, we update the intermediate variable as
\begin{equation}
	\label{neumann_series_iteration}
	\small
	\begin{aligned}
		\boldsymbol{v}^{i+1} = \boldsymbol{v}^{i}\Big(\boldsymbol{I} - 
         \nabla^{2}_{\boldsymbol{w}} \mathcal{L}_{\mathcal{S}}(\boldsymbol{w}^{\ast}, c)\Big), \qquad \boldsymbol{p}^{i+1}  =\boldsymbol{p}^{i} + \boldsymbol{v}^{i+1}.
	\end{aligned}
\end{equation}
After $J$ iterations, $\boldsymbol{U}_{1}^{'}$ is given by
\begin{equation}
	\label{equation:G_1_final}
	\begin{aligned}
		\boldsymbol{U}_{1}^{'} = -\boldsymbol{p}^{J}\nabla^{2}_{\boldsymbol{w}c} \mathcal{L}_{\mathcal{S}}(\boldsymbol{w}^{\ast}, c).
	\end{aligned}
\end{equation}
Obviously, the intermediate variable keeps the vector during iterations, thus computing 
$\boldsymbol{U}_{1}^{'}$ only needs Hessian-vector product and Jacobian-vector product~\citep{baydin2018automatic}, eliminating the need of time-consuming and memory-consuming Hessian and Jacobian.

% The second term $\boldsymbol{U}_{2}$ is computed as
% \begin{equation}
% 	\label{equation:gradient_G_2}
% 	\begin{aligned}
% 		\boldsymbol{U}_{2} &= 2K(1-\Vert \nabla_{\boldsymbol{w}} \mathcal{L}_{\mathcal{S}}(\hat{\boldsymbol{w}}, c) \Vert)^{K-1} \nabla_{\boldsymbol{w}} \mathcal{L}_{\mathcal{S}}(\hat{\boldsymbol{w}}, c) \nabla^{2}_{\boldsymbol{w}} \mathcal{L}_{\mathcal{S}}(\hat{\boldsymbol{w}}, c) \\
% 		&\times [\nabla^{2}_{\boldsymbol{w}} \mathcal{L}_{\mathcal{S}}(\boldsymbol{w}^{\ast}, c)]^{-1} 
% 		\nabla^{2}_{\boldsymbol{w}c} \mathcal{L}_{\mathcal{S}}(\boldsymbol{w}^{*}, c)
% 	\end{aligned}
% \end{equation}
\noindent \textit{Approximation of $\boldsymbol{U}_{2}$.}
We utilize $\nabla^{2}_{\boldsymbol{w}} \mathcal{L}_{\mathcal{S}}(\hat{\boldsymbol{w}}, c)$ to approximate $ \nabla^{2}_{\boldsymbol{w}} \mathcal{L}_{\mathcal{S}}(\boldsymbol{w}^{\ast}, c)$ in Eq.~\eqref{equation:gradient_G_2}, and $\boldsymbol{U}_{2}$ is approximated as
\begin{equation}
	\label{equation:gradient_G_2_approx}
 % \footnotesize
 \small
	\begin{aligned}
		\boldsymbol{U}_{2} &\approx \boldsymbol{U}^{'}_{2} \triangleq
        -2K\Big(1-\Vert  \nabla_{\boldsymbol{w}} \mathcal{L}_{\mathcal{S}}(\hat{\boldsymbol{w}}, c) \Vert\Big)^{K-1} \\
        & \times
        \nabla_{\boldsymbol{w}} \mathcal{L}_{\mathcal{S}}(\hat{\boldsymbol{w}}, c)
		\nabla^{2}_{\boldsymbol{w}c} \mathcal{L}_{\mathcal{S}}(\boldsymbol{w}^{\ast}, c).
	\end{aligned}
\end{equation}
By substituting Eq.~\eqref{equation:G_1:approx}, Eq.~\eqref{equation:gradient_G_2_approx}, and Eq.~\eqref{equation:gradient_F_W_1_2} into Eq.~\eqref{equation:gradient_c_first}, the gradient of the curvature $c$ is computed as 
\begin{equation}
\label{equation:gradient_c_final}
\footnotesize
	\begin{aligned}
		&\frac{d \mathcal{F}(\boldsymbol{w}^{\ast}(c), c)}{d c} =  \nabla_{c}  \mathcal{F}(\boldsymbol{w}^{\ast}(c), c)  + \boldsymbol{U}_{1} + \boldsymbol{U}_{2}
		\\
		&\approx
		 \nabla_{c}  \mathcal{F}(\boldsymbol{w}^{\ast}(c), c) - \nabla_{\boldsymbol{w}}\mathcal{L}_{\mathcal{V}}
              (\boldsymbol{w}^{\ast}, c) \\
              & \times
                \sum_{j=0}^{J}\Big(\boldsymbol{I} - 
                \nabla^{2}_{\boldsymbol{w}} \mathcal{L}_{\mathcal{S}}(\boldsymbol{w}^{\ast}, c)
                \Big)^{j}	\nabla^{2}_{\boldsymbol{w}c} \mathcal{L}_{\mathcal{S}}(\boldsymbol{w}^{\ast}, c)
		\\
		&-
		   2K\Big(1-\Vert  \nabla_{\boldsymbol{w}} \mathcal{L}_{\mathcal{S}}(\hat{\boldsymbol{w}}, c) \Vert\Big)^{K-1} 	
        \nabla_{\boldsymbol{w}} \mathcal{L}_{\mathcal{S}}(\hat{\boldsymbol{w}}, c)
		\nabla^{2}_{\boldsymbol{w}c} \mathcal{L}_{\mathcal{S}}(\boldsymbol{w}^{\ast}, c)
		.
	\end{aligned}
\end{equation}
Compared with the gradient of $c$ in Eq.~\eqref{equation:gradient_c_first}, the estimated gradient in Eq.~\eqref{equation:gradient_c_final} significantly reduces time complexity. 
The overall algorithm is summarized in Algorithm~\ref{algorithm:overall_algorithm}.

\section{Theoretical Guarantees of Our Method}
In this section, we present theoretical guarantees of the proposed method.
Specifically, in Section~\ref{section:approximation_error_ours}, we conduct approximation error analyses to present error bounds for the approximation ways used in our method, supporting the validity of the introduced approximation algorithms.
% Specifically, we present the approximation error analyses to 
% present the error bound of the approximation ways in our method in Section~\ref{section:approximation_error_ours}, 
% demonstrating the rationality of the introduced approximation algorithms.
In Section~\ref{section:convergence_analysis}, we further establish the convergence analyses to demonstrate that the weights  and curvatures in HNNs can converge.
Then, in Section~\ref{ea}, we conduct efficiency analyses for proving the efficiency of the approximation algorithms.
\subsection{Assumptions}
To quantify the following approximation errors and analyze the convergence performance, we introduce the following assumptions.
% \vspace{-0.5em}
\begin{assumption} (Lipschitz continuity).
	Assume that $\mathcal{L}_{\mathcal{S}}$, $\mathcal{L}_{\mathcal{V}}$, $\nabla_{\boldsymbol{w}} \mathcal{L}_{\mathcal{S}}(\boldsymbol{w}, c)$, $\nabla_{\boldsymbol{w}} \mathcal{L}_{\mathcal{V}}(\boldsymbol{w}, c)$ , $\nabla^{2}_{\boldsymbol{w}} \mathcal{L}_{\mathcal{S}}(\boldsymbol{w}, c)$, $\nabla^{2}_{\boldsymbol{w}c} \mathcal{L}_{\mathcal{S}}(\boldsymbol{w}, c)$, and $\nabla  \mathcal{F}(c)$  are Lipschitz continuous \textit{w.r.t.} $\boldsymbol{w}$ with constants $L_{G}, L_{\bar{G}}, H, L_{H_{2}}, L_{G^{'}}, L_{G_{2}}, L_{H_{c}}$.
    Assume that $\nabla_{\boldsymbol{w}} \mathcal{L}_{\mathcal{S}}(\boldsymbol{w}, c)$ is Lipschitz continuous \textit{w.r.t.} $c$ with constant $L_{G_{3}}$.
    Assume that $\frac{\partial \hat{\text{sn}}}{\partial \boldsymbol{w}}$ is Lipschitz continuous with constant $L_{G_{4}}$.
\end{assumption}
% \vspace{-0.5em}
\noindent \textbf{Remark.} The assumptions are commonly used in the convergence analysis of SGD-based methods~\citep{kingma2014adam, reddi2019convergence, zhuang2022surrogate} and bi-level optimization methods~\citep{ rajeswaran2019meta, ji2022theoretical}.

\begin{algorithm}[H]
	\small
	\caption{Training process of the proposed method.}
	\label{algorithm:overall_algorithm}
	\begin{algorithmic}[1]
		\renewcommand{\algorithmicrequire}{\textbf{Input:}}
		\renewcommand{\algorithmicensure}{\textbf{Output:}}
		\REQUIRE 
  Initial parameters $\boldsymbol{w}$ and curvature $c$, maximum iterations for inner level $T$ and outer level $\mathcal{T}$, perturbation radius $\hat{\rho}$, inner-level step-size $\eta$, and outer-level step-size $\eta_{c}$.
  % Initial parameters of HNNs $\boldsymbol{w}$, initial curvature $c$, maximum iteration of inner level $T$, maximum iteration of outer level $\mathcal{T}$, perturbation radius $\hat{\rho}$, step-size of inner level $\eta$, step-size of outer level $\eta_{c}$.
		\STATE $I_{\text{outer}} =0$.
		\WHILE{$I_{\text{outer}} < \mathcal{T}$}
		\STATE $I_{\text{inner}} = 0$
		\WHILE {$I_{\text{inner}} \leq T$}
		\STATE Update parameters $\boldsymbol{w}$ via Eq.~\eqref{equation:inner_level _update}.
		\ENDWHILE
		\STATE Compute loss function $\mathcal{F}(\boldsymbol{w}^{\ast}(c),c)$ in Eq.~\eqref{equation:objective_approx}.
		\STATE Compute the direct gradient $\nabla_{c} \mathcal{F}(\boldsymbol{w}^{\ast}(c), c)$.
		\STATE$\boldsymbol{v}^{0}=\boldsymbol{p}^{0} = \nabla_{\boldsymbol{w}}\mathcal{L}_{\mathcal{V}}(\boldsymbol{w}^{\ast}(c),c)$.
		\STATE $i=0$
		\WHILE{$i \le J$}
		\STATE Update intermediate vector $\boldsymbol{v}_{i}$ and $\boldsymbol{p}_{i}$ via Eq.~\eqref{neumann_series_iteration}.
		\ENDWHILE
		% \STATE Compute $\boldsymbol{U}_{1}$ via Eq.~\eqref{equation:G_1_final}, and compute $\boldsymbol{U}_{2}$ via Eq.\eqref{equation:gradient_G_2_approx}.
		\STATE Compute gradients of curvatures via Eq.~\eqref{equation:gradient_c_final}.
  % $\frac{d \mathcal{F}(\boldsymbol{w}^{\ast}(c), c)}{d c} = \frac{\partial \mathcal{F}(\boldsymbol{w}^{\ast}(c), c)}{\partial c} + \boldsymbol{U}_{1} + \boldsymbol{U}_{2}$.
		\STATE Update curvatures via Eq.~\eqref{equation:curvature_update}.
		\ENDWHILE
		\STATE Return the updated curvature $c$ and  updated parameters $\boldsymbol{w}$.
	\end{algorithmic}
\end{algorithm}

\subsection{Approximation Error Analyses in Our Method}
\label{section:approximation_error_ours}
We utilize 
$\hat{\rm{sn}}(\hat{\boldsymbol{w}}(c))$ to efficiently compute $\text{SN}(\boldsymbol{w}^{\ast}(c))$.
Moreover,  we utilize $\boldsymbol{U}_{1}^{'}$ and $\boldsymbol{U}_{2}^{'}$ to approximate $\boldsymbol{U}_{1}$ and $\boldsymbol{U}_{2}$ respectively, aiming to simplify the gradient of $c$. 
In this subsection, we present theoretical analyses of approximation error of scope sharpness measure and approximation error of gradients of curvatures. These analyses indicate that the approximation errors are upper-bounded, showing the rationality of approximation ways in the proposed method.

% \subsubsection{Approximation error of $\text{SN}(\boldsymbol{w}^{\ast}(c))$}
\subsubsection{Approximation Error of Scope Sharpness Measure}

% In our method, we adopt the 
In Theorem~\ref{Theorem:approximation_error_sn}, we present the approximation error of scope sharpness measure, \textit{i.e.}, the upper bound of   $| \rm{SN}(\boldsymbol{w}^{\ast}(c)) - \hat{\rm{sn}}(\hat{\boldsymbol{w}}(c))  |$.
\begin{theorem}
\label{Theorem:approximation_error_sn}
% [Approximation error of  $\rm{SN}(\boldsymbol{w}(c))$]
	Suppose  the above assumptions  hold, $\hat{\rm{sn}}(\hat{\boldsymbol{w}}(c))$ satisfies
 % \vspace{-0.5em}
	\begin{equation}
		\small
		\begin{aligned}
			| 
\rm{SN}(\boldsymbol{w}^{\ast}(c)) - 
			\hat{\rm{sn}}(\hat{\boldsymbol{w}}(c))  
			|
			\le \mathcal{E}_{mea},
		\end{aligned}
	\end{equation}
	where $\mathcal{E}_{mea}$ is a  constant and computed as
 \begin{equation}
        \begin{aligned}
            \mathcal{E}_{mea}  \triangleq 1 + \rho^{2}L_{G_{4}}.
        \end{aligned}
    \end{equation}

\end{theorem}

\begin{proof}
    Recall that $\hat{\rm{sn}}(\hat{\boldsymbol{w}}(c))$ and $\rm{SN}(\boldsymbol{w}^{\ast
    }(c))$ are computed as
    \begin{equation}
    \small
        \begin{aligned}
           & \hat{\rm{sn}}(\hat{\boldsymbol{w}}(c)) = 1-\Big(1-\Big\|
   \nabla_{\boldsymbol{w}} \mathcal{L}_{\mathcal{S}}(\hat{\boldsymbol{w}}(c), c)
   \Big\|^{2}\Big)^{K+1},
         \\
           & \rm{SN}(\boldsymbol{w}^{\ast}(c)) = \underset{\Vert \boldsymbol{\epsilon} \Vert \le \rho}{\max} \text{sn}(\boldsymbol{w}^{\ast}(c)+\boldsymbol{\epsilon}).
        \end{aligned}
    \end{equation}
    We adopt two approximation ways to approximate $\rm{SN}(\boldsymbol{w}^{\ast}(c))$  as  $\hat{\rm{sn}}(\hat{\boldsymbol{w}}(c))$.
    (1) We utilize $ \hat{\text{sn}}(\boldsymbol{w}) \triangleq 1-\Big(1-\Big\|  \nabla_{\boldsymbol{w}} \mathcal{L}_{\mathcal{S}}(\boldsymbol{w}, c) \Big\|^{2}\Big)^{K+1}$ to approximate $\text{sn}(\boldsymbol{w}) \triangleq \sum_{i=0}^{\infty}\nabla_{\boldsymbol{w}} \mathcal{L}_{\mathcal{S}}(\boldsymbol{w}, c)^{\top}\Big(\boldsymbol{I} - \nabla_{\boldsymbol{w}} \mathcal{L}_{\mathcal{S}}(\boldsymbol{w}, c)
    \nabla_{\boldsymbol{w}} \mathcal{L}_{\mathcal{S}}(\boldsymbol{w}, c)^{\top}\Big)^{i}
		\nabla_{\boldsymbol{w}} \mathcal{L}_{\mathcal{S}}(\boldsymbol{w}, c)$.
    (2) We utilize $ \hat{\text{sn}}(\boldsymbol{w}^{*}({c} )) +  \boldsymbol{\epsilon}^{\top}  \frac{\partial \hat{\text{sn}}(\boldsymbol{w}^{*}({c} ))}{\partial \boldsymbol{w}^{*}}$ to approximate 
    $ \hat{\text{sn}}(\boldsymbol{w}^{\ast}(c)+\boldsymbol{\epsilon})$.

    We first present the error bound for the first approximation. From the construction of $\text{sn}(\boldsymbol{w})$ and $\hat{\rm{sn}}(\boldsymbol{w})$, we have
    \begin{equation}
    % \footnotesize
    \small
    \label{equation:sn_error_bound}
        \begin{aligned}
            & \bigg| \text{sn}(\boldsymbol{w}) - \hat{\rm{sn}}(\boldsymbol{w}) \bigg| 
  %           &=  \big|
  %           \sum_{i=0}^{K}(\nabla_{\boldsymbol{w}} \mathcal{L}_{\mathcal{S}}(\boldsymbol{w}, c))^{\top}\Big(\boldsymbol{I} - 
  %           \nabla_{\boldsymbol{w}} \mathcal{L}_{\mathcal{S}}(\boldsymbol{w}, c)
  %           \nabla_{\boldsymbol{w}} \mathcal{L}_{\mathcal{S}}(\boldsymbol{w}, c)^{\top}\Big)^{i}
		% (\nabla_{\boldsymbol{w}} \mathcal{L}_{\mathcal{S}}(\boldsymbol{w}, c)) \\
  %   &- 
  %       \sum_{i=0}^{\infty}(\nabla_{\boldsymbol{w}} \mathcal{L}_{\mathcal{S}}(\boldsymbol{w}, c))^{\top}\Big(\boldsymbol{I} - \nabla_{\boldsymbol{w}} \mathcal{L}_{\mathcal{S}}(\boldsymbol{w}, c)\nabla_{\boldsymbol{w}} \mathcal{L}_{\mathcal{S}}(\boldsymbol{w}, c)^{\top}\Big)^{i}
		% (\nabla_{\boldsymbol{w}} \mathcal{L}_{\mathcal{S}}(\boldsymbol{w}, c))
  %           \big|\\
            = \bigg|
                 \sum_{i=K+1}^{\infty}(\nabla_{\boldsymbol{w}} \mathcal{L}_{\mathcal{S}}(\boldsymbol{w}, c))^{\top}
                 \\
                 & \times
                 \Big(\boldsymbol{I} - \nabla_{\boldsymbol{w}} \mathcal{L}_{\mathcal{S}}(\boldsymbol{w}, c) 
                 \nabla_{\boldsymbol{w}} \mathcal{L}_{\mathcal{S}}(\boldsymbol{w}, c)^{\top}\Big)^{i}
		(\nabla_{\boldsymbol{w}} \mathcal{L}_{\mathcal{S}}(\boldsymbol{w}, c))
            \bigg| \\
           & =\sum_{i=K+1}^{\infty} \bigg|  
                \Big\|\nabla_{\boldsymbol{w}} \mathcal{L}_{\mathcal{S}}(\boldsymbol{w}, c)\Big\|^{2} \Big(1-\Big\|\nabla_{\boldsymbol{w}} \mathcal{L}_{\mathcal{S}}(\boldsymbol{w}, c)\Big\|^{2}\Big)^{i}
                \bigg| \le 1.
        \end{aligned}
    \end{equation}
  The last inequality holds because 
  \begin{equation}
   \small
      \begin{aligned}
           \|\nabla_{\boldsymbol{w}} \mathcal{L}_{\mathcal{S}}(\boldsymbol{w}, c)\|^{2} <1,
      \end{aligned}
  \end{equation}
  and 
   \begin{equation}
   \small
      \begin{aligned}
          &\sum_{i=K+1}^{\infty} |  
                \|\nabla_{\boldsymbol{w}} \mathcal{L}_{\mathcal{S}}(\boldsymbol{w}, c)\|^{2} (1-\|\nabla_{\boldsymbol{w}} \mathcal{L}_{\mathcal{S}}(\boldsymbol{w}, c)\|^{2})^{i}
                |\\
                &= (1-\|\nabla_{\boldsymbol{w}} \mathcal{L}_{\mathcal{S}}(\boldsymbol{w}, c)\|^{2})^{K+1}\le 1
      \end{aligned}
  \end{equation}

    % We first provide the  error bound for the first approximation.
    % \begin{equation}
    %     \begin{aligned}
    %         &\text{sn}(\boldsymbol{w}) - \hat{\rm{sn}}(\boldsymbol{w}) \\
    %         &=  1-\Big(1-\Big\|\frac{\partial \mathcal{L}_{\mathcal{S}}}{\partial \boldsymbol{w}}\Big\|^{2}\Big)^{K} - \underset{i \rightarrow \infty}{\lim}  1-\Big(1-\Big\|\frac{\partial \mathcal{L}_{\mathcal{S}}}{\partial \boldsymbol{w}}\Big\|^{2}\Big)^{i} \\
    %         & \le 1-(1-G^{2})^{K},
    %     \end{aligned}
    % \end{equation}
    % following by $\Vert \frac{\partial \mathcal{L}_{\mathcal{S}}}{\partial \boldsymbol{w}} \Vert \le G$, from assumptions.
    
    As to the second approximation, we mainly adopt the Taylor expansion for approximation. 
    The upper bound of error is given by
    \begin{equation}
    \small
        \begin{aligned}
             &\bigg|\hat{\text{sn}}(\boldsymbol{w}^{\ast}(c)+\boldsymbol{\epsilon}) - \hat{\text{sn}}(\boldsymbol{w}^{*}({c} )) -  \boldsymbol{\epsilon}^{\top}  \frac{\partial \hat{\text{sn}}(\boldsymbol{w}^{*}({c} ))}{\partial \boldsymbol{w}^{*}} \bigg| \\
             & \le \frac{1}{2} \left| \boldsymbol{\epsilon}^{\top}\frac{\partial^{2} \rm{sn}(\boldsymbol{w}^{'})}{\partial \boldsymbol{w}^{'} 
               \partial {\boldsymbol{w}^{'}}^{\top}         }\boldsymbol{\epsilon} \right| \le \frac{1}{2}\Vert \boldsymbol{\epsilon} \Vert^{2} \left\Vert \frac{\partial^{2} \rm{sn}(\boldsymbol{w}^{'})}{\partial \boldsymbol{w}^{'} 
               \partial {\boldsymbol{w}^{'}}^{\top}} \right \Vert
               \le \rho^{2}L_{G_{4}}.
        \end{aligned}
    \end{equation}
    By denoting $\mathcal{E}_{mea}$ as $ 1 + \rho^{2}L_{G_{4}}$, we prove the theorem.
    
    % In this way, $\mathcal{E}_{mea}$ is computed by
    % \begin{equation}
    %     \begin{aligned}
    %         \mathcal{E}_{mea} = 1 + \rho^{2}L_{G_{4}}.
    %     \end{aligned}
    % \end{equation}
\end{proof}

\subsubsection{Approximation Error of Gradients of Curvatures}
We denote the exact gradients of $c$ as $\nabla c \triangleq \frac{d \mathcal{F}(\boldsymbol{w}^{\ast}(c), c)}{d c} $, while we denote the approximated gradients of $c$ in Eq.~\eqref{equation:gradient_c_final} as $\hat{\nabla }c$.  The approximation error has an upper bound, as shown in Theorem~\ref{Theorem:approximation_error_c}.
% The upper bound demonstrates the rationality of the proposed approximation ways.
We first present Lemmas~\ref{Theorem:approximation error_G_1} and ~\ref{Theorem:approximation error_G_2}  to analyze the approximation error of $\boldsymbol{U}_{1}$ and $\boldsymbol{U}_{2}$, respectively.

\noindent \emph{Approximation error for $\boldsymbol{U}_{1}$.} 
We adopt two approximation ways for computing $\boldsymbol{U}_{1}$. 

(1) $\boldsymbol{U}_{1}$ deriving with implicit differentiation algorithm requires $\nabla_{\boldsymbol{w}} \mathcal{L}_{\mathcal{S}}(\boldsymbol{w}^{*}, c) = 0$. While it is non-trival to optimize $\boldsymbol{w}$ to let the gradient of $\boldsymbol{w}^{*}$  equals to 0, \textit{i.e.}, $\nabla_{\boldsymbol{w}} \mathcal{L}_{\mathcal{S}}(\boldsymbol{w}^{*}, c)\neq 0$ in practice. Suppose that  $\nabla_{\boldsymbol{w}} \mathcal{L}_{\mathcal{S}}(\boldsymbol{w}^{**}, c) = 0$. 
The exact term $\boldsymbol{U}_{1}^{*}$ needs to computed  using  $\boldsymbol{w}^{**}$, \textit{i.e.},
\begin{equation}
\small
	\begin{aligned}
		\boldsymbol{U}_{1}^{*} \triangleq \nabla_{\boldsymbol{w}} \mathcal{L}_{\mathcal{V}}(\boldsymbol{w}^{**}(c),c) [\nabla^{2}_{\boldsymbol{w}} \mathcal{L}_{\mathcal{S}}(\boldsymbol{w}^{**}, c)]^{-1}\nabla^{2}_{\boldsymbol{w}c} \mathcal{L}_{\mathcal{S}}(\boldsymbol{w}^{**}, c).
	\end{aligned}
\end{equation}

(2) We utilize the first $J$ terms Neumann series to approximate the inverse of the Hessian matrix in Eq.~\eqref{equation:G_1:approx}.

The error of the first approximation way, \textit{i.e.}, the upper bound of $\Vert \boldsymbol{U}_{1}^{*} - \boldsymbol{U}_{1}^{'} \Vert$, is presented in Lemma~\ref{Theorem:approximation error_G_1}.

\begin{lemma}
	\label{Theorem:approximation error_G_1}
	Assume that $\Vert \boldsymbol{w}^{**} - \boldsymbol{w}^{*} \Vert \le  \delta$. $\boldsymbol{U}_{1}^{'}$ satisfies
	\begin{equation}
		\small
		\begin{aligned}
			\Vert 
			\boldsymbol{U}_{1}^{*} - \boldsymbol{U}_{1}^{'}
			\Vert
			\le \mathcal{E}_{cur_{1}},
		\end{aligned}
	\end{equation}
    where $\mathcal{E}_{cur_{1}}$ is given by
    \begin{equation}
    \small
        \begin{aligned}
         \mathcal{E}_{cur_{1}} =    \mathcal{E}_{cur_{1}}^{'} \delta + \mathcal{E}_{cur_{1}}^{''}.
        \end{aligned}
    \end{equation}
	Here, $\mathcal{E}_{cur_{1}}^{'}$ and $\mathcal{E}_{cur_{1}}^{''}$ are constants that only depend on the  Lipschitz constants, which are computed as
 \begin{equation}
 \small
            \begin{aligned}
                 &\mathcal{E}_{cur_{1}}^{'} \triangleq \frac{L_{\bar{G}}L_{G^{'}}L_{G_{3}} + L_{G_{3}}L_{H_{2}}L_{H} + L_{\bar{G}}L_{G_{2}}L_{H}}{L_{H_{s}}^{2}} \\
                &\mathcal{E}_{cur_{1}}^{''} \triangleq 
                \frac{L_{\bar{G}}L_{G_{3}}\left(L_{H}-L_{H_{s}}+L_{H_{s}}(1+L_{H})^{J}\right)}{L_{H}L_{H_{s}}}.
            \end{aligned}
        \end{equation}
\end{lemma}

\begin{proof}
To quantify the error bound of $\Vert  \boldsymbol{U}_{1}^{*} - \boldsymbol{U}_{1}^{'}  \Vert$, we utilize the the Lipschitz continuity  assumptions of $\mathcal{L}_{\mathcal{V}}$, $\nabla_{\boldsymbol{w}} \mathcal{L}_{\mathcal{S}}(\boldsymbol{w}, c)$, $\nabla \mathcal{L}_{\mathcal{V}}(\boldsymbol{w})$, $\nabla^{2}_{\boldsymbol{w}} \mathcal{L}_{\mathcal{S}}(\boldsymbol{w}, c)$, $\nabla^{2}_{\boldsymbol{w}c} \mathcal{L}_{\mathcal{S}}(\boldsymbol{w}, c)$
, and $\nabla_{\boldsymbol{w}} \mathcal{L}_{\mathcal{S}}(\boldsymbol{w}, c)$.

    Recall that $\boldsymbol{U}_{1}^{*}$ and $\boldsymbol{U}_{1}^{'}$ are computed as 
    \begin{equation}
    % \small
    \footnotesize
        \begin{aligned}
            \boldsymbol{U}_{1}^{*} &= -\nabla_{\boldsymbol{w}} \mathcal{L}_{\mathcal{V}}(\boldsymbol{w}^{**}, c)[\nabla^{2}_{\boldsymbol{w}} \mathcal{L}_{\mathcal{S}}(\boldsymbol{w}^{**}, c)]^{-1}\nabla^{2}_{\boldsymbol{w}c} \mathcal{L}_{\mathcal{S}}(\boldsymbol{w}^{**}, c)\\
            \boldsymbol{U}_{1}^{'} &= -\nabla_{\boldsymbol{w}} \mathcal{L}_{\mathcal{V}}(\boldsymbol{w}^{*}, c)
            \sum_{j=0}^{J}[\boldsymbol{I} - \nabla^{2}_{\boldsymbol{w}} \mathcal{L}_{\mathcal{S}}(\boldsymbol{w}^{\ast}, c)]^{j}
            \nabla^{2}_{\boldsymbol{w}c} \mathcal{L}_{\mathcal{S}}(\boldsymbol{w}^{*}, c).
        \end{aligned}
    \end{equation}
    Thus we have
    \begin{equation}
    \small
    \label{equation:G_1_G}
        \begin{aligned}
           & \Vert  \boldsymbol{U}_{1}^{*} - \boldsymbol{U}_{1}^{'}  \Vert=\\
           %  & =\Vert
           %   \nabla_{\boldsymbol{w}} \mathcal{L}_{\mathcal{V}}(\boldsymbol{w}^{**}, c)[\nabla^{2}_{\boldsymbol{w}} \mathcal{L}_{\mathcal{S}}(\boldsymbol{w}^{**}, c)]^{-1}\nabla^{2}_{\boldsymbol{w}c} \mathcal{L}_{\mathcal{S}}(\boldsymbol{w}^{**}, c)- \\
           % & \nabla_{\boldsymbol{w}} \mathcal{L}_{\mathcal{V}}(\boldsymbol{w}^{*}, c)
           %   \sum_{j=0}^{J}[\boldsymbol{I} -\nabla^{2}_{\boldsymbol{w}} \mathcal{L}_{\mathcal{S}}(\boldsymbol{w}^{\ast}, c)]^{j}
           %  \nabla^{2}_{\boldsymbol{w}c} \mathcal{L}_{\mathcal{S}}(\boldsymbol{w}^{*}, c)
           %  \Vert\\
             & \Vert
            \nabla_{\boldsymbol{w}} \mathcal{L}_{\mathcal{V}}(\boldsymbol{w}^{**}, c)[\nabla^{2}_{\boldsymbol{w}} \mathcal{L}_{\mathcal{S}}(\boldsymbol{w}^{**}, c)]^{-1}\nabla^{2}_{\boldsymbol{w}c} \mathcal{L}_{\mathcal{S}}(\boldsymbol{w}^{**}, c)- \\
             & 
          \nabla_{\boldsymbol{w}} \mathcal{L}_{\mathcal{V}}(\boldsymbol{w}^{*}, c)[\nabla^{2}_{\boldsymbol{w}} \mathcal{L}_{\mathcal{S}}(\boldsymbol{w}^{\ast}, c)]^{-1}\nabla^{2}_{\boldsymbol{w}c} \mathcal{L}_{\mathcal{S}}(\boldsymbol{w}^{*}, c)+ \\
             & 
          \nabla_{\boldsymbol{w}} \mathcal{L}_{\mathcal{V}}(\boldsymbol{w}^{*}, c)[\nabla^{2}_{\boldsymbol{w}} \mathcal{L}_{\mathcal{S}}(\boldsymbol{w}^{\ast}, c)]^{-1}\nabla^{2}_{\boldsymbol{w}c} \mathcal{L}_{\mathcal{S}}(\boldsymbol{w}^{*}, c)- \\
             & \nabla_{\boldsymbol{w}} \mathcal{L}_{\mathcal{V}}(\boldsymbol{w}^{*}, c) 
             \sum_{j=0}^{J}[\boldsymbol{I} -\nabla^{2}_{\boldsymbol{w}} \mathcal{L}_{\mathcal{S}}(\boldsymbol{w}^{\ast}, c)]^{j}
            \nabla^{2}_{\boldsymbol{w}c} \mathcal{L}_{\mathcal{S}}(\boldsymbol{w}^{*}, c)
            \Vert\\
            &\le
            \Vert
             \nabla_{\boldsymbol{w}} \mathcal{L}_{\mathcal{V}}(\boldsymbol{w}^{**}, c)[\nabla^{2}_{\boldsymbol{w}} \mathcal{L}_{\mathcal{S}}(\boldsymbol{w}^{**}, c)]^{-1}\nabla^{2}_{\boldsymbol{w}c} \mathcal{L}_{\mathcal{S}}(\boldsymbol{w}^{**}, c) \\
             & 
           - \nabla_{\boldsymbol{w}} \mathcal{L}_{\mathcal{V}}(\boldsymbol{w}^{*}, c)[\nabla^{2}_{\boldsymbol{w}} \mathcal{L}_{\mathcal{S}}(\boldsymbol{w}^{\ast}, c)]^{-1}\nabla^{2}_{\boldsymbol{w}c} \mathcal{L}_{\mathcal{S}}(\boldsymbol{w}^{*}, c)
            \Vert\\
            &+
            \Vert
 \nabla_{\boldsymbol{w}} \mathcal{L}_{\mathcal{V}}(\boldsymbol{w}^{*}, c)[\nabla^{2}_{\boldsymbol{w}} \mathcal{L}_{\mathcal{S}}(\boldsymbol{w}^{\ast}, c)]^{-1}\nabla^{2}_{\boldsymbol{w}c} \mathcal{L}_{\mathcal{S}}(\boldsymbol{w}^{*}, c)- \\
             & \nabla_{\boldsymbol{w}} \mathcal{L}_{\mathcal{V}}(\boldsymbol{w}^{*}, c)
             \sum_{j=0}^{J}[\boldsymbol{I} -\nabla^{2}_{\boldsymbol{w}} \mathcal{L}_{\mathcal{S}}(\boldsymbol{w}^{\ast}, c)]^{j}
            \nabla^{2}_{\boldsymbol{w}c} \mathcal{L}_{\mathcal{S}}(\boldsymbol{w}^{*}, c)
            \Vert.
        \end{aligned}
    \end{equation}
We first focus on the  first term  to the right of inequality and have
\begin{equation}
\small
\label{equation:G_{1}_G_fisrt_term}
    \begin{aligned}
     & \Vert
             \nabla_{\boldsymbol{w}} \mathcal{L}_{\mathcal{V}}(\boldsymbol{w}^{**}, c)[\nabla^{2}_{\boldsymbol{w}} \mathcal{L}_{\mathcal{S}}(\boldsymbol{w}^{**}, c)]^{-1}\nabla^{2}_{\boldsymbol{w}c} \mathcal{L}_{\mathcal{S}}(\boldsymbol{w}^{**}, c)- \\
             & 
          \nabla_{\boldsymbol{w}} \mathcal{L}_{\mathcal{V}}(\boldsymbol{w}^{*}, c)[\nabla^{2}_{\boldsymbol{w}} \mathcal{L}_{\mathcal{S}}(\boldsymbol{w}^{\ast}, c)]^{-1}\nabla^{2}_{\boldsymbol{w}c} \mathcal{L}_{\mathcal{S}}(\boldsymbol{w}^{*}, c)
            \Vert \\
           % & =
           % \Vert
           %    \nabla_{\boldsymbol{w}} \mathcal{L}_{\mathcal{V}}(\boldsymbol{w}^{**}, c)[\nabla^{2}_{\boldsymbol{w}} \mathcal{L}_{\mathcal{S}}(\boldsymbol{w}^{**}, c)]^{-1}\nabla^{2}_{\boldsymbol{w}c} \mathcal{L}_{\mathcal{S}}(\boldsymbol{w}^{**}, c)- \\
           %   &\nabla_{\boldsymbol{w}} \mathcal{L}_{\mathcal{V}}(\boldsymbol{w}^{*}, c)[\nabla^{2}_{\boldsymbol{w}} \mathcal{L}_{\mathcal{S}}(\boldsymbol{w}^{\ast}, c)]^{-1}\nabla^{2}_{\boldsymbol{w}c} \mathcal{L}_{\mathcal{S}}(\boldsymbol{w}^{**}, c)+ \\
           %   &\nabla_{\boldsymbol{w}} \mathcal{L}_{\mathcal{V}}(\boldsymbol{w}^{*}, c)[\nabla^{2}_{\boldsymbol{w}} \mathcal{L}_{\mathcal{S}}(\boldsymbol{w}^{\ast}, c)]^{-1}\nabla^{2}_{\boldsymbol{w}c} \mathcal{L}_{\mathcal{S}}(\boldsymbol{w}^{**}, c)- \\
           %    & 
           %\nabla_{\boldsymbol{w}} \mathcal{L}_{\mathcal{V}}(\boldsymbol{w}^{*}, c)[\nabla^{2}_{\boldsymbol{w}} \mathcal{L}_{\mathcal{S}}(\boldsymbol{w}^{\ast}, c)]^{-1}\nabla^{2}_{\boldsymbol{w}c} \mathcal{L}_{\mathcal{S}}(\boldsymbol{w}^{*}, c)
           % \Vert\\
           &\le
            \Vert
              \nabla_{\boldsymbol{w}} \mathcal{L}_{\mathcal{V}}(\boldsymbol{w}^{**}, c)[\nabla^{2}_{\boldsymbol{w}} \mathcal{L}_{\mathcal{S}}(\boldsymbol{w}^{**}, c)]^{-1}\nabla^{2}_{\boldsymbol{w}c} \mathcal{L}_{\mathcal{S}}(\boldsymbol{w}^{**}, c)- \\
             &\nabla_{\boldsymbol{w}} \mathcal{L}_{\mathcal{V}}(\boldsymbol{w}^{*}, c)[\nabla^{2}_{\boldsymbol{w}} \mathcal{L}_{\mathcal{S}}(\boldsymbol{w}^{\ast}, c)]^{-1}\nabla^{2}_{\boldsymbol{w}c} \mathcal{L}_{\mathcal{S}}(\boldsymbol{w}^{**}, c)
             \Vert
             \\
             &+ \Vert \nabla_{\boldsymbol{w}} \mathcal{L}_{\mathcal{V}}(\boldsymbol{w}^{*}, c)[\nabla^{2}_{\boldsymbol{w}} \mathcal{L}_{\mathcal{S}}(\boldsymbol{w}^{\ast}, c)]^{-1}\nabla^{2}_{\boldsymbol{w}c} \mathcal{L}_{\mathcal{S}}(\boldsymbol{w}^{**}, c)- \\
              & 
          \nabla_{\boldsymbol{w}} \mathcal{L}_{\mathcal{V}}(\boldsymbol{w}^{*}, c)[\nabla^{2}_{\boldsymbol{w}} \mathcal{L}_{\mathcal{S}}(\boldsymbol{w}^{\ast}, c)]^{-1}\nabla^{2}_{\boldsymbol{w}c} \mathcal{L}_{\mathcal{S}}(\boldsymbol{w}^{*}, c)
           \Vert  .
    \end{aligned}
\end{equation}
Here, the first term to the right of inequality satisfies that 
\begin{equation}
\small
    \begin{aligned}
        &\Vert
              \nabla_{\boldsymbol{w}} \mathcal{L}_{\mathcal{V}}(\boldsymbol{w}^{**}, c)[\nabla^{2}_{\boldsymbol{w}} \mathcal{L}_{\mathcal{S}}(\boldsymbol{w}^{**}, c)]^{-1}\nabla^{2}_{\boldsymbol{w}c} \mathcal{L}_{\mathcal{S}}(\boldsymbol{w}^{**}, c)- \\
             &\nabla_{\boldsymbol{w}} \mathcal{L}_{\mathcal{V}}(\boldsymbol{w}^{*}, c)[\nabla^{2}_{\boldsymbol{w}} \mathcal{L}_{\mathcal{S}}(\boldsymbol{w}^{\ast}, c)]^{-1}\nabla^{2}_{\boldsymbol{w}c} \mathcal{L}_{\mathcal{S}}(\boldsymbol{w}^{**}, c)
             \Vert\\
             &\le
             \Vert
              \nabla_{\boldsymbol{w}} \mathcal{L}_{\mathcal{V}}(\boldsymbol{w}^{**}, c)[\nabla^{2}_{\boldsymbol{w}} \mathcal{L}_{\mathcal{S}}(\boldsymbol{w}^{**}, c)]^{-1} \\
            & - 
             \nabla_{\boldsymbol{w}} \mathcal{L}_{\mathcal{V}}(\boldsymbol{w}^{*}, c)[\nabla^{2}_{\boldsymbol{w}} \mathcal{L}_{\mathcal{S}}(\boldsymbol{w}^{\ast}, c)]^{-1}
             \Vert
             \Vert
             \nabla^{2}_{\boldsymbol{w}c} \mathcal{L}_{\mathcal{S}}(\boldsymbol{w}^{**}, c) 
             \Vert,
    \end{aligned}
\end{equation}
    where 
    \begin{equation}
    \small
        \begin{aligned}
            &\Vert
              \nabla_{\boldsymbol{w}} \mathcal{L}_{\mathcal{V}}(\boldsymbol{w}^{**}, c)[\nabla^{2}_{\boldsymbol{w}} \mathcal{L}_{\mathcal{S}}(\boldsymbol{w}^{**}, c)]^{-1}- \\
               &
             \nabla_{\boldsymbol{w}} \mathcal{L}_{\mathcal{V}}(\boldsymbol{w}^{*}, c)[\nabla^{2}_{\boldsymbol{w}} \mathcal{L}_{\mathcal{S}}(\boldsymbol{w}^{\ast}, c)]^{-1}
             \Vert \le\\
             % &\le 
             % \Vert
             %      \nabla_{\boldsymbol{w}} \mathcal{L}_{\mathcal{V}}(\boldsymbol{w}^{**}, c)[\nabla^{2}_{\boldsymbol{w}} \mathcal{L}_{\mathcal{S}}(\boldsymbol{w}^{**}, c)]^{-1}-
             %       \nabla_{\boldsymbol{w}} \mathcal{L}_{\mathcal{V}}(\boldsymbol{w}^{**}, c)[\nabla^{2}_{\boldsymbol{w}} \mathcal{L}_{\mathcal{S}}(\boldsymbol{w}^{\ast}, c)]^{-1}
             % \Vert\\
             % &+
             % \Vert
             %     \nabla_{\boldsymbol{w}} \mathcal{L}_{\mathcal{V}}(\boldsymbol{w}^{**}, c)[\nabla^{2}_{\boldsymbol{w}} \mathcal{L}_{\mathcal{S}}(\boldsymbol{w}^{\ast}, c)]^{-1}-
             %  \nabla_{\boldsymbol{w}} \mathcal{L}_{\mathcal{V}}(\boldsymbol{w}^{*}, c)[\nabla^{2}_{\boldsymbol{w}} \mathcal{L}_{\mathcal{S}}(\boldsymbol{w}^{\ast}, c)]^{-1}
             % \Vert \\
             &
             \Vert
              \nabla_{\boldsymbol{w}} \mathcal{L}_{\mathcal{V}}(\boldsymbol{w}^{**}, c)
             \Vert
             \Vert
                 [\nabla^{2}_{\boldsymbol{w}} \mathcal{L}_{\mathcal{S}}(\boldsymbol{w}^{**}, c)]^{-1}-
                  [\nabla^{2}_{\boldsymbol{w}} \mathcal{L}_{\mathcal{S}}(\boldsymbol{w}^{\ast}, c)]^{-1}
             \Vert\\
             &+
             \Vert
                   \nabla_{\boldsymbol{w}} \mathcal{L}_{\mathcal{V}}(\boldsymbol{w}^{**}, c) -\nabla_{\boldsymbol{w}} \mathcal{L}_{\mathcal{V}}(\boldsymbol{w}^{*}, c)
             \Vert
             \Vert
             [\nabla^{2}_{\boldsymbol{w}} \mathcal{L}_{\mathcal{S}}(\boldsymbol{w}^{\ast}, c)]^{-1}
             \Vert.
        \end{aligned}
    \end{equation}
    From assumptions, we have that
    \begin{equation}
    \small
        \begin{aligned}
            \Vert
                 [\nabla^{2}_{\boldsymbol{w}} \mathcal{L}_{\mathcal{S}}(\boldsymbol{w}^{\ast}, c)]^{-1}
            \Vert
              \le \frac{1}{L_{H_{s}}}.
        \end{aligned}       
    \end{equation}
    % From the Neumann series, we have
    % \begin{equation}
    %     \begin{aligned}
    %         \Vert
    %              [\nabla^{2}_{\boldsymbol{w}} \mathcal{L}_{\mathcal{S}}(\boldsymbol{w}^{\ast}, c)]^{-1}
    %         \Vert
    %           \le
    %         \sum_{i=1}^{\infty}
    %         \Vert
    %          (\boldsymbol{I} - \nabla^{2}_{\boldsymbol{w}} \mathcal{L}_{\mathcal{S}}(\boldsymbol{w}^{\ast}, c))^{i}
    %         \Vert.
    %     \end{aligned}       
    % \end{equation}
    % \begin{equation}
    %     \begin{aligned}
    %         \Vert
    %              [\nabla^{2}_{\boldsymbol{w}} \mathcal{L}_{\mathcal{S}}(\boldsymbol{w}^{\ast}, c)]^{-1}
    %         \Vert = 
    %         \Vert
    %         \sum_{i=1}^{\infty} (\boldsymbol{I} - \nabla^{2}_{\boldsymbol{w}} \mathcal{L}_{\mathcal{S}}(\boldsymbol{w}^{\ast}, c))^{i}
    %         \Vert\\
    %           \le
    %         \sum_{i=1}^{\infty}
    %         \Vert
    %          (\boldsymbol{I} - \nabla^{2}_{\boldsymbol{w}} \mathcal{L}_{\mathcal{S}}(\boldsymbol{w}^{\ast}, c))^{i}
    %         \Vert.
    %     \end{aligned}       
    % \end{equation}
    % From assumptions, we have
    % \begin{equation}
    %     \begin{aligned}
    %         \Vert \boldsymbol{I} - \nabla^{2}_{\boldsymbol{w}} \mathcal{L}_{\mathcal{S}}(\boldsymbol{w}^{\ast}, c)) \Vert \le 1-L_{H_{s}}.
    %     \end{aligned}
    % \end{equation}
    % Thus we have
    %  \begin{equation}
    %     \begin{aligned}
    %         \Vert
    %              [\nabla^{2}_{\boldsymbol{w}} \mathcal{L}_{\mathcal{S}}(\boldsymbol{w}^{\ast}, c)]^{-1}
    %         \Vert
    %           \le  \sum_{i=1}^{\infty} (1-H)^{i} = \frac{1}{H}.
    %     \end{aligned}       
    % \end{equation}
    Then as to the error bound of $ \Vert
                 [\nabla^{2}_{\boldsymbol{w}} \mathcal{L}_{\mathcal{S}}(\boldsymbol{w}^{**}, c)]^{-1}-
                  [\nabla^{2}_{\boldsymbol{w}} \mathcal{L}_{\mathcal{S}}(\boldsymbol{w}^{\ast}, c)]^{-1}
             \Vert$, we have
             \begin{equation}
             \small
                 \begin{aligned}
                    & \Vert
                 [\nabla^{2}_{\boldsymbol{w}} \mathcal{L}_{\mathcal{S}}(\boldsymbol{w}^{**}, c)]^{-1}-
                  [\nabla^{2}_{\boldsymbol{w}} \mathcal{L}_{\mathcal{S}}(\boldsymbol{w}^{\ast}, c)]^{-1}
             \Vert\\
            & \le
            \Vert
            [\nabla^{2}_{\boldsymbol{w}} \mathcal{L}_{\mathcal{S}}(\boldsymbol{w}^{**}, c)]^{-1}
            \Vert
            \Vert
            [ \nabla^{2}_{\boldsymbol{w}} \mathcal{L}_{\mathcal{S}}(\boldsymbol{w}^{\ast}, c)]^{-1}
             \Vert
          \\
             &\times
               \Vert
                 \nabla^{2}_{\boldsymbol{w}} \mathcal{L}_{\mathcal{S}}(\boldsymbol{w}^{**}, c)-
                  \nabla^{2}_{\boldsymbol{w}} \mathcal{L}_{\mathcal{S}}(\boldsymbol{w}^{\ast}, c)
             \Vert
             \le \frac{L_{G^{'}}\delta}{L_{H_{s}}^{2}}
                 \end{aligned}
             \end{equation} 
             Moreover, we have that
             \begin{equation}
             \small
                 \begin{aligned}
                    & \Vert
                   \nabla_{\boldsymbol{w}} \mathcal{L}_{\mathcal{V}}(\boldsymbol{w}^{**}, c) -\nabla_{\boldsymbol{w}} \mathcal{L}_{\mathcal{V}}(\boldsymbol{w}^{*}, c)
             \Vert\\
            & \times
             \Vert
             [\nabla^{2}_{\boldsymbol{w}} \mathcal{L}_{\mathcal{S}}(\boldsymbol{w}^{\ast}, c)]^{-1}
             \Vert \le \frac{L_{H_{2}}\delta}{L_{H_{s}}},
                 \end{aligned}
             \end{equation} 
           \begin{equation}
           \small
               \begin{aligned}
                    & \Vert
              \nabla_{\boldsymbol{w}} \mathcal{L}_{\mathcal{V}}(\boldsymbol{w}^{**}, c)[\nabla^{2}_{\boldsymbol{w}} \mathcal{L}_{\mathcal{S}}(\boldsymbol{w}^{**}, c)]^{-1} 
              \\
              &- 
             \nabla_{\boldsymbol{w}} \mathcal{L}_{\mathcal{V}}(\boldsymbol{w}^{*}, c)[\nabla^{2}_{\boldsymbol{w}} \mathcal{L}_{\mathcal{S}}(\boldsymbol{w}^{\ast}, c)]^{-1}
             \Vert 
             \\
             &\le \frac{L_{\bar{G}}L_{G^{'}}\delta}{L_{H_{s}}^{2}} + \frac{L_{H_{2}}\delta}{L_{H_{s}}},
               \end{aligned}
           \end{equation}
       and
        \begin{equation}
         \begin{aligned}
        &\Vert
              \nabla_{\boldsymbol{w}} \mathcal{L}_{\mathcal{V}}(\boldsymbol{w}^{**}, c)[\nabla^{2}_{\boldsymbol{w}} \mathcal{L}_{\mathcal{S}}(\boldsymbol{w}^{**}, c)]^{-1}\nabla^{2}_{\boldsymbol{w}c} \mathcal{L}_{\mathcal{S}}(\boldsymbol{w}^{**}, c)- \\
             &\nabla_{\boldsymbol{w}} \mathcal{L}_{\mathcal{V}}(\boldsymbol{w}^{*}, c)[\nabla^{2}_{\boldsymbol{w}} \mathcal{L}_{\mathcal{S}}(\boldsymbol{w}^{\ast}, c)]^{-1}\nabla^{2}_{\boldsymbol{w}c} \mathcal{L}_{\mathcal{S}}(\boldsymbol{w}^{**}, c)
             \Vert\\
             &\le \frac{L_{\bar{G}}L_{G^{'}}L_{G_{3}}\delta}{L_{H_{s}}^{2}} + \frac{L_{G_{3}}L_{H_{2}}\delta}{L_{H_{s}}}.
            \end{aligned}
        \end{equation}
Now we consider the last terms in Eq.~\eqref{equation:G_{1}_G_fisrt_term},
\begin{equation}
\small
    \begin{aligned}
        & \Vert \nabla_{\boldsymbol{w}} \mathcal{L}_{\mathcal{V}}(\boldsymbol{w}^{*}, c)[\nabla^{2}_{\boldsymbol{w}} \mathcal{L}_{\mathcal{S}}(\boldsymbol{w}^{\ast}, c)]^{-1}\nabla^{2}_{\boldsymbol{w}c} \mathcal{L}_{\mathcal{S}}(\boldsymbol{w}^{**}, c)- \\
              & 
          \nabla_{\boldsymbol{w}} \mathcal{L}_{\mathcal{V}}(\boldsymbol{w}^{*}, c)[\nabla^{2}_{\boldsymbol{w}} \mathcal{L}_{\mathcal{S}}(\boldsymbol{w}^{\ast}, c)]^{-1}\nabla^{2}_{\boldsymbol{w}c} \mathcal{L}_{\mathcal{S}}(\boldsymbol{w}^{*}, c) \Vert
            \le \\
             &\Vert
           \nabla_{\boldsymbol{w}} \mathcal{L}_{\mathcal{V}}(\boldsymbol{w}^{*}, c)
             \Vert
             \Vert
             [\nabla^{2}_{\boldsymbol{w}} \mathcal{L}_{\mathcal{S}}(\boldsymbol{w}^{\ast}, c)]^{-1} \Vert\\
            &\times \Vert
            \nabla^{2}_{\boldsymbol{w}c} \mathcal{L}_{\mathcal{S}}(\boldsymbol{w}^{**}, c) - 
            \nabla^{2}_{\boldsymbol{w}c} \mathcal{L}_{\mathcal{S}}(\boldsymbol{w}^{*}, c)
            \Vert \le
            \frac{L_{\bar{G}}L_{G_{2}}\delta}{L_{H_{s}}}.
    \end{aligned}
\end{equation}
        In this way, the error bound of the first term to the right of inequality in Eq.~\eqref{equation:G_1_G} is
        \begin{equation}
            \begin{aligned}
               & \Vert
             \nabla_{\boldsymbol{w}} \mathcal{L}_{\mathcal{V}}(\boldsymbol{w}^{**}, c)[\nabla^{2}_{\boldsymbol{w}} \mathcal{L}_{\mathcal{S}}(\boldsymbol{w}^{**}, c)]^{-1}\nabla^{2}_{\boldsymbol{w}c} \mathcal{L}_{\mathcal{S}}(\boldsymbol{w}^{**}, c)- \\
             & 
          \nabla_{\boldsymbol{w}} \mathcal{L}_{\mathcal{V}}(\boldsymbol{w}^{*}, c)[\nabla^{2}_{\boldsymbol{w}} \mathcal{L}_{\mathcal{S}}(\boldsymbol{w}^{\ast}, c)]^{-1}\nabla^{2}_{\boldsymbol{w}c} \mathcal{L}_{\mathcal{S}}(\boldsymbol{w}^{*}, c)
            \Vert
            \\
            &\le 
            \frac{L_{\bar{G}}L_{G^{'}}L_{G_{3}}\delta}{L_{H_{s}}^{2}} + \frac{L_{G_{3}}L_{H_{2}}\delta}{L_{H_{s}}} +  \frac{L_{\bar{G}}L_{G_{2}}\delta}{L_{H_{s}}}.
            \end{aligned}
        \end{equation}
        Next, we consider the second term to the right of inequality in Eq.~\eqref{equation:G_1_G}, \textit{i.e.},
         \begin{equation}
        \small
            \begin{aligned}
                &\Vert
           \nabla_{\boldsymbol{w}} \mathcal{L}_{\mathcal{V}}(\boldsymbol{w}^{*}, c)[\nabla^{2}_{\boldsymbol{w}} \mathcal{L}_{\mathcal{S}}(\boldsymbol{w}^{\ast}, c)]^{-1}\nabla^{2}_{\boldsymbol{w}c} \mathcal{L}_{\mathcal{S}}(\boldsymbol{w}^{*}, c)- \\
             & \nabla_{\boldsymbol{w}} \mathcal{L}_{\mathcal{V}}(\boldsymbol{w}^{*}, c)
             \sum_{j=0}^{J}[\boldsymbol{I} -\nabla^{2}_{\boldsymbol{w}} \mathcal{L}_{\mathcal{S}}(\boldsymbol{w}^{\ast}, c)]^{j}
            \nabla^{2}_{\boldsymbol{w}c} \mathcal{L}_{\mathcal{S}}(\boldsymbol{w}^{*}, c)
            \Vert\\
           & \le 
            \Vert
              \nabla_{\boldsymbol{w}} \mathcal{L}_{\mathcal{V}}(\boldsymbol{w}^{*}, c)
            \Vert
            \Vert
            [\nabla^{2}_{\boldsymbol{w}} \mathcal{L}_{\mathcal{S}}(\boldsymbol{w}^{\ast}, c)]^{-1} \\
            &-
            \sum_{j=0}^{J}[\boldsymbol{I} -\nabla^{2}_{\boldsymbol{w}} \mathcal{L}_{\mathcal{S}}(\boldsymbol{w}^{\ast}, c)]^{j}
            \Vert
            \Vert
           \nabla^{2}_{\boldsymbol{w}c} \mathcal{L}_{\mathcal{S}}(\boldsymbol{w}^{*}, c)
            \Vert,
            \end{aligned}
        \end{equation}
        and then it satisfies that
         \begin{equation}
        \small
            \begin{aligned}
                &\Vert
           \nabla_{\boldsymbol{w}} \mathcal{L}_{\mathcal{V}}(\boldsymbol{w}^{*}, c)[\nabla^{2}_{\boldsymbol{w}} \mathcal{L}_{\mathcal{S}}(\boldsymbol{w}^{\ast}, c)]^{-1}\nabla^{2}_{\boldsymbol{w}c} \mathcal{L}_{\mathcal{S}}(\boldsymbol{w}^{*}, c)- \\
             & \nabla_{\boldsymbol{w}} \mathcal{L}_{\mathcal{V}}(\boldsymbol{w}^{*}, c)
             \sum_{j=0}^{J}[\boldsymbol{I} -\nabla^{2}_{\boldsymbol{w}} \mathcal{L}_{\mathcal{S}}(\boldsymbol{w}^{\ast}, c)]^{j}
            \nabla^{2}_{\boldsymbol{w}c} \mathcal{L}_{\mathcal{S}}(\boldsymbol{w}^{*}, c)
            \Vert\\
            & \le
            L_{\bar{G}}L_{G_{3}} \Vert
            [\nabla^{2}_{\boldsymbol{w}} \mathcal{L}_{\mathcal{S}}(\boldsymbol{w}^{\ast}, c)]^{-1}
            -
            \sum_{j=0}^{J}[\boldsymbol{I} -\nabla^{2}_{\boldsymbol{w}} \mathcal{L}_{\mathcal{S}}(\boldsymbol{w}^{\ast}, c)]^{j}
            \Vert\\
           & \le L_{\bar{G}}L_{G_{3}}\left(\frac{1}{L_{H_{s}}}+\frac{(1+L_{H})^{J}-1}{L_{H}}\right)\\
             &= \frac{L_{\bar{G}}L_{G_{3}}\left(L_{H}-L_{H_{s}}+L_{H_{s}}(1+L_{H})^{J}\right)}{L_{H}L_{H_{s}}}.
            \end{aligned}
        \end{equation}

        In this way, the error bound between $\boldsymbol{U}_{1}^{*}$ and $\boldsymbol{U}_{1}^{'}$
        is given by
        \begin{equation}
        % \footnotesize
        \small
            \begin{aligned}
               & \Vert \boldsymbol{U}_{1}^{*} - \boldsymbol{U}_{1}^{'} \Vert\\
                 &\le
                 \frac{L_{\bar{G}}L_{G^{'}}L_{G_{3}} + L_{G_{3}}L_{H_{2}}H + L_{\bar{G}}L_{G_{2}}L_{H}}{L_{H_{s}}^{2}} \delta \\
                 &+ \frac{L_{\bar{G}}L_{G_{3}}\left(L_{H}-L_{H_{s}}+L_{H_{s}}(1+L_{H})^{J}\right)}{L_{H}L_{H_{s}}}.
            \end{aligned}
        \end{equation}
        By denoting
        \begin{equation}
        \small
            \begin{aligned}
                &\mathcal{E}_{cur_{1}}^{'} \triangleq \frac{L_{\bar{G}}L_{G^{'}}L_{G_{3}} + L_{G_{3}}L_{H_{2}}L_{H} + L_{\bar{G}}L_{G_{2}}L_{H}}{L_{H_{s}}^{2}} \\
                &\mathcal{E}_{cur_{1}}^{''} \triangleq 
                \frac{L_{\bar{G}}L_{G_{3}}\left(L_{H}-L_{H_{s}}+L_{H_{s}}(1+L_{H})^{J}\right)}{L_{H}L_{H_{s}}}
            \end{aligned},
        \end{equation}
     we have proved Lemma~\ref{Theorem:approximation error_G_1}.
   %   {
   %      \color{blue}
   %      By analyzing the upper bound of $ \Vert 
			% \boldsymbol{U}_{1}^{*} - \boldsymbol{U}_{1}^{'}
			% \Vert$, we demonstrate that the approximation way of $\boldsymbol{U}_{1}^{'}$ in our method is rational.
   %   }
\end{proof}

\noindent \emph{Approximation error of $\boldsymbol{U}_{2}$.}
We adopt an approximation way to simplify the computation of $\boldsymbol{U}_{2}$, which has been presented in Eq.~\eqref{equation:gradient_G_2_approx}. The generalization error for $\boldsymbol{U}_{2}$ is presented in the following lemma.
\begin{lemma}
\label{theorem:approximation_error_2}
	Supposing that the assumptions holds, the approximated $\boldsymbol{U}_{2}^{'}$ satisfies that
	\begin{equation}
    \small
		\begin{aligned}
			\Vert \boldsymbol{U}_{2} -\boldsymbol{U}_{2}^{'}  \Vert
			\le
			\mathcal{E}_{cur_{2}},
		\end{aligned}
	\end{equation}
	where $\mathcal{E}_{cur_{2}}$ is a constant that only depends on the Lipschitz constants, which is computed as 
 \begin{equation}
 \small
    \begin{aligned}
        \mathcal{E}_{cur_{2}} \triangleq \frac{2 \rho KL_{G} L_{G^{'}}L_{G_{3}}}{L_{H_{s}}}.
    \end{aligned}
\end{equation}
	\label{Theorem:approximation error_G_2}
\end{lemma}

\begin{proof}

In the following proof, we mainly leverage the Lipschitz continuity assumptions of $\mathcal{L}_{\mathcal{S}}$,  $\nabla^{2}_{\boldsymbol{w}} \mathcal{L}_{\mathcal{S}}(\boldsymbol{w}, c)$, and $\nabla_{\boldsymbol{w}} \mathcal{L}_{\mathcal{S}}(\boldsymbol{w}, c)$, and the upper bound of the norm of $\boldsymbol{\epsilon}$. 

    Review that $\boldsymbol{U}_{2}$ is computed as
    \begin{equation}
    \small
        \begin{aligned}
            \boldsymbol{U}_{2} &= -2K(1-\Vert \nabla_{\boldsymbol{w}} \mathcal{L}_{\mathcal{S}}(\hat{\boldsymbol{w}}, c) \Vert)^{K-1} \nabla_{\boldsymbol{w}} \mathcal{L}_{\mathcal{S}}(\hat{\boldsymbol{w}}, c)  \\
		&\times \nabla^{2}_{\boldsymbol{w}} \mathcal{L}_{\mathcal{S}}(\hat{\boldsymbol{w}}, c) [\nabla^{2}_{\boldsymbol{w}} \mathcal{L}_{\mathcal{S}}(\boldsymbol{w}^{\ast}, c)]^{-1} 
		\nabla^{2}_{\boldsymbol{w}c} \mathcal{L}_{\mathcal{S}}(\boldsymbol{w}^{*}, c),
        \end{aligned}
    \end{equation} 
    while $\boldsymbol{U}_{2}^{'}$ is computed as
    \begin{equation}
    \small
	\label{equation:gradient_G_2_approx_sp}
	\begin{aligned}
		  \boldsymbol{U}^{'}_{2} &=
		-2K(1-\Vert \nabla_{\boldsymbol{w}} \mathcal{L}_{\mathcal{S}}(\hat{\boldsymbol{w}}, c) \Vert)^{K-1} \\
        & \times
        \nabla_{\boldsymbol{w}} \mathcal{L}_{\mathcal{S}}(\hat{\boldsymbol{w}}, c)  
		\nabla^{2}_{\boldsymbol{w}c} \mathcal{L}_{\mathcal{S}}(\boldsymbol{w}^{*}, c).
	\end{aligned}
\end{equation}
The approximation error is computed as 
\begin{equation}
\footnotesize
    \begin{aligned}
       & \Vert \boldsymbol{U}_{2} -\boldsymbol{U}_{2}^{'}  \Vert
	% &=
 %        \Big \Vert
 %        2K(1-\Vert \nabla_{\boldsymbol{w}} \mathcal{L}_{\mathcal{S}}(\hat{\boldsymbol{w}}, c) \Vert)^{K-1} \nabla_{\boldsymbol{w}} \mathcal{L}_{\mathcal{S}}(\hat{\boldsymbol{w}}, c) \nabla^{2}_{\boldsymbol{w}} \mathcal{L}_{\mathcal{S}}(\hat{\boldsymbol{w}}, c) \\
	% 	&\times [\nabla^{2}_{\boldsymbol{w}} \mathcal{L}_{\mathcal{S}}(\boldsymbol{w}^{\ast}, c)]^{-1} 
	% 	\nabla^{2}_{\boldsymbol{w}c} \mathcal{L}_{\mathcal{S}}(\boldsymbol{w}^{*}, c) \\
 %        &- 2K(1-\Vert \nabla_{\boldsymbol{w}} \mathcal{L}_{\mathcal{S}}(\hat{\boldsymbol{w}}, c) \Vert)^{K-1} \nabla_{\boldsymbol{w}} \mathcal{L}_{\mathcal{S}}(\hat{\boldsymbol{w}}, c)  
	% 	\nabla^{2}_{\boldsymbol{w}c} \mathcal{L}_{\mathcal{S}}(\boldsymbol{w}^{*}, c)
 %        \Big \Vert \\
        =
       \big (  2K(1-\Vert \nabla_{\boldsymbol{w}} \mathcal{L}_{\mathcal{S}}(\hat{\boldsymbol{w}}, c) \Vert)^{K-1}  \big) \times \\
      & \Big \Vert 
        \nabla_{\boldsymbol{w}} \mathcal{L}_{\mathcal{S}}(\hat{\boldsymbol{w}}, c)  
       \nabla^{2}_{\boldsymbol{w}} \mathcal{L}_{\mathcal{S}}(\hat{\boldsymbol{w}}, c) [\nabla^{2}_{\boldsymbol{w}} \mathcal{L}_{\mathcal{S}}(\boldsymbol{w}^{\ast}, c)]^{-1} 
		\nabla^{2}_{\boldsymbol{w}c} \mathcal{L}_{\mathcal{S}}(\boldsymbol{w}^{*}, c) \\
  &- \nabla_{\boldsymbol{w}} \mathcal{L}_{\mathcal{S}}(\hat{\boldsymbol{w}}, c)  
   \nabla^{2}_{\boldsymbol{w}} \mathcal{L}_{\mathcal{S}}(\boldsymbol{w}^{\ast}, c) [\nabla^{2}_{\boldsymbol{w}} \mathcal{L}_{\mathcal{S}}(\boldsymbol{w}^{\ast}, c)]^{-1} 
		\nabla^{2}_{\boldsymbol{w}c} \mathcal{L}_{\mathcal{S}}(\boldsymbol{w}^{*}, c)
       \Big  \Vert.
    \end{aligned}
\end{equation}
Then the approximation error satisfies that
\begin{equation}
\small
    \begin{aligned}
     &\Vert \boldsymbol{U}_{2} -\boldsymbol{U}_{2}^{'}  \Vert 
      \le
        \big (  2K(1-\Vert \nabla_{\boldsymbol{w}} \mathcal{L}_{\mathcal{S}}(\hat{\boldsymbol{w}}, c) \Vert)^{K-1}  \big)
       \\
        & \times
        \Vert
            \nabla^{2}_{\boldsymbol{w}} \mathcal{L}_{\mathcal{S}}(\hat{\boldsymbol{w}}, c) -
             \nabla^{2}_{\boldsymbol{w}} \mathcal{L}_{\mathcal{S}}(\boldsymbol{w}^{\ast}, c)
        \Vert
        \Vert
        [\nabla^{2}_{\boldsymbol{w}} \mathcal{L}_{\mathcal{S}}(\boldsymbol{w}^{\ast}, c)]^{-1} 
        \Vert\\
        &\times
        \Vert
            \nabla^{2}_{\boldsymbol{w}c} \mathcal{L}_{\mathcal{S}}(\boldsymbol{w}^{*}, c)
        \Vert   \Vert 
            \nabla_{\boldsymbol{w}} \mathcal{L}_{\mathcal{S}}(\hat{\boldsymbol{w}}, c)  
        \Vert \\
        & \le
        2K\frac{L_{G} L_{G^{'}}L_{G_{3}}}{L_{H_{s}}}\Vert \hat{\boldsymbol{w}} - \boldsymbol{w}^{*} \Vert \le   \frac{2 \rho KL_{G} L_{G^{'}}L_{G_{3}}}{L_{H_{s}}}.
    \end{aligned}
\end{equation}
By denoting 
\begin{equation}
\small
    \begin{aligned}
        \mathcal{E}_{cur_{2}} \triangleq \frac{2 \rho KL_{G} L_{G^{'}}L_{G_{3}}}{L_{H_{s}}},
    \end{aligned}
\end{equation}
we have proved  Lemma~\ref{Theorem:approximation error_G_2}.
\end{proof}

After proving Lemma~\ref{Theorem:approximation error_G_1} and Lemma~\ref{Theorem:approximation error_G_2}, we can obtain the upper bound of the approximation error in $\hat{\nabla }c$, as shown in Theorem~\ref{Theorem:approximation_error_c}.
\begin{theorem}
	\label{Theorem:approximation_error_c}
	Suppose assumptions hold, the approximated gradients of curvatures $\hat{\nabla }c$ satisfy
	\begin{equation}
		\small
		\begin{aligned}
			\Vert 
			\nabla c - \hat{\nabla} c
			\Vert
			&\le \Vert \boldsymbol{U}_{1} - \boldsymbol{U}_{1}^{'} \Vert + \Vert \boldsymbol{U}_{2} - \boldsymbol{U}_{2}^{'} \Vert \\
            &\le \mathcal{E}_{cur_{1}} + \mathcal{E}_{cur_{2}},
		\end{aligned}
	\end{equation}
	where $\mathcal{E}_{cur_{1}}$ and $\mathcal{E}_{cur_{2}}$ are constants and have been defined in Lemma~\ref{Theorem:approximation error_G_1} and Lemma~\ref{Theorem:approximation error_G_2}, respectively. 
        % \begin{equation}
        % \small
        %     \begin{aligned}
        %         \mathcal{E}_{cur_{1}} & \triangleq  \frac{L_{\bar{G}}L_{G^{'}}L_{G_{3}} + L_{G_{3}}L_{H_{2}}L_{H} + L_{\bar{G}}L_{G_{2}}L_{H}}{L_{H_{s}}^{2}} \delta \\
        %         &+ 
        %         \frac{L_{\bar{G}}L_{G_{3}}\left(L_{H}-L_{H_{s}}+L_{H_{s}}(1+L_{H})^{J}\right)}{HL_{H_{s}}}, \\
        %         \mathcal{E}_{cur_{2}} &\triangleq \frac{2 \rho KL_{G} L_{G^{'}}L_{G_{3}}}{L_{H_{s}}}.
        %     \end{aligned}
        % \end{equation}
\end{theorem}

By the aforementioned analyses, we demonstrate the approximation error in gradients of curvatures is upper-bounded, thus approximation ways in our method are rational.

\subsection{Convergence Analyses}
\label{section:convergence_analysis}
We establish the convergence analyses of the parameters $\boldsymbol{w}$ and the curvatures $c$, which are presented in Theorem~\ref{Theorem:Convergence_analysis_M} and  Theorem~\ref{Theorem:Convergence_analysis_c}, respectively.
Theoretical analyses show that parameters $\boldsymbol{w}$ and the curvatures $c$ can converge, thereby demonstrating the effectiveness of our method.

In Theorem~\ref{Theorem:Convergence_analysis_M}, we bound the gradient of parameters $\boldsymbol{w}$ to analyze the convergence. 

\begin{theorem}
	Set the iteration number of training parameters of HNNs as $T$, and choose step-size of parameters $\boldsymbol{w}$ as $\eta^{(t)} = \frac{\eta^{(0)}}{\sqrt{t}}$ and perturbation radius $\hat{\rho}^{(t)}$ satisfies $ \frac{\hat{\rho}^{(0)}}{\sqrt{t}} $, parameters $\boldsymbol{w}$ satisfy
	\begin{equation}
		\small
		\begin{aligned}
			\min_{t\in [1,...,T]} \big\Vert \nabla_{\boldsymbol{w}} \mathcal{L}_{\mathcal{S}} (\boldsymbol{w}^{(t)}, c) \big\Vert^{2} & \le \frac{\mathcal{C}_{w_{1}}}{\ln T} + \frac{\mathcal{C}_{w_{2}}(1+\ln T)}{\ln T},
		\end{aligned}
	\end{equation}
	where $\mathcal{C}_{w_{1}}$ and $\mathcal{C}_{w_{2}}$  are constants and computed as
        \begin{equation}
% \footnotesize
\small
    \begin{aligned}
        &\mathcal{C}_{w_{1}} =  \frac{\beta_{\max}}{\eta^{(0)}} \left( \mathcal{L}_{\mathcal{S}}(\boldsymbol{w}^{(0)}, c) - \mathcal{L}_{\mathcal{S}}(\boldsymbol{w}^{(T+1)}, c)\right)  \\
        & \mathcal{C}_{w_{2}} = \frac{\beta_{\max}}{\eta^{(0)}} \left( \frac{1}{\beta_{\min}}L_{H}L_{G}\eta^{(0)}\hat{\rho}^{(0)} + \frac{1}{2} L_{H}L_{G}^{2} \eta^{(0)2}\right).
    \end{aligned}
\end{equation}

% \begin{equation}
% \small
%     \begin{aligned}
%         &\mathcal{C}_{w_{1}} = \mathcal{L}_{\mathcal{S}}(\boldsymbol{w}^{(0)}, c) -  \mathcal{L}_{\mathcal{S}}^{\min}  + \frac{1}{\beta_{\min}}HG\eta^{(0)}\hat{\rho}^{(0)} + \frac{1}{2} HG^{2} \eta^{(0)2}  \\
%         & \mathcal{C}_{w_{2}} = \frac{1}{\beta_{\min}}HG\eta^{(0)}\hat{\rho}^{(0)} + \frac{1}{2} HG^{2} \eta^{(0)2},
%     \end{aligned}
% \end{equation}

	\label{Theorem:Convergence_analysis_M}
\end{theorem}

\begin{proof}
    Recall that the parameters of HNNs are updated as
            \begin{equation}
            \small
        	\begin{aligned}
        		&\boldsymbol{w} \leftarrow \boldsymbol{w} - \eta \nabla_{\boldsymbol{w}} \mathcal{L}_{\mathcal{S}} (\boldsymbol{w}^{'}, c),\\
         &  \text{where } \boldsymbol{w}^{'} = \boldsymbol{w} + \hat{\rho}
         \nabla_{\boldsymbol{w}} \mathcal{L}_{\mathcal{S}} (\boldsymbol{w}, c)/\Vert  \nabla_{\boldsymbol{w}} \mathcal{L}_{\mathcal{S}} (\boldsymbol{w}, c) \Vert.
        	\end{aligned}
        \end{equation}
    To bound the gradients of parameters $\boldsymbol{w}$, we utilize the Lipschitz continuity assumptions of 
    $\mathcal{L}_{\mathcal{S}}$, $\mathcal{L}_{\mathcal{V}}$ and $\nabla_{\boldsymbol{w}} \mathcal{L}_{\mathcal{S}}(\boldsymbol{w}, c)$, and establish the connection between the update vector $\nabla_{\boldsymbol{w}} \mathcal{L}_{\mathcal{S}} (\boldsymbol{w}^{'}, c)$ and gradients $ \nabla_{\boldsymbol{w}} \mathcal{L}_{\mathcal{S}} (\boldsymbol{w}, c)$  by decomposing the update vector.
    It follows that
    \begin{equation}
    \small
        \begin{aligned}
            &\mathcal{L}_{\mathcal{S}}(\boldsymbol{w}^{(t+1)}, c) \le  \mathcal{L}_{\mathcal{S}}(\boldsymbol{w}^{(t)}, c) + \frac{L_{H}}{2} \Vert \boldsymbol{w}^{(t+1)} - \boldsymbol{w}^{(t)} \Vert^{2} \\
            &
             +
             (\nabla_{\boldsymbol{w}} \mathcal{L}_{\mathcal{S}} (\boldsymbol{w}^{(t)}, c))^{\top} (\boldsymbol{w}^{(t+1)} - \boldsymbol{w}^{(t)}) \\
             & = \mathcal{L}_{\mathcal{S}}(\boldsymbol{w}^{(t)}, c) +
             \frac{L_{H}}{2} \Vert \eta^{(t)} \nabla_{\boldsymbol{w}} \mathcal{L}_{\mathcal{S}} (\boldsymbol{w}^{'}, c) \Vert^{2}
              \\
             &- (\nabla_{\boldsymbol{w}} \mathcal{L}_{\mathcal{S}} (\boldsymbol{w}^{(t)}, c))^{\top} 
             \eta^{(t)} \nabla_{\boldsymbol{w}} \mathcal{L}_{\mathcal{S}} (\boldsymbol{w}^{'}, c) .
        \end{aligned}
    \end{equation}
    Then, we have
    \begin{equation}
    \small
        \begin{aligned}
            & \mathcal{L}_{\mathcal{S}}(\boldsymbol{w}^{(t+1)}, c)
            -
            \mathcal{L}_{\mathcal{S}}(\boldsymbol{w}^{(t)}, c) \le
            \frac{L_{H}}{2}\eta^{(t)2} \Vert  \nabla_{\boldsymbol{w}} \mathcal{L}_{\mathcal{S}} (\boldsymbol{w}^{'}, c) \Vert^{2}
            \\
            & -\eta^{(t)} (\nabla_{\boldsymbol{w}} \mathcal{L}_{\mathcal{S}} (\boldsymbol{w}^{(t)}, c))^{\top} 
              \nabla_{\boldsymbol{w}} \mathcal{L}_{\mathcal{S}} (\boldsymbol{w}^{'}, c) 
             .
        \end{aligned}
    \end{equation}
    We decompose vectors $\nabla_{\boldsymbol{w}} \mathcal{L}_{\mathcal{S}} (\boldsymbol{w}^{'}, c)$ into parallel direction $\nabla f_{\parallel}^{(t)}$ of vector$ \nabla_{\boldsymbol{w}} \mathcal{L}_{\mathcal{S}} (\boldsymbol{w}, c)$ and orthogonal direction $\nabla f_{\perp}^{(t)}$, \textit{i.e.},
    \begin{equation}
    \small
        \begin{aligned}
            \nabla f_{\parallel}^{(t)} = \beta_{t}  \nabla_{\boldsymbol{w}} \mathcal{L}_{\mathcal{S}} (\boldsymbol{w}, c), \nabla f_{\perp}^{(t)} =  \nabla_{\boldsymbol{w}} \mathcal{L}_{\mathcal{S}} (\boldsymbol{w}, c) - \nabla f_{\parallel}^{(t)}.
        \end{aligned}
    \end{equation}
We have that
\begin{equation}
\small
    \begin{aligned}
        &\left\langle \nabla_{\boldsymbol{w}} \mathcal{L}_{\mathcal{S}} (\boldsymbol{w}^{(t)}, c),  \nabla_{\boldsymbol{w}} \mathcal{L}_{\mathcal{S}} (\boldsymbol{w}^{'}, c) \right \rangle \\
        % &=\left\langle \nabla_{\boldsymbol{w}} \mathcal{L}_{\mathcal{S}} (\boldsymbol{w}^{(t)}, c),  \frac{1}{\beta}\nabla f_{\parallel}^{(t)} \right \rangle\\
        & = \left\langle \nabla_{\boldsymbol{w}} \mathcal{L}_{\mathcal{S}} (\boldsymbol{w}^{(t)}, c),  \frac{1}{\beta_{t}} \big( \nabla_{\boldsymbol{w}} \mathcal{L}_{\mathcal{S}} (\boldsymbol{w}, c) - f_{\perp}^{(t)}\big) \right \rangle\\
        % &=\left\langle \nabla_{\boldsymbol{w}} \mathcal{L}_{\mathcal{S}} (\boldsymbol{w}^{(t)}, c),  \frac{1}{\beta_{t}}  \nabla_{\boldsymbol{w}} \mathcal{L}_{\mathcal{S}} (\boldsymbol{w}, c) \right \rangle
        %     -
        %     \left\langle \nabla_{\boldsymbol{w}} \mathcal{L}_{\mathcal{S}} (\boldsymbol{w}^{(t)}, c), \frac{1}{\beta_{t}} f_{\perp}^{(t)}\right \rangle\\
        % &= \frac{1}{\beta_{t}} \Vert  \nabla_{\boldsymbol{w}} \mathcal{L}_{\mathcal{S}} (\boldsymbol{w}^{(t)}, c) \Vert^{2}  -   \frac{1}{\beta_{t}}\left\langle \nabla_{\boldsymbol{w}} \mathcal{L}_{\mathcal{S}} (\boldsymbol{w}^{(t)}, c), f_{\perp}^{(t)}\right \rangle\\
        % & = \frac{1}{\beta_{t}} \Vert  \nabla_{\boldsymbol{w}} \mathcal{L}_{\mathcal{S}} (\boldsymbol{w}^{(t)}, c) \Vert^{2}  -   \frac{1}{\beta_{t}}\left\langle \nabla_{\boldsymbol{w}} \mathcal{L}_{\mathcal{S}} (\boldsymbol{w}^{(t)}, c),   \nabla_{\boldsymbol{w}} \mathcal{L}_{\mathcal{S}} (\boldsymbol{w}^{(t)}, c) \rm{sin}(\theta_{t}) \right \rangle\\
        &=\frac{1}{\beta_{t}} \Vert  \nabla_{\boldsymbol{w}} \mathcal{L}_{\mathcal{S}} (\boldsymbol{w}^{(t)}, c) \Vert^{2}  -   \frac{1}{\beta_{t}}   \Vert  \nabla_{\boldsymbol{w}} \mathcal{L}_{\mathcal{S}} (\boldsymbol{w}^{(t)}, c) \Vert^{2} \\
        &\times  (|\rm{tanh}(\theta_{t})|+ O(\theta^{2}_{t})),
    \end{aligned}
\end{equation}
    where $\rm{sin}(\theta_{t})$ is the angle between $ \nabla_{\boldsymbol{w}} \mathcal{L}_{\mathcal{S}} (\boldsymbol{w}^{(t)}, c)$ and   $ \nabla_{\boldsymbol{w}} \mathcal{L}_{\mathcal{S}} (\boldsymbol{w}^{'(t)}, c)$.
    Moreover, from assumptions, we have
    \begin{equation}
    \small
        \begin{aligned}
            |\rm{tan} \theta_{t}| &\le \frac{\Vert \nabla_{\boldsymbol{w}} \mathcal{L}_{\mathcal{S}} (\boldsymbol{w}^{(t)}, c) - \nabla_{\boldsymbol{w}} \mathcal{L}_{\mathcal{S}} (\boldsymbol{w}^{'(t)}, c)\Vert}{
            \Vert 
            \nabla_{\boldsymbol{w}} \mathcal{L}_{\mathcal{S}} (\boldsymbol{w}^{(t)}, c)
            \Vert
            }\\
            &\le \frac{\hat{\rho}^{t} L_{H} \Vert \frac{  \nabla_{\boldsymbol{w}} \mathcal{L}_{\mathcal{S}} (\boldsymbol{w}^{(t)}, c)}{\Vert 
            \nabla_{\boldsymbol{w}} \mathcal{L}_{\mathcal{S}} (\boldsymbol{w}^{(t)}, c)
            \Vert} \Vert}
            { \Vert 
            \nabla_{\boldsymbol{w}} \mathcal{L}_{\mathcal{S}} (\boldsymbol{w}^{(t)}, c)
            \Vert} =
            \frac{\hat{\rho}^{t}L_{H}}
            { \Vert \nabla_{\boldsymbol{w}} \mathcal{L}_{\mathcal{S}} (\boldsymbol{w}^{(t)}, c) \Vert  }.
        \end{aligned}
    \end{equation}
    Moreover, we have that
    \begin{equation}
    \small
        \begin{aligned}
            &\mathcal{L}_{\mathcal{S}}(\boldsymbol{w}^{(t+1)}, c) -  \mathcal{L}_{\mathcal{S}}(\boldsymbol{w}^{(t)}, c) 
            \\
            &\le
            -\eta^{(t)} \left\langle \nabla_{\boldsymbol{w}} \mathcal{L}_{\mathcal{S}} (\boldsymbol{w}^{(t)}, c),  \nabla_{\boldsymbol{w}} \mathcal{L}_{\mathcal{S}} (\boldsymbol{w}^{'(t)}, c) \right \rangle \\
            &+ \frac{1}{2}L_{H} \eta^{(t)2}\Vert \nabla_{\boldsymbol{w}} \mathcal{L}_{\mathcal{S}} (\boldsymbol{w}^{'(t)}, c) \Vert^{2}.
        \end{aligned}
    \end{equation}
    Hence, we can obtain that
    \begin{equation}
    \footnotesize
        \begin{aligned}
            &\mathcal{L}_{\mathcal{S}}(\boldsymbol{w}^{(t+1)}, c) -  \mathcal{L}_{\mathcal{S}}(\boldsymbol{w}^{(t)}, c) 
            \\
            &\le
              \frac{\eta^{(t)}}{\beta_{t}} \Vert  \nabla_{\boldsymbol{w}} \mathcal{L}_{\mathcal{S}} (\boldsymbol{w}^{(t)}, c)  \Vert^{2} |\rm{tan}(\theta_{t})| \\
            &-\frac{\eta^{(t)}}{\beta_{t}} \Vert \nabla_{\boldsymbol{w}} \mathcal{L}_{\mathcal{S}} (\boldsymbol{w}^{(t)}, c) \Vert^{2} +\frac{1}{2}L_{H}\eta^{(t)2} \Vert \nabla_{\boldsymbol{w}} \mathcal{L}_{\mathcal{S}} (\boldsymbol{w}^{'(t)}, c) \Vert^{2}\\
            &\le 
             \frac{\eta^{(t)}}{\beta_{t}} L_{H} \hat{\rho}^{(t)}  \Vert  \nabla_{\boldsymbol{w}} \mathcal{L}_{\mathcal{S}} (\boldsymbol{w}^{(t)}, c)  \Vert \\
             &-\frac{\eta^{(t)}}{\beta_{t}} \Vert \nabla_{\boldsymbol{w}} \mathcal{L}_{\mathcal{S}} (\boldsymbol{w}^{(t)}, c) \Vert^{2} 
             +\frac{1}{2}L_{H}\eta^{(t)2} \Vert \nabla_{\boldsymbol{w}} \mathcal{L}_{\mathcal{S}} (\boldsymbol{w}^{'(t)}, c) \Vert^{2}.
        \end{aligned}
    \end{equation}
We further assume that 
\begin{equation}
\small
    \begin{aligned}
        0< \beta_{\min} \le \beta_{t} \le \beta_{\max},
    \end{aligned}
\end{equation}
and thus have
\begin{equation}
\label{equation:final_equation_1}
\small
    \begin{aligned}
       & \mathcal{L}_{\mathcal{S}}(\boldsymbol{w}^{(t+1)}, c) -  \mathcal{L}_{\mathcal{S}}(\boldsymbol{w}^{(t)}, c) \\
       & \le
             -\frac{\eta^{(t)}}{\beta_{\max}} \Vert \nabla_{\boldsymbol{w}} \mathcal{L}_{\mathcal{S}} (\boldsymbol{w}^{(t)}, c) \Vert^{2} 
              \\
             &+ \frac{1}{\beta_{\min}}L_{H}\hat{\rho}^{(t)}\eta^{(t)}L_{G}+ \frac{1}{2} L_{H}\eta^{(t)2}L_{G}^{2}.
    \end{aligned}
\end{equation}
By rearranging the term in Eq.~\eqref{equation:final_equation_1},
we have
\begin{equation}
\small
    \begin{aligned}
       & \frac{\eta^{(t)}}{\beta_{\max}} \Vert \nabla_{\boldsymbol{w}} \mathcal{L}_{\mathcal{S}} (\boldsymbol{w}^{(t)}, c) \Vert^{2} \\
        &\le    \mathcal{L}_{\mathcal{S}}(\boldsymbol{w}^{(t)}, c) -  \mathcal{L}_{\mathcal{S}}(\boldsymbol{w}^{(t+1)}, c) \\
        & +  \frac{1}{\beta_{\min}}L_{H}\hat{\rho}^{(t)}\eta^{(t)}L_{G}  + \frac{1}{2} L_{H}\eta^{(t)2}L_{G}^{2}.
    \end{aligned}
\end{equation}
Telescoping from t=1 to T, we have
\begin{equation}
\footnotesize
    \begin{aligned}
        &\frac{1}{\beta_{\max}} \sum_{t=1}^{T} \eta^{(t)} \Vert \nabla_{\boldsymbol{w}} \mathcal{L}_{\mathcal{S}} (\boldsymbol{w}^{(t)}, c) \Vert^{2} 
        \\
        &\le \mathcal{L}_{\mathcal{S}}(\boldsymbol{w}^{(1)}, c) - \mathcal{L}_{\mathcal{S}}(\boldsymbol{w}^{(T+1)}, c)\\
        & + \frac{1}{\beta_{\min}}L_{H}\hat{\rho}^{(t)}\eta^{(t)}L_{G} \sum_{t=1}^{T} \hat{\rho}^{(t)}\eta^{(t)} + \frac{1}{2} L_{H}L_{G}^{2} \sum_{t=1}^{T} \eta^{(t)2}.
    \end{aligned}
\end{equation}
Let $\eta^{(t)} = \frac{\eta^{(0)}}{\sqrt{t}}$ and $\hat{\rho}^{(t)} = \frac{\hat{\rho}^{(0)}}{\sqrt{t}}$, we have that 
\begin{equation}
\footnotesize
    \begin{aligned}
          &\frac{\eta^{(0)}}{\beta_{\max}}  \sum_{t=1}^{T} \frac{1}{\sqrt{t}} \Vert \nabla_{\boldsymbol{w}} \mathcal{L}_{\mathcal{S}} (\boldsymbol{w}^{(t)}, c) \Vert^{2}  \\
          & \le \mathcal{L}_{\mathcal{S}}(\boldsymbol{w}^{(0)}, c) -  \mathcal{L}_{\mathcal{S}}(\boldsymbol{w}^{(T+1)}, c)  \\
          &\frac{1}{\beta_{\min}}L_{H}L_{G}\eta^{(0)}\hat{\rho}^{(0)} (1+\ln T) + \frac{1}{2} L_{H}L_{G}^{2} \eta^{(0)2}(1+\ln T).
    \end{aligned}
\end{equation}
Because we have 
\begin{equation}
\small
    \begin{aligned}
        &\sum_{t=1}^{T} \frac{1}{\sqrt{t}} \Vert \nabla_{\boldsymbol{w}} \mathcal{L}_{\mathcal{S}} (\boldsymbol{w}^{(t)}, c) \Vert^{2} \\
        &\ge  \sum_{t=1}^{T} \frac{1}{\sqrt{t}} \min_{t\in [1,...,T]}\Vert \nabla_{\boldsymbol{w}} \mathcal{L}_{\mathcal{S}} (\boldsymbol{w}^{(t)}, c) \Vert^{2} \\
        & \ge \ln T \min_{t\in [1,...,T]}\Vert \nabla_{\boldsymbol{w}} \mathcal{L}_{\mathcal{S}} (\boldsymbol{w}^{(t)}, c) \Vert^{2},
    \end{aligned}
\end{equation}
it holds that
\begin{equation}
\footnotesize
    \begin{aligned}
        &\min_{t\in [1,...,T]} \Vert \nabla_{\boldsymbol{w}} \mathcal{L}_{\mathcal{S}} (\boldsymbol{w}^{(t)}, c) \Vert^{2} \le \\
       & +
        \frac{\beta_{\max}}{\eta^{(0)}} \frac{1}{\ln T} \left(  
        \mathcal{L}_{\mathcal{S}}(\boldsymbol{w}^{(0)}, c) - \mathcal{L}_{\mathcal{S}}(\boldsymbol{w}^{(T+1)}, c) \right) \\
        &+\frac{\beta_{\max}}{\eta^{(0)}} \frac{(1+\ln T)}{\ln T}\left(\frac{1}{\beta_{\min}}L_{H}L_{G}\eta^{(0)}\hat{\rho}^{(0)} 
          + \frac{1}{2} L_{H}L_{G}^{2} \eta^{(0)2} \right)
       .
    \end{aligned}
\end{equation}
By denoting 
\begin{equation}
% \footnotesize
\small
    \begin{aligned}
        &\mathcal{C}_{w_{1}} =  \frac{\beta_{\max}}{\eta^{(0)}} \left( \mathcal{L}_{\mathcal{S}}(\boldsymbol{w}^{(0)}, c) -\mathcal{L}_{\mathcal{S}}(\boldsymbol{w}^{(T+1)}, c)\right)  \\
        & \mathcal{C}_{w_{2}} = \frac{\beta_{\max}}{\eta^{(0)}} \left( \frac{1}{\beta_{\min}}L_{H}L_{G}\eta^{(0)}\hat{\rho}^{(0)} + \frac{1}{2} L_{H}L_{G}^{2} \eta^{(0)2}\right),
    \end{aligned}
\end{equation}
 we obtain the theorem.
\end{proof}

In Theorem~\ref{Theorem:Convergence_analysis_c}, we present the convergence performance of the curvatures $c$ by providing the upper bound of $\frac{1}{\mathcal{T}} \sum_{t = 1}^{\mathcal{T}}
			\Big\| \nabla c^{(t)}  \Big\|^2$.
% The following theorem presents the convergence performance of the curvature $c$.
\begin{theorem}
	Set the iteration number of training curvatures as $\mathcal{T}$, and choose the step-size as $\frac{1}{8L_{H_{c}}} \le \eta^{(t)}_{c} \le \frac{1}{4L_{H_{c}}} $,  the obtained curvature $c$ satisfies
	\begin{equation}
		\small
            % \footnotesize
		\begin{aligned}
			\frac{1}{\mathcal{T}} \sum_{t = 1}^{\mathcal{T}}
			\Big\| \nabla c^{(t)}  \Big\|^2 & \le 
			\frac{64}{3} L_{H_{c}} \frac{\mathcal{F}(c^{(1)}) - \mathcal{F}(c^{(\mathcal{T}+1)})}{\mathcal{T}} + \mathcal{C}_{c},
		\end{aligned}
	\end{equation}
	where $\mathcal{C}_{c}$  is a constant and computed as 
 $\mathcal{C}_{c} \triangleq 9\mathcal{E}_{cur_{1}}^{2} + 9\mathcal{E}_{cur_{2}}^{2} $.
	\label{Theorem:Convergence_analysis_c}
\end{theorem}

\begin{proof}
To analyze the bound of gradients of curvatures, we utilize the Lipschitz continuity assumptions, and then leverage the connection between the update vector $\hat{\nabla}c^{(\tau)}$ and the gradients of curvatures $\nabla c^{(\tau)} $ by Theorem~\ref{Theorem:approximation_error_c}.
    From the assumption, we have that
    \begin{equation}
    \small
    \label{equation:convergence_c_0}
        \begin{aligned}
            &\mathcal{F}(c^{(\tau + 1)})
            % & \le \mathcal{F}(c^{(\tau))}) + \langle \nabla \mathcal{F}(c^{(\tau)}) , c^{(\tau+1)} - c^{(\tau)}  \rangle\\
            % &+ \frac{1}{2} ( c^{(\tau+1)} - c^{(\tau)})^{\top} \nabla^{2} \mathcal{F}(c^{(\tau)})  ( c^{(\tau+1)} - c^{(\tau)})\\
             \le 
            \mathcal{F}(c^{(\tau))}) - \eta^{(t)}_{c}\langle \nabla c^{(\tau)} ,  \hat{\nabla}c^{(\tau)} \rangle\\
            &+ \frac{1}{2} L_{H_{c}}\eta^{(t)2}_{c} \Vert c^{(\tau+1)} - c^{(\tau)} \Vert^{2}\\
            &=\mathcal{F}(c^{(\tau))}) + \frac{1}{2} L_{H_{c}}\eta^{(t)2}_{c} \Vert  \hat{\nabla}c^{(\tau)} \Vert^{2} \\
            &- \eta^{(t)}_{c}\langle \nabla c^{(\tau)} ,  \hat{\nabla}c^{(\tau)} -\nabla c^{(\tau)} +\nabla c^{(\tau)}  \rangle,\\
        \end{aligned}
    \end{equation}
    and we further have that
    \begin{equation}
    \small
    \label{equation:convergence_c_1}
        \begin{aligned}
            &\mathcal{F}(c^{(\tau + 1)})
            % & \le \mathcal{F}(c^{(\tau))}) + \langle \nabla \mathcal{F}(c^{(\tau)}) , c^{(\tau+1)} - c^{(\tau)}  \rangle\\
            % &+ \frac{1}{2} ( c^{(\tau+1)} - c^{(\tau)})^{\top} \nabla^{2} \mathcal{F}(c^{(\tau)})  ( c^{(\tau+1)} - c^{(\tau)})\\
             \le  \mathcal{F}(c^{(\tau))}) + \frac{1}{2} L_{H_{c}}\eta^{(t)2}_{c} \Vert  \hat{\nabla}c^{(\tau)} \Vert^{2} \\
            &- \eta^{(t)}_{c}\langle \nabla c^{(\tau)} ,  \hat{\nabla}c^{(\tau)} -\nabla c^{(\tau)}  \rangle - \eta^{(t)2}_{c} \Vert  \nabla c^{(\tau)}  \Vert^{2}\\
            &\le  \mathcal{F}(c^{(\tau))}) - (\frac{1}{2} \eta_{c}^{(t)} - \eta_{c}^{(t)2} L_{H_{c}})\Vert  \nabla c^{(\tau)}  \Vert^{2} \\
            &+ (\frac{1}{2} \eta_{c}^{(t)} + \eta_{c}^{(t)2} L_{H_{c}})\Vert  \nabla c^{(\tau)} -   \hat{\nabla}c^{(\tau)}   \Vert^{2}.
        \end{aligned}
    \end{equation}
    From Theorem~\ref{Theorem:approximation_error_c}, we have that
    \begin{equation}
    \small
        \begin{aligned}
            \Vert  \nabla c^{(\tau)} -   \hat{\nabla}c^{(\tau)}   \Vert^{2} \le 3\mathcal{E}_{cur_{1}}^{2} + 3\mathcal{E}_{cur_{2}}^{2}.
        \end{aligned}
    \end{equation}
    Thus Eq.~\eqref{equation:convergence_c_1} can be rewritten  as 
    \begin{equation}
    \small
        \begin{aligned}
            &(\frac{1}{2} \eta_{c}^{(t)} - \eta_{c}^{(t)2} L_{H_{c}})\Vert  \nabla c^{(\tau)}  \Vert^{2} \le \mathcal{F}(c^{(\tau))}) - \mathcal{F}(c^{(\tau+1))}) 
             \\
            &
            +(\frac{1}{2} \eta_{c}^{(t)} +\eta_{c}^{(t)2} L_{H_{c}}) (3\mathcal{E}_{cur_{1}}^{2} + 3\mathcal{E}_{cur_{2}}^{2}).
        \end{aligned}
    \end{equation}
    Telescoping from t=1 to $\mathcal{T}$ and according to assumptions, we have
    \begin{equation}
    \small
        \begin{aligned}
             &(\frac{1}{2} \eta_{c} - \eta_{c}^{2} L_{H_{c}}) \frac{1}{\mathcal{T}}
             \sum_{t=1}^{\mathcal{T}}
             \Vert  \nabla c^{(\tau)}  \Vert^{2} \le \frac{\mathcal{F}(c^{(1)}) - \mathcal{F}(c^{(\mathcal{T}+1)})}{\mathcal{T}} \\
            &+ (\frac{1}{2} \eta_{c}^{2} 
            + \eta_{c}^{2} L_{H_{c}}) (3\mathcal{E}_{cur_{1}}^{2} + 3\mathcal{E}_{cur_{2}}^{2}).
        \end{aligned}
    \end{equation}
    By setting $\frac{1}{8L_{H_{c}}} \le \eta_{c} \le \frac{1}{4 L_{H_{c}}}$, we have
    \begin{equation}
    \small
        \begin{aligned}
            \frac{1}{\mathcal{T}}
             \sum_{t=1}^{\mathcal{T}}
             \Vert  \nabla c^{(\tau)}  \Vert^{2} & \le \frac{64}{3} L_{H_{c}} \frac{\mathcal{F}(c^{(1)}) - \mathcal{F}(c^{(\mathcal{T}+1)})}{\mathcal{T}} \\
             &+ 9(\mathcal{E}_{cur_{1}}^{2} + \mathcal{E}_{cur_{2}}^{2}).
        \end{aligned}
    \end{equation}
    By denoting $\mathcal{C}_{c} \triangleq 9\mathcal{E}_{cur_{1}}^{2} +9\mathcal{E}_{cur_{2}}^{2} $,
    we have proved the theorem.
\end{proof}

Based on Theorem~\ref{Theorem:Convergence_analysis_M} and  Theorem~\ref{Theorem:Convergence_analysis_c}, we analyze the convergence performance of the parameters of HNNs $\boldsymbol{w}$ and curvatures $c$, which again demonstrate the rationality of our algorithm.

\subsection{Efficiency Analyses}
\label{ea}
\subsubsection{Analysis of Scope Sharpness Measure}

 We adopt two approximation ways to approximate $\rm{SN}(\boldsymbol{w})$  as  $\hat{\rm{sn}}(\hat{\boldsymbol{w}}(c))$.
    (1) We utilize $ \hat{\text{sn}}(\boldsymbol{w}^{*}(c))$ to approximate $\text{sn}(\boldsymbol{w}^{*}(c))$.
    (2) We utilize $ \hat{\text{sn}}(\boldsymbol{w}^{*}({c} )) +  \boldsymbol{\epsilon}^{\top}  \frac{\partial \hat{\text{sn}}(\boldsymbol{w}^{*}({c} ))}{\partial \boldsymbol{w}^{*}}$ to approximate 
    $ \hat{\text{sn}}(\boldsymbol{w}^{\ast}(c)+\boldsymbol{\epsilon})$.
    Here, we present the efficiency analysis for the first approximation way.
    Suppose that the number of the parameter $\boldsymbol{w}$ as $d$, then the complexity of computing $\rm{sn}(\cdot)$ equals to
    \begin{equation}
    \small
        \begin{aligned}
            \mathcal{O}(\text{sn}(\cdot)) = \mathcal{O}( d^{3} +d^{2} + 2d).
        \end{aligned}
    \end{equation}
    % since the complexity of matrix inverse is $\mathcal{O}(d^{3})$.
    In contrast, the complexity of  $\hat{\rm{sn}}(\cdot)$ is
    \begin{equation}
    \small
        \begin{aligned}
            \mathcal{O}(\hat{\rm{sn}}(\cdot)) = \mathcal{O}(d).
        \end{aligned}
    \end{equation}
    Obviously, the approximated $\hat{\rm{sn}}(\cdot)$ significantly reduce the computation complexity of $\rm{sn}(\cdot)$.

    % by following that
    % \begin{equation}
    %     \begin{aligned}
    %         &\mathcal{O}(\nabla_{\boldsymbol{w}} \mathcal{L}_{\mathcal{S}}(\boldsymbol{w}, c) \nabla_{\boldsymbol{w}} \mathcal{L}_{\mathcal{S}}(\boldsymbol{w}, c) ) = n^{2}\\
    %         & \mathcal{O}( [\nabla_{\boldsymbol{w}} \mathcal{L}_{\mathcal{S}}(\boldsymbol{w}, c) \nabla_{\boldsymbol{w}} \mathcal{L}_{\mathcal{S}}(\boldsymbol{w}, c)]^{-1}) = n^{3}\\
    %         &
    %     \end{aligned}
    % \end{equation}
    As to the second approximation way, it is non-trivial to analyze efficiency since the complexity and computation time for the maximization problem are impossible to analyze. Intuitively, the approximation way can efficiently solve the maximization problem, only needing a one-step update.

\subsubsection{Analysis of Implicit Differentiation}
We also provide a comparison of the computation complexity of different algorithms to compute gradient with respect to $c$.
Some works show that the complexity of computing gradients or Jacobian-vector products of a differentiable function is no more than five times the complexity of computing the function itself, and the complexity of computing Hessian-vector products is no more than five times the complexity of computing gradients, see~\citep{griewank1993some,griewank2008evaluating} for details. 

Recall that 
the gradient of $c$ is computed by
\begin{equation}
\small
% \footnotesize
	\label{equation:gradient_c_first_sp}
	\begin{aligned}
		\frac{d \mathcal{F}(\boldsymbol{w}^{\ast}(c), c)}{d c}  &= \nabla_{c}  \mathcal{F}(\boldsymbol{w}^{\ast}(c), c) \\
        &+ \nabla_{\boldsymbol{w}} \mathcal{F}(\boldsymbol{w}^{\ast}(c), c) \nabla_{c} \boldsymbol{w}^{\ast}(c),
	\end{aligned}
\end{equation}
where
$ \nabla_{\boldsymbol{w}} \mathcal{F}(\boldsymbol{w}^{\ast}(c), c) $ is expanded as
\begin{equation}
\small
% \footnotesize
	\label{equation:gradient_F_W_sp}
	\begin{aligned}
		&\nabla_{\boldsymbol{w}} \mathcal{F}(\boldsymbol{w}^{\ast}(c), c)  =\nabla_{\boldsymbol{w}} \mathcal{L}_{\mathcal{V}}(\boldsymbol{w}^{*}, c)+ \\
         &2K\Big(1-\Vert \nabla_{\boldsymbol{w}} \mathcal{L}_{\mathcal{S}}(\hat{\boldsymbol{w}}, c) \Vert\Big)^{K-1} 
       \nabla_{\boldsymbol{w}} \mathcal{L}_{\mathcal{S}}(\hat{\boldsymbol{w}}, c)
         \nabla^{2}_{\boldsymbol{w}} \mathcal{L}_{\mathcal{S}}(\hat{\boldsymbol{w}}, c)
        ,
	\end{aligned}
\end{equation}
and $\nabla_{c} \boldsymbol{w}^{\ast}(c)$ can be expanded as 
 \begin{equation}
 \label{equation:w_c_sp}
 \small
     \begin{aligned}
         \nabla_{c} \boldsymbol{w}^{\ast}(c) &= -\sum_{j\le T}\Big(\prod_{k<j}\boldsymbol{I} - \nabla^{2}_{\boldsymbol{w}} \mathcal{L}_{\mathcal{S}}(\boldsymbol{w}^{(T-k)}, c)\Big)\\
         &\times \nabla^{2}_{\boldsymbol{w}c}\mathcal{L}_{\mathcal{S}}(\boldsymbol{w}^{(T-j)}, c).
     \end{aligned}
 \end{equation}
 By utilizing the above-mentioned principles, the complexity of computing $\nabla_{\boldsymbol{w}} \mathcal{F}(\boldsymbol{w}^{\ast}(c), c) $ satisfies that
 \begin{equation}
 \small
     \begin{aligned}
         \mathcal{O}(\nabla_{\boldsymbol{w}} \mathcal{F}(\boldsymbol{w}^{\ast}(c), c) ) \le 35 \mathcal{O}(\mathcal{L}(\cdot)) + \mathcal{O}(2d).
     \end{aligned}
 \end{equation}
 The complexity of computing $\frac{\partial \boldsymbol{w}^{\ast}(c)}{\partial c}$ satisfies that
 \begin{equation}
 \small
     \begin{aligned}
         &\mathcal{O}( \nabla_{c} \boldsymbol{w}^{\ast}(c)) \le \\
         &\mathcal{O}\left(\frac{T(T+1)}{2}\left(25\mathcal{O}(\mathcal{L}(\cdot)) + d\right)+25T\mathcal{O}(\mathcal{L}(\cdot))\right).
     \end{aligned}
 \end{equation}
  By utilizing the above-mentioned principles, the complexity of the gradient in Eq.~\eqref{equation:gradient_c_first_sp}
satisfies that
\begin{equation}
\small
    \begin{aligned}
       \mathcal{O}&\left(\frac{d \mathcal{F}(\boldsymbol{w}^{\ast}(c), c)}{d c}\right) \le    \frac{T(T+1)+6}{2}\mathcal{O}\left(d\right)   \\
        &+ \left(\frac{25T(T+1)}{2}+25T+40 \right)\mathcal{O}(\mathcal{L}(\cdot)).
    \end{aligned}
\end{equation}

We denote the gradient of $c$ computed by implicit differentiation as $\frac{d \mathcal{F}_{i}(\boldsymbol{w}^{\ast}(c), c)}{d c}$.
Recall that it is computed as 
\begin{equation}
\small
	% \label{equation:implicit_differentiation_part}
	\begin{aligned}
		&\frac{d \mathcal{F}_{i}(\boldsymbol{w}^{\ast}(c), c)}{d c} = \nabla_{c}  \mathcal{F}(\boldsymbol{w}^{\ast}(c), c) \\
        &- \nabla_{\boldsymbol{w}} \mathcal{F}(\boldsymbol{w}^{\ast}(c), c) \Big(\nabla^{2} \mathcal{L}_{\mathcal{S}}(\boldsymbol{w}^{\ast} )\Big)^{-1} 
       \nabla_{\boldsymbol{w}c}^{2} \mathcal{L}_{\mathcal{S}}(\boldsymbol{w}^{\ast}, c).
	\end{aligned}
\end{equation}
Its complexity satisfies that
\begin{equation}
\small
    \begin{aligned}
        &\mathcal{O}\left(\frac{d \mathcal{F}_{i}(\boldsymbol{w}^{\ast}(c), c)}{d c}\right) \\
        &\le 65\mathcal{O}(\mathcal{L}(\cdot)) + \mathcal{O}(d^{3}) + 2\mathcal{O}(d^{2}) + 3\mathcal{O}(d).
    \end{aligned}
\end{equation}
Obviously, computing  $\frac{d \mathcal{F}_{i}(\boldsymbol{w}^{\ast}(c), c)}{d c}$ requires $\mathcal{O}(d^{2})$ and $\mathcal{O}(d^{3})$, which are intractable for deep neural works.

We further adopt $\boldsymbol{U}_{1}^{'}$ and $\boldsymbol{U}_{2}^{'}$ to approximate $\boldsymbol{U}_{1}$ and $\boldsymbol{U}_{2}$. 
Specifically, $\boldsymbol{U}_{1}^{'}$ and $\boldsymbol{U}_{2}^{'}$ are computed by
\begin{equation}
\small
    \begin{aligned}
       \boldsymbol{U}_{1}^{'} 
       &=
       -\nabla_{\boldsymbol{w}} \mathcal{L}_{\mathcal{V}}(\boldsymbol{w}^{*}, c)
       \\
		&\times \sum_{j=0}^{J}\Big(\boldsymbol{I} - \nabla^{2}_{\boldsymbol{w}} \mathcal{L}_{\mathcal{S}}(\boldsymbol{w}^{\ast}, c)\Big)^{j}
		\nabla^{2}_{\boldsymbol{w}c} \mathcal{L}_{\mathcal{S}}(\boldsymbol{w}^{*}, c)\\
        \boldsymbol{U}_{2}^{'} 
       &=-2K\Big(1-\Big\| \nabla_{\boldsymbol{w}} \mathcal{L}_{\mathcal{S}}(\hat{\boldsymbol{w}}, c) \Big\|\Big)^{K-1}
       \\
       &\times \nabla_{\boldsymbol{w}} \mathcal{L}_{\mathcal{S}}(\hat{\boldsymbol{w}}, c)  
		\nabla^{2}_{\boldsymbol{w}c} \mathcal{L}_{\mathcal{S}}(\boldsymbol{w}^{*}, c).
    \end{aligned}
\end{equation}
The complexity of $\boldsymbol{U}_{1}^{'}$ satisfies that
\begin{equation}
\small
    \begin{aligned}
        \mathcal{O}(\boldsymbol{U}_{1}^{'}) \le (25J+30) \mathcal{O}(\mathcal{L}(\cdot)) + J\mathcal{O}(d),
    \end{aligned}
\end{equation}
while the complexity of $\boldsymbol{U}_{2}^{'}$ satisfies that
\begin{equation}
\small
    \begin{aligned}
        \mathcal{O}(\boldsymbol{U}_{2}^{'}) \le 30 \mathcal{O}(\mathcal{L}(\cdot)) + \mathcal{O}(d).
    \end{aligned}
\end{equation}
We denote the gradient of c approximated with the proposed method as  $\frac{d \mathcal{F}_{o}(\boldsymbol{w}^{\ast}(c), c)}{d c}$. The complexity of $\frac{d \mathcal{F}_{o}(\boldsymbol{w}^{\ast}(c), c)}{d c}$ satisfies that
\begin{equation}
\small
\label{equation:complexity_ours}
    \begin{aligned}
        &\mathcal{O}\left(\frac{d \mathcal{F}_{o}(\boldsymbol{w}^{\ast}(c), c)}{d c}\right)\le\\
        & (25J + 60)\mathcal{O}(\mathcal{L}(\cdot)) + (J+1) \mathcal{O}(d).
    \end{aligned}
\end{equation}
Here, $J$ usually is set as $2$. By substituting $J=2$ into Eq.~\eqref{equation:complexity_ours}, the complexity of our method is 
\begin{equation}
\small
    \begin{aligned}
        \mathcal{O}\left(\frac{d \mathcal{F}_{o}(\boldsymbol{w}^{\ast}(c), c)}{d c}\right) \le 110\mathcal{O}(\mathcal{L}(\cdot)) + 3 \mathcal{O}(d).
    \end{aligned}
\end{equation}
\noindent \textbf{Remark.}
In summary, the complexity of $\frac{d \mathcal{F}(\boldsymbol{w}^{\ast}(c), c)}{d c}$ is 
\begin{equation}
% \footnotesize
\small
    \begin{aligned}
        &\mathcal{O}(\frac{d \mathcal{F}(\boldsymbol{w}^{\ast}(c), c)}{d c}) \le \frac{T(T+1)+6}{2}\mathcal{O}\left(d\right)   \\
        &+ \left(\frac{25T(T+1)}{2}+25T+40 \right)\mathcal{O}(\mathcal{L}(\cdot)),
    \end{aligned}
\end{equation}
the complexity of $\frac{d \mathcal{F}(\boldsymbol{w}^{\ast}(c), c)}{d c}$ is 
\begin{equation}
% \footnotesize
\small
    \begin{aligned}
         &\mathcal{O}(\frac{d \mathcal{F}_{i}(\boldsymbol{w}^{\ast}(c), c)}{d c}) 
         \\
         &\le 65\mathcal{O}(\mathcal{L}(\cdot)) + \mathcal{O}(d^{3}) + 2\mathcal{O}(d^{2}) + 3\mathcal{O}(d),
    \end{aligned}
\end{equation}
while the complexity of the approximated gradient $\frac{d \mathcal{F}_{o}(\boldsymbol{w}^{\ast}(c), c)}{d c}$ in our method is 
\begin{equation}
% \footnotesize
\small
    \begin{aligned}
        \mathcal{O}(\frac{d \mathcal{F}_{o}(\boldsymbol{w}^{\ast}(c), c)}{d c}) \le 110\mathcal{O}(\mathcal{L}(\cdot)) + 3 \mathcal{O}(d).
    \end{aligned}
\end{equation}
% \begin{equation}
% \footnotesize
% \left\{
%     \begin{aligned}
%         &\mathcal{O}(\frac{d \mathcal{F}(\boldsymbol{w}^{\ast}(c), c)}{d c}) \le \frac{T(T+1)+6}{2}\mathcal{O}\left(d\right)   \\
%         &+ \left(\frac{25T(T+1)}{2}+25T+40 \right)\mathcal{O}(\mathcal{L}(\cdot)) 
%         \\
%         & \mathcal{O}(\frac{d \mathcal{F}_{i}(\boldsymbol{w}^{\ast}(c), c)}{d c}) \le 65\mathcal{O}(\mathcal{L}(\cdot)) + \mathcal{O}(d^{3}) + 2\mathcal{O}(d^{2}) + 3\mathcal{O}(d)\\
%         &\mathcal{O}(\frac{d \mathcal{F}_{o}(\boldsymbol{w}^{\ast}(c), c)}{d c}) \le 110\mathcal{O}(\mathcal{L}(\cdot)) + 3 \mathcal{O}(d)
%     \end{aligned}
%     \right..
% \end{equation}
Compared with  $\frac{d \mathcal{F}(\boldsymbol{w}^{\ast}(c), c)}{d c}$, the proposed gradient $\frac{d \mathcal{F}_{o}(\boldsymbol{w}^{\ast}(c), c)}{d c}$ avoids to unroll the inner-level optimization process. The complexity of the proposed gradient is independent of the inner-level update step $T$, thereby reducing the complexity.
Compared with $\frac{d \mathcal{F}_{i}(\boldsymbol{w}^{\ast}(c), c)}{d c}$, $\frac{d \mathcal{F}_{o}(\boldsymbol{w}^{\ast}(c), c)}{d c}$ avoids the complexity of $\mathcal{O}(d^{3})$ by avoiding the inverse of the Hessian matrix, thereby further reducing the complexity.

\section{Experiments}
\subsection{Settings}
We conduct experiments on four settings: classification, learning from long-tailed data, learning from noisy data, and few-shot learning to evaluate the performance of our method.
In this subsection, we present the task descriptions and implementation details of these four experimental settings, including the dataset descriptions, baseline introduction, and hyper-parameters.

\noindent \textbf{Classification.} Following C-HNNs~\citep{guo2022clipped} that is a popular hyperbolic learning method, 
% we clip the Euclidean features before projecting the feature to hyperbolic spaces, and utilize hyperbolic MLR~\citep{shimizu2020hyperbolic} as the classifier to classify the hyperbolic feature.
we conduct experiments on three commonly used datasets: CIFAR10~\citep{krizhevsky2009learning}, CIFAR100~\citep{krizhevsky2009learning}, and ImageNet~\citep{deng2009imagenet}. 
For the CIFAR10 and CIFAR100, we utilize ResNet18~\citep{he2016deep}, WideResNet28-2~\citep{zagoruyko2016wide}, and PyramidNet110~\citep{han2017deep} as backbones to evaluate the performance of the proposed method. As to ImageNet, we utilize ResNet18 as the backbone. 
We follow training setups of C-HNNs~\citep{guo2022clipped} to conduct experiments. For experiments on CIFAR10 and CIFAR100 datasets, we set the initial learning rate as $0.1$, use the cosine learning rate scheduler, and set the batch size as 128. As to the perturbation radius $\hat{\rho}$, we set $\hat{\rho} = 0.005$ for CIFAR100, and set $\hat{\rho} = 0.01$ for CIFAR10. For ImageNet, we set the initial learning rate as $0.1$ and decay it by 10 every $30$ epochs. The batch size is $256$. We set $\hat{\rho} = 10^{-4}$ for ImageNet.

\noindent \textbf{Learning from long-tailed Data.}
We conduct classification on the long-tailed data to evaluate the generalization of our method. 
The long-tailed data refers to the fact that a few classes in the training set contain a major number of samples, while the remaining classes have only a small number of samples~\citep{ma2024geometric}. The number of samples in each category in the test set is consistent. It is challenging to learn using the long-tailed data since the data distribution of the training set is different from that of the test set \citep{yang2022survey}.

We conduct experiments on three major long-tailed datasets: CIFAR10-LT, CIFAR100-LT, and ImageNet-LT.  Following the setting of~\citep{cao2019learning}, we construct the CIFAR10-LT and CIFAR100-LT datasets by sampling from the original datasets with different imbalance
ratios $\text{IR} = \frac{N_{\text{max}}}{N_{\text{min}}}$, where $N_{\text{max}}$ and $N_{\text{min}}$ are the corresponding number of the most and least frequent classes. We set the imbalance ratio as $\{200,100,50\}$ for evaluation. We utilize the same way to construct the ImageNet-LT dataset, with the imbalance ratio being 256.
We adopt ResNet32~\citep{he2016deep} as the backbone for CIFAR10-LT and CIFAR100-LT, and adopt ResNeXt-50~\citep{xie2017aggregated} as the backbone for ImageNet-LT. 

% Training details of long-tailed classification are provided in \textbf{appendix}~\ref{ed}.

We apply our method to the cross-entropy (CE) method,  rebalanced classifier methods: DRW~\citep{cao2019learning}, LDAM-DRW~\citep{cao2019learning}, GCL~\citep{li2022long}, and contrastive learning methods: GL-Mixture~\citep{du2023global}.
To the best of our knowledge, the ability of hyperbolic spaces has not been explored in the long-tailed setting. To develop the proposed method to these five baseline methods,  we first project the feature to hyperbolic spaces and utilize the hyperbolic MLR to classify the images. Then, we utilize the SAM method to train HNNs and apply the proposed method to learn curvatures to improve the generalization of HNNs.

\begin{table*}[t]
	\centering
	% \small
	%\footnotesize
        \caption{Hyper-parameters for CIFAR10-LT and CIFAR100-LT datasets.}
        \label{table:CIFAR10_CIFAR100_hyperparameters}	
        % \vskip 0.15in
	% \setlength{\tabcolsep}{0.3mm}{
	% 	\resizebox{8.3cm}{!}{
			\begin{tabular}{cccccc}
				\midrule
				\bfseries Baseline &  CE & DRW & LDAM & GCL & GL-Mixture   \\
				\toprule
    	Learning rate & 0.1/0.1  & 0.1/0.1 & 0.1/0.1 & 0.1/0.1 &  0.01/0.01   \\
				% \midrule
		Perturbation radius $\hat{\rho}$ &  0.1/0.2 & 0.5/0.8 & 0.8/0.8 & 0.5/0.5 &  0.05/0.05   \\
				\midrule
			\end{tabular}
	% 	}	
	% }
\end{table*}

\begin{table*}[t]
	\centering
	% \normalsize
	%\footnotesize
	\setlength{\tabcolsep}{0.5mm}{
         \begin{threeparttable}
         \caption{Test Accuracy (\%)  on the classification task.}
		% \resizebox{8.3cm}{!}{
			\begin{tabular}{ccccccc}
				\midrule
				& \bfseries Method &  ENNs & HNNs& C-HNNs  & \bfseries Ours+HNNs & \bfseries Ours+C-HNNs  \\
    \toprule
    \multirow{3}{*}{CIFAR10 
                     }
				&	ResNet18 & $95.32 \pm 0.13$  & $93.83 \pm 0.09$ & $95.20 \pm 0.30$ &  $96.03 \pm 0.07$ & $\mathbf{96.50 \pm 0.04}$ \\
				
				& WideResNet28-2 & $94.81 \pm 0.42$  & $88.82 \pm 0.51 $ & $94.76 \pm 0.44$  & $95.96 \pm 0.21$  & $\mathbf{96.11 \pm 0.13}$ \\
				
			&	PyramidNet110 & $96.19 \pm 0.11$ &  $95.89 \pm 0.21$ & $96.03\pm0.14$ & $97.26 \pm 0.05$  & $\mathbf{97.48 \pm 0.06}$   \\
				\midrule
     \multirow{3}{*}{CIFAR100 
                     }
			&	ResNet18 & $78.32 \pm 0.32$ & $69.97 \pm 0.20$ & $78.43 \pm 0.15$ & $79.51 \pm  0.03$   & $\mathbf{80.61 \pm 0.02}$   \\
				
				& WideResNet28-2 & $75.45 \pm 0.25$  & $72.26 \pm 0.41$ & $75.88 \pm 0.38$ & $76.88 \pm  0.05$ & $\mathbf{78.35 \pm  0.03}$   \\
				
			&	PyramidNet110 & $82.74 \pm 0.12$  & $80.29 \pm 0.26$ & $82.16 \pm 0.08$ & $84.39 \pm 0.06$  & $\mathbf{85.62 \pm 0.02}$  \\
				\midrule
     \multirow{1}{*}{ImageNet 
                     }
			& ResNet18 & 69.82 & 65.74 & 68.45 & $68.23$    & $\mathbf{70.53}$  \\
			\midrule
			\end{tabular}
   \begin{tablenotes} % 添加 tablenotes 环境用于脚注
            % \footnotesize
            \item The best results are shown in bold. The experiments on CIFAR10 and CIFAR100 datasets are repeated for 5 times, and we report both average accuracy and standard deviation.
        \end{tablenotes}
		\label{table:CIFAR100}	
             \end{threeparttable}
	}
\end{table*}

As to CIFAR10-LT and CIFAR100-LT datasets, we set the batch size as $64$ for five baselines. The other hyper-parameters of different baselines are various, which are shown in Table~\ref{table:CIFAR10_CIFAR100_hyperparameters}. We utilize the cosine learning rate schedule to decay the learning rate for these five baselines.
% \textit{CE.} The learning rate for two datasets are set as $0.1$. We utilize the cosine learning rate schedule to decay the learning rate. We set the perturbation radius $\hat{\rho}$ as  0.1 for CIFAR10-LT and set the perturbation radius $\hat{\rho}$ as  0.2 for CIFAR100-LT. \textit{DRW.}
% The learning rates for two datasets are set as $0.1$. We utilize the cosine learning rate schedule to decay the learning rate. We set the perturbation radius $\hat{\rho}$ as  0.5 for CIFAR10-LT, and set the perturbation radius $\hat{\rho}$ as  0.8 for CIFAR100-LT. \textit{LDAM.}
% The learning rates for two datasets are set as $0.1$. We utilize the cosine learning rate schedule to decay the learning rate. We set the perturbation radius $\hat{\rho}$ as  0.8 for CIFAR10-LT and CIFAR100-LT. \textit{GCL.}
% The learning rates for two datasets are set as $0.1$. We utilize the cosine learning rate schedule to decay the learning rate. We set the perturbation radius $\hat{\rho}$ as  0.5 for CIFAR10-LT and CIFAR100-LT.  \textit{GL-Mixture.}
% The learning rates for two datasets are set as $0.01$. We utilize the cosine learning rate schedule to decay the learning rate. We set the perturbation radius $\hat{\rho}$ as  0.05 for CIFAR10-LT and CIFAR100-LT when IR=50/100, and we set the perturbation radius $\hat{\rho}$ as  0.01 for CIFAR10-LT and CIFAR100-LT when IR=200.
In terms of the experiment on the ImageNet-LT, we set the batch size as $256$, set the initial learning rate as 0.01, decay it by 10 every 30 epochs, and set the perturbation radius $\hat{\rho}$ as 1e-4.

\noindent \textbf{Learning from noisy data.}
We conduct classification on noisy data to evaluate the robustness and generalization of our method, where the noisy data refers to some labels in the datasets corrupted from ground-truth labels. We conduct experiments on the Clothing1M dataset~\citep{xiao2015learning} that is a large-scale clothing dataset obtained by crawling images from several online shopping websites. 
The Clothing1M dataset comprises 14 classes and contains over a million noisy labeled samples, along with a small set of approximately 50,000 clean samples used for constructing validation and testing datasets. In our study, we adopt the EMLC method~\citep{taraday2023enhanced} and GENKL~\citep{huang2023genkl} as baselines, and we employ HNNs as the backbone for evaluating the proposed method.
% Clothing1M dataset consists of 14 classes and  contains more than a million noisy labeled samples as well as a small set (around 50K) of clean samples that is used to construct validation and testing datasets.
% Here, we adopt the EMLC method~\citep{taraday2023enhanced} and GENKL~\citep{huang2023genkl} as the baselines and utilize HNNs as the backbone to evaluate our method.
 Following prior works, we use HNNs with ResNet50~\citep{he2016deep}  as the backbone and pre-train the backbone on ImageNet. The HNNs are trained for 5 epochs. 
 Hyper-parameters are consistent with the baseline method ELMC and GENKL.
In terms of the ELMC baseline, we set the learning rate
as 0.1, the batch size as 512, and the $\hat{\rho}$ as 0.8.
In terms of the GENKL baseline, we set the learning rate
as 0.001, the batch size as 32, and the $\hat{\rho}$ as 0.02.

\noindent \textbf{Few-shot learning.}
The few-shot learning problem aims to recognize samples from unseen classes, given very few labeled examples. Studies such as~\citep{khrulkov2020hyperbolic, hong2023hyperbolic, li2023euclidean} have shown that modeling data on hyperbolic spaces leads to better performance. 
Inspired by it, we project the features onto hyperbolic spaces and build a hyperbolic classifier via hyperbolic distances following the work of~\citep{khrulkov2020hyperbolic}. 
Following the setting of MAML~\citep{finn2017model}, we learn the hyperbolic classifier for few-shot learning. 
We conduct experiments on two popular datasets: mini-ImageNet~\citep{vinyals2016matching} and tiered-ImageNet~\citep{ren2018meta}.
In terms of the few-shot learning task, as to the mini-Imagenet, 
we set the learning rate as 0.1, the batch size as 32, and the $\hat{\rho}$ as 0.01; as to the tiered-Imagenet, we set the learning rate
as 0.05, the batch size as 32, and the $\hat{\rho}$ as 0.01.

\noindent \textbf{Hierarchical structures of datasets.}
The utilized datasets (including CIFAR, mini-ImageNet, tiered-ImageNet, ImageNet, Clothing1M) all exhibit  inherent hierarchical structures that can be well captured by the hyperbolic spaces.
Existing works have shown that the CIFAR, mini-ImageNet, tiered-ImageNet, and ImageNet datasets have hierarchical structures, and the four datasets are commonly used to evaluate hyperbolic algorithms~\citep{Khrulkov_2020_CVPR,Fang_2021_ICCV,9749838,Yan_2021_CVPR,Guo_2022_CVPR,Ermolov_2022_CVPR}. 
For example, the work~\citep{NIPS2017_59dfa2df} suggests that the semantics of images in ImageNet have a hierarchical structure, with details illustrated in a prominent visual analysis conducted by~\citet{bostock_imagenet_hierarchy}.
% \footnote{\href{https://observablehq.com/@mbostock/imagenet-hierarchy/}{https://observablehq.com/@mbostock/imagenet-hierarchy/}}. 
CIFAR100 also has a two-level hierarchical structure, with 20 root nodes and 100 leaf nodes, with details described in the work~\citep{cifar10}.
% \footnote{\href{http://labs.criteo.com/2014/02/kaggle-display-advertisingchallenge-dataset/}{http://labs.criteo.com/2014/02/kaggle-display-advertisingchallenge-dataset/}}.
Although Clothing1M dataset has not been utilized to evaluate the performance of hyperbolic algorithms, this dataset also presents the hierarchical structure with four root nodes (including Tops, Bottoms, Footwear, and Accessories) and fourteen leaf nodes.

To further confirm the above points, we compute  $\delta$-Hyperbolicity 
~\citep{Khrulkov_2020_CVPR,6729484,10.1145/3442381.3449872} 
to measure whether the used datasets have hierarchical structures. 
A $\delta$-Hyperbolicity value closer to $0$ indicates a stronger hyperbolic structure of a dataset. The values of $\delta$-Hyperbolicity are $\bold{0.24}$, $\bold{0.21}$, $\bold{0.25}$, $\bold{0.22}$, $\bold{0.23}$, and $\bold{0.24}$ on the mini-ImageNet, tiered-ImageNet, CIFAR10, CIFAR100, ImageNet, and Clothing1M datasets, respectively. 
These results reveal that the utilized datasets are appropriate for evaluating the performance of hyperbolic algorithms.

\subsection{Main Results}

In this subsection, we present the main results, compared methods, and experimental analyses of four settings. Overall, our method surpassed all compared baseline methods and achieves state-of-the-art performance in some settings, demonstrating the effectiveness and superiority of our method.

\begin{table*}
	\centering
	% \small
	%\footnotesize
         \caption{Top-1 Accuracy  (\%) on CIFAR10-LT and CIFAR100-LT datasets.} 	\label{table:CIFAR10_CIFAR100}	
	\setlength{\tabcolsep}{3mm}{
                    \begin{threeparttable}
		% \resizebox{8.6cm}{!}{
			\begin{tabular}{ccccccc}
				\midrule
				\bfseries & \multicolumn{3}{c}{CIFAR10-LT} & \multicolumn{3}{c}{CIFAR100-LT} \\
				\cmidrule{2-4}\cmidrule{5-7}
                % \midrule
				Imbalance Ratio &  200 & 100 & 50 & 200  & 100 & 50   \\
				\toprule
				CE & 65.68 & 70.70 &  74.81 & 34.84 & 38.43 & 43.90    \\
				CE+HNNs & 64.50 & 68.09 & 74.3 & 33.60 & 37.44 &  42.66   \\
				CE+C-HNNs & 65.25 & 69.96 & 76.53 & 34.44 & 38.34 & 42.97    \\
                  CE+HNNs+Ours &  65.03 & 71.07 &  \bfseries 77.90  &  35.19 &  39.02 &  43.79   \\
				\bfseries  CE+C-HNNs+Ours  & \bfseries 66.26 & \bfseries 72.3 & \bfseries 78.99  & \bfseries 35.99 & \bfseries 39.67 &  \bfseries 44.66   \\
				\midrule
				DRW (Cao et al, 2019)  & 67.85 & 75.47 & 79.67  & 37.21 & 40.66 & 46.41\\
				DRW+HNNs & 68.12 & 73.03 & 79.17  & 35.01 & 38.37 & 44.65 \\
				DRW+C-HNNs & 68.12 & 75.52 & 81.1  & 37.16 & 41.29 &  46.33\\
            DRW+HNNs+Ours & 69.07 &  78.99 & \bfseries 81.83  &  39.15 &  43.11 &   48.56   \\
				\bfseries DRW+C-HNNs+Ours & \bfseries 69.97  & \bfseries 80.48 & \bfseries 82.64  &  \bfseries 40.62 & \bfseries 44.77 & \bfseries 49.5\\
				\midrule
				LDAM (Cao et al, 2019)     & 73.52 & 77.03 & 81.03 & 38.91 & 42.04 & 47.62 \\
				LDAM+HNNs     & 72.29 & 76.96 & 80.54 & 34.27 & 38.94 & 46.35 \\
				LDAM+C-HNNs     & 72.81 & 76.96 & 81.54 & 35.1 & 39.94 & 46.99 \\
                  LDAM+HNNs+Ours &  76.66 &  79.92 &  83.99  &  41.75 &  45.78 &  50.33   \\
				\bfseries  LDAM+C-HNNs+Ours     & \bfseries 78.89 & \bfseries 82.01 & \bfseries 85.05 & \bfseries 43.03 &\bfseries  47.13 &\bfseries  51.02 \\
				\midrule
				GCL (Li et al, 2022a)     & 79.03 & 82.68 & 85.46 & 44.88 & 48.71 &  53.55 \\
				GCL+HNNs     & 78.76 & 82.67 & 84.88 & 41.76 & 46.25 & 50.77 \\
				GCL+C-HNNs     & 80.19 & 82.71 & 85.19 & 43.53 & 45.27 & 52.61 \\
                GCL+HNNs+Ours &  80.16 &  83.93 & 86.30 &  44.84 &  49.50 &   53.04   \\
				\bfseries  GCL+C-HNNs+Ours     & \bfseries 80.82 & \bfseries 84.25 &\bfseries  86.42 & \bfseries 45.65 & \bfseries 49.84 & \bfseries 53.96 \\
				\midrule
				GL-Mixture (Du et al, 2023)     & 81.86 & 88.5 & 91.04 & 44.79 & 57.97 & 63.78 \\
				GL-Mixture+HNNs     & 82.72 & 85.97 & 89.26 & 49.31 & 56.52 & 60.81 \\
				GL-Mixture+C-HNNs     & 70.71 & 86.01 & 89.2 & 33.43 & 49.93 & 58.53 \\
                 \bfseries GL-Mixture+HNNs+Ours  & \bfseries 85.07 & \bfseries 88.96 &\bfseries  91.72 & \bfseries 52.33 &\bfseries  59.43 & \bfseries 64.11 \\
				 GL-Mixture+C-HNNs+Ours   &   81.86 &  87.64 &  90.75  &  45.18 &  57.80 &   62.8   \\
				\midrule
			\end{tabular}
             \begin{tablenotes} % 添加 tablenotes 环境用于脚注
            \item The best results of each baseline and its variations are shown in bold.
        \end{tablenotes}
		% }
	
   \end{threeparttable}
	}
\end{table*}

\begin{table*}[t]
	\centering
        \normalsize
	% \small
	%\footnotesize
  \caption{Top-1 Accuracy  (\%) on CIFAR10-LT and CIFAR100-LT datasets compared with the state-of-the-art methods.}
	\setlength{\tabcolsep}{2mm}{
		% \resizebox{8.6cm}{!}{
			\begin{tabular}{cccccccc}
				\midrule
				 & \multirow{2}*{\bfseries Method} & \multicolumn{3}{c}{CIFAR10-LT} & \multicolumn{3}{c}{CIFAR100-LT} \\
				% \cline{3-5}
    \cmidrule{3-5}\cmidrule{6-8}
				&  &  200 & 100 & 50 & 200  & 100 & 50   \\
				\toprule
                     \multirow{5}{*}{Rebalanced 
                     }   
                    &  CB-Focal~\citep{cui2019class} & 68.89 & 74.60 & 79.30 & 36.23 & 39.60 & 45.20 \\
                    & LDAM~\citep{cao2019learning}    & 73.52 & 77.03 & 81.03 & 38.91 & 42.04 & 47.62 \\
                    \multirow{3}{*}{ 
                     Classifier} 
                     & MiSLAS~\citep{zhong2021improving}  & 77.31 & 82.06  &  85.16 & 42.33 & 47.50 & 52.62\\
                    & BGP~\citep{wang2022balanced} & - & -  &  - & 41.2 & 45.2 & 50.5 \\
                    & GCL~\citep{li2022long}   & 79.03 & 82.68 & 85.46 & 44.88 & 48.71 &  53.55 \\
                     & CDB-S~\citep{sinha2022class} & - & -  &  - & 37.99 & 42.59 & 46.82 \\
                    & AREA~\citep{chen2023area} & 74.99 & 78.88  &  82.68 & 43.85 & 48.83 & 51.77\\
                    &FUR~\citep{ma2024geometric} & 79.80 & 83.70  &  86.20 & 46.20 & 50.90 & 54.10\\
                    \midrule
                     \multirow{5}{*}{Contrastive}  
				& TSC~\citep{li2022targeted} & - & 79.70 & 82.90 & - & 42.80 & 46.30 \\
                & KCL~\citep{kang2020exploring}     & - & 77.60 & 81.70 & - & 42.80 & 46.30 \\
                \multirow{3}{*}{learning}
               & Paco~\citep{cui2021parametric}     & - & - & - & - & 52.0 & 56.0 \\
               & SSD~\citep{li2021self}     & - & - & - & - & 46.0 & 50.50 \\
               & GL-Mixture~\citep{du2023global}     & 81.86 & 88.5 & 91.04 & 44.79 & 57.97 & 63.78 \\
               & SBCL~\citep{hou2023subclass}     & - & - & - & - & 44.9 & 48.7 \\
               \midrule
               \multirow{2}{*}{SAM-based} 
               & VS-SAM~\citep{rangwani2022escaping}  & -&  82.4 &- &  - & 46.6 & - \\
               & CC-SAM~\citep{zhou2023class}  & 80.94 &  83.92 & 86.22 &  45.66 & 50.83 & 53.91 \\
				\midrule
			&	\bfseries Ours     & \bfseries 85.07 & \bfseries 88.96 & \bfseries 91.72 & \bfseries 52.33 & \bfseries 59.43 & \bfseries 64.11 \\
				\midrule
				
			\end{tabular}
    \begin{tablenotes} % 添加 tablenotes 环境用于脚注
            % \footnotesize
            \item The best results are shown in bold. 
        \end{tablenotes}
		}
		\label{table:CIFAR10_CIFAR100_2}	
	% }
\end{table*}

\begin{table*}[t]
	\centering
	% \small
	% \footnotesize
	
  \caption{Top-1 Accuracy (\%)  on ImageNet-LT dataset.} \label{table:Imagenet-LT}
  %\resizebox{8.3cm}{!}{
	\setlength{\tabcolsep}{3mm}{
            \begin{threeparttable}
		\begin{tabular}{cccccc}
			\midrule
			\bfseries Method &  Backbone & All & Many & Med  & Few   \\
			\toprule
			% CB-Focal~\citep{cui2019class}     & ResNet-50 & 33.2 & 39.6 & 32.7 & 16.8 \\
			% LDAM~\citep{cao2019learning}     & ResNet-50 & 49.8 & 60.4 & 46.9 & 30.7 \\
   %              GCL~\citep{li2022long} & ResNet-50 & 54.9 & - & - & - \\
   %              TSC~\citep{li2022targeted} & ResNet-50 & 52.4 & 63.5 & 49.7 & 30.4 \\
   %              LDAM+SAM~\citep{rangwani2022escaping} & ResNet-50 & 53.1  &  62.0 & 52.1 & 34.8 \\
			LADE~\citep{hong2021disentangling}  & ResNetXt-50 & 52.3 & 64.4 & 47.7 & 34.3 \\
			DisAlign~\citep{zhang2021distribution} & ResNetXt-50 & 53.4 & 62.7 & 52.1 & 31.4 \\
			PaCo~\citep{li2021self} & ResNeXt-50 & 56 & 64.4 & \bfseries 55.7 & 33.7 \\
                SSD~\citep{li2021self} & ResNeXt-50 & 56 & 66.8 & 53.1 & 35.4 \\
			ResLT~\citep{cui2022reslt} & ResNeXt-50 & 52.9 & 63.0 & 53.3 & 35.5 \\
			GL-Mixture~\citep{du2023global} & ResNeXt-50 & 56.3  &  $\mathbf{70.1}$ & 52.4 & 30.4 \\
			% \hline
			CC-SAM~\citep{zhou2023class}  & ResNeXt-50 & 55.4  &  63.1 & 53.4 & \bfseries  41.1 \\
			\midrule
			\bfseries Ours & ResNeXt-50 & $\mathbf{56.8}$  &  66.9 &  55.5 & 34.3 \\
			\midrule
		\end{tabular}
          \begin{tablenotes} % 添加 tablenotes 环境用于脚注
            \item The best results are shown in bold. 
        \end{tablenotes}
         \end{threeparttable}
		}
\end{table*}

\begin{table}[]
    \centering
    \caption{Comparison with state-of-the-art methods in test
accuracy (\%) on Clothing1M. 
% The methods in the lower part of the table use extra clean data, while the methods in the upper part do not (above an below the dividing line).
}\label{tabel:clothing1m}
\setlength{\tabcolsep}{0.1mm}{
    \begin{tabular}{ccc}
    \midrule
          \bfseries Method & \bfseries Extra Data & \bfseries  Acc \\
          \toprule
          Cross-entropy  & \XSolidBrush & 69.21\\
          Joint-Optim~\citep{tanaka2018joint}  & \XSolidBrush & 72.16\\
          P-correction~\citep{PENCIL_CVPR_2019} & \XSolidBrush & 73.49\\
          C2D~\citep{zheltonozhskii2022contrast} & \XSolidBrush & 74.30\\
          DivideMix~\citep{li2020dividemix} & \XSolidBrush & 74.76\\
          ELR+~\citep{liu2020early} & \XSolidBrush & 74.81\\
          AugDesc~\citep{nishi2021augmentation} & \XSolidBrush & 75.11\\
          SANM~\citep{tu2023learning} & \XSolidBrush & 75.63\\
          \midrule
          Meta Cleaner~\citep{zhang2019metacleaner} & \Checkmark & 72.50 \\
          Meta-Learning~\citep{li2019learning} & \Checkmark & 73.47 \\
          MW-Net~\citep{shu2019meta}  &  \Checkmark & 73.72 \\
          FaMUS~\citep{xu2021faster} &  \Checkmark & 74.4\\
          MLC~\citep{zheng2021meta} & \Checkmark & 75.78\\
          MSLG~\citep{algan2021meta} & \Checkmark & 76.02\\
          Self Learning~\citep{han2019deep} & \Checkmark & 76.44\\
          FasTEN~\citep{kye2022learning} & \Checkmark & 77.83\\
          EMLC~\citep{taraday2023enhanced} & \Checkmark & 79.35 \\
           GENKL~\citep{huang2023genkl} & \Checkmark & 81.34 \\
         \midrule
          EMLC+HNNs & \Checkmark &  79.40 \\
          EMLC+C-HNNs & \Checkmark &  79.70 \\
           EMLC+HNNs+Ours & \Checkmark &  80.07 \\
          \bfseries EMLC+C-HNNs+Ours & \Checkmark & \bfseries 80.38 \\
          GENKL+HNNs & \Checkmark &  81.48 \\
          GENKL+C-HNNs & \Checkmark &  81.56 \\
           GENKL+HNNs+Ours & \Checkmark &  82.04 \\
          \bfseries GENKL+C-HNNs+Ours & \Checkmark & \bfseries 82.27 \\
          \midrule
    \end{tabular}}
     \begin{tablenotes} % 添加 tablenotes 环境用于脚注
            % \footnotesize
            \item The results of our method are shown in bold. The proposed method not only outperforms the baseline methods, but also achieves the best performance. 
        \end{tablenotes}

\end{table}

\begin{table*}[t]
	\small
	\centering
	\begin{threeparttable} % 使用 threeparttable 环境
        \caption{Accuracy (\%) on the mini-ImageNet and tiered-ImageNet datasets.}
        \label{table:miniimagenet}
        \begin{tabular}{cccccc}
			\midrule
			\multirow{2}*{\bfseries Method} & \multirow{2}*{\bfseries Backbone} &  \multicolumn{2}{c}{\bfseries mini-ImageNet} &  \multicolumn{2}{c}{\bfseries tiered-ImageNet} \\
			\cmidrule{3-4}\cmidrule{5-6}
			& & \bfseries 1-shot 5-way & \bfseries 5-shot 5-way  & \bfseries 1-shot 5-way & \bfseries 5-shot 5-way \\
			\toprule
			MAML~\citep{finn2017model} & ResNet12 & $51.03 \pm 0.50$ & $68.26 \pm 0.47$ & $58.58 \pm 0.49$ & $71.24 \pm 0.43$ \\
			L2F~\citep{baik2020learning} & ResNet12 & $57.48 \pm 0.49$ & $74.68 \pm 0.43$ & $63.94 \pm 0.48$ & $77.61 \pm 0.41$ \\
			ALFA~\citep{baik2020meta} & ResNet12 & $60.06 \pm 0.49$ & $77.42 \pm 0.42$ & $64.43 \pm 0.49$ & $81.77 \pm 0.39$ \\
			MetaFun~\citep{xu2020metafun} & ResNet12 & $62.12 \pm 0.30$ & $78.20 \pm 0.16$  & $67.72 \pm 0.14$ & $78.20 \pm 0.16$ \\
			DSN~\citep{simon2020adaptive} & ResNet12 & $62.64 \pm 0.66$ & $78.83 \pm 0.45$ & $66.22 \pm 0.75$ & $82.79 \pm 0.48$ \\
			Chen \emph{et al.}~\citep{chen2021meta} & ResNet12 & $63.17 \pm 0.23$ & $79.26 \pm 0.17$ & $68.62 \pm 0.27$ & $83.74 \pm 0.18$ \\
			MeTAL~\citep{baik2021meta} & ResNet12 & $59.64 \pm 0.38$ & $76.20 \pm 0.19$ & $63.89 \pm 0.43$ & $80.14 \pm 0.40$ \\
			C-HNNs~\citep{guo2022clipped} & ResNet12 & $53.01 \pm 0.22$ & $72.66 \pm  0.15$ & - & - \\
			CurAML~\citep{gao2023curvature} & ResNet12 & $63.13 \pm 0.41$ & $81.04 \pm  0.39$ & $68.46 \pm 0.56$ & $83.84 \pm  0.40$ \\
			Hyper ProtoNet~\citep{khrulkov2020hyperbolic} & ResNet18 & $59.47 \pm 0.20$ & $76.84 \pm 0.14$ & $62.28 \pm 0.23$ & $74.50 \pm 0.21$ \\
			Hyperbolic kernel~\citep{fang2021kernel} & ResNet18 & $61.04 \pm 0.21$ & $77.33 \pm  0.15$ & $61.04 \pm 0.21$ & $77.33 \pm  0.15$ \\
			Poincaré radial kernel~\citep{fang2023poincare} & ResNet18 & $62.15 \pm 0.20$ & $77.81 \pm  0.15$ & $65.33 \pm 0.21$ & $77.48 \pm 0.20$ \\
			\midrule
			 Ours+HNNs & ResNet12 & $64.06 \pm 0.22$ & $81.50 \pm 0.17$ & $70.40 \pm 0.23$  & $84.60 \pm 0.21$ \\
			\bfseries Ours+C-HNNs & ResNet12 & $\mathbf{65.30 \pm 0.19}$ & $\mathbf{82.20 \pm 0.15}$ & $\mathbf{72.10 \pm 0.25}$  & $\mathbf{86.25 \pm 0.16}$ \\
			\midrule
		\end{tabular}
        \begin{tablenotes} % 添加 tablenotes 环境用于脚注
            % \footnotesize
            \item The best results are shown in bold.  The experiments are repeated for 5 times, and we report both average accuracy and standard deviation
        \end{tablenotes}
	\end{threeparttable}
\end{table*}

\noindent \textbf{Classification.}
We compare our method with methods on hyperbolic spaces: HNNs~\citep{ganea2018hyperbolic}, HWe compare our method with methods on hyperbolic spaces: HNNs~\citep{guo2022clipped}, and the methods on the Euclidean spaces.
Results on CIFAR10, CIFAR100, and ImageNet datasets are shown in Table~\ref{table:CIFAR100}. 
Our method, when applied to HNNs (`Ours + HNNs'), outperforms the standard HNNs, and when applied to C-HNNs (`Ours + C-HNNs'), it surpasses the standard C-HNNs.
 % The experimental results show that our method applied to HNNs (`Ours + HNNs') outperforms standard HNNs, and our method applied to C-HNNs (`Ours + C-HNNs') surpasses standard C-HNNs. 
`Ours + C-HNNs' achieves improvements of about $2.67\%$, $7.29\%$, $1.59\%$   over HNNs, and about $1.3\%$, $1.35\%$, $1.45\%$ over C-HNNs on the ResNet18, WideResNet28-2, PyramidNet110 backbones, respectively. on the CIFAR100 dataset, `Ours + C-HNNs' achieves about $10.64\%$, $6.12\%$, $5.32\%$ improvement over HNNs and about $2.17\%$, $1.47\%$, $1.23\%$ improvement over C-HNNs on the ResNet18, WideResNet28-2, PyramidNet110 backbones, respectively;  
on the ImageNet dataset, our method also surpasses the other methods, where it achieves about $5\%$  improvement over HNNs and about $2\%$ improvement over C-HNNs. These results show that our method improves the performance and generalization of HNNs and C-HNNs.
This demonstrates that our method enhances both the performance and generalization of HNNs and C-HNNs.

\noindent \textbf{Learning from long-tailed data.}
  % We compare our method with five baseline methods on Euclidean spaces, five baseline methods developed with HNNs and C-HNNs on the CIFAR10-LT and CIFAR100-LT. 
 We compare our method with five baselines using Euclidean neural networks, HNNs, and C-HNNs on the CIFAR10-LT and CIFAR100-LT datasets.
 To the best of our knowledge, these five baselines have not been developed to the hyperbolic spaces.
  For comparison, we reproduce these five baseline methods on the hyperbolic spaces with HNNs and C-HNNs.
   Results are shown in Table~\ref{table:CIFAR10_CIFAR100}. 
   The performance of our method applied to HNNs (`Ours + HNNs') and C-HNNs (`Ours + C-HNNs') surpasses that of the standard HNNs and C-HNNs, respectively. These results further demonstrate that our method enhances both the performance and generalization of HNNs and C-HNNs when the distributions of training set and test set  differ.
  Notably, our method significantly improves the performance with IR set as $200$.  
  This improvement is attributed to the fact that baseline methods converge to the very sharp regions with IR set as $200$, while our method can significantly reduce the sharpness of the loss curve and improve generalization. Further evidence is presented in section~\ref{section:altd}.
  
  % The reason is that local minima obtained by baseline methods converge to the very sharp region with IR set as $200$, while our method can significantly reduce the sharpness of the loss curve and improve generalization.  
  
  We also compare our method, based on the GL-Mixture method~\citep{du2023global}, with other long-tailed methods from three categories: rebalanced classifier methods,  contrastive learning methods, and SAM-based methods. Results on CIFAR10-LT, CIFAR100-LT, and ImageNet-LT are shown in Table~\ref{table:CIFAR10_CIFAR100_2} and Table~\ref{table:Imagenet-LT}, respectively.
  Our method achieves superior performance compared to existing methods. Notably, when $\text{IR} = 200$, our method exceeds the current best method by $2.3\%$ on CIFAR10-LT and $6.4\%$ on CIFAR100-LT. On ImageNet-LT, our method also achieves state-of-the-art performance, demonstrating the superiority of our method.

\noindent \textbf{Learning from noisy data.}
We compare our method with the baseline methods on the Euclidean spaces: ELMC~\citep{taraday2023enhanced} and GENKL~\citep{huang2023genkl}. Additionally, we compare our method with baseline methods on hyperbolic spaces: ELMC+HNNs and GENKL+HNNs.
To the best of our knowledge, EMLC and GENKL have not been developed to the hyperbolic spaces; thus, we reproduce the EMLC and GENKL methods using the HNNs backbone.
Moreover, we compared our method with the state-of-the-art methods that use extra clean data and those that do not.
Results are shown in Table~\ref{tabel:clothing1m}.
Compared to HNNs and C-HNNs, our method achieves superior performance, underscoring its effectiveness in handling noisy data scenarios.

% which demonstrate that our method can improve the generalization and the robustness on noisy data.

\noindent \textbf{Few-shot learning.}

 Table~\ref{table:miniimagenet} presents the accuracy on the mini-Imagenet and tiered-Imagenet for few-shot learning. 
  Our method consistently achieves the best performance compared to both HNNs and C-HNNs, highlighting its ability to improve the generalization of hyperbolic neural networks in few-shot learning tasks.

\begin{figure*}[h!]
	\centering
        % \vskip 0.2in
	\vspace{-0.5em}
	\begin{center}
        
		\subfloat{\label{fig:resnet18_sharpness_cifar10} \includegraphics[width=0.24\textwidth]{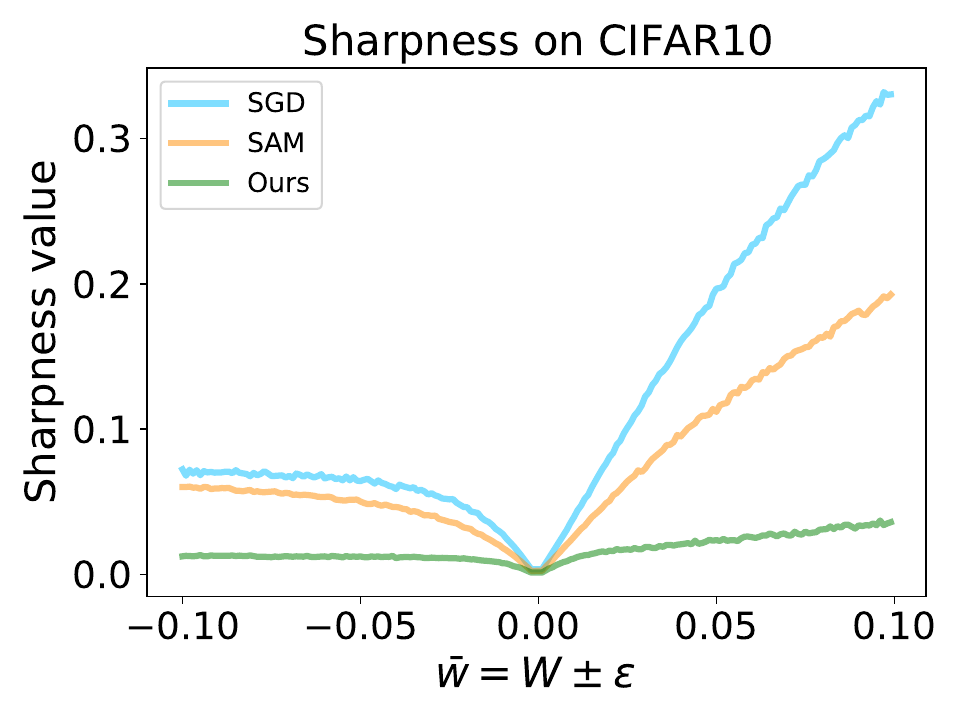}}
		\subfloat{\label{fig:resnet18_eigenvalue_cifar10} \includegraphics[width=0.24\textwidth]{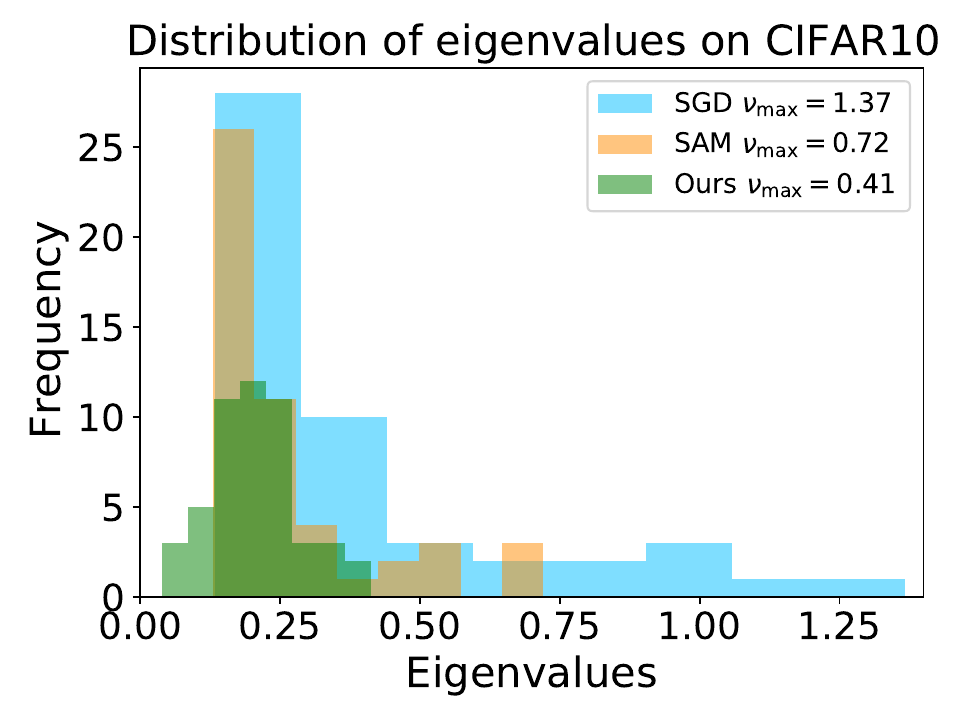}}
            \subfloat{\label{fig:resnet18_sharpness_cifar100} \includegraphics[width=0.24\textwidth]{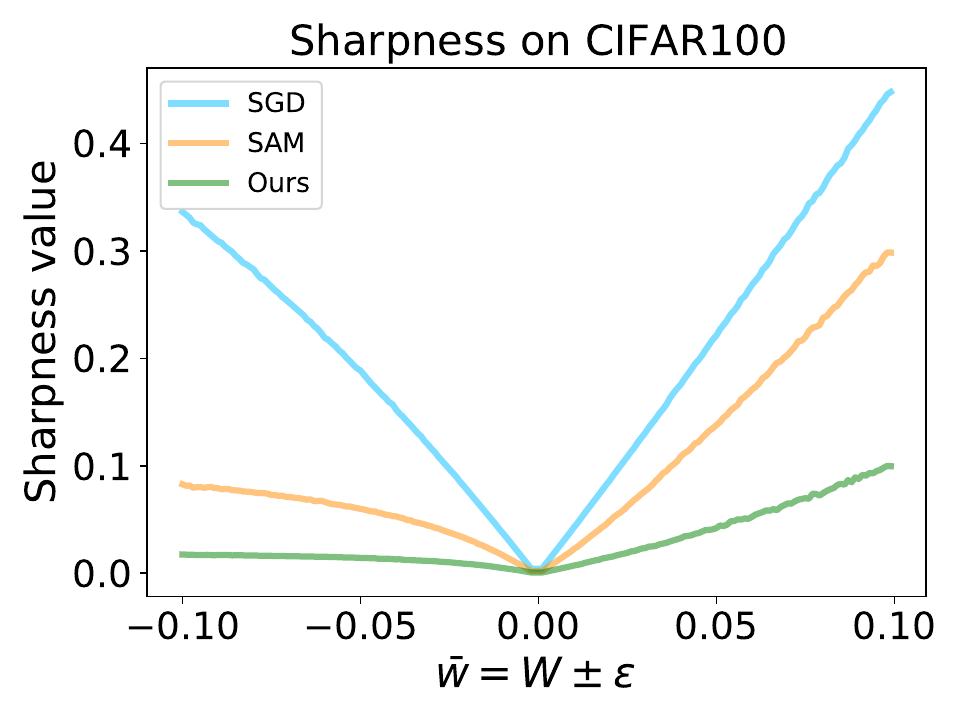}}
		\subfloat{\label{fig:resnet18_eigenvalue_cifar100} \includegraphics[width=0.24\textwidth]{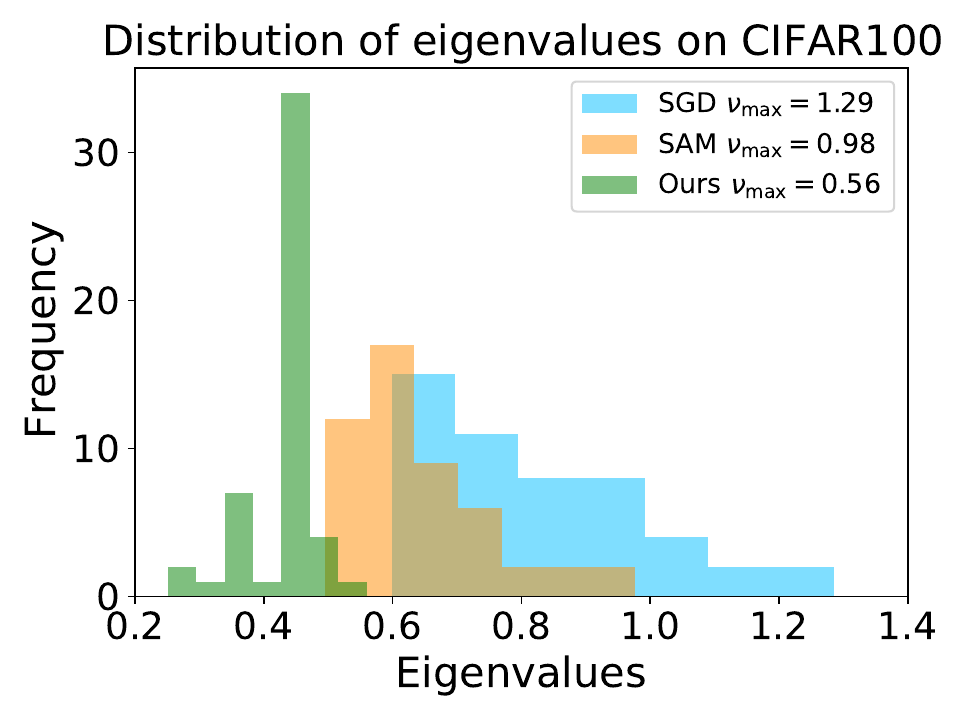}}
        % \vspace{-0.5em}
        \caption{Sharpness and Eigenvalues on the ResNet18 Backbone}
        \label{fig:VIS_resnet18}
	\end{center}
\end{figure*}

\begin{figure*}[h!]
	\centering
        % \vskip 0.2in
	\vspace{-0.5em}
	\begin{center}
        
		\subfloat{\label{fig:wide28-2_sharpness_cifar10} \includegraphics[width=0.24\textwidth]{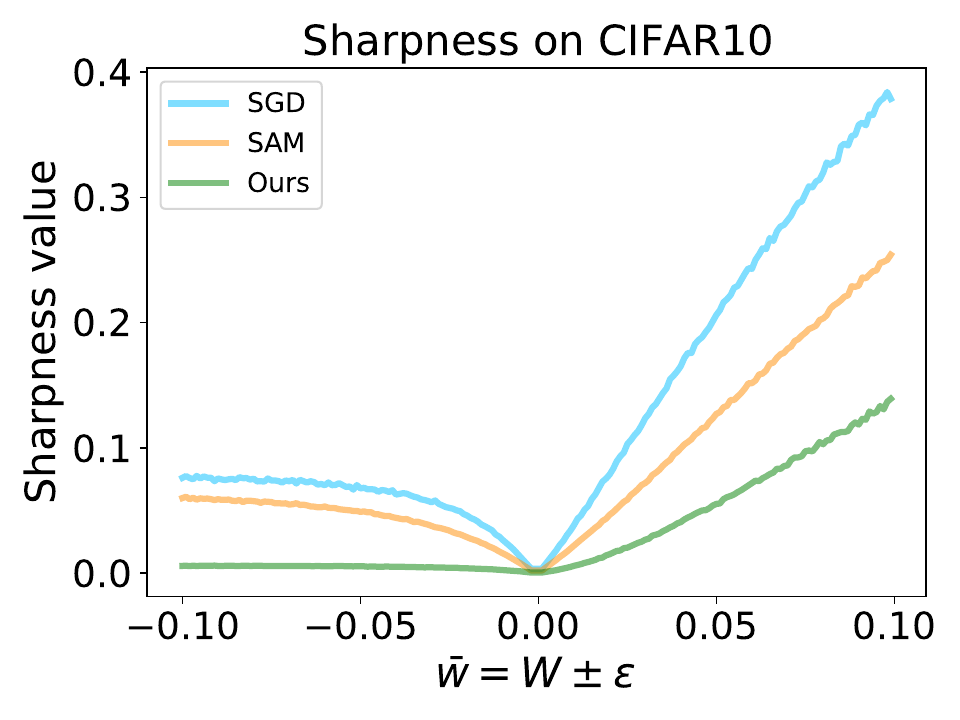}}
		\subfloat{\label{fig:wide28-2_eigenvalue_cifar10} \includegraphics[width=0.24\textwidth]{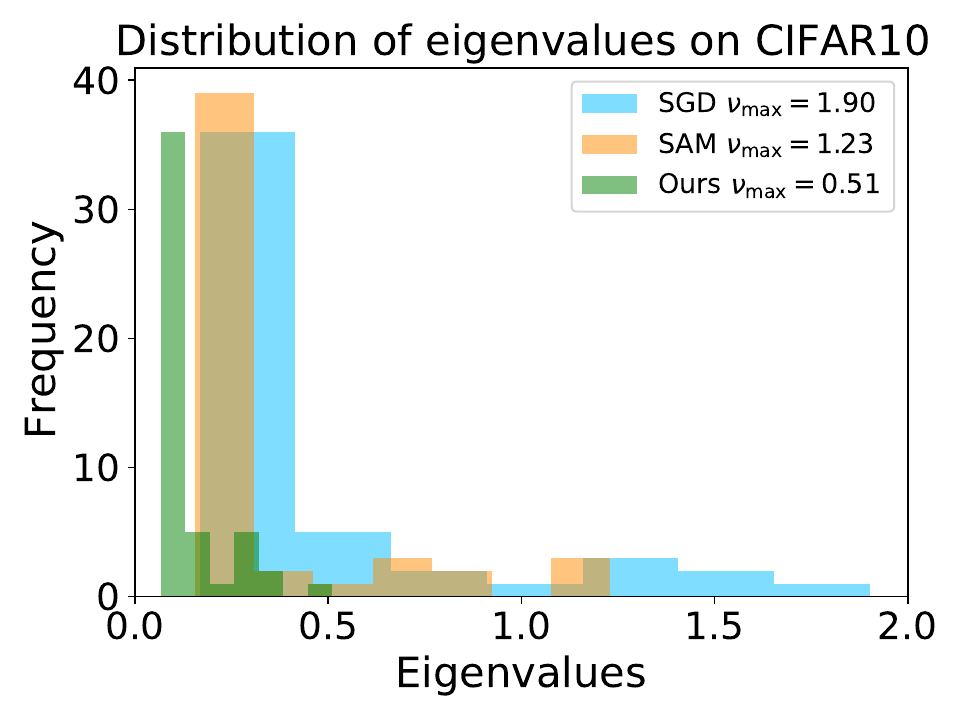}}
            \subfloat{\label{fig:wide28-2_sharpness_cifar100} \includegraphics[width=0.24\textwidth]{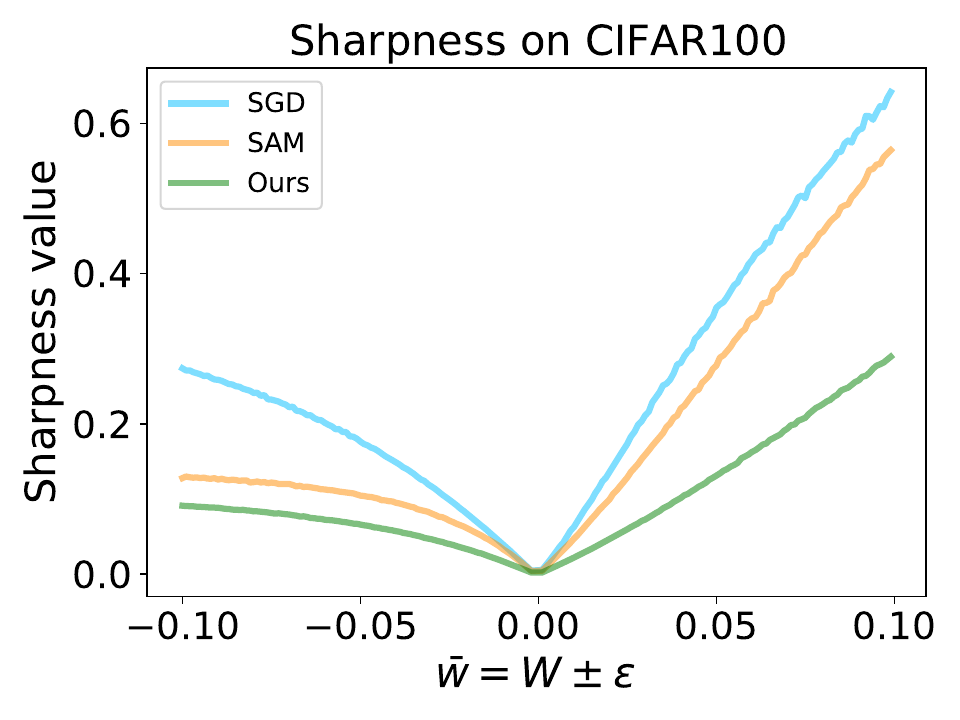}}
		\subfloat{\label{fig:wide28-2_eigenvalue_cifar100} \includegraphics[width=0.24\textwidth]{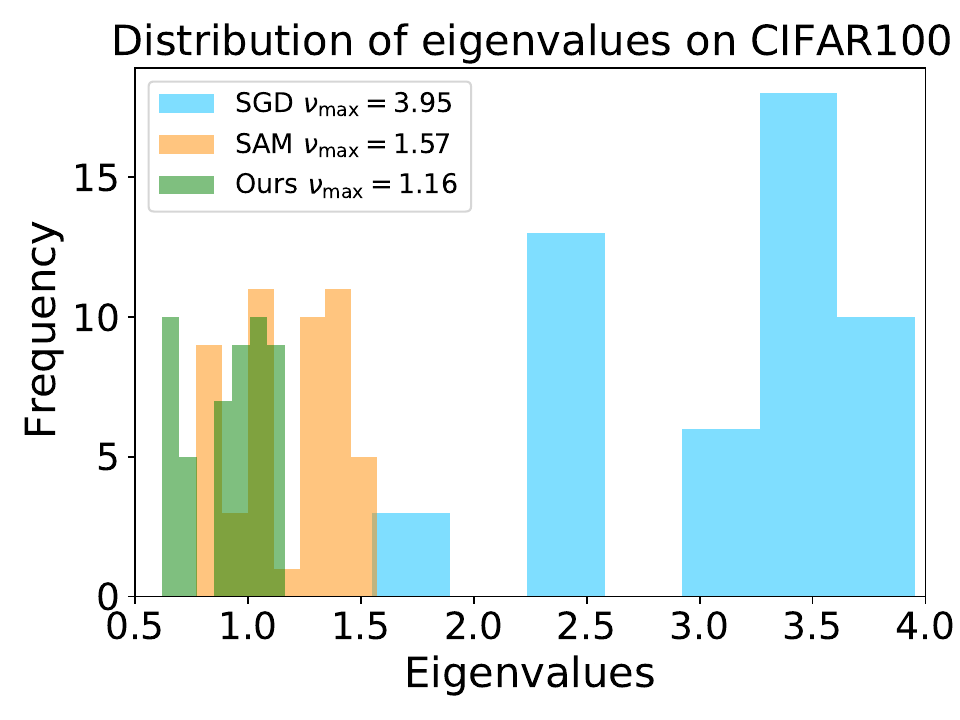}}
        % \vspace{-0.5em}
        \caption{Sharpness and Eigenvalues on the WideResNet28-2 Backbone}
        \label{fig:VIS_wide28-2}
	\end{center}
\end{figure*}

\subsection{Visualization}
% In order to further analyze the effectiveness of our method, we conduct  smoothness analysis on the afore-mentioned settings, which demonstrates that our method effectively smooth the loss landscape of the local minima of HNNs by learning appropriate curvatures. Moreover, we conduct convergence analysis, demonstrating that our method achieves faster convergence speed and obtains the best optima.
In order to further analyze the effectiveness of our method, we conduct  smoothness analysis and convergence analysis on the afore-mentioned settings.

\subsubsection{Smoothness Analysis}
\noindent \textbf{Analysis on classification task.}
We compare the sharpness of our method to that of the SAM and SGD methods with the fixed curvature $c=0.01$. To compute the sharpness term 
$\mathcal{L}_{\mathcal{S}}^{\rm sharp} = \underset{\Vert \boldsymbol{\epsilon} \Vert \le \rho}{\max} \mathcal{L}_{\mathcal{S}}(\boldsymbol{w} + \boldsymbol{\epsilon}, c) - \mathcal{L}_{\mathcal{S}}(\boldsymbol{w}, c)$, we approximate $\epsilon$ as $\rho\nabla_{\boldsymbol{w}} \mathcal{L}_{\mathcal{S}}(\boldsymbol{w}, c)/\Vert \mathcal{L}_{\mathcal{S}}(\boldsymbol{w}, c) \Vert$. We set the increasing radii $\rho$ to compute $\epsilon$, and compute  perturbed weight as $\bar{\boldsymbol{w}} = \boldsymbol{w} \pm \epsilon$.  We also visualize the Hessian spectra of HNNs trained with SGD, SAM, and our method for sharpness comparison, where lower Hessian spectra indicate lower sharpness. We use power iteration~\citep{yao2018hessian} to compute the top eigenvalues of the Hessian and present the histogram of the top-50 eigenvalues for each method.
Our experiments are conducted on the CIFAR10 and CIFAR100 datasets using ResNet18 and WideResNet28-2 backbones.
% We conduct experiments on the CIFAR10 and CIFAR100 datasets with the ResNet18 and WideResNet28-2 backbones.

Results on the ResNet18 backbone are shown in Fig.~\ref{fig:VIS_resnet18}.
% As illustrated in Fig.~\ref{fig:resnet18_sharpness_cifar10} and Fig.~\ref{fig:resnet18_sharpness_cifar100}, 
Sharpness values illustrated in Fig.~\ref{fig:VIS_resnet18} reveal that  the proposed method achieves the lowest  sharpness value compared to the SGD and SAM methods. Notably,  on the CIFAR100 dataset with $\rho = 0.1$, the sharpness of the proposed method is $0.09$, significantly lower than $0.29$ of the SAM method and $0.45$ of  C-HNNs.
Distributions on eigenvalues presented in  Fig.~\ref{fig:VIS_resnet18} reveal that  HNNs 
trained with the proposed method exhibit the lowest Hessian spectra compared to those trained with SAM and SGD with fixed curvatures.
% trained with our method lead to the lowest Hessian spectra compared to SAM and SGD methods with fixed curvatures. 
Results on the ResNet18 backbone demonstrate that our method effectively  learns optimal curvatures, thereby reducing the sharpness and  smoothing the loss landscape of the local minima of HNNs.

Similar results are observed on the WideResNet backbone, as shown in Fig.~\ref{fig:VIS_wide28-2}.
The proposed method consistently achieves the lowest sharpness value compared to the SGD and SAM methods on the CIFAR10 and CIFAR100 datasets. 
Distributions of  eigenvalues show that  HNNs trained with the proposed method lead to the lowest Hessian spectra compared to SAM and SGD methods with fixed curvatures. 
This confirms that our method can effectively smooth the loss landscape of the local minima of HNNs by learning appropriate curvatures, regardless of the backbone used.
\begin{table*}[t]
	\small
	\centering
 % \normalsize
	%\footnotesize
	\caption{Curvature analysis on the tiered-ImageNet dataset.}\label{table:tieredimagenet_ablation}
	\setlength{\tabcolsep}{0.3mm}
	% {
 %            \resizebox{8.3cm}{!}{
        \begin{threeparttable}
		\begin{tabular}{ccccccccc}
			\midrule
			\bfseries Curvature & 0 & $1.0 \times 10^{-4}$ & $1.0 \times 10^{-3}$ & $1.0 \times 10^{-2}$ & $1.0 \times 10^{-1}$ & $5.0 \times 10^{-1}$ & $1.0$  & \bfseries Ours (Converged Curvature)\\
			\toprule
			\bfseries 1-shot 5-way  & $56.33$  &  $50.97$ & $44.70$ & $38.31$ & $70.25$ & $69.94$ & $69.10$ & $\mathbf{72.10} (1.13 \times 10^{-1})$  \\
			\bfseries 5-shot 5-way &  $74.79$  & $74.40$ & $72.40$ & $64.30$ &  $85.15$ & $84.75$  &  $83.75$ &$\mathbf{86.25} (1.27 \times 10^{-1})$ \\
			\midrule
		\end{tabular}
            \begin{tablenotes} % 添加 tablenotes 环境用于脚注
            % \footnotesize
            \item The best results are shown in bold. The contents in `$()$' represent the learned curvatures.
        \end{tablenotes}
         \end{threeparttable}
	% 	}

	% }
\end{table*}

\begin{table*}[t]
	\small
	\centering
	%\footnotesize
	%\resizebox{8.3cm}{!}{
  \caption{Curvature analysis on the CIFAR10-LT and CIFAR100-LT datasets.}
  \begin{threeparttable}
		\begin{tabular}{ccccccc}
			\midrule
			\bfseries Curvature & 0 & $1.0 \times 10^{-4}$ & $5.0 \times 10^{-4}$ & $1.0 \times 10^{-3}$ & $5.0 \times 10^{-3}$ & \bfseries Ours (Converged Curvature)\\
			\toprule
			\bfseries CIFAR10-LT  &  $82.77$ &  $81.77$ & $83.52$ & $80.97$  &$77.49$  & $\mathbf{85.07} (5.36 \times 10^{-4})$   \\
                % \hline
			\bfseries CIFAR100-LT &  $45.92$  & $49.38$ & $48.6$ & $46.48$   &  $43.73$ &$\mathbf{52.33} (3.55 \times 10^{-4})$ \\
			\midrule
		\end{tabular}
		%}
         \begin{tablenotes} % 添加 tablenotes 环境用于脚注
            % \footnotesize
            \item The best results are shown in bold. We adopt the GL-mixture as the baseline. The contents in `$()$' represent the learned curvatures.
        \end{tablenotes}
        \end{threeparttable}
		\label{table:longtailed_ablation}	
\end{table*}

\begin{figure}[h]
	\centering
 % \vspace{-0.35cm}
	% \vspace{-0.5em}
	\begin{center}
		\subfloat{\label{fig:IR=100} \includegraphics[width=0.24\textwidth]{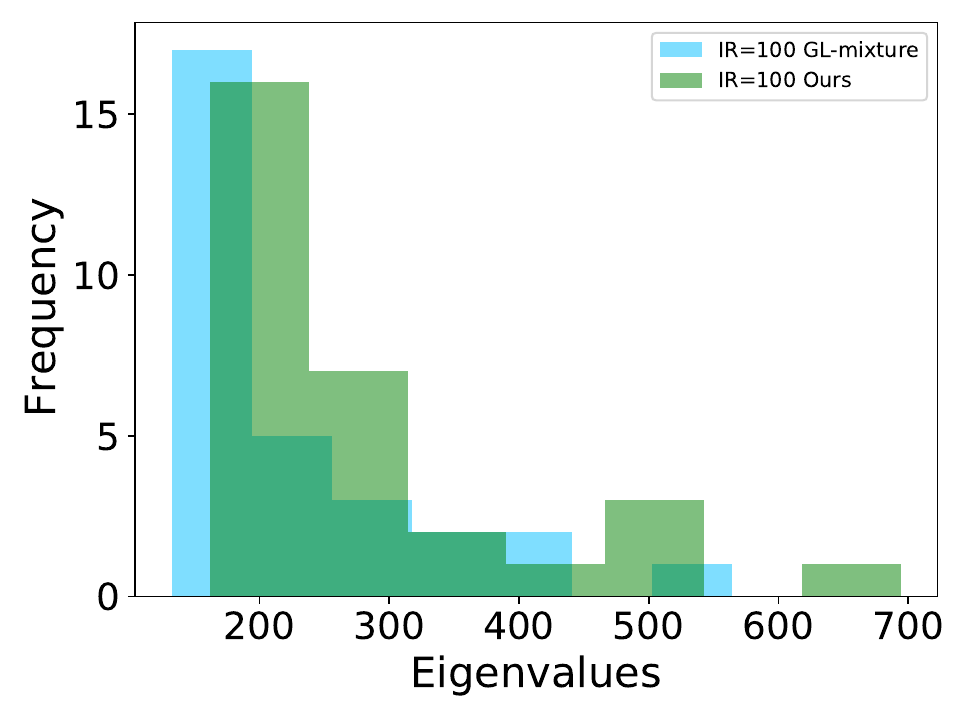}}
		\subfloat{\label{fig:IR=200} \includegraphics[width=0.24\textwidth]{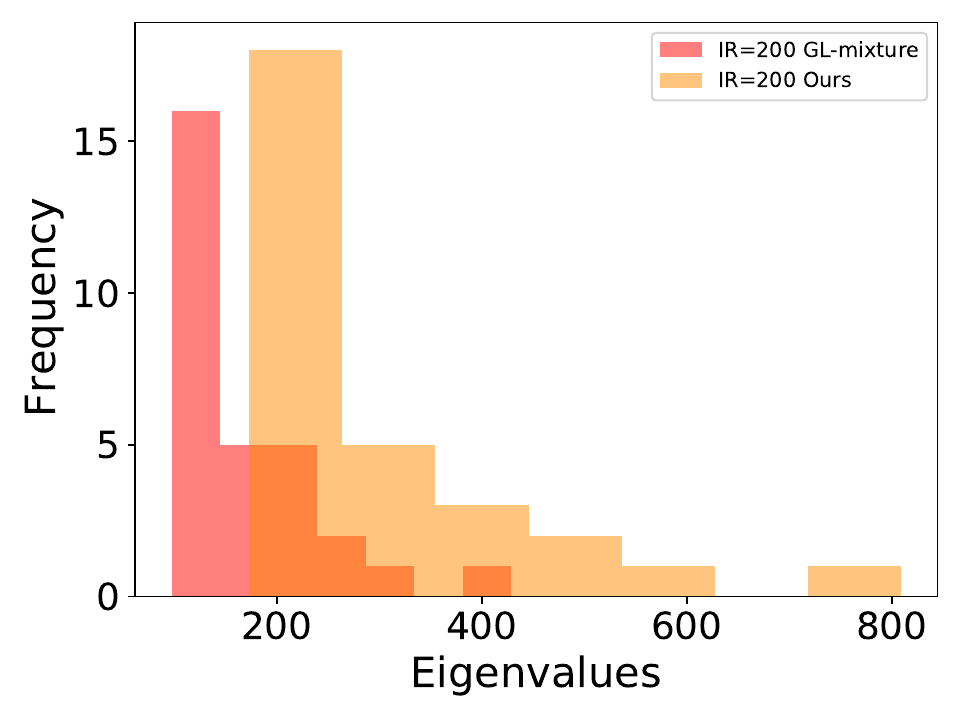}}
		\quad
		% \subfloat{\includegraphics[width=0.44\textwidth]{pdf/legend_unified_end.pdf}}
	\end{center}
 % \vspace{-1.5em}
	\caption{Distribution of eigenvalues on the CIFAR100-LT dataset}
	\label{fig:smoothness_long_tailed}
\end{figure}

\noindent \textbf{Analysis on long-tailed data.}
\label{section:altd}
We visualize the Hessian spectra of models trained with GL-mixture and the proposed method for sharpness comparison. We use power iteration to compute the top eigenvalues of Hessian and report the histogram of the distribution of top 30 Hessian eigenvalues for each method. We compare models trained with GL-mixture and the proposed method on the CIFAR100-LT with Imbalance Ratio (IR) $ = 100$ and IR $= 200$. 
As shown in Fig.~\ref{fig:smoothness_long_tailed}, models trained with our method lead to the lowest Hessian spectra on both IR $= 100$ and IR $= 200$. This again demonstrates that our method can smooth the loss landscape and further improve the generalization. 

For the GL-mixture method, we observe that with IR$ = 200$, the maximum eigenvalues is $8.09 \times 10^{2}$, while the maximum eigenvalues is $6.94 \times 10^{2}$ with IR $= 100$. This demonstrates that the local minima obtained by the GL-mixture method converges to sharper regions when IR is set to $200$ compared to IR$ = 100$.

\begin{figure}[h]
	\centering
 % \vspace{-0.35cm}
	% \vspace{-0.5em}
	\begin{center}
		\subfloat{\label{fig:his_difference} \includegraphics[width=0.24\textwidth]{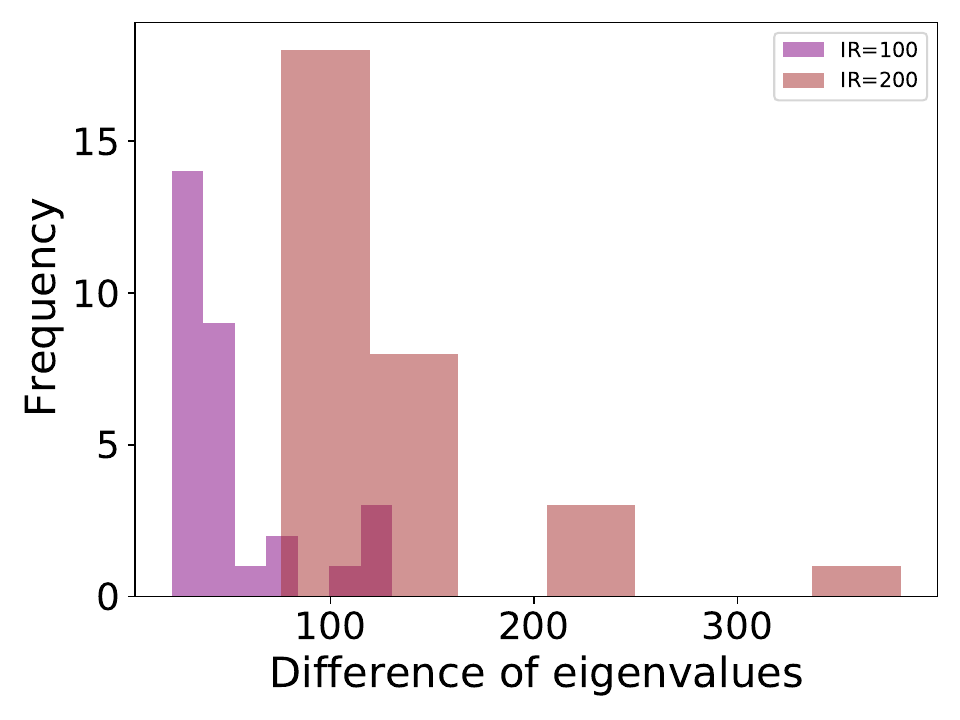}}
		\subfloat{\label{fig:plot_difference} \includegraphics[width=0.24\textwidth]{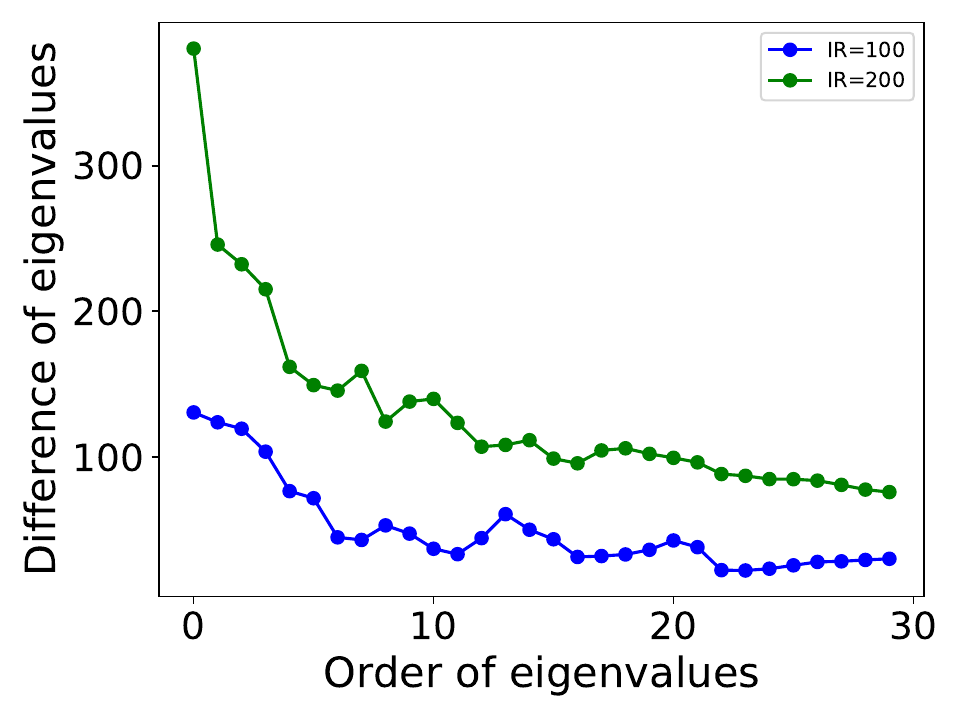}}
		\quad
	\end{center}
 % \vspace{-1.5em}
	\caption{Analysis of differences in eigenvalues}
	\label{fig:smoothness_long_tailed_cha}
\end{figure}

Moreover, we observe that the proposed method reduces more sharpness of the loss landscape with IR $ = 200$, compared with IR $= 100$.
To prove this point, we analyze the differences in eigenvalues between GL-mixture and our method. Specifically, we subtract the eigenvalues of the proposed method from the corresponding ranks of the baseline method.  
We visualize the distributions of differences in eigenvalues, and we plot a line graph of these differences, where the X-axis represents the order of eigenvalues, with '0' denoting the maximum eigenvalue. As shown in Fig.~\ref{fig:smoothness_long_tailed_cha},  the differences in eigenvalues for IR $= 200$  are significantly larger than for IR $= 00$.
These results demonstrate that our method can introduce greater improvements with IR$ = 200$ compared to IR$ = 100$, which analyzes the phenomenon in Table~\ref{table:CIFAR10_CIFAR100}.

% \vspace{-0.35cm}
% \vspace{-1cm}

% \vspace{-1cm}
\begin{figure}[h]
	\centering
 % \vspace{-0.35cm}
	% \vspace{-0.5em}
	\begin{center}
		\subfloat{\label{fig:IR=100} \includegraphics[width=0.24\textwidth]{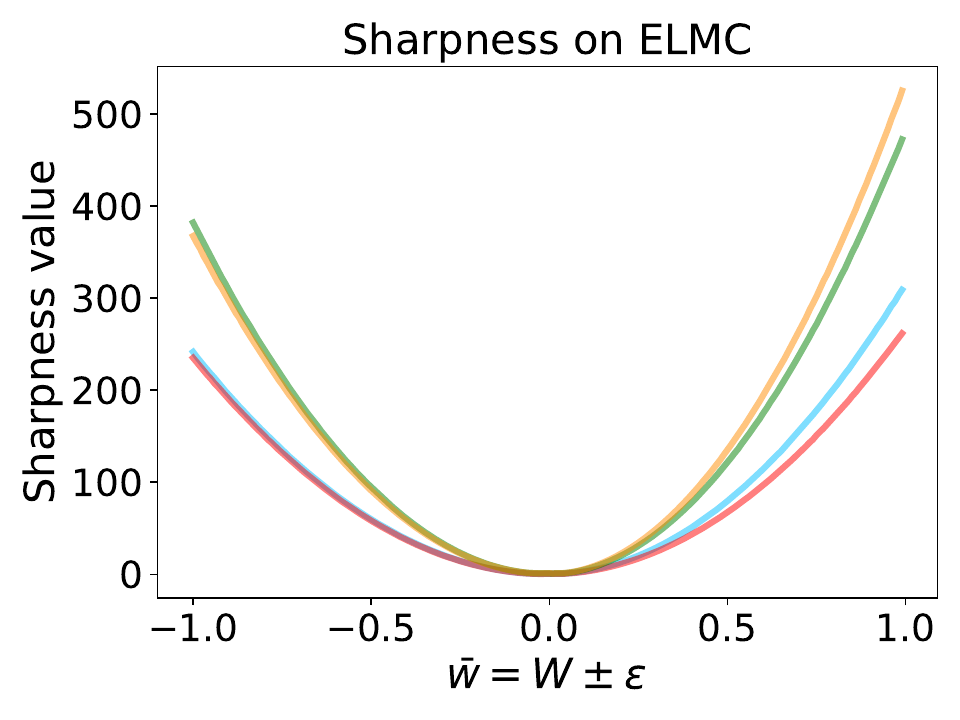}}
		\subfloat{\label{fig:IR=200} \includegraphics[width=0.24\textwidth]{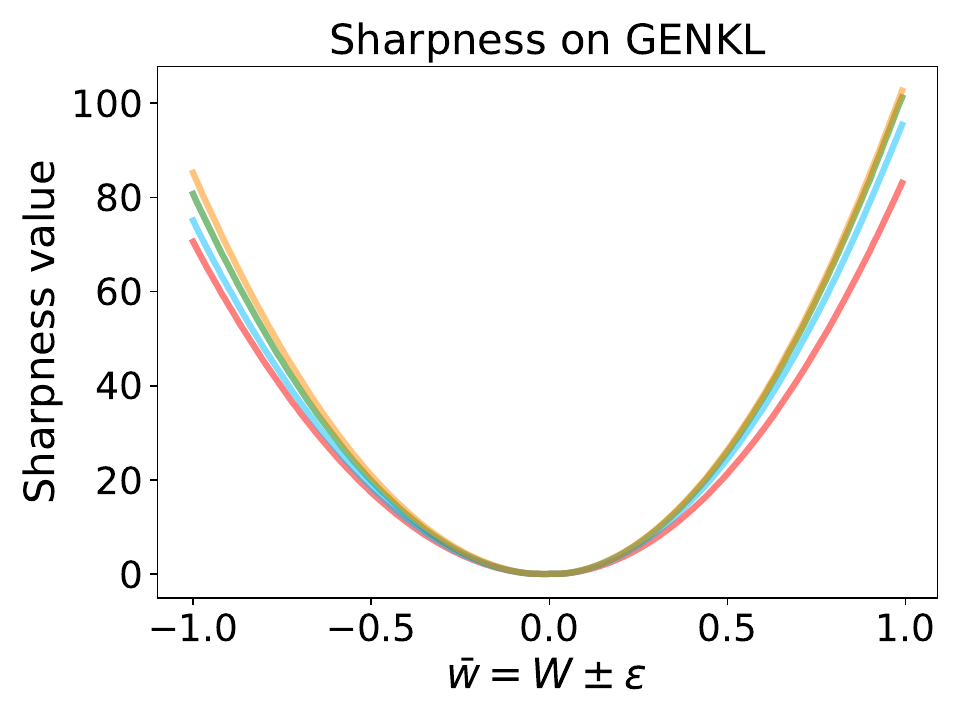}}
		\quad
		\subfloat{\includegraphics[width=0.48\textwidth]{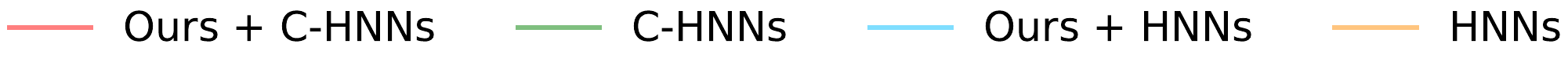}}
	\end{center}
 % \vspace{-1.5em}
	\caption{Sharpness on the learning on noisy data task}
	\label{fig:smoothnes_noisy}
\end{figure}

\begin{figure}[h]
	\centering
 % \vspace{-0.35cm}
	% \vspace{-0.5em}
	\begin{center}
		\subfloat{\label{fig:IR=100} \includegraphics[width=0.24\textwidth]{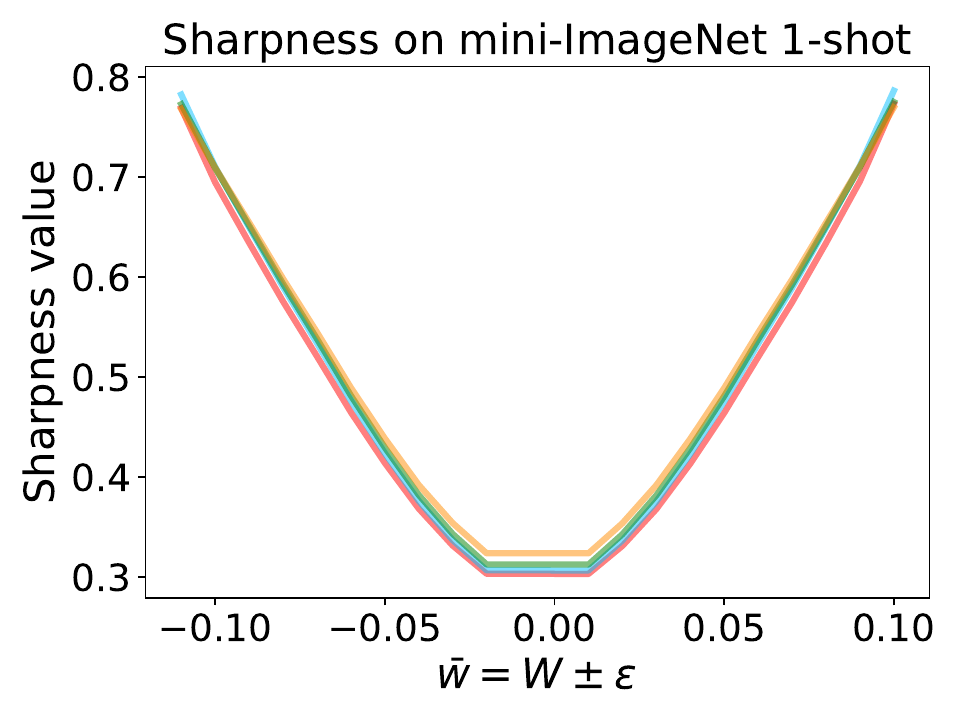}}
		\subfloat{\label{fig:IR=200} \includegraphics[width=0.24\textwidth]{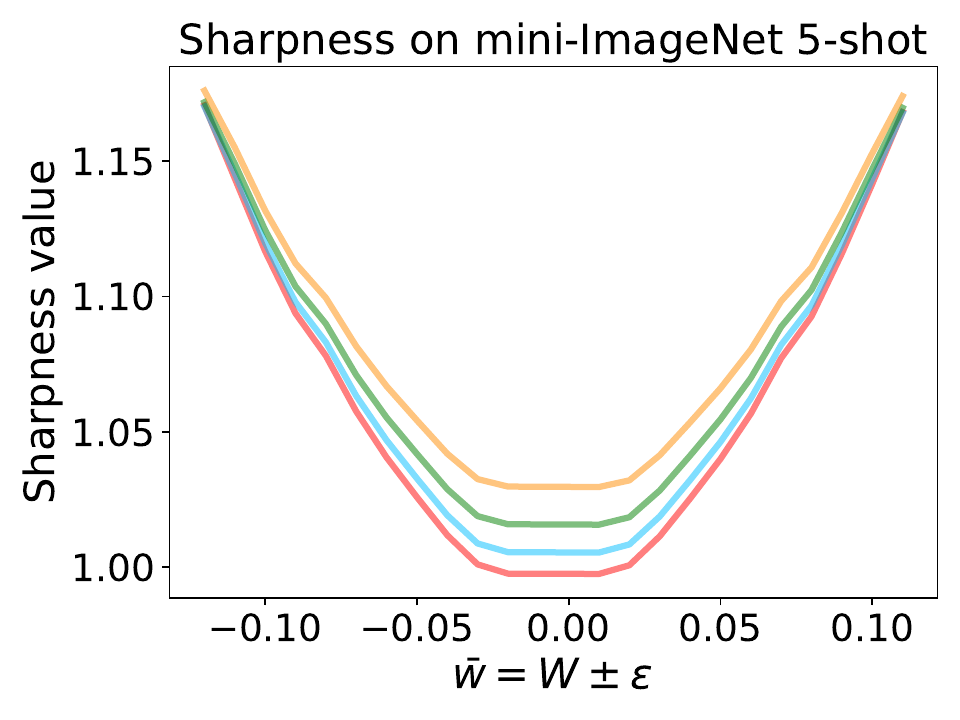}}
		\quad
		\subfloat{\includegraphics[width=0.48\textwidth]{loss_pertube_legend_end.pdf}}
	\end{center}
 % \vspace{-1.5em}
	\caption{Sharpness on the few-shot learning task}
	\label{fig:smoothnes_few_shot}
\end{figure}

\noindent \textbf{Analysis on  noisy data.}
We conduct visualization analysis on two baselines, ELMC and GenKL with HNNs and C-HNNs backbones, which are trained with a fixed curvature $0.1$.
Sharpness values on the ELMC baseline and GenKL baseline illustrated in Fig.~\ref{fig:smoothnes_noisy} reveal that the proposed method achieves the lowest sharpness value compared to the HNNs and C-HNNs methods. 
Visualization results demonstrate that our method effectively learns optimal curvatures, thereby reducing the sharpness and smoothing the loss landscape of the local minima of HNNs, improving the generalization
performance.

\noindent \textbf{Analysis on few-shot learning task.}
We conduct visualization analysis on the mini-imagenet dataset with 1-shot 5-way and 5-shot 5-way two settings.
We compare our method with HNNs and C-HNNs two methods, which are trained with a fixed curvature $1$.
As illustrated in Fig.~\ref{fig:smoothnes_few_shot}, our method again achieves the lowest sharpness across both settings. These visualization results further validate that our method effectively learns optimal curvatures, reducing sharpness and smoothing the loss landscape.

\begin{figure}[h]
	\centering
 % \vspace{-0.35cm}
	% \vspace{-0.5em}
	\begin{center}
		\subfloat{\label{fig:IR=100} \includegraphics[width=0.24\textwidth]{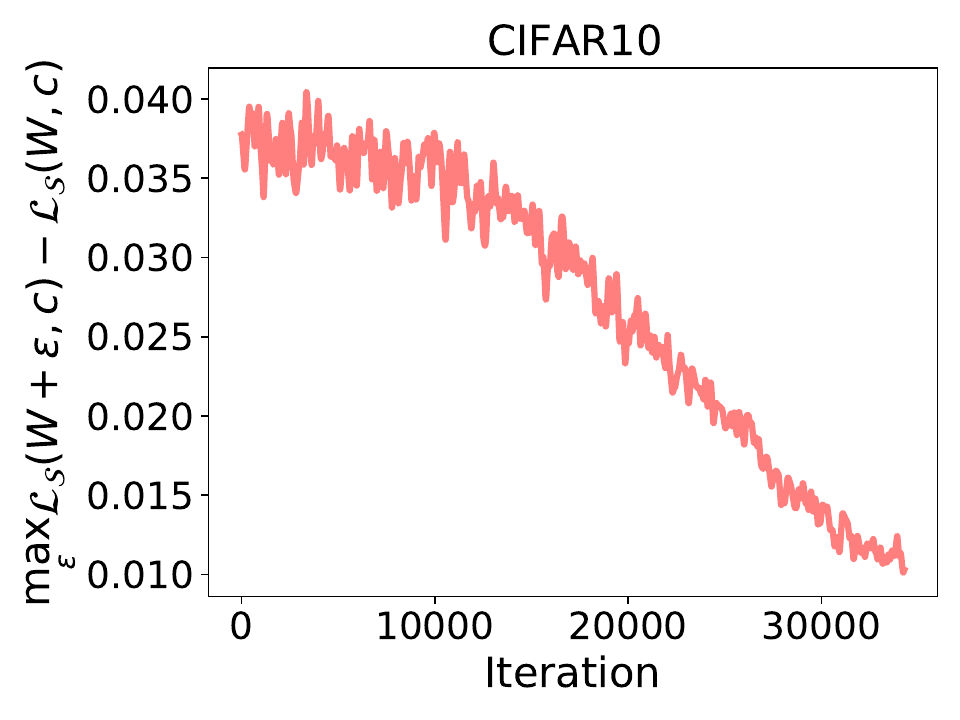}}
		\subfloat{\label{fig:IR=200} \includegraphics[width=0.24\textwidth]{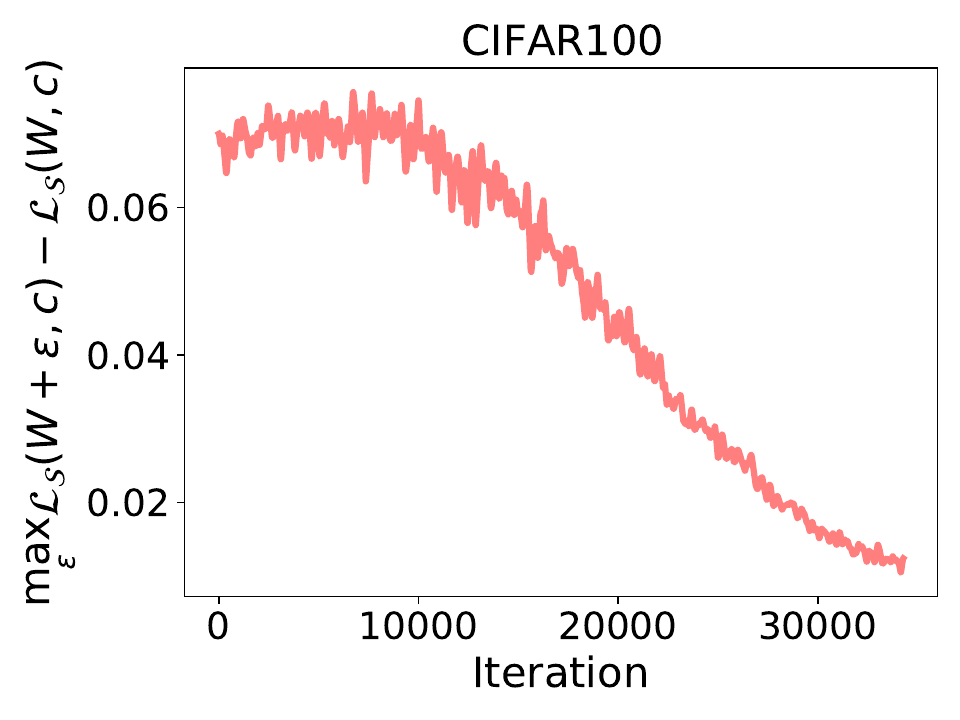}}
		% \quad
		% \subfloat{\includegraphics[width=0.6\textwidth]{pdf/loss_pertube_legend_end.pdf}}
	\end{center}
 % \vspace{-1.5em}
	\caption{Evolution of $\max_{\boldsymbol{\epsilon}} \mathcal{L}_{\mathcal{S}}(\boldsymbol{w} + \boldsymbol{\epsilon}, c) - \mathcal{L}_{\mathcal{S}}(\boldsymbol{w}, c)$ in curvature training step.}
	\label{fig:smoothness_cifar_10_100_resnet18}
\end{figure}

\noindent \textbf{Analysis of bi-level framwork.}
We measure the sharpness of models during training to analyze the effect of bi-level optimization framework. We train the curvature and estimate the sharpness $ \mathcal{L}^{\text{sharp}}_{\mathcal{S}} =  \max_{\boldsymbol{\epsilon}} \mathcal{L}_{\mathcal{S}}(\boldsymbol{w} + \boldsymbol{\epsilon}, c) - \mathcal{L}_{\mathcal{S}}(\boldsymbol{w}, c)$, where experiments are conducted by using the ResNet18 as the backbone on CIFAR datasets. 
We report the evolution of $\max_{\boldsymbol{\epsilon}} \mathcal{L}_{\mathcal{S}}(\boldsymbol{w} + \boldsymbol{\epsilon}, c) - \mathcal{L}_{\mathcal{S}}(\boldsymbol{w}, c)$ during training, where $\mathcal{L}_{\mathcal{S}}(\cdot)$ stands for the training loss computed on the current batch.
As shown in Fig.~\ref{fig:smoothness_cifar_10_100_resnet18}, as the curvature is trained, the sharpness of HNNs $\max_{\boldsymbol{\epsilon}} \mathcal{L}_{\mathcal{S}}(\boldsymbol{w} + \boldsymbol{\epsilon}, c)- \mathcal{L}_{\mathcal{S}}(\boldsymbol{w}, c)$ decreases, indicating that the generalization of HNNs improves.

% \vspace{-1cm}

% \begin{figure}[h]
% 	\centering
%  % \vspace{-0.35cm}
% 	% \vspace{-0.5em}
% 	\begin{center}
% 		\subfloat{\label{fig:IR=100} \includegraphics[width=0.24\textwidth]{loss_pertube_c_iteration_cifar10.pdf}}
% 		\subfloat{\label{fig:IR=200} \includegraphics[width=0.24\textwidth]{loss_pertube_c_iteration_cifar100.pdf}}
% 		% \quad
% 		% \subfloat{\includegraphics[width=0.6\textwidth]{pdf/loss_pertube_legend_end.pdf}}
% 	\end{center}
%  % \vspace{-1.5em}
% 	\caption{Evolution of $\max_{\boldsymbol{\epsilon}} \mathcal{L}(\boldsymbol{w} + \boldsymbol{\epsilon}) - \mathcal{L}(\boldsymbol{w})$ in curvature training step.}
% 	\label{fig:smoothness_cifar_10_100_resnet18}
% \end{figure}

\subsubsection{Convergence Analysis}
\begin{figure}[h!]
	\centering
        % \vskip 0.2in
	
	\begin{center}
        
		\subfloat{ \includegraphics[width=0.24\textwidth]{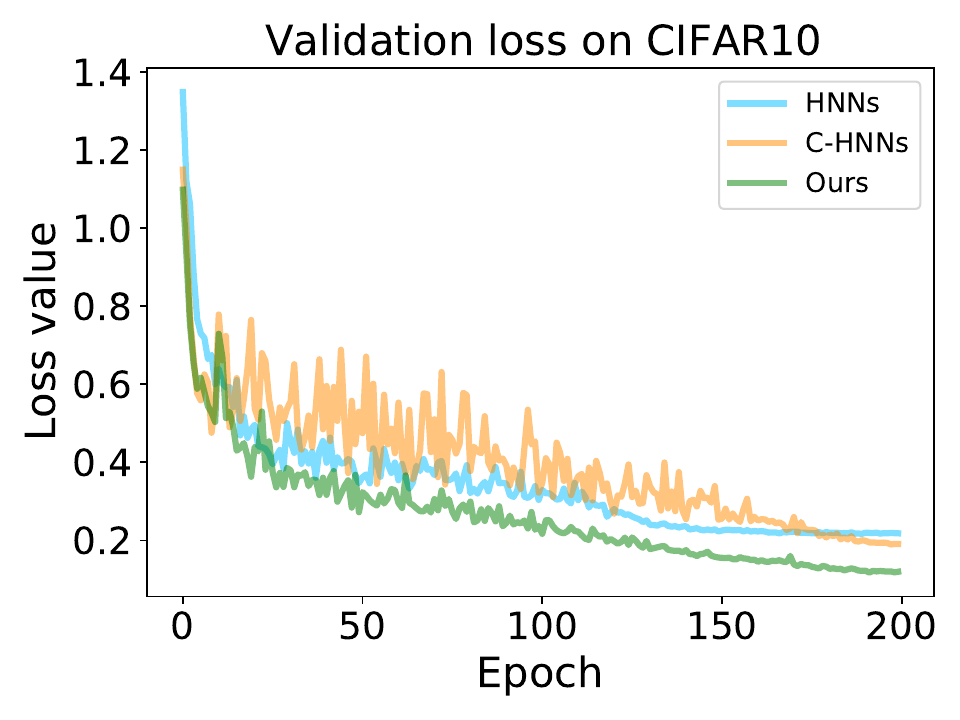}}
		\subfloat{\includegraphics[width=0.24\textwidth]{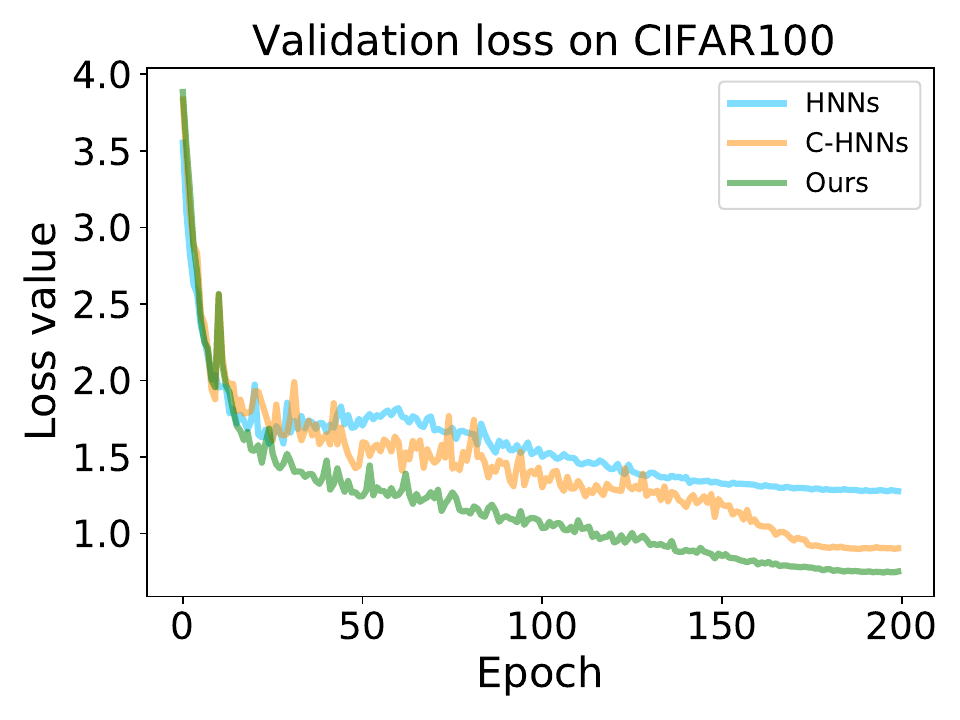}}
        % \vspace{-0.5em}
        \caption{Loss plots for classification task on the ResNet18 Backbone}
        \label{fig:VIS_loss_convergence_resnet18}
	\end{center}
 % \vspace{-1.5em}
 % \vskip -0.2in
	
\end{figure}

\begin{figure}[h!]
	\centering
        % \vskip 0.2in
	\begin{center}
        
		\subfloat{\includegraphics[width=0.24\textwidth]{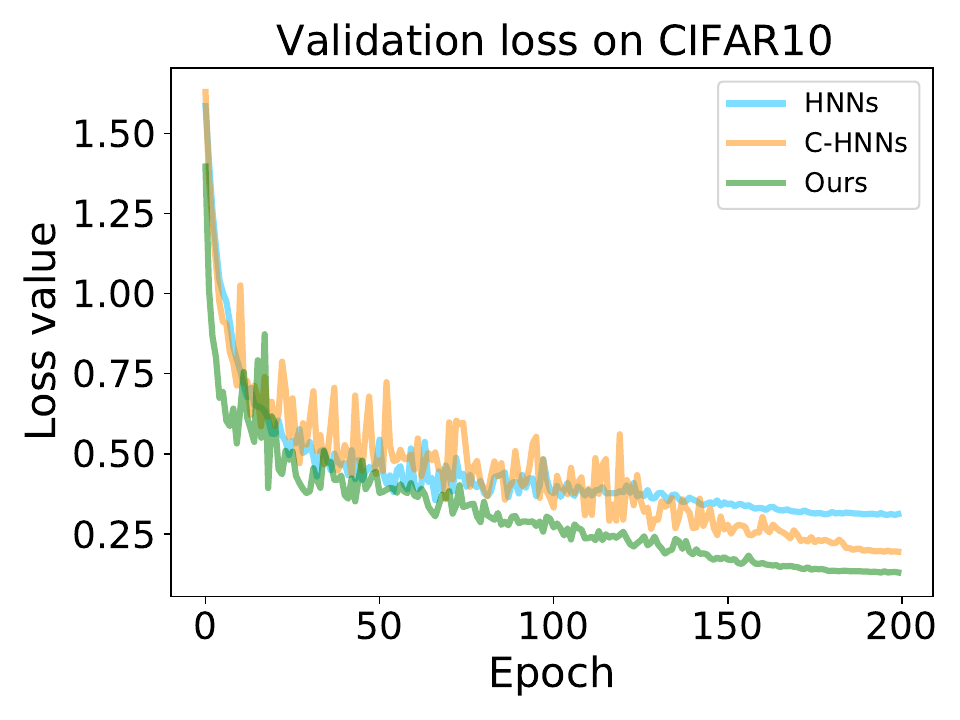}}
		\subfloat{ \includegraphics[width=0.24\textwidth]{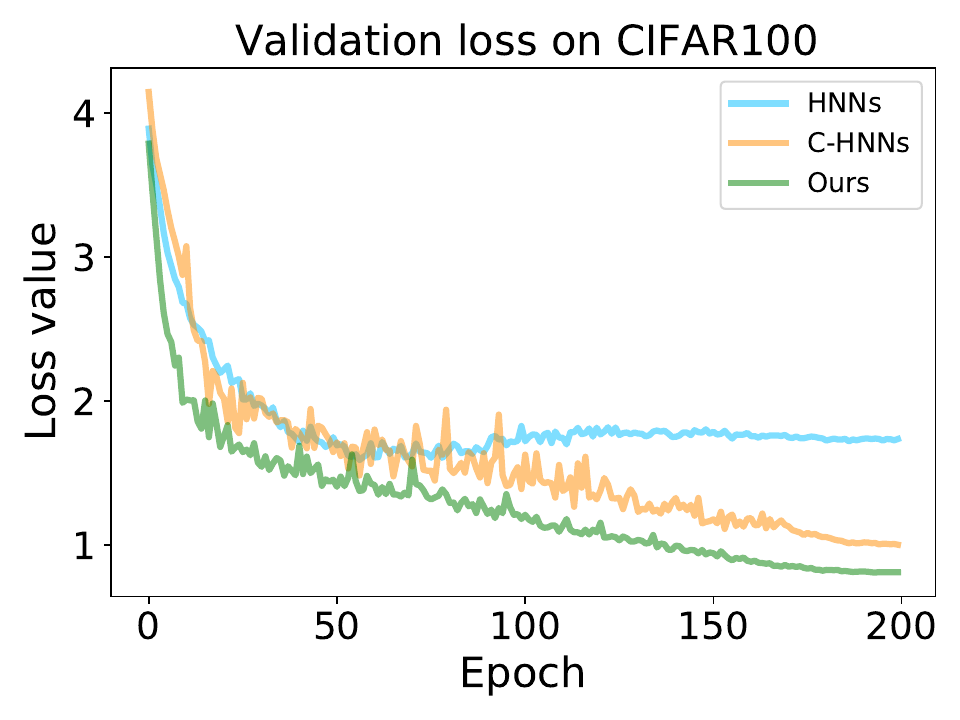}}
        % \vspace{-0.5em}
        \caption{Loss plots for classification task on the WideResNet28-2 Backbone}
        \label{fig:VIS_loss_convergence_wide28-2}
	\end{center}
	
\end{figure}

We visualize the loss plots on the test dataset during the training stage.  Experiments are conducted on the CIFAR10 and CIFAR100 datasets using ResNet18 and WideResNet28-2 backbones, where results are shown in Fig.~\ref{fig:VIS_loss_convergence_resnet18} and Fig.~\ref{fig:VIS_loss_convergence_wide28-2}, respectively.
Results show that the proposed method outperforms other methods on hyperbolic spaces. From the experimental results, we observe that the proposed method achieves faster convergence speed and obtains the best optima on the unseen data. This further demonstrates that our method can improve the generalization of HNNs.

% This involved subtracting the maximum feature value of the baseline method from the maximum feature value of our method, the second-largest feature value of the baseline method from the second-largest feature value of our method, and so on.

% Distribution of  differences
% Line graph of  differences

\begin{table*}[t]
	\centering
	\small
	%\footnotesize
        \caption{Test Accuracy (\%)  on the CIFAR100-LT dataset.}
        % \vskip 0.15in
	\setlength{\tabcolsep}{1mm}{
		\resizebox{13.3cm}{!}{
                \begin{threeparttable} 
			\begin{tabular}{ccccccccccc}
				\midrule
				\bfseries perturbation radius $\hat{\rho}$ &  0.01 & 0.02 & 0.03 & 0.04 & 0.05 & 0.06 & 0.07 & 0.08 & 0.09 & 0.1   \\
				\toprule
    	IR = 50 & $61.62$  & $63.18$ & $62.83$ & $63.53$ & $\mathbf{64.11}$ & $62.28$ & $63.55$ & $62.84$ & $61.59$ & $62.07$   \\
				% \midrule
				IR = 100 & $53.34$  & $57.25$ & $57.00$ & $54.25$ & $\mathbf{59.43}$ & $56.76$ & $58.02$ & $57.50$ & $56.84$ & $56.18$   \\
				% \midrule
    IR = 200 & $\mathbf{52.33}$  & $51.30$ & $ 51.87$ & $51.13$ & $49.49$ & $50.12$ & $49.29$ & $45.94$ & $48.40$ & $46.00$  \\
				\midrule
			\end{tabular}	
     \begin{tablenotes} % 添加 tablenotes 环境用于脚注
            % \footnotesize
            \item The best results are shown in bold. We adopt the GL-mixture as the baseline.
        \end{tablenotes}	
        \end{threeparttable}
  }
		\label{table:CIFAR100_rho_analysis}	
         
	}
\end{table*}

\begin{table}[t]
	\centering
	\small
	%\footnotesize
        \caption{Test Accuracy (\%)  on the CIFAR100 dataset.}\label{table:CIFAR100_T_analysis}
        % \vskip 0.15in
	% \setlength{\tabcolsep}{0.3mm}{
	% 	\resizebox{8.3cm}{!}{
			\begin{tabular}{cccc}
				\midrule
				\bfseries Iterations of inner level $T$ &  2 & 5 & 10   \\
				\toprule
				Resnet18 & $80.61$  & $80.32$ & $80.38$    \\
				% \hline
				WideResNet28-2 & $78.35$  & $77.96$ & $78.16$   \\
				\midrule
			\end{tabular}
                \begin{tablenotes} % 添加 tablenotes 环境用于脚注
            % \footnotesize
            \item The best results are shown in bold.
        \end{tablenotes}	
	% 	}

	% }
\end{table}

\begin{table*}[t]
	\centering
	\normalsize
	%\footnotesize
	\setlength{\tabcolsep}{3mm}{
         \begin{threeparttable}
         \caption{ Training time (seconds) on the classification task.}
		% \resizebox{8.3cm}{!}{
			\begin{tabular}{cccccc}
				\midrule
				& \bfseries BackBone &  ResNet18  & WideResNet28-2 & PyramidNet110  \\
    \toprule
    \multirow{2}{*}{CIFAR10 
                     }
                     % & Parameters &  1485392 &  11237056  &  18461457 \\
			&	C-HNNs & $1675.732$  & $2509.873$ & $26260.523$  \\
				
				& Ours & $2032.880$  & $2880.888$ & $29886.101$   \\
				\midrule
     \multirow{2}{*}{CIFAR100 
                     }
                       % & Parameters &  1508432 &  11260096  &  18461457 \\
			&	C-HNNs & $1699.631$ & $2534.344$ & $26338.488$    \\
				& Ours & $2058.302$  & $2908.640$ & $29986.054$   \\
				\midrule
			\end{tabular}
   % \begin{tablenotes} % 添加 tablenotes 环境用于脚注
   %          % \footnotesize
   %          \item The best results are shown in bold. The experiments on CIFAR10 and CIFAR100 datasets are repeated for 5 times, and we report both average accuracy and standard deviation.
   %      \end{tablenotes}
		\label{table:CIFAR100_time}	
             \end{threeparttable}
	}
\end{table*}

\begin{table*}[t]
	\centering
	\normalsize
	%\footnotesize
	\setlength{\tabcolsep}{3mm}{
         \begin{threeparttable}
         \caption{ Training memory (MB)  on the classification task.}
		% \resizebox{8.3cm}{!}{
			\begin{tabular}{cccccc}
				\midrule
				& \bfseries BackBone &  WideResNet28-2  &  ResNet18 & PyramidNet110  \\
    \toprule
    \multirow{2}{*}{CIFAR10 
                     }
               % & Parameters &  1485392 &  11237056  &  18461457 \\
			&	C-HNNs & $1192$  & $1456$ & $13426$  \\
				
				& Ours & $1411$  & $1816$ & $15592$   \\
				\midrule
     \multirow{2}{*}{CIFAR100 
                     }
                       % & Parameters &  1508432 &  11260096  &  18461457 \\
			&	C-HNNs & $1344$ & $1458$ & $13560$    \\
				& Ours & $1606$  & $1818$ & $15890$   \\
				\midrule
			\end{tabular}
   % \begin{tablenotes} % 添加 tablenotes 环境用于脚注
   %          % \footnotesize
   %          \item The best results are shown in bold. The experiments on CIFAR10 and CIFAR100 datasets are repeated for 5 times, and we report both average accuracy and standard deviation.
   %      \end{tablenotes}
		\label{table:CIFAR100_memory}	
             \end{threeparttable}
	}
\end{table*}

\subsection{Curvature Analysis for the Generalization of HNNs}

% \subsection{Ablation}

% 待做的实验：
% 1. 在cifar10上做可视化实验
% 2. 在wideresnet上做可视化实验
% 3. 在resnet18和wideresnet上做收敛性分析的实验

We set different curvatures to train the HNNs to demonstrate the effects of curvatures for the generalization of HNNs. For the few-shot learning task on the tiered-imagenet dataset, we set the curvatures as $\{0, 0.0001, 0.001, 0.01, 0.1, 0.5,1\}$, where $0$ denotes Euclidean spaces. Results are shown in Table ~\ref{table:tieredimagenet_ablation}. For the task of learning from long-tailed data, we set the curvatures as $\{0, 0.0001,0.0005,0.001,0.005\}$ and set the imbalance ratio as 200. 
The converged curvatures of the few-shot learning task on tiered-imagenet dataset on 1-shot 5-way and 5-shot 5-way settings are $1.13 \times 10^{-1}$ and  $1.27 \times 10^{-1}$, respectively.
As to the long-tail classification task, the converged curvatures on the  CIFAR10-LT and CIFAR100-LT datasets are $5.36 \times 10^{-4}$ and $3.55 \times 10^{-4}$, respectively.
Results are shown in Table~\ref{table:longtailed_ablation}. The experimental results indicate that different curvatures lead to varying generalization performances, highlighting the importance of significance for the generalization of HNNs.
Besides, the results demonstrate that the proposed method achieves the best generalization performance, demonstrating its capability to learn optimal curvatures to improve the generalization of HNNs.

\subsection{Analyses of Hyper-Parameters}
In this subsection, we analyze the effect of some important hyper-parameters, \textit{i,e.},  the perturbation radius $\hat{\rho}$ and the inner-level iteration $T$.

\subsubsection{Analysis of the Perturbation Radius $\hat{\rho}$}

We utilize the perturbation radius $\hat{\rho}$
to update the parameters of HNNs in Eq.~\eqref{equation:inner_level _update}.
% , following by the work of sharpness aware minimization~\citep{foret2020sharpness}.  
We adopt the task of learning from long-tailed data as an example, using the GL-mixture as the baseline. 
We set $\hat{\rho}$ as $\{0.01,0.02,0.03,0.04,0.05,0.06,0.07,0.08,0.09,0.1\}$.
Results are shown in Table~\ref{table:CIFAR100_rho_analysis}.

Experimental results show that the overall algorithm is somewhat sensitive to $\hat{\rho}$, which is consistent with the observation in existing works, \textit{i.e.}, the SAM method is also sensitive to the perturbation radius $\hat{\rho}$~\citep{foret2020sharpness, wang2024enhancing, kwon2021asam}.
The key reason for this sensitivity is the scale-dependency of sharpness, \textit{i.e.}, scaling the model parameters alters the sharpness without affecting the optimal value~\citep{kwon2021asam}.
As a result, small changes in parameter scale may significantly influence optimization, leading to large performance variations when the radius is varied.

In the future, we plan to develop a scale-invariant SAM algorithm by introducing normalization strategies and regularization terms that ensure consistent sharpness behavior under parameter rescaling,  improving robustness to the choice of $\hat{\rho}$. Furthermore, we will explore learning-to-learn strategies to automatically learn the optimal perturbation radius during training, reducing manual hyperparameter tuning and improving the practicality.

We also observe that compared to IR=50 and IR=100, the performances in the setting IR=200 is more sensitive to the perturbation radius $\hat{\rho}$.
This is because that
the obtained local minima usually converge to the sharper regions with IR set as $200$, compared with IR $= 100$, as discussed in section~\ref{section:altd}

% Hence,  IR $= 200$ is sensitive about the  perturbation radius compared with IR $= 100/50$.

\subsubsection{Analysis of the Inner-Level Iteration $T$}
In our experimental settings, we set $T$ as 2, meaning that we first update the parameters of HNNs and then update the curvature. Notably, more inner-level iterations do not correspond to more updates of HNNs or more epochs.   The iterations for optimizing HNNs remain the same for different values of $T$, but the iterations for optimizing curvature vary. For example, with $T=5$, the iterations for optimizing curvature are one-fifth of those with $T=1$.
To analyze the effect of $T$, we set  $T$ as 5 and 10 for the classification task and CIFAR100 dataset.
Results are shown in Table~\ref{table:CIFAR100_T_analysis}.
We can find that our method is insensitive to the inner-level iterations $T$.
Some works~\citep{wu2018understanding} argue that more inner-level iterations contribute to better performance theoretically.
Our experimental results are contrary to this conclusion. We consider there might be two possible reasons. 
(1) The implicit differentiation method used to compute the gradients of curvatures is independent of the inner-level iterations. (2) More inner-level optimization iterations result in fewer optimization iterations for curvatures.

% (1) We utilize the proposed implicit differentiation method to update the curvature, where the computation of gradient is independent of the inner-level iteration. (2) More iterations of inner-level optimization mean that less optimization iterations for curvature.

% We employ the $T=5$ as the example to explain the reason. $T=5$ means that we update the parameters of HNNs for 5 iterations, and then update the curvature. Thus the iterations of optimizing HNNs are same for different $T$, but the iterations of optimizing curvature are different. As to $T=5$, the iterations of optimizing curvature is one fifth of $T=1$. \textit{If we set same iteration of curvature optimization for different $T$, the training iterations and epochs for optimization of HNNs are different. This leads to unfair comparisons with comparative methods.} \textbf{Thus our experimental settings are fair with other methods and are reasonable.  }

% The effect of the perturbation radius on HNNs is similar to that of the perturbation radius on neural networks in the Euclidean spaces.

\subsection{Time and Memory Consumption}
We conduct experiments to evaluate the time and memory consumption on the CIFAR datasets using three backbones, \textit{i.e.} WideResNet28-2, ResNet18, and PyramidNet110.
C-HNNs are used as the baseline, and we measure the time consumption over 200 epochs.
Training time and memory consumption are shown in Table~\ref{table:CIFAR100_time} and Table~\ref{table:CIFAR100_memory}, respectively. 
Results indicate that the proposed method increases $10\%-20\%$ time consumption and memory consumption on the three backbones.
Given that our method yields obvious performance improvement over C-HNNs, the additional time and memory costs are acceptable.

\section{Conclusion}
% In this paper, we have presented that generalization  bound on the hyperbolic spaces that connects the curvature with the generalization ability of HNNs. 
In this paper, we have presented that curvatures of hyperbolic spaces significantly affect the generalization of HNNs, as evidenced by the derived PAC-Bayesian
generalization bound of HNNs.
The proposed sharpness-aware curvature learning method can learn optimal curvatures to improve the generalization of HNNs. The introduced implicit differentiation algorithm can efficiently approximate gradients of curvatures. We have presented theoretical analyses, demonstrating that the approximate error in our method is bounded and our algorithm can converge. Experiments 
on four settings: classification, learning from long-tailed data, learning from noisy data, and few-shot learning
show that our method can learn curvatures to improve the generalization of HNNs.
 	
In this work, we study generalization on a single hyperbolic space with a single curvature. Actually, real-world data with complex hierarchical structures should be represented on product manifolds with multiple curvatures. In the future, we will study generalization for product manifolds.

\noindent \textbf{Acknowledgements.} This work was supported by the Natural Science Foundation of China (NSFC) under Grant No.62406009, the Shenzhen Science and Technology Program under Grant No. JCYJ20241202130548062, the Natural Science Foundation of Shenzhen under Grant No. JCYJ20230807142703006,  the NSFC under Grant No. 62172041, and the Key Research Platforms and Projects of the Guangdong Provincial Department of Education under Grant No.2023ZDZX1034.

\noindent \textbf{Data Availibility.} All datasets used in this study are open-access and have been cited in the paper.

% \begin{IEEEbiographynophoto}{John Doe}
% Use $\backslash${\tt{begin\{IEEEbiographynophoto\}}} and the author name as the argument followed by the biography text.
% \end{IEEEbiographynophoto}

% \vfill

\bibliography{Tangentspace}% common bib file
%% if required, the content of .bbl file can be included here once bbl is generated
%%\input sn-article.bbl

\end{document}